\documentclass{ifm-tr}

\usepackage{amsmath,amsfonts,bm}

\def\eqref#1{equation~\ref{#1}}

\def\1{\bm{1}}

\DeclareMathAlphabet{\mathsfit}{\encodingdefault}{\sfdefault}{m}{sl}
\SetMathAlphabet{\mathsfit}{bold}{\encodingdefault}{\sfdefault}{bx}{n}

\usepackage{amsmath,amsfonts,amssymb,mathtools,bm}  %
\usepackage{thmtools}          %
\usepackage{algorithm}         %
\usepackage{algpseudocode}     %
\usepackage{enumitem}          %
\usepackage{siunitx}           %
\usepackage{tikz}              %

\makeatletter
\let\jais@kernelbegin\begin
\makeatother
\usepackage{arabtex}
\usepackage{utf8}
\setcode{utf8}
\makeatletter
\let\jais@arabbegin\begin
\def\begin#1{\def\jais@a{#1}\def\jais@b{document}%
  \ifx\jais@a\jais@b\expandafter\jais@kernelbegin\else\expandafter\jais@arabbegin\fi{#1}}
\makeatother

\usetikzlibrary{arrows.meta,positioning}
\usepackage{booktabs}          %
\usepackage{subcaption}        %
\usepackage{booktabs,tabularx,array}
\usepackage{todonotes} %
\usepackage{pgfplots}
\usepackage{url}            %

\usepackage{xurl}          
\usepackage{nicefrac}       %
\usepackage{microtype}      %
\usepackage{graphicx}
\usepackage{xspace}
\usepackage[frozencache,cachedir=.]{minted}
\usepackage{longtable}
\usepackage{nicematrix}
\usepackage{multirow}
\usepackage{makecell}
\usepackage{soul}  
\usepackage{wrapfig}
\usepackage{pdflscape}

\usetikzlibrary{arrows.meta, positioning, fit, backgrounds, calc}
\usepackage[bottom]{footmisc}

\usepackage[table]{xcolor}
\definecolor{jaisrow}{RGB}{235,235,235} %

\usepackage{colortbl}
\usepackage[normalem]{ulem}
\usepackage[most]{tcolorbox}

\newtcolorbox{promptbox}[1][]{
    enhanced,
    breakable,
    colback=gray!5,
    colframe=gray!55,
    boxrule=0.6pt,
    arc=2pt,
    left=6pt,
    right=6pt,
    top=6pt,
    bottom=6pt,
    fonttitle=\bfseries,
    title={#1}
}

\newtcolorbox{takeawaysbox}{
    enhanced,
    breakable,
    colback=blue!3,
    colframe=blue!40!black,
    boxrule=0.6pt,
    arc=2pt,
    left=6pt,
    right=6pt,
    top=4pt,
    bottom=4pt,
    fonttitle=\bfseries,
    title={Key Takeaways}
}
\newenvironment{takeaways}{\begin{takeawaysbox}\begin{itemize}[leftmargin=*,nosep]}{\end{itemize}\end{takeawaysbox}}

\newcommand{\huggingface}{\raisebox{-1.5pt}{\includegraphics[height=1.05em]{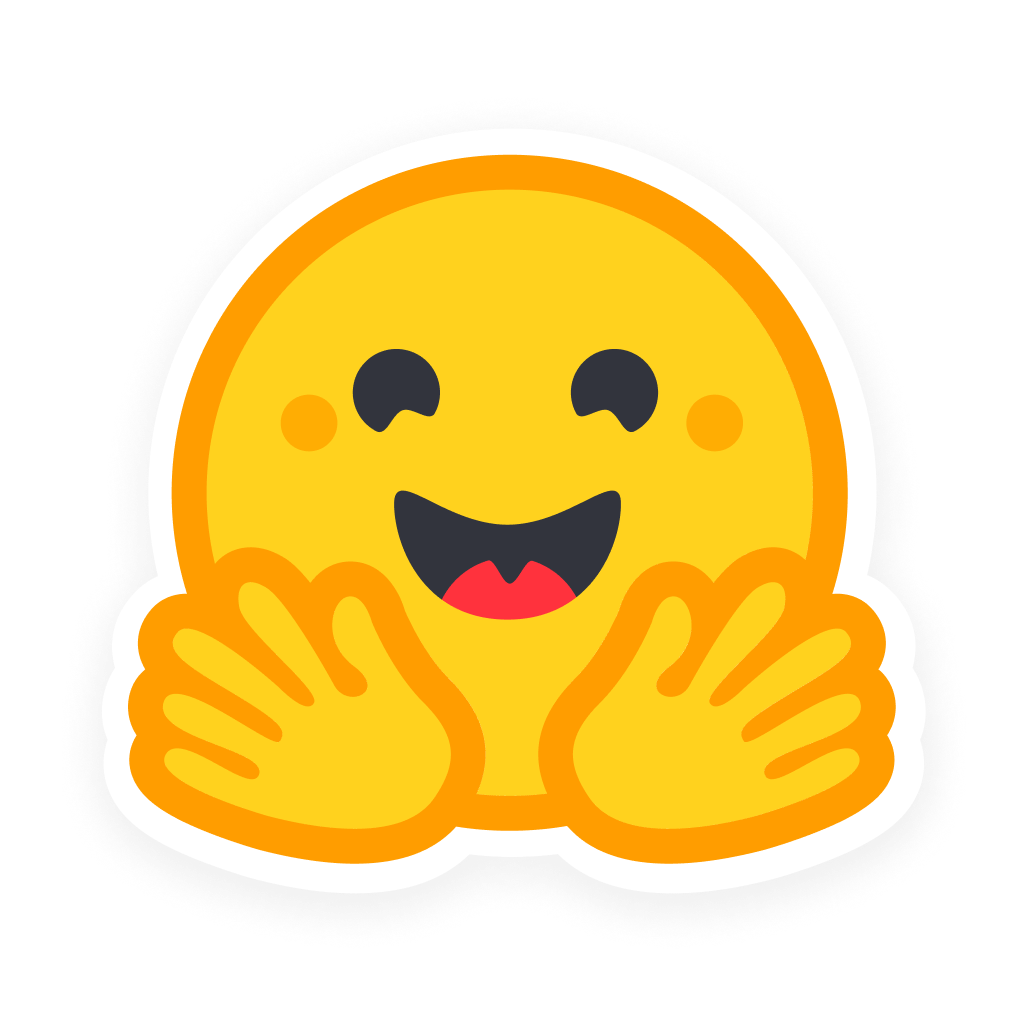}}\xspace}
\newcommand{\github}{\raisebox{-1.5pt}{\includegraphics[height=1.05em]{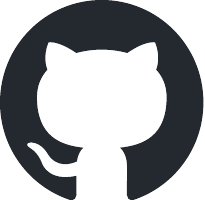}}\xspace}
\newcommand{\logo}{\raisebox{-1.5pt}{\includegraphics[height=1.05em]{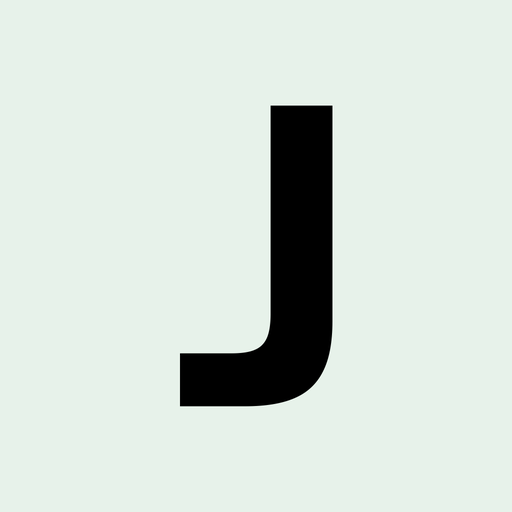}}\xspace}
\newcommand{\wandb}{\raisebox{-1.5pt}{\includegraphics[height=1.05em]{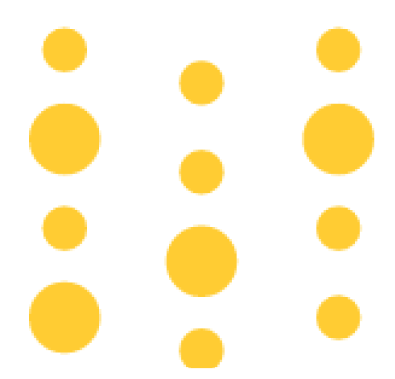}}\xspace}

\colorlet{chkdani}{violet}
\colorlet{chknik}{magenta}
\colorlet{chkmaaz}{blue}
\colorlet{chkshivam}{teal}
\colorlet{chkamr}{olive}
\colorlet{chkTBD}{red}
\colorlet{chkfrikha}{cyan!60!black}
\colorlet{chkanwar}{olive!70!black}
\colorlet{chkgeorge}{brown}
\colorlet{chkneha}{purple}
\colorlet{chkhadi}{red!60!black}
\colorlet{chkhaonan}{red!60!black}
\colorlet{chkomar}{blue!60!black}
\colorlet{chkbilal}{green!50!black}
\colorlet{chkahmad}{orange!60!black}
\colorlet{chknurkhan}{blue}
\colorlet{chkhasan}{magenta!60!black}
\colorlet{chkmostafa}{teal!70!black}

\definecolor{cellHighlight}{HTML}{dbefff}

\newcommand\rurl[1]{%
    \href{https://#1}{\nolinkurl{#1}}%
}

\ExplSyntaxOn
\tl_new:N \l__jais_model_tl
\cs_new_protected:Npn \jais_model_inner:n #1
  {
    \tl_set:Nn \l__jais_model_tl { #1 }
    \tl_replace_all:Nnn \l__jais_model_tl { - } { - \penalty \exhyphenpenalty \hskip \c_zero_skip }
    \texttt { \l__jais_model_tl }
  }
\cs_new_eq:NN \jaismodelinner \jais_model_inner:n
\ExplSyntaxOff
\newcommand{\model}[1]{\jaismodelinner{#1}\xspace}
\AtBeginDocument{\ifdefined\pdfstringdefDisableCommands\pdfstringdefDisableCommands{\def\model#1{#1}}\fi}
\newcommand{\benchmark}[1]{#1\xspace}
\newcommand{\modelname}{\model{Jais~3}}
\newcommand{\araLingBmarks}{Arabic Linguistics\xspace} %
\newcommand{\tashekeelBMark}{\araLingBmarks{} -- Tashkeel\xspace}
\newcommand{\parsingBMark}{\araLingBmarks{} -- I'rab\xspace}
\title{\bfseries Jais 3: Introducing Arabic-centric Agentic AI}
\hypersetup{pdftitle={Jais 3: Introducing Arabic-centric Agentic AI}, pdfauthor={Jais Team}}

\author[*]{Jais Team}

\affiliation[]{Institute of Foundation Models, Mohamed bin Zayed University of Artificial Intelligence\\} 
\affiliation[]{Cerebras Systems, Inception42}

\makeatletter
\reportnumber{\the\year-\two@digits\month-\two@digits\day}
\makeatother

\abstract{
We present \modelname, a fleet of six open, Arabic-centric agentic language models ranging from 0.9B to 375B parameters. \modelname is built to complete tasks for Arabic speakers, not only to answer questions in plain text: it is trained to plan multi-step tasks, call tools and Model Context Protocol (MCP) servers, operate a terminal, search the web, and read, edit, and produce PDF, spreadsheet, and presentation files, while communicating with the user in Arabic. Instead of pretraining from scratch, we adapt each model of the fully open \model{K2 Horizon} fleet to Arabic through continual pretraining, followed by supervised fine-tuning (SFT), final token preference optimization (FTPO), and reinforcement learning with verifiable rewards (RLVR). We combine natural Arabic text with translated, transliterated, and synthetic Arabic, and replay the base model's own training data to keep its reasoning, agentic, and English abilities. \modelname covers Modern Standard Arabic and the major regional dialects, follows instructions with Arabic-specific constraints, and inherits the 512K-token context window of \model{K2 Horizon} (all models except the 0.9B). To our knowledge, its 375B model is the largest open Arabic-centric language model to date. For evaluation, we revisit how Arabic models are assessed, moving to fully generative evaluation on audited Arabic benchmarks. Across model sizes, \modelname matches or outperforms open-weight models of comparable size on Arabic and agentic benchmarks, with particularly strong agentic capabilities and Arabic instruction following, and far fewer hallucinations. We release the weights and the checkpoints from every training stage of all six models, together with samples of the training data, the data recipes and training configurations, the training logs, the audited benchmarks, and the evaluation code.
}

\begin{document}
\maketitle

\renewcommand{\thefootnote}{\fnsymbol{footnote}}
\footnotetext[1]{Correspondence to: \texttt{\{haonan.li,hector.liu,eric.xing\}@mbzuai.ac.ae}}
\renewcommand{\thefootnote}{\arabic{footnote}}

\begin{minipage}{\textwidth}
    \centering
    \includegraphics[width=0.3\linewidth]{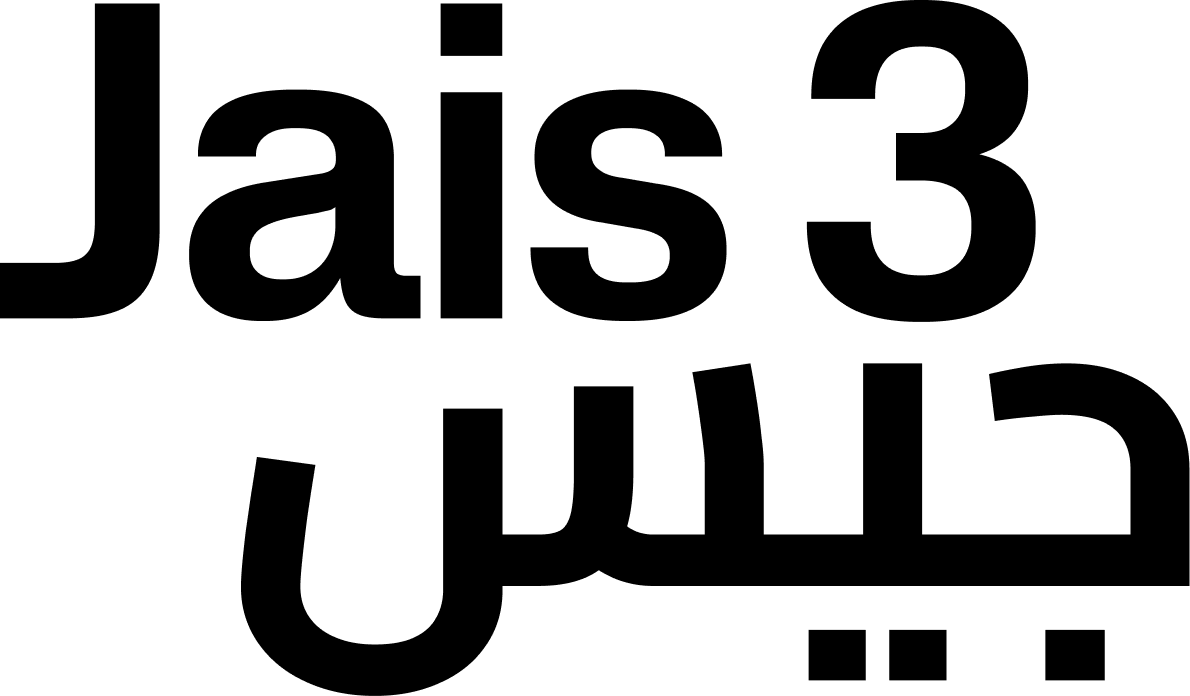}
\end{minipage}

\hfill
\begin{minipage}{\textwidth}
\centering
\begin{tabular}{ l l l }
\huggingface & \modelname & \rurl{hf.co/collections/IFM/jais-3}  \\
\logo & \modelname (Web) & \rurl{jaischat.ai} \\
\github & \modelname (Training Code) & \rurl{github.com/ifm-ai/xllm} \\
\github & \modelname (RL Code) & \rurl{github.com/ifm-ai/RL360} \\
\github & \modelname (Eval Code) & \rurl{github.com/ifm-ai/gen-eval-arabic} \\
\huggingface & \modelname (Audited Benchmarks) & \rurl{hf.co/collections/IFM/arabic-benchmark-auditing} \\
\wandb & \modelname (Training Logs) & \href{https://wandb.ai/mbzuai-llm/Jais-3}{W\&B Logs} \\
\end{tabular}
\end{minipage}

\newpage
\setcounter{tocdepth}{2}
\tableofcontents
\newpage

\section{Introduction}\label{sec:intro}

Arabic is used daily by more than 400 million people~\citep{un2026arabicday}, yet it remains under-served by open large language models (LLMs). Part of the reason lies in the language itself. Arabic is a family of varieties: Modern Standard Arabic (MSA) for writing and formal speech, and regional dialects (Gulf, Egyptian, Levantine, Maghrebi, and others) for everyday use, increasingly also written in Latin script as Arabizi. Natural Arabic text on the web is limited and noisy, and much of what exists is classical, religious, or news prose. Arabic-centric models such as \model{Jais}~\citep{sengupta2023jais,anwar2026jais2}, \model{AceGPT}~\citep{huang2023acegpt}, \model{ALLaM}~\citep{bari2024allam}, and \model{Fanar}~\citep{fanarteam2025,fanar2026fanar2} have narrowed the gap. These models, like the benchmarks used to compare them, are judged mainly on whether they can answer a question in Arabic. Meanwhile, open agentic models, which plan multi-step work, call tools, and handle files, are trained and benchmarked mostly in English, and how well they serve Arabic users is rarely a design goal. As a result, open models that can complete real tasks for Arabic speakers, rather than only converse with them, are largely missing.

\begin{figure}[htbp]
    \includegraphics[width=\textwidth]{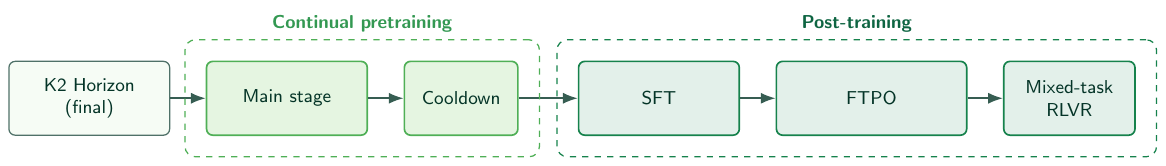}
    
    \caption{Training pipeline of \modelname. Each model starts from the
    final, post-trained \model{K2 Horizon} checkpoint of the same size and passes
    through Arabic and English continual pretraining (ending with a short
    cooldown), supervised fine-tuning (SFT), final token preference optimization (FTPO), and
    mixed-task reinforcement learning with verifiable rewards (RLVR).}
    \label{fig:training-pipeline}
\end{figure}

This report presents \modelname, a fleet of six Arabic-centric agentic models that range from 0.9B to 375B parameters: four dense models (0.9B, 3.7B, 7B, 32B) and two sparse Mixture-of-Experts models: a 36B model with Mixture-of-Value Attention (36B MoVA, 4B active parameters) and a 375B model (23B active). Instead of pretraining from scratch, we build \modelname on \model{K2 Horizon}~\citep{k2horizon2026}, a fully open model fleet pretrained on up to about 22T tokens with extensive reasoning and agentic data. Each \modelname model starts from the final, post-trained checkpoint of the \model{K2 Horizon} model of the same size and keeps its architecture and tokenizer unchanged (Section~\ref{subsec:bg-k2horizon}). We adapt it to Arabic through continual pretraining, which ends with a short cooldown, then post-train it with supervised fine-tuning (SFT), final token preference optimization (FTPO), and reinforcement learning with verifiable rewards (RLVR) (Figure~\ref{fig:training-pipeline}). Our goal is to add Arabic language, dialectal, and cultural competence while keeping the reasoning, long-context, agentic, and English abilities of the base fleet. To our knowledge, \modelname is the first language adaptation of \model{K2 Horizon}, and its 375B model is the largest open Arabic-centric language model released to date. 

\paragraph{Contributions.} The main contributions of this work are:
\begin{itemize}
    \item \textbf{An open Arabic-centric agentic model fleet.} We release six models from 0.9B to 375B parameters, covering on-device to large-scale server deployments. All models except the 0.9B inherit the 512K-token context window of \model{K2 Horizon}. On Arabic, \modelname leads or matches the best open models at every size, with the largest margins at small sizes, and the 375B model is competitive with \model{Claude Sonnet 5} (Section~\ref{subsec:eval-arabic}).

    \item \textbf{Agentic capabilities for Arabic users.} \modelname models can plan multi-step tasks, call tools and Model Context Protocol (MCP) servers, operate a terminal, and search the web while communicating with the user in Arabic. Connected to file tools, they can read, edit, and produce PDF, spreadsheet, and presentation files directly in Arabic (Sections~\ref{subsec:data-files} and~\ref{subsec:agentic-eval}). They also follow complex instructions with Arabic-specific constraints on orthography, diacritics, and morphology (Section~\ref{subsec:data-instruct}).

    \item \textbf{A documented recipe for adapting a strong base model to a new language.} We describe the full adaptation pipeline and the decisions that mattered (Sections~\ref{sec:training} and~\ref{sec:posttrain}). Natural Arabic text runs out long before the token budget, so translated, transliterated, and synthetic Arabic make up a large share of continual pretraining. A fixed share of replay from the base model's own mid-training data limits forgetting. Each data source needs its own filters, because rules tuned on English web text silently discard large amounts of valid Arabic.

    \item \textbf{A generative, audited Arabic evaluation suite.} Most Arabic leaderboards score multiple-choice log-probabilities, which do not fit reasoning models, and many widely used benchmarks contain wrong, ambiguous, or outdated ground truths. We move all evaluation to a generative protocol, and we audit the Arabic benchmarks in our suite with three LLM judges, removing the items that all three flag as flawed (Section~\ref{subsubsec:eval-ar-audit}). We also introduce new Arabic benchmarks: AraIFBench for instruction following, two linguistic benchmarks for grammatical analysis (\textit{i'rab}) and vocalization (\textit{tashkeel}), Arabic versions of AA-LCR and SimpleQA-Verified (Section~\ref{subsubsec:eval-new-bench}), and an Arabic file-processing benchmark for agents (Section~\ref{subsubsec:eval-file-bench}). We release the cleaned benchmarks, the discarded items, and the evaluation code.

    \item \textbf{An open release beyond the weights.} For each model we release checkpoints from every training stage: continual pretraining (including the cooldown), SFT, FTPO, and RLVR. We also release samples of the training data, the data recipes and training configurations, the training and RL code, the W\&B training logs, the audited benchmarks, and the evaluation code, so that others can study, reproduce, or branch from any stage of the pipeline (Section~\ref{sec:release}).
\end{itemize}

\paragraph{Outline of the report.}
Section~\ref{sec:background} reviews related work on Arabic and multilingual models, language adaptation, \model{K2 Horizon}, and Arabic evaluation. Section~\ref{sec:data_preprocess} describes the data processing shared across training stages. Sections~\ref{sec:training} and~\ref{sec:posttrain} present continual pretraining, SFT, FTPO, and RLVR, together with their data and training recipes; the FTPO and RLVR subsections also include stage-level results. Section~\ref{sec:eval} presents the Arabic, agentic, English, factuality, and safety evaluations, including our benchmark auditing and newly introduced benchmarks. Section~\ref{sec:release} discusses release details, intended use, and limitations, and Section~\ref{sec:conclusion} concludes the report. Additional details are provided in the appendices, including multi-turn data details (Appendix~\ref{app:multiturn}), and the complete evaluation setup (Appendix~\ref{app:jais3-evaluation-details}).

\section{Background and Related Work}\label{sec:background}
\subsection{Arabic-Centric and Multilingual LLMs}

Frontier open and closed models have demonstrated strong performance across many tasks, but multilingual evaluations still show gaps between English and other languages~\citep{xuan2025mmluprox,caciolai2026omnilingualgaia2}. Arabic, as a medium-resource language, remains below English for several evaluated models and tasks~\citep{xuan2025mmluprox}. 
These gaps are especially clear in Arabic dialects. Recent evaluations report lower performance in dialectal settings than in Modern Standard Arabic (MSA), including cultural reasoning, translation, and dialect-controlled generation~\citep{alkautsar2026arabicdialogues}.
At the same time, efforts to create Arabic-centric models have progressed to address these gaps. \model{Jais~1} introduced Arabic-centric bilingual models, with pretrained and instruction-tuned versions of up to 30B parameters. These models combine large-scale Arabic, mainly MSA, with English and code during pretraining, followed by Arabic and English instruction tuning~\citep{sengupta2023jais,inception2024jaisfamily}. The \model{Jais-adapted} series adds Arabic capability to English-focused \model{Llama~2} models through vocabulary expansion and continual bilingual pretraining, followed by bilingual instruction tuning~\citep{gosal2024bilingual,inception2024jaisfamily}. \model{Jais~2} extends this line with 8B and 70B models trained from scratch. The family places greater emphasis on dialects and Arabic cultural knowledge, including poetry, cuisine, and Islamic question answering~\citep{anwar2026jais2}.

Other Arabic-centric efforts include \model{Fanar 2.0}, which combines Arabic model adaptation with multimodal and tool-use support~\citep{fanarteam2025,fanar2026fanar2}; \model{AceGPT}, which focuses on Arabic adaptation and cultural alignment~\citep{huang2023acegpt}; and \model{ALLaM}, which combines vocabulary expansion, bilingual training, and preference alignment~\citep{bari2024allam}. \model{Atlas-Chat} targets Moroccan Darija written in Arabic script \citep{shang-etal-2025-atlas}, whereas \model{Nile-Chat} targets Egyptian Arabic in both Arabic and Latin scripts~\citep{shang2025nilechat}. \model{Falcon-H1-Arabic} uses a hybrid Mamba--Transformer architecture~\citep{boussaha2026falconh1arabic}, and \model{Command R7B Arabic} focuses on retrieval-augmented generation and contextual faithfulness~\citep{alnumay2025commandr7barabic}.

\model{BLOOM} was one of the early open efforts in multilingual language modeling, with pretraining across 46 natural languages, including Arabic~\citep{bigscience2022bloom}. \model{BLOOMZ} followed with multilingual instruction tuning to improve generalization across tasks and languages~\citep{muennighoff2023bloomz}. Later releases include \model{Llama~3}~\citep{grattafiori2024llama3}, \model{Gemma~3}, with reported support for over 140 languages~\citep{gemmateam2025gemma3}, and \model{Gemma~4}~\citep{gemmateam2026gemma4}. The \model{Qwen} family expanded support from 119 languages and dialects in \model{Qwen3} to 201 in \model{Qwen3.5}~\citep{qwenteam2025qwen3,qwenteam2026qwen35}, followed by \model{Qwen3.8}~\citep{qwenteam2026qwen38}. However, broad language coverage does not ensure equal performance. On MMLU-ProX, Arabic accuracy trails English by 14.7, 7.8, and 6.9 percentage points for \model{Llama-3.3-70B}, \model{Gemma-3-27B}, and \model{Qwen3-32B}, respectively~\citep{xuan2025mmluprox}. Evaluations of earlier models from these families also show gaps between Modern Standard Arabic and Arabic dialects~\citep{alkautsar2026arabicdialogues}.

Gains in Arabic knowledge and generation do not by themselves establish reliable Arabic tool calling or multi-step agent performance. A recent Arabic adaptation of BFCL reports lower tool-calling accuracy with Arabic prompts for \model{Llama~3.3}, \model{Qwen3}, and \model{gpt-oss}~\citep{kubrak2026arabictools}. Telco-GAIA extends evaluation to Arabic and English tool-using agents, and finds remaining weaknesses in tasks involving documents and images~\citep{khizbullin2026telcogaia}. However, these studies cover limited sets of models and tasks, and neither reports a systematic evaluation of tool use across Arabic dialects~\citep{kubrak2026arabictools,khizbullin2026telcogaia}. 
\modelname addresses both gaps: it is an Arabic-centric agentic model trained for multi-step tool use in Arabic and bilingual settings.

\subsection{Language Adaptation and Continual Pretraining}

Continual pretraining (CPT) adapts an existing language model by continuing its language-modeling objective on a new data distribution, typically next-token prediction on target-language text~\citep{gururangan2020dont}. It has become a practical way to bring open foundation models to new languages without pretraining from scratch. For Japanese, \model{Swallow} combines vocabulary expansion with large-scale CPT and reports continued improvements on Japanese benchmarks up to 100 billion training tokens, alongside decreases on English benchmarks~\citep{fujii2024swallow}. For Arabic, \citet{gosal2024bilingual} combine vocabulary expansion, an embedding-alignment stage, and full-model bilingual CPT, obtaining substantial Arabic gains while preserving or improving aggregate English performance. Both works expand the vocabulary because the inherited tokenizer segments the target language poorly, which requires initializing new embeddings~\citep{minixhofer2022wechsel,dobler2023focus}. \modelname does not need this step, because the \model{K2 Horizon} tokenizer already covers Arabic well (Section~\ref{subsec:bg-k2horizon}). We also train all parameters, rather than restricting updates to adapters or low-rank modules~\citep{pfeiffer2020madx,hu2022lora}.

A central challenge in CPT is balancing the acquisition of new capabilities with the retention of existing ones. Mixing target-language text with source-language or general-domain data is a common mitigation, and its benefits extend beyond source-language perplexity: \citet{elhady2025emergent} show that excluding English during language adaptation can impair in-context learning and downstream target-language performance despite similar target-language validation perplexity. Optimization interacts with these data choices: \citet{ibrahim2024continual} show that learning-rate rewarming and decay combined with replay are effective under distribution shift, and the CMR scaling law models the relationship between mixture proportions, training duration, and general-domain versus target-domain loss~\citep{gu2024cmr}. Following this line of work, \modelname replays a fixed share of \model{K2 Horizon}'s own mid-training data in every stage and ends continual pretraining with a short cooldown (Section~\ref{sec:training}).

Most prior adaptations start from a base model and add instruction tuning afterward. Continuing to train an already post-trained model carries a further risk: fine-tuning can erode a model's general and instruction-following abilities~\citep{luo2023forgetting}, and even benign fine-tuning can weaken safety alignment~\citep{qi2024finetuning}. A second common ingredient for under-resourced languages is machine-translated data, used both in pretraining~\citep{sengupta2023jais} and in instruction tuning~\citep{lai-etal-2023-okapi,ustun2024aya}. \modelname combines these settings. It applies full-model CPT to a post-trained, agentic \model{K2 Horizon} checkpoint with an unchanged tokenizer, mixes natural Arabic with translated, transliterated, and synthetic Arabic, and replays reasoning and agentic traces so that it keeps these capabilities, not only its English language modeling.

\subsection{\texorpdfstring{\model{K2 Horizon}}{K2 Horizon}}\label{subsec:bg-k2horizon}
\model{K2 Horizon}~\citep{k2horizon2026} is a fully open fleet of six language models developed at the Institute of Foundation Models (IFM) and released under the Apache~2.0 license. The models use a standard Transformer backbone with grouped-query attention, rotary position embeddings, and SwiGLU feed-forward layers, and add components of their own: grouped RMSNorm across the fleet and, in the 36B model, Mixture-of-Value Attention (MoVA) with a softplus-activated attention gate. The fleet has four dense models (0.9B, 3.7B, 7B, and 32B) and two sparse ones (Table~\ref{tab:k2horizon-configs}). The 375B model is a fine-grained Mixture-of-Experts (MoE) model with 192 routed experts and one shared expert per layer, of which eight routed experts are active per token. The 36B MoVA model combines MoE feed-forward layers with MoVA, which applies expert routing to the value projections in attention: each token selects 4 of 64 value experts, at no additional KV-cache cost. All models share a 250,624-token vocabulary except the 0.9B, which uses a 64,256-token vocabulary to save memory on device.

\begin{table}[H]
\centering
\small
\setlength{\tabcolsep}{5pt}
\begin{tabular}{lrrrrrr}
\toprule
 & \textbf{0.9B} & \textbf{3.7B} & \textbf{7B} & \textbf{32B} & \textbf{36B MoVA} & \textbf{375B} \\
\midrule
Architecture & Dense & Dense & Dense & Dense & MoE + MoVA & MoE \\
Active backbone params (B) & 0.88 & 3.78 & 6.95 & 32.21 & 4.67 & 23.59 \\
Total backbone params (B) & 0.88 & 3.78 & 6.95 & 32.21 & 36.16 & 376.09 \\
Layers & 28 & 36 & 36 & 64 & 48 & 61 \\
Hidden dimension & 1,536 & 2,560 & 4,096 & 5,120 & 2,560 & 6,144 \\
Query / KV heads & 32 / 8 & 32 / 8 & 32 / 8 & 64 / 8 & 32 / 8 & 48 / 8 \\
Vocabulary size & 64,256 & 250,624 & 250,624 & 250,624 & 250,624 & 250,624 \\
Pretraining tokens (T) & 5.0 & 21.9 & 21.9 & 21.9 & 21.9 & 15.1 \\
Context after mid-training & 128K & 512K & 512K & 512K & 512K & 512K \\
\bottomrule
\end{tabular}
\caption{\model{K2 Horizon} configurations~\citep{k2horizon2026}, the starting point of \modelname. Backbone parameter counts exclude the input embedding and output projection. The 375B model was pretrained in two phases (7.05T and 8.05T tokens).}
\label{tab:k2horizon-configs}
\end{table}

\paragraph{Training.} The models are pretrained with an 8K-token context on about 22T tokens~\citep{k2horizon2026} (15T for the 375B model and 5T for the 0.9B model). The pretraining mixture combines web, code, mathematics, science, multilingual, and domain-specific data with about 10T synthetic tokens, and nearly 17\% of it consists of problem-solving trajectories with explicit reasoning. Mid-training then extends the context in stages from 32K to 128K and 512K tokens (1.1T, 500B, 110B, and 200B tokens over four stages), while adding reasoning and agentic data; the 0.9B model follows a shorter two-stage schedule that ends at 128K. Post-training trains domain-specific RL specialists, merges them into one policy, and adds supervised fine-tuning, with a short final token preference optimization (FTPO) stage on the small models to remove degenerate repetition. The release includes final and intermediate checkpoints, training code, data or data recipes, and training logs.

\paragraph{Arabic in \model{K2 Horizon}.} Arabic is already well represented in \model{K2 Horizon}~\citep{k2horizon2026}. Multilingual text makes up about 7\% of the pretraining mixture across 17 languages, and Arabic is the second-largest of them. The tokenizer was also trained on a corpus that is roughly a quarter Arabic, to give Arabic text an efficient segmentation. Even so, Arabic is not a target language of \model{K2 Horizon}, and adapting it to Arabic pays off: \modelname raises the Arabic average by 9 to 35 points over the \model{K2 Horizon} model it starts from. 

\paragraph{From \model{K2 Horizon} to \modelname.} Each \modelname model starts from the final, post-trained checkpoint of the \model{K2 Horizon} model of the same size, with its architecture and tokenizer unchanged. \modelname therefore inherits the long-context, reasoning, and agentic abilities of its base, including the 512K-token context window of \model{K2 Horizon} (all models except the 0.9B), and the adaptation described in Sections~\ref{sec:training} and~\ref{sec:posttrain} aims to add Arabic competence without losing them. Because adaptation starts from a post-trained model rather than a pretrained one, continual pretraining must also preserve instruction-following and agentic behavior. To this end, we replay \model{K2 Horizon}'s own mid-training data, which includes reasoning traces and agentic trajectories, in continual pretraining, cooldown, and SFT (Sections~\ref{subsec:data-cpt} and~\ref{subsec:train-sft}). To our knowledge, \modelname is the first adaptation of \model{K2 Horizon} to a new language, and since every part of the starting point is public, the adaptation can be repeated or extended by others.

\subsection{Problems with Current Arabic Evaluation}
Most Arabic leaderboards, such as the Open Arabic LLM Leaderboard~\citep{elfilali2024oall}, score models by the log-probability of multiple-choice options, which does not reflect how modern reasoning models are used. Such models respond freely: they may state the answer first and justify it afterward, or reason at length before concluding, in plain text or Markdown. Instructions to answer in a fixed format are not reliably followed, so rigid answer extraction can mark correct responses as wrong. We therefore convert all benchmarks to generative evaluations and use an LLM judge to identify and grade the final answer, using the original benchmark gold label as a reference, regardless of its position or format.

Several widely used benchmarks also contain defective items that penalize correct answers. For instance, an ArabicMMLU~\citep{koto-etal-2024-arabicmmlu} item keys 15 as the number of NATO members, although the alliance now has 32; an Arabic GSM8K~\citep{arabic-gsm8k} reference solution miscounts the coins in its own question; and an ArabCulture~\citep{sadallah-etal-2025-commonsense} item offers three equally plausible options but keys only one as correct. We filter such items with a three-judge LLM audit, which removes 5.3\% of the items in the 15 benchmark sets we use (Section~\ref{subsubsec:eval-ar-audit}).

\section{Data Processing}\label{sec:data_preprocess}

\begin{takeaways}
  \item \textbf{Plan for the natural-text ceiling.} Natural Arabic web text and literature supply about 19\% of the continual-pretraining mixture; translated and synthetic Arabic supply about 38\% of the run, of which about 16 points is Arabizi (Latin script) and about 22 points Arabic script. The parallel dialect text is an LLM translation of English conversations, and the English side of those conversations adds a further 6 points.
  \item \textbf{Replay runs through every stage.} Data from the base model's own mid-training (web, reasoning, and agentic traces) makes up 25\% of continual pretraining and SFT and is part of every cooldown mixture, where it is drawn from the most recent mid-training phase of the replay data used in continual pretraining.
  \item \textbf{Every source needs its own filters and quality checks.} One shared rule set removed 12\% of a dialect corpus and 34\% of a diacritized one; per-source rules kept over 99\% of both.
\end{takeaways}

\modelname is trained by continuing the pretraining of \model{K2 Horizon} (Section~\ref{subsec:bg-k2horizon}), so the data work has a different goal from building a corpus from scratch: add Arabic knowledge, dialects, and assistant behavior without eroding what the base model already does well. This shapes every decision in this section. Arabic data is selected and weighted for what the base model lacks, and a share of the base model's own mid-training data (web text, reasoning traces, and agentic trajectories) is replayed in each stage. \modelname builds on the Arabic data work of \model{Jais~2}~\citep{anwar2026jais2}. For this release we name each source, report tokens under one tokenizer, and state what was filtered, deduplicated, and decontaminated, including the cases where a step was not run.

The inventory we draw from holds approximately 1.63\,T tokens, counted with the \modelname tokenizer, which is unchanged from \model{K2 Horizon}: 1.41\,T of pretraining-style text across 39 datasets and 224\,B of chat-formatted data across 35 datasets. These are sizes of what we prepared, not what the model saw; the training mixtures sample from the inventory with per-source weights, and we report inventory and exposure separately throughout. The Arabic side combines \texttt{ArabicTxT360} and style rewrites of it, Arabic translations of the base model's reasoning data, a synthetic
Arabizi corpus, parallel dialect text, and small diacritization and search-trajectory sets. Coverage therefore spans Modern Standard Arabic, regional dialects, Arabizi, and diacritized text.

The data description is organized by training stage (Table~\ref{tab:data-stages}). Continual pretraining samples 201.5\,B tokens, with each source weighted for about one pass, 75\% Arabic-side and 25\% replay. Cooldown combines small, high-quality sets (chat-formatted cooldown data, monthly Arabic Wikipedia, and knowledge-infusion and explanation-enriched sets; Section~\ref{subsec:data-cooldown} describes their decontamination and its limits) with mid-training replay; it samples 3.47\,B tokens (4.0\,B with the 0.9B's tokenizer) in a three-to-one ratio. SFT samples about 41\,B tokens: Arabic and English instruction data in high- and low-thinking modes, long-context Arabic question answering (about 17\% of the mixture), agentic trajectories, most of them Arabic translations (about 14\%), dialect and Arabizi conversations, culturally grounded sets (Islamic QA, inheritance law, poetry, UAE and country knowledge), safety data, and 25\% replay. Data for reinforcement learning with verifiable rewards (RLVR) is described last.

\begin{table}[t]
  \centering
  \footnotesize\rmfamily
  \renewcommand{\arraystretch}{1.15}
  \setlength{\tabcolsep}{2.2pt}
  \begin{tabular*}{\linewidth}{@{\extracolsep{\fill}}l>{\raggedright\arraybackslash}p{3.2cm}>{\raggedright\arraybackslash}p{6.4cm}l@{}}
    \toprule
    \textbf{Stage} & \textbf{Used in training} & \textbf{Mixture} & \textbf{Section} \\
    \midrule
    Continual pretraining & 201.5\,B tokens & 151.1\,B Arabic-side tokens from 32 sources, one pass (75\%); 50.4\,B mid-training replay (25\%) & \S\ref{subsec:data-cpt} \\
    Cooldown & 3.47\,B tokens & chat-format data, Arabic Wikipedia, knowledge and task QA sets (75\%); mid-training replay (25\%) & \S\ref{subsec:data-cooldown} \\
    SFT & 40.8\,B tokens per epoch, 22.2\,M examples & 138 mixture entries, one epoch; includes mid-training replay (25\%) & \S\ref{sec:posttrain-data} \\
    RLVR & 19{,}769 prompts & five verifiable tasks: instruction following, grammatical analysis, mathematics, diacritization, code (plus about 10.9\,K instruction-following prompts for the 375B) & \S\ref{subsec:rl-data} \\
    \bottomrule
  \end{tabular*}
  \caption{Data used in each training stage. Token counts are under the \modelname tokenizer; shares are fractions of tokens sampled in the stage.}
  \label{tab:data-stages}
\end{table}

Section~\ref{subsec:data-pipeline} describes the processing shared across sources: extraction to one document per record, quality filtering, MinHash deduplication within and across sources, screening against evaluation sets, and token accounting. The data used in each training stage is described with that stage: continual pretraining and cooldown data in Sections~\ref{subsec:data-cpt} and~\ref{subsec:data-cooldown}, and SFT, FTPO, and RLVR data in Section~\ref{sec:posttrain}. Section~\ref{subsec:data-limits} lists limitations.

\subsection{Processing Pipeline}\label{subsec:data-pipeline}

\begin{figure}[t]
  \centering
  \resizebox{\textwidth}{!}{\begin{tikzpicture}[
  font=\small,
  stage/.style={draw, rounded corners=2pt, fill=blue!6, minimum height=9mm, minimum width=26mm, align=center, font=\small\bfseries},
  knob/.style={text width=27mm, align=left, font=\scriptsize, anchor=north},
  arr/.style={-{Latex[length=2mm]}, thick},
  bar/.style={fill=blue!45, draw=none},
  loss/.style={fill=orange!45, draw=none},
  lbl/.style={font=\scriptsize, anchor=east},
  pct/.style={font=\scriptsize, anchor=west}
]
  \node[stage] (ex) at (0,0) {Extract};
  \node[stage, right=6mm of ex] (fi) {Filter};
  \node[stage, right=6mm of fi] (de) {Deduplicate};
  \node[stage, right=6mm of de] (dc) {Decontaminate};
  \node[stage, right=6mm of dc] (ct) {Count \& mix};
  \draw[arr] (ex) -- (fi); \draw[arr] (fi) -- (de); \draw[arr] (de) -- (dc); \draw[arr] (dc) -- (ct);

  \node[knob, below=1.5mm of ex] {one document per line; chat keeps roles, \texttt{think}, \texttt{tool\_calls}};
  \node[knob, below=1.5mm of fi] {23 rules, tuned thresholds; disabled per source after sampling drops};
  \node[knob, below=1.5mm of de] {MinHash 128 perms, banding per source; key per source (content, alignment pair, Q+A+tools)};
  \node[knob, below=1.5mm of dc] {n-gram overlap vs.\ eval sets; Arabic folding; Quran/hadith hits set aside};
  \node[knob, below=1.5mm of ct] {training tokenizer, full count; content, reasoning, tool fields separately};

\end{tikzpicture}}
  \caption{Processing pipeline. The five stages are shared by most sources; the settings under each stage are chosen per source.}
  \label{fig:data-pipeline}
\end{figure}
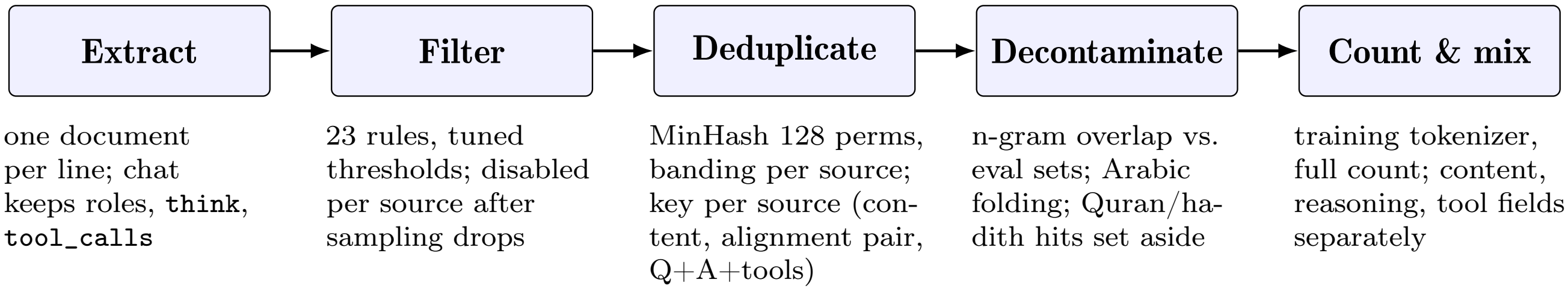

Most sources pass through the same five stages: extraction, quality filtering, deduplication, decontamination, and token accounting (Figure~\ref{fig:data-pipeline}), with configurations tailored to each source. These configurations account for differences in linguistic content and corpus structure: boilerplate removal heuristics can exclude valid diacritized classical prose, while deduplication criteria can collapse parallel dialect sentences whose high similarity is intentional. We therefore treat the pipeline as a fixed set of tools and a per-source profile, and we accept a profile only after inspecting samples of what it keeps and what it removes. The rest of this section describes the tools; the figures quoted are the cases that forced a profile change.

\paragraph{Extraction.} Each source is converted to JSONL with one self-contained document per line, \texttt{\{id, text, metadata\}}, so that later stages can hash, filter, and count documents independently of the vendor's packaging. Chat and agentic records keep their structure (\texttt{messages} with roles, and \texttt{think} and \texttt{tool\_calls} fields where present); for document-oriented tools we project them to text by joining the turn contents, and the projection never feeds \texttt{think}, \texttt{tool\_calls}, or tool schemas into that text. Generation and annotation metadata, including the evaluation questions used to seed synthetic data, stay outside \texttt{text}.

\paragraph{Quality filtering.} The Arabic filter applies 23 document-level rules: length (word count, mean word length), composition (alphabetic fraction, Arabic-character ratio, symbol-to-word ratio, stop-word count against 55 Arabic function words), repetition (duplicate lines, paragraphs, and 2- to 10-gram fractions), boilerplate (navigation phrases, booking listings, bullet and ellipsis lines, short-line tables, lorem ipsum), and content (an adult-URL list and a clitic-aware bad-word list). Measurements are taken on a normalized copy, with diacritics stripped and clitics such as \emph{wa-} separated, while the emitted text is unchanged.

Defaults were calibrated on high-quality Arabic text rather than imported from English pipelines: lowering the mean-word-length floor from 3.5 to 3.2 characters reduced rejection of classical and religious prose from 19.7\% to zero; lowering the alphabetic-fraction floor from 0.80 to 0.65 reduced rejection from 54.4\% to 9.6\% on the same subset. Inspection of rejected documents revealed three defects. The undiacritized stop-word list failed on fully vocalized text, rejecting 56\% of a hadith collection. The uppercase check considered only cased characters, so acronyms such as \emph{CNN} or \emph{OPEC} triggered rejection of Arabic paragraphs, removing 22.4\% of one news source. The bad-word list matched inside ordinary words after clitic expansion (for example \emph{zab\=un}, customer), and correcting it recovered 76.4\% of the dialect documents it had removed.

\begin{figure}[t]
  \centering
  \includegraphics[width=0.85\textwidth]{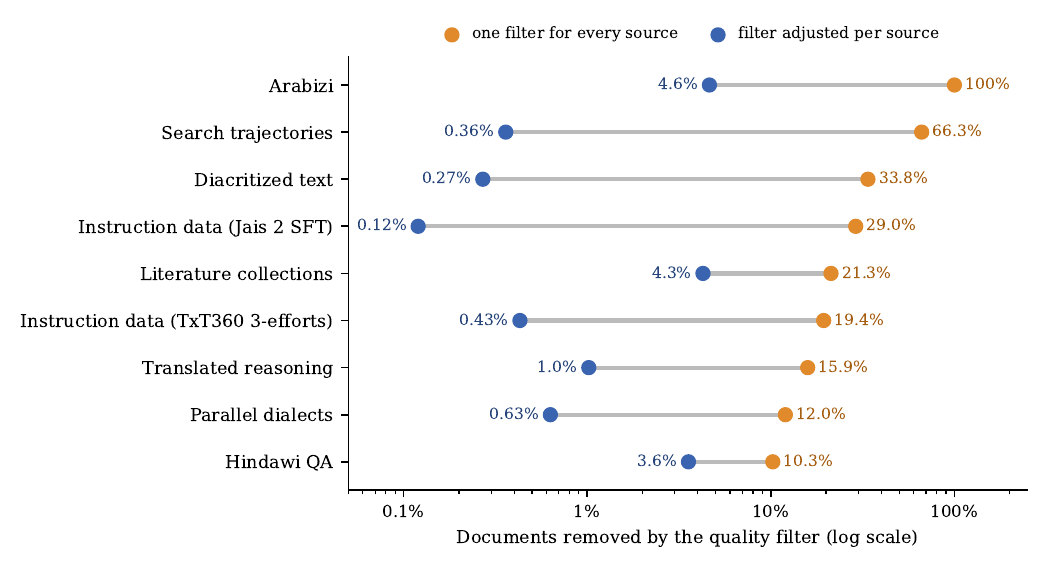}
  \caption{Documents removed by the quality filter when one filter is applied to every source (orange) and when the filter is adjusted per source (blue), for the nine sources whose audits measured both. Log scale. In each audit, samples of the documents the one-size filter removed were read; they were overwhelmingly valid text (classical passages, dictionaries, Latin-script Arabic, reasoning traces), which is why the adjusted profiles were accepted.}
  \label{fig:filter-profiles}
\end{figure}

Each source then gets its own profile (Figure~\ref{fig:filter-profiles}). The diacritization corpus lost 33.8\% of its documents to the sentence-count rule alone, because classical passages are written on one line without terminators, and its final profile removes 0.27\%. The parallel-dialect corpus lost 12\% at defaults, e.g., records on drug information or scientific abstracts whose repetition is structural, and retained 99.2\% after nine rules were disabled. For search-trajectory data, the Arabic-ratio rule would have rejected 66.3\% of documents because of their English reasoning traces; the adjusted profile instead adds a stuck tool-call loop check, which removes 305 trajectories (0.36\%).

\paragraph{Deduplication.} We run MinHash~\citep{broder1997resemblance} locality-sensitive hashing with 128 permutations over 13-character shingles on lowercased, punctuation-stripped text, banded per source: 9 bands of 13 rows for most sources (candidate pairs at Jaccard $\approx$0.85), and 16 bands of 8 rows ($\approx$0.71) for the style rewrites and Arabizi. Candidate pairs are grouped into connected components and one representative per component is kept. The comparison key is chosen per source. For search-trajectory data the default content key would have removed 2,177 records with a 99.95\% false-positive rate, because retrieved Wikipedia passages dominate the hash even when the question, reasoning, and answer differ; keying on question, reasoning, answer, and tool calls, and excluding retrieved text, flagged two records, one of them a false positive, so one record was removed. For parallel dialect text, each record holds an English conversation together with its dialect translation, and we hash the whole record rather than either side, so two translations of the same conversation are not collapsed into one. We do not deduplicate questions alone in synthetic QA sets, where one question with several distinct answers is intended.

\paragraph{Decontamination.} \label{par:decontamination}The training sets we prepared for continual pretraining, cooldown,
and SFT (Table~\ref{tab:decon}) are screened against our evaluation benchmarks
before use with AllenAI's \texttt{decon} n-gram overlap detector\footnote{\url{https://github.com/allenai/decon}} (Rust, with pyo3 bindings)
wrapped in a DataTrove~\citep{penedo2024datatrove} pipeline,\footnote{Token 8-grams under the \modelname tokenizer, testing an
n-gram every 3 tokens, with hits expanded to the full matching span and scored as an
IDF-weighted overlap of question, answer and passage; anything at or above 0.8 is reported,
the upstream default, unchanged. The stride of 3 is the largest with no blind spot, since a
region is certain to be found when $m \le \ell_{\min} - n + 1$, where $m$ is the stride, $n$
the n-gram length, and $\ell_{\min}$ the length in tokens of the shortest indexed evaluation
span; it leaves 0.00\% of indexed
evaluation entries at risk against 7.61\% at the tool's default of 10. Using the model's own
tokenizer matters here: \texttt{cl100k} needs 4.17 tokens per Arabic word against 1.86 for the \modelname tokenizer,
so an 8-token n-gram would span under two Arabic words.} after folding Arabic orthographic
variants (alef and ya forms, ta marbuta, tatweel, diacritics) on both the evaluation index and
the scanned text. Overlap is not the same as leakage: Quranic verses, hadith and shared
instruction boilerplate appear in thousands of documents and in the benchmarks alike, and the
single largest such item matched 11{,}485 separate documents. These are excluded on review
before anything is removed. Table~\ref{tab:decon} shows information after this exclusion. All replay was screened: the replay used in SFT is drawn from the same \model{K2 Horizon} mid-training data as the 50.4\,B continual-pretraining replay, so the continual-pretraining scan covers both.

\begin{table}[t]
\centering
\small
\begin{tabular}{llrr}
\toprule
Scan & Coverage & Removed & Rate \\
\midrule
Continual pretraining & 13 datasets, 149.4\,M lines  & 15{,}689 & 0.0105\% \\
Cooldown              & 11 datasets, 4.3\,M lines    & 1{,}952  & 0.045\%  \\
SFT                   & 33 datasets, 33.1\,M samples & 3{,}272  & 0.0099\% \\
\bottomrule
\end{tabular}
\caption{Contamination screening of the three training stages against the evaluation
benchmarks. Counts are lines, one JSONL record each, not tokens; line length varies roughly
130-fold across datasets, from 2{,}784 bytes to 362{,}075 per line, so rates are not comparable
across rows.}
\label{tab:decon}
\end{table}

In the continual-pretraining scan, 111{,}089 lines matched: 15{,}689 matched at least one discriminative item and were removed from the training copies, and the other 95{,}400 matched only non-discriminative text;
the rewrite touched 3{,}447 files, whose cleaned versions were verified byte for byte against
the originals. The overlap concentrates in poetry benchmarks with none in the diacritization or search-context sets.
The cooldown scan ran against 554 of the 567 delivered evaluation files, covering 210{,}269
questions indexed as 287{,}416 entries after deduplication; of 1{,}955 flagged lines, 3 were
kept after individual review as false positives and 1{,}952 were removed, and about half of
the Arabic Wikipedia hits were canonical religious text quoted by both the article and the
evaluation item. The SFT scan removed 3{,}272 samples across 33 datasets, with the poetry set
highest at 1.45\%, and every scanned file is written through byte-identical to its source
apart from the removed records. A re-scan of the cleaned cooldown release flags only the three
kept false positives and nothing else. Screening is lexical throughout and operates on whole
records, so a flagged record is dropped in full.

\paragraph{Token accounting.} After processing we count every document with the training tokenizer, without subsampling. We recompute all counts, as source-provided values differed from audited counts by up to a factor of eight; in one case, undecoded binary files were assigned token counts equal to their character lengths. For chat data we count assistant turns as the number of assistant messages and tokenize content, reasoning, and tool-call fields separately: a content-only count missed 28\% of the tokens of the translated agentic data and 631\,M reasoning tokens of one Arabic instruction set. These are inventory counts; exposure depends on mixture weights and repetition and is reported per stage.

\paragraph{Chat-format data.}
Conversational data, the SFT pool and the chat-formatted cooldown set, is stored as \texttt{messages[]} with no \texttt{text} field, so the document filter above drops all of it, and it would in any case reject turns that are correct for chat: a bare option letter as the answer, or English reasoning on an Arabic prompt. These sets go through a chat-aware pipeline instead, with five checks:
\begin{itemize}
  \item \emph{Format gate.} Every assistant turn must carry exactly one of \texttt{think} or \texttt{think\_faster}. A stratified sample is read before any detector is trusted.
  \item \emph{Deduplication.} Runs over six keys (byte-identical record, conversation after Unicode and diacritic normalization, first user prompt, final answer, the generator's seed hash, and MinHash near-duplicate prompts), with exact families reduced to one record per cluster and near-duplicate clusters judged as complete records.
  \item \emph{Content.} A role-aware bad-word pass flags assistant turns for a judge and a human read.
  \item \emph{Identity.} Scans assistant content and reasoning for another model's name next to a first-person cue.
  \item \emph{Consistency.} Every reasoning turn with a delivered answer is judged for agreement between the end of its trace and that answer.
\end{itemize}
Within the SFT and chat-format cooldown sets, flags are candidates, never removals: each drop is decided after reading the whole record, refusals and educational answers to sensitive prompts are kept, and a prompt that recurs with different answers is kept in full.

On the chat cooldown set this removed 1,233 of 1,844,798 conversations (0.067\%): 453 exact duplicates, 154 explicit completions out of 348 candidates read, 126 reasoning--answer contradictions, 68 records with the reasoning trace serialized into the answer, 430 tutor and multiple-choice dialogues that were wrong on reading, and 2 overlong failed traces; 69,225 system prompts naming the base model were rewritten to the \modelname identity. The reasoning--answer judge ran over 10.4\,M reasoning turns of the SFT pool and flagged 2.5\%. Most flags, 87\%, sat in the two Arabic and Arabizi instruction sets, which were republished without 3.55\% and 2.40\% of their rows (contradictions, substituted personal names, output-language and keyword-constraint failures, and 14,738 Arabizi answers with Arabic letters fused inside words); 19,193 flagged records were removed from 28 further sets before decontamination. The agentic sets, which hold about 14,000 of the consistency flags, were not filtered for consistency.

Identity leakage concentrates in synthetic data and inside reasoning traces. The mid-training chat replay was handled differently: records were removed on the name match alone, so all 789,812 of 13.7\,M records (5.75\%) that carried another model's name in an assistant turn were dropped, and a judge confirmed 27,120 of them as self-identification, 84\% inside the reasoning trace and 81\% naming ChatGPT. A regenerated TxT360~\citep{txt360} delivery carried the teacher's name in 1,554 traces and the generation prompt's own instructions in 23,440, and was rebuilt without 2,883 rows.

\subsection{Limitations and Exclusions}\label{subsec:data-limits}
Screening is lexical. The contamination scan matches token n-grams after orthographic folding, so translated or paraphrased benchmark items are not detected, and MinHash over character shingles measures lexical overlap only, so paraphrased duplicates survive deduplication.

\section{Continual Pretraining}\label{sec:training}

\begin{takeaways}
  \item \textbf{One main-stage dataset for every size.} Each model starts from its post-trained \model{K2 Horizon} checkpoint and trains on the same main-stage data (201.5\,B tokens; 233.6\,B with the 0.9B's tokenizer), 75\% Arabic-side data and 25\% replay, then a short cooldown; batch size, learning rate, and weight decay change with size, and the 0.9B also differs in tokenizer and position encoding.
  \item \textbf{Constant learning rate, then an annealing cooldown.} The main stage keeps the learning rate flat, so its last checkpoint is not annealed; the cooldown decays it to a tenth of the peak on curated data, much of it in chat format, that is closer in shape to SFT.
  \item \textbf{Benchmark-derived cooldown data uses training splits only.} One small set adds written explanations to the training splits of HellaSwag, ARC, and DROP (Table~\ref{tab:cooldown-sources}).
\end{takeaways}

Continual pretraining adapts each \model{K2 Horizon} model to Arabic in two stages: a main stage and a cooldown. Sections~\ref{subsec:data-cpt} and~\ref{subsec:data-cooldown} describe the data of each stage, and Section~\ref{subsec:train-cpt} the training recipe and the runs.

\subsection{Main-Stage Data}\label{subsec:data-cpt}

The main stage samples 201.5\,B tokens, with each Arabic source weighted for one pass over its retained tokens. On the Arabic side the mixture has 32 entries: the 19 subsets of the translated reasoning data, each weighted separately, and the 13 other sources of Table~\ref{tab:cpt-sources}, counting the targeted and instruction rows as their individual sets. A quarter is replay from \model{K2 Horizon}'s own mid-training data, and three quarters come from the sources in Table~\ref{tab:cpt-sources}. Natural Arabic web text and literature make up 19\% of the run. Translated and synthetic sources make up 44\%, of which about 6 points is the English side of the parallel dialect data. The remaining 12\% is small targeted sets, Arabic instruction data, English domain data, and further Arabic texts. Every source went through the pipeline of Section~\ref{subsec:data-pipeline} with its own filter settings; Table~\ref{tab:cpt-processing} records the settings and removal rates for each one.

\begin{table}[!htbp]
  \centering
  \small\rmfamily
  \renewcommand{\arraystretch}{1.15}
  \setlength{\tabcolsep}{2.2pt}
  \begin{tabular*}{\linewidth}{@{\extracolsep{\fill}}ll>{\raggedright\arraybackslash}p{5.2cm}cc@{}}
    \toprule
    \textbf{Group} & \textbf{Source} & \textbf{Origin} & \textbf{Tokens} & \textbf{Share} \\
    \midrule
    Natural Arabic & ArabicTxT360 & 83 web crawl snapshots, 2013--2024 & 28.4\,B & 14.1\% \\
     & Literature collections & Classical texts, poetry, news & 9.65\,B & 4.8\% \\
    \addlinespace[2pt]\midrule
    Translated and synthetic & Translated reasoning & 19 subsets of \model{K2 Horizon}'s pretraining data & 30.9\,B & 15.3\% \\
     & Parallel dialects & English conversations paired with dialect translations from \model{GLM-5} and \model{Jais~2}, judge-filtered for dialect fidelity, meaning and naturalness & 26.9\,B & 13.4\% \\
     & Arabizi & English TxT360 rendered as Arabizi in four dialect groups: Gulf, Egyptian, Levantine and Darija & 25.1\,B & 12.5\%$^{\ddagger}$ \\
     & Style rewrites & ArabicTxT360 passages rewritten in three styles & 4.43\,B & 2.2\% \\
     & QA augmentation & Web and Hindawi passages with generated questions and answers & 1.42\,B & 0.7\% \\
    \addlinespace[2pt]\midrule
    Targeted & Diacritization, search, exams & Sadeed/Tashkeela diacritization, and Arabic exam Q\&A & 0.58\,B & 0.3\% \\
    \addlinespace[2pt]
    Instruction data & Arabic chat data & TxT360 instruction data at three reasoning-effort levels (Arabic); \model{Jais~2} SFT with regenerated answers & 5.5\,B & 2.7\% \\
    \addlinespace[2pt]
    English domain & Investment and economy & Finance and economics documents & 6.33\,B & 3.1\% \\
    \addlinespace[2pt]
    Other & Further Arabic documents & Partner-supplied Arabic books, cross-document deduplicated and de-watermarked & 11.9\,B & 5.9\% \\
    \addlinespace[2pt]
    Replay & Mid-training replay & \model{K2 Horizon} mid-training data & 50.4\,B & 25.0\% \\
    \bottomrule
  \end{tabular*}
  \caption{Main-stage sources. Tokens are the inventory after filtering and deduplication, counted with the training tokenizer; share is the source's fraction of the 201.5\,B tokens sampled in the main stage. $^{\ddagger}$A further $\sim$3.5 points of Arabizi is inside the parallel-dialect data.}
  \label{tab:cpt-sources}
\end{table}

\begin{table}[!htbp]
  \centering
  \footnotesize\rmfamily
  \renewcommand{\arraystretch}{1.05}
  \setlength{\tabcolsep}{2.2pt}
  \begin{tabular*}{\linewidth}{@{\extracolsep{\fill}}>{\raggedright\arraybackslash}p{2.6cm}>{\raggedright\arraybackslash}p{5.8cm}>{\raggedright\arraybackslash}p{3.0cm}>{\raggedright\arraybackslash}p{2.1cm}c@{}}
    \toprule
    \textbf{Source} & \textbf{Filter settings (changes from Section~\ref{subsec:data-pipeline} defaults)} & \textbf{Removed by filter (default settings in brackets)} & \textbf{Removed by dedup} & \textbf{Final docs} \\
    \midrule
    ArabicTxT360 & nothing off; added page-template boilerplate rules; three passes over the Section~\ref{subsec:data-pipeline} defaults & 24.6\% of documents, 33.5\% of tokens & upstream, global & 73.8\,M \\
    \addlinespace[2pt]
    Literature collections & word cap off; stop-word rule matched on undiacritized text & 4.3\% (21.3\%) & 23.5\% & 9.03\,M \\
    \addlinespace[2pt]
    Further Arabic documents & no changes & 0.0\% (0.9\% of characters) & 20.6\% & 199\,K \\
    \addlinespace[2pt]
    Investment and economy & Arabic-ratio rule off; added binary-content rule; English thresholds & 56.9\% of web pages; 3.1\% of words (64.7\%) & 53.6\% & 4.37\,M \\
    \addlinespace[2pt]
    Translated reasoning & line cleaning off outside web text; six per-subset profiles & 1.0\% (15.9\%) & 0.04\% & 17.6\,M \\
    \addlinespace[2pt]
    Parallel dialects & nine repetition and content rules off; bad-word list corrected & 0.6\% (12\%) & 0.2\% & 8.73\,M \\
    \addlinespace[2pt]
    Arabizi & Arabic-ratio, stop-word and alphabetic-fraction rules off & 4.6\% (100\%) & 0.08\% & 11.8\,M \\
    \addlinespace[2pt]
    Style rewrites & nothing off; added rules for model refusals, data dumps and script splices & 12.6\% & 0.005\% & 12.4\,M \\
    \addlinespace[2pt]
    Web QA & bullet rule off; checks scoped to the passage only & 10.7\% & 0.2\% & 2.97\,M \\
    \addlinespace[2pt]
    Hindawi QA & bullet and toxic-word rules off; one-sentence minimum & 3.6\% (10.3\%) & 5.3\% & 462\,K \\
    \addlinespace[2pt]
    Exams & line cleaning, ellipsis, bullet and short-table rules off & 38.7\% & 4.0\% & 24\,K \\
    \addlinespace[2pt]
    Diacritization & sentence-count and eight other rules off & 0.27\% (33.8\%) & 0.08\% & 925\,K \\
    \addlinespace[2pt]
    Search trajectories & default rules relaxed for English reasoning traces; added a stuck tool-call loop check & 0.36\% (66.3\%) & 1 record & 85\,K \\
    \addlinespace[2pt]
    Instruction (TxT360) & checks off for reasoning text; thresholds per category and effort level & 0.4\% (19.4\%) & 0.7\% & 2.82\,M \\
    \addlinespace[2pt]
    Instruction (\model{Jais~2} SFT) & nothing off; added a judge-based semantic filter; thresholds per task family & 0.1\% (29.0\%), then 21\% by judge & 1.1\% & 1.81\,M \\
    \addlinespace[2pt]
    Mid-training replay & identity-leak removal & 5.8\% & not run & 12.9\,M \\
    \bottomrule
  \end{tabular*}
  \caption{Processing record per main-stage source. Percentages are of documents entering each stage unless noted. The bracketed figure is what the default settings removed on the same input; it is omitted where no default run was measured. One step outside the quality filter is not listed: 16 documents carrying personal data were dropped from Further Arabic documents.}
  \label{tab:cpt-processing}
\end{table}

\paragraph{Natural Arabic.} ArabicTxT360 is our Arabic extraction of 83 web crawl snapshots, refiltered with the Arabic rule set; older crawls were markedly noisier, with 25--44\% of documents removed from the 2014--2017 snapshots against about 14\% from 2023--2024. The literature collections include, among others, classical texts, poetry, news, and religious scholarship with the longest record being 16.3\,M tokens long.

\paragraph{Translated and synthetic Arabic.} This is the largest group, because natural Arabic runs out long before the token budget. We translated 19 subsets of \model{K2 Horizon}'s pretraining data (reasoning, planning, question answering, instruction following, and code reasoning). A pilot showed that long mathematical derivations were often truncated in translation, so we cut mathematics from 31\% to about 14\% of the tokens planned for translation. A human rating covered only 50 short samples, which averaged 4.95 out of 5.

The Arabizi corpus transliterates 30\,B tokens of English TxT360 into Gulf, Egyptian, Levantine, and Darija Arabizi (about 27.5\,B Arabizi tokens before our filtering and deduplication, 25.1\,B after). For the parallel dialect corpus, \model{GLM-5} translates English conversations into the target dialect and script, and \model{Jais~2} writes a second version. An LLM judge scores both versions from 1 to 4 on dialect accuracy, meaning, and naturalness, and an automated check flags unwanted switches into other languages. To build fluency in both scripts, a subset was also translated from the English source into Arabizi. The style rewrites turn web passages into formal, humorous, and simplified versions with \model{Qwen} to add register variety; the humorous style is about 88\% dialectal. Finally, QA augmentation appends generated question-answer pairs to web and Hindawi passages. We did not deduplicate these against the original web text, because the passage is meant to appear again in question-answer form; only 17.5\% of the web QA tokens are new.

\paragraph{Targeted sets.} Three small sets target specific skills. The diacritization set is instruction-formatted over fully vocalized classical and modern text, with loss on the answer only. The search-trajectory set records an agent that searches a bilingual Wikipedia index to answer factual questions the model had previously answered wrong. Only 17\% of its tokens are written by the model; the rest are retrieved passages. The third set holds questions and answers from exam papers of 13 Arab countries extracted with \model{Gemma~4}, and judged faithful to the source in 94.6--99.9\% of cases.

\paragraph{Instruction data and replay.} About 3\% of the run is Arabic instruction data in chat format, drawn from the same sets used later in SFT. Because each model starts from a post-trained checkpoint, this data keeps its chat behavior exercised during document-style training. The replay quarter is \model{K2 Horizon}'s own mid-training data, which holds English web text, reasoning traces, and agentic trajectories; it is there to limit forgetting of English, code, and reasoning while the model takes in new Arabic text. Before using it we removed every record in which the assistant might identify as another model: name matching flagged 790\,K records, a judge confirmed 27,120 of them as real leaks (84\% inside the reasoning trace), and we removed all 790\,K.

\paragraph{Mixture.} Every Arabic source is weighted for exactly one pass over its retained tokens. The QA-augmentation sources are converted to one chat per question--answer pair, so a passage with several questions appears once per question. The only data seen more than once is replay, where the data of the second phase of the final mid-training stage is repeated about twice. The 75/25 split between new data and replay is the same in continual pretraining and SFT; the cooldown mixture is described in Section~\ref{subsec:data-cooldown}. The English domain set came to us inside an Arabic crawl; after filtering it is 99\% English.

\subsection{Cooldown Data}\label{subsec:data-cooldown}

The cooldown mixes curated cooldown sets with replay from the most recent mid-training phase; Table~\ref{tab:cooldown-sources} gives the composition used for every model, three quarters cooldown sets and a quarter replay, 3.47\,B tokens (4.0\,B with the 0.9B's tokenizer). The cooldown sets are small and mostly curated or synthetic: chat-format instruction data, monthly Arabic Wikipedia, English QA, and multiple-choice questions with written explanations. Table~\ref{tab:cooldown-processing} records the processing of each set.

\begin{table}[!htbp]
  \centering
  \small\rmfamily
  \renewcommand{\arraystretch}{1.15}
  \setlength{\tabcolsep}{2.2pt}
  \begin{tabular*}{\linewidth}{@{\extracolsep{\fill}}l>{\raggedright\arraybackslash}p{5.4cm}lcc@{}}
    \toprule
    \textbf{Source} & \textbf{Content} & \textbf{Language} & \textbf{Tokens} & \textbf{Share} \\
    \midrule
    Chat-format cooldown & instruction conversations, half with reasoning traces & Ar, En & 1.64\,B & 47.4\% \\
    \addlinespace[3pt]
    Arabic Wikipedia & monthly snapshot, redirects and disambiguation pages removed & Ar & 0.61\,B & 17.5\% \\
    \addlinespace[3pt]
    Knowledge infusion & synthetic texts and questions on 47 subjects at four levels & En & 0.15\,B & 4.3\% \\
    \addlinespace[3pt]
    Synthetic multiple choice & new questions with explanations, modeled on Arabic school-exam and general-knowledge benchmarks and on MMLU & Ar, En & 0.17\,B & 5.0\% \\
    \addlinespace[3pt]
    Benchmark training items & training splits of HellaSwag, ARC, and DROP with written explanations & Ar, En & 0.03\,B & 0.8\% \\
    \addlinespace[3pt]
    Mid-training replay & most recent phase of the main-stage replay data & En & 0.87\,B & 25.0\% \\
    \bottomrule
  \end{tabular*}
  \caption{Cooldown sources. Tokens after processing, counted with the training tokenizer; share is the fraction of the 3.47\,B tokens sampled in the stage.}
  \label{tab:cooldown-sources}
\end{table}

\paragraph{Chat-format data.} The largest cooldown set is 1.8\,M instruction conversations, split almost evenly between Arabic and English, half of them with a reasoning trace. More than half are multiple-choice questions. It went through the chat-aware checks used for SFT data instead of the document filter; these checks removed 0.07\% of conversations: exact duplicates, overlong records, conversations with inconsistent reasoning and answers, and explicit content. Where the same prompt appeared with different answers we kept all of them.

\paragraph{Arabic Wikipedia.} Wikipedia is where the web-text tools failed most visibly. The stock filter would have removed 38\% of articles; reading samples of what it removed showed that only about 0.2\% of the removed tokens were actually junk. The rest were valid articles tripped by rules that assume web boilerplate: repeated category lines, Latin citation titles read as shouting, and a rule meant for JavaScript notices that matched the Arabic word for JavaScript. Deduplication failed the same way: the stock MinHash run removed 63,485 articles, because similar stub articles chained together (one component linked 12,706 articles about Moroccan villages), while a pairwise check found 6 true duplicates. We therefore applied neither the filter nor MinHash, and removed only redirects, disambiguation pages, empty shells, and those 6 duplicates, 5.8\% of records and 0.8\% of tokens.

\paragraph{Knowledge infusion.} This English set pairs synthetic text sections with generated questions and answers, on 47 subjects at four levels from high school to researcher. A three-pass audit with a large judge model estimated that 18.6\% of answers were wrong, rising to about 75\% for researcher-level abstract algebra and physics against about 10\% for high-school humanities. We repaired 22,577 answers, dropped 23,667, and estimate the remaining error rate at no less than 9\%.

\paragraph{Synthetic multiple choice.} These sets contain new multiple-choice questions with worked explanations, generated from Arabic materials and the training sets of general-knowledge benchmarks in variants written for a generic reader and for a student who chose a wrong answer. We found that generation makes questions easier and leaves a shortcut. The correct option is the longest one in 38--57\% of items, against about 25\% by chance; in the source Arabic benchmark it is the longest less often than chance. Generated questions are also shorter and more definitional than their sources. We removed byte-identical copies but kept rephrasings of the same question, which are intended as augmentation. We did not rebalance option lengths, which would require rewriting the options; the bias remains in the data. Another small cooldown set attaches written explanations to the training splits of HellaSwag, ARC, and DROP.

\begin{table}[!htbp]
  \centering
  \small\rmfamily
  \renewcommand{\arraystretch}{1.15}
  \setlength{\tabcolsep}{2.2pt}
  \begin{tabular*}{\linewidth}{@{\extracolsep{\fill}}>{\raggedright\arraybackslash}p{3.3cm}>{\raggedright\arraybackslash}p{7.0cm}>{\raggedright\arraybackslash}p{2.6cm}c@{}}
    \toprule
    \textbf{Source} & \textbf{Processing (changes from Section~\ref{subsec:data-pipeline} defaults)} & \textbf{Removed} & \textbf{Final records} \\
    \midrule
    Chat-format cooldown & chat-aware SFT checks instead of the document filter; reasoning-aware exact dedup & 0.07\% & 1.84\,M \\
    \addlinespace[2pt]
    Arabic Wikipedia & stock filter and MinHash not applied; structural removal and verified pairwise dedup & 5.8\% (stock filter: 38\%) & 1.27\,M \\
    \addlinespace[2pt]
    Knowledge infusion & judge-based answer verification, repair, and removal & 14.3\% of answers & 142\,K answers \\
    \addlinespace[2pt]
    Synthetic multiple choice & minimal length filter; structural validator; byte-identical dedup only & 0.1--13.6\% by set & 0.84\,M \\
    \addlinespace[2pt]
    Benchmark training items & Arabic-ratio rule only; byte-identical dedup & 5.1\% & 99\,K \\
    \bottomrule
  \end{tabular*}
  \caption{Processing record per cooldown source. Removed is the share of records removed by processing, or of answers where noted.}
  \label{tab:cooldown-processing}
\end{table}

\paragraph{Decontamination.} The released cooldown data was scanned against 554 evaluation files: 1,955 of 4.3\,M lines were flagged and 1,952 removed (0.045\%). About half of the Wikipedia matches were Quranic verses and other shared religious text rather than copied benchmark items. The scan could not cover HellaSwag, ARC, and DROP, whose reference files were not available to it, and it cannot see translated or paraphrased items.

\subsection{Training Recipe and Runs}\label{subsec:train-cpt}

Table~\ref{tab:cpt-settings} summarizes the settings of each model for the main stage and the cooldown. The main-stage data is identical across sizes.

\begin{table}[t]
\centering
\scriptsize\rmfamily
\renewcommand{\arraystretch}{1.15}
\setlength{\tabcolsep}{2.2pt}
\begin{tabular*}{\linewidth}{@{\extracolsep{\fill}}l*{10}{c}@{}}
\toprule
 & & \multicolumn{2}{c}{\textbf{Main stage}} & \multicolumn{2}{c}{\textbf{Cooldown}} & & & & & \\
\cmidrule(lr){3-4}\cmidrule(lr){5-6}
\textbf{Model} & \textbf{Vocab} & \textbf{Tokens} & \textbf{Updates}
  & \textbf{Tokens} & \textbf{Updates}
  & \textbf{Batch} & \textbf{Peak}
  & \textbf{Weight} & \textbf{RoPE}
  & \textbf{Max} \\
 & & & & & & \textbf{(tok / seqs)} & \textbf{LR}
  & \textbf{decay}
  & & \textbf{context} \\
\midrule
\model{Jais-3-0.9B}
  & 64{,}256$^{\dagger}$ & 233.6\,B$^{\dagger}$ & 11{,}141 & 4.0\,B$^{\dagger}$ & 189
  & 21.0\,M / 2{,}560 & $8.0\times10^{-5}$
  & 0.1 & $10^{6}$ + YaRN & 128K \\
\addlinespace[2pt]
\model{Jais-3-3.7B}
  & 250{,}624 & 201.5\,B & 10{,}248 & 3.47\,B & 177
  & 19.7\,M / 2{,}400 & $4.0\times10^{-5}$
  & 0.06 & $10^{7}$ & 512K \\
\addlinespace[2pt]
\model{Jais-3-7B}
  & 250{,}624 & 201.5\,B & 10{,}248 & 3.47\,B & 177
  & 19.7\,M / 2{,}400 & $3.0\times10^{-5}$
  & 0.06 & $10^{7}$ & 512K \\
\addlinespace[2pt]
\model{Jais-3-32B}
  & 250{,}624 & 201.5\,B & 9{,}606 & 3.47\,B & 166
  & 21.0\,M / 2{,}560 & $1.5\times10^{-5}$
  & 0.06 & $10^{7}$ & 512K \\
\addlinespace[2pt]
\model{Jais-3-MoVA-36B-A4B}
  & 250{,}624 & 201.5\,B & 9{,}606 & 3.47\,B & 166
  & 21.0\,M / 2{,}560 & $4.0\times10^{-5}$
  & 0.06 & $10^{7}$ & 512K \\
\addlinespace[2pt]
\model{Jais-3-375B-A23B}
  & 250{,}624 & 201.5\,B & 6{,}004 & 3.47\,B & 104
  & 33.6\,M / 4{,}096 & $1.4\times10^{-5}$
  & 0.073 & $10^{7}$ & 512K \\
\bottomrule
\end{tabular*}

\caption{Continual-pretraining settings
for each \modelname model. Tokens and updates are given
for the main stage and the cooldown. All models are trained
with AdamW on 8K-token sequences; the main stage uses a short
warmup and a constant learning rate, and the cooldown decays it
linearly to a tenth of the peak. Max context is the context window
inherited from \model{K2 Horizon}; YaRN uses a scaling factor of
16 from an original 8K context.
$^{\dagger}$Token counts depend on the tokenizer: the
0.9B's smaller vocabulary splits the same text
into more tokens.}
\label{tab:cpt-settings}
\end{table}

\paragraph{Recipe.}
Every model starts from its \model{K2 Horizon} checkpoint, which is the final post-trained model rather than the pretrained base (Section~\ref{subsec:bg-k2horizon}). We train all parameters with the next-token prediction loss and start from a fresh optimizer state rather than reusing the one saved with the checkpoint. The optimizer is AdamW~\citep{loshchilov2017decoupled} with $\beta_1{=}0.9$ and $\beta_2{=}0.95$, and gradients are clipped at a global norm of 1.0. Each update covers 2{,}400 to 4{,}096 sequences of 8{,}192 tokens
(19.7--33.6\,M tokens; Table~\ref{tab:cpt-settings}), so one
pass over the main-stage data takes 6{,}004 to 11{,}141
updates.

The learning rate rises linearly over the first 48 updates
(30 for the 375B), about 1B tokens, and then stays flat until the
cooldown, so the main stage ends at full learning rate and
the annealing is left to the cooldown and its own data.
The peak ranges from $1.4\times10^{-5}$ for the 375B
to $8.0\times10^{-5}$ for the 0.9B, and weight decay is
0.06, except 0.073 for the 375B and 0.1 for the 0.9B
(Table~\ref{tab:cpt-settings}).

The mixture weights are fixed for the whole stage
(Section~\ref{subsec:data-cpt}). The replay portion, 50.4\,B tokens, is drawn from \model{K2 Horizon}'s final two mid-training stages, which comprise high-quality, SFT-like data. We chose the 75/25 split after ablations
with 90/10, 75/25, and 50/50 mixtures of new data and
replay, in which 75/25 gave the best balance between
learning Arabic and retaining what \model{K2 Horizon} already
knew. The token budget and the batch size follow our
earlier continual-pretraining runs. We also conduct ablation studies on the learning rate and batch size, confirming that the hyperparameters inspired by \model{K2 Horizon} are optimal.

\paragraph{Cooldown.}
The cooldown anneals the model on curated data and
re-exposes it to the chat format before SFT begins. It
picks up from the last main-stage checkpoint; the learning
rate decays linearly from the main-stage peak to 10\%, with no warmup, and every other setting stays the
same. For every model, it is a single pass over
the cooldown mixture; Table~\ref{tab:cpt-settings} gives
the tokens and updates of each model.

Following the same 75/25 mixing ratio, the cooldown dataset (Section~\ref{subsec:data-cooldown}) combines replay data from the main stage with a mixture of Wikipedia and chat-formatted data across both thinking modes; incorporating these during the annealing phase both strengthens the model's factual knowledge and familiarizes it with the thinking formats prior to SFT.

\paragraph{Infrastructure.}
We trained the 0.9B, 32B, 36B MoVA, and 375B models, in
both the main stage and the cooldown, with our in-house
framework (xLLM) in bfloat16 with fully sharded data
parallelism. The 0.9B ran on NVIDIA H100 80GB GPUs
without tensor parallelism; the 32B, 36B MoVA, and 375B
ran on NVIDIA H200 141GB GPUs, with 2-way tensor
parallelism for the 36B MoVA and 8-way for the 375B.
The 3.7B and 7B were trained on Cerebras CS-2 systems
in cbfloat16.

To fit the batch at 8K context, we recompute the first and
third MLP projections and the attention value projection
in the backward pass instead of keeping their activations.
The training data is tokenized, packed into 8{,}192-token
sequences, and shuffled across the full dataset before the
run starts, so the data loader only reads pre-built
shards. Padding accounts for less than 0.02\% of tokens,
and about 5\% of sequences end in a document that was cut
at the sequence boundary. Checkpoints hold the model
weights and the optimizer state. They were written every
250 updates in the main stage and every 100 in the
cooldown, and we kept the last two.

\paragraph{Runs.}
The main stage took about 1.8\,K H100 GPU hours for the
0.9B (233.6\,B tokens), 10.9\,K H200 GPU hours for the
36B MoVA, and 76.9\,K H200 GPU hours for the 375B; on
Cerebras, it took 3.3\,K and 4.7\,K CS-2 system hours for
the 3.7B and 7B. The 32B main-stage run did not
log its runtime.

The 0.9B main stage ran its 11{,}141 updates in one
uninterrupted job, and the cooldown took another 30 GPU
hours. Training loss dropped quickly at first,
from 2.81 at the first update to an average of 1.74 around
update 500, as the model adjusted to the new mixture.
After that it fell slowly, reaching 1.61 by update 2{,}500,
1.57 by update 5{,}000, and 1.54 at the end of the stage.
There were no loss spikes: no update exceeded the average
of the preceding 50 by more than 10\%, and after the first
500 updates the gradient norm stayed between about 0.2 and 0.6.
Over the cooldown, the loss went from 1.44 to 1.22 as the
learning rate came down.

The larger models ended the main
stage at a training loss of 1.56 (3.7B), 1.49 (7B), 1.33
(32B), 1.35 (36B MoVA), and 1.21 (375B), and the
cooldown at 1.27, 1.21, 0.98, 0.99, and 0.88, each averaged
over the last 5\% of updates. For SFT we used the final cooldown checkpoint of
each model and did not select among intermediate
checkpoints.

\section{Post-Training}\label{sec:posttrain}

\label{subsec:posttrain-overview}
Post-training turns each base model into a chat assistant. Starting from the cooldown checkpoint, we apply three stages in sequence (Figure~\ref{fig:training-pipeline}). Each stage targets a different gap left by the one before it: supervised fine-tuning (SFT) establishes the desired behaviors, final token preference optimization (FTPO) corrects a failure mode that SFT introduces, and reinforcement learning with verifiable rewards (RLVR) sharpens performance on tasks whose outputs can be checked automatically. The following sections describe the training data and method for each stage, and report stage-level results for FTPO and RLVR.

The first stage, SFT (Sections~\ref{subsec:sft-data} and~\ref{subsec:train-sft}), teaches the model the \modelname{} chat format, its high- and low-thinking modes, and the Arabic, agentic, and safety behaviors we target. To keep the base models' existing abilities intact, a quarter of the SFT tokens are replayed from \model{K2 Horizon}'s mid-training data.

SFT alone, however, leaves the model prone to degenerate repetition loops. The second stage, FTPO (Section~\ref{subsec:train-ftpo}), is a short preference-optimization round that addresses this directly, reducing the average loop rate from 13.8\% to 1.2\%.

Finally, RLVR (Section~\ref{subsec:train-rl}) trains the model with programmatic rewards across five domains: instruction following, mathematics, code, Arabic grammatical analysis, and diacritization.

\subsection{SFT Data}\label{subsec:sft-data}\label{sec:posttrain-data}

SFT trains for one epoch on a mixture of \num{40.8}\,B tokens (\num{44.9}\,B with the 0.9B tokenizer) in \num{22.2}\,M examples from 138 mixture entries, a quarter of it replay of \model{K2 Horizon} mid-training data. The mixture is sampled from a pool of 33 datasets with \num{174.4}\,B tokens (Table~\ref{tab:posttrain-data-summary}). Most of the pool is the general chat and replay family (\num{153.0}\,B), of which the run uses only a small part. Most of the pool's tokens are in completions, and approximately half of the completion tokens are reasoning traces. Reasoning is present throughout the pool. Nearly every dataset is sampled in paired high- and low-thinking variants, 98.2\% of rows come from datasets containing at least some reasoning trace, and only six datasets contain none at all.

Table~\ref{tab:posttrain-data-summary} summarizes the pool by dataset family. Each of the 138 mixture entries is a dataset or one of its splits with its own sampling weight; Table~\ref{tab:sft-mixture} gives the mixture's composition by category, and Section~\ref{subsec:train-sft} describes the training run. The mixture subsets split further by reasoning mode and file shard and add the replay band. The subsections below describe the datasets we built for this release; families without a subsection are summarized only in Table~\ref{tab:posttrain-data-summary}.

\begin{table}[t]
  \centering
  \small\rmfamily
  \renewcommand{\arraystretch}{1.15}
  \setlength{\tabcolsep}{2.2pt}
  \begin{tabular*}{0.8\linewidth}{@{\extracolsep{\fill}}lccc@{}}
    \toprule
    \textbf{Family} & \textbf{Samples} & \textbf{Tokens} & \textbf{Trace} \\
    \midrule
    General chat and replay              & 19,263,061 & 153.03\,B & 54.3\% \\
    \addlinespace[2pt]
    Long context                         &   150,890 &   7.11\,B & 99.9\% \\
    \addlinespace[2pt]
    Agentic and tool use                 &   425,251 &   5.83\,B &  8.3\% \\
    \addlinespace[2pt]
    Dialect (Arabic script, Arabizi, ID) & 5,919,316 &   3.54\,B & 58.8\% \\
    \addlinespace[2pt]
    Translation and transliteration      &  1,033,966 &   0.72\,B & 20.6\% \\
    \addlinespace[2pt]
    Instruction following and multi-turn &  2,543,981 &   2.22\,B & 61.3\% \\
    \addlinespace[2pt]
    Regional and religious knowledge     &  2,858,783 &   1.40\,B & 44.4\% \\
    \addlinespace[2pt]
    Robustness and factuality            &   475,503 &   0.26\,B & 90.9\% \\
    \addlinespace[2pt]
    Safety and identity                  &  259,813 &   0.11\,B & 64.3\% \\
    \addlinespace[2pt]
    Arabic linguistics                   &  285,921 &   0.07\,B & 35.0\% \\
    \addlinespace[2pt]
    Synthetic ArabicMMLU                 &   102,988 &   0.07\,B & 78.8\% \\
    \addlinespace[2pt]
    Law, finance, medical                &     98,430 &   0.04\,B & 98.2\% \\
    \midrule
    Total                                & 33,417,903 & 174.40\,B & 53.3\% \\
    \bottomrule
  \end{tabular*}
  \caption{The SFT pool by family. Tokens are prompts plus completions;
  \emph{trace} is the share of completion tokens inside a reasoning trace.}
  \label{tab:posttrain-data-summary}
\end{table}

\graphicspath{{figures/}{./}}
\subsubsection{Multi-Turn Data}
\label{sec:multiturn}

\begin{figure*}[t]
  \centering
  \includegraphics[width=\textwidth]{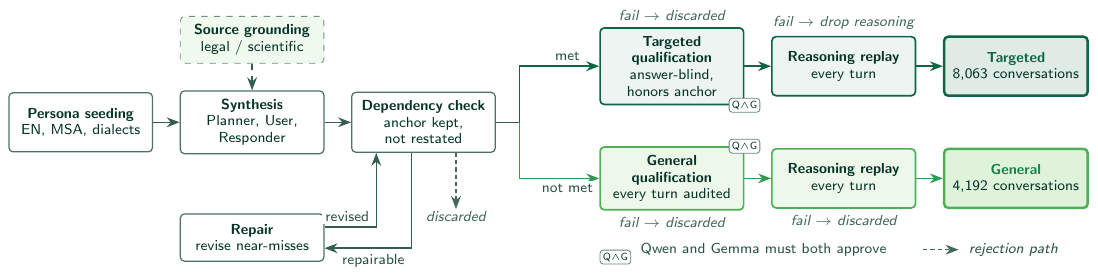}
  \caption{Multi-turn data pipeline. Conversations that realize a long-range
  dependency, directly or after repair, enter the targeted lane; those that do
  not, but are otherwise sound, enter the general lane. The lanes differ when a
  reasoning replay fails: a targeted conversation keeps its original answer
  without a reasoning trace, whereas a general conversation is discarded.}
  \label{fig:mt-pipeline}
\end{figure*}

Many failures of chat assistants are failures to carry state across a conversation: an early instruction is dropped, a fact the user shared is ignored, or a correct answer is abandoned when the user pushes back. We synthesize English and Arabic multi-turn conversations of two kinds. \emph{Targeted conversations} are built around these failures: each ends with a request whose correct answer depends on a specific earlier turn. \emph{General conversations} carry no such dependency but pass the same quality checks. Both come from one pipeline (Figure~\ref{fig:mt-pipeline}), which produced 8{,}063 targeted and 4{,}192 general conversations, 12{,}255 in total, each with 4--10 user turns. The SFT mixture (Table~\ref{tab:sft-mixture}) includes all of them once, across three entries totaling \num{79.8}\,M tokens (0.19\% of the mixture).

\paragraph{Synthesis.}
We follow the structure of the MultiChallenge pipeline~\citep{deshpande-etal-2025-multichallenge}. Each conversation starts from a persona sampled from a pool of personas%
, localized to English, MSA, or an Arabic dialect. A \emph{Planner} writes a blueprint specifying the scenario, the dependency type, and where the anchoring information is placed. A \emph{User} simulator and a \emph{Responder} then realize the blueprint as a dialogue of 8--20 messages; the Responder sees only the visible conversation, never the blueprint. \model{Qwen3.6-27B}~\citep{qwenteam2026qwen36} and \model{gemma-4-31B-it}~\citep{gemmateam2026gemma4} serve as both generators and judges. All conversations are synthesized from scratch, without benchmark prompts as seeds, and pass our decontamination pipeline (Section~\ref{par:decontamination}).

\paragraph{Targeted conversations.}
We adopt the four dependency types of \citet{deshpande-etal-2025-multichallenge}: instruction retention, user-information recall, versioned editing, and self-coherence. An \emph{anchor} turn introduces an instruction, fact, document version, or commitment, and the final user turn depends on it without restating it (Appendix~\ref{app:mt-anatomy}). A structural audit of the visible conversation checks that the anchor is present, that at least one exchange separates it from the final turn, that the final turn does not restate it, and that the anchor is unambiguous. Under strict qualification, the audit also confirms that changing the anchor would change the correct answer. Conversations that narrowly fail the audit are repaired and audited again rather than discarded. Legal and scientific conversations can additionally be grounded in pinned source passages.

\paragraph{General conversations.}
Many conversations that fail the dependency audit are still good conversations, so we re-qualify them for general quality instead of discarding them. Without the structural guarantee, quality must be established turn by turn: both judges audit every assistant turn for correctness, coherence, language quality, factual accuracy, safety, privacy, and consistent model identity, and check that no turn uses information introduced only later in the conversation.

\paragraph{Verification.}
Three mechanisms apply to both kinds of data. First, \emph{answer-blind regeneration}: the final answer is regenerated from the visible conversation alone, so the judges evaluate an independent attempt rather than approve an existing one. Second, \emph{unanimous cross-family judging}: a conversation is accepted only if both judges, one from each model family, approve it. This favors precision, since an error specific to one family is unlikely to be shared by the other, and no conversation is accepted on its generator's judgment alone. Third, \emph{reasoning replay}: every assistant turn is replayed with explicit reasoning, again answer-blind. The replayed answer must be of good quality, consistent with its reasoning, and in agreement with the original answer, because later turns build on it. If a replayed turn fails, a targeted conversation keeps the original answer without a reasoning trace, whereas a general conversation is rejected. Finally, the assembled conversation receives a whole-chat review; Appendix~\ref{app:multiturn} gives the full criteria and fallbacks.

\subsubsection{Country Knowledge Data}
\label{sec:data-country-knowledge}

We built a question-answer dataset on the 22 member states of the Arab League, covering politics, history, geography, the economy, religion, food, sports, famous people, landmarks, folklore, customs, and local expressions. Approximately 70\% of the examples are in the dialect of the country concerned and 30\% in MSA; Somalia, Djibouti, and Comoros are covered in MSA only. Every example includes a reasoning trace. Of the 1.38\,M examples, the final SFT run uses 1.37\,M (0.64\,B tokens, 1.6\% of the SFT tokens), and 220{,}511 of the set went through evidence-based verification.

\paragraph{Generation.} Questions come from a taxonomy of 29 categories and their subtopics; subtopics name areas of knowledge rather than entities, so the generator chooses what to ask about. For each country, register, and category--subtopic pair, \model{gemma-4-31B-it} writes first-person questions with an asker profile (e.g., resident, student, traveler, researcher), a target difficulty, and one of six forms: factual question, explanation, comparison, scenario, opinion, or list. Questions are deduplicated within each country and register by exact match and MinHash. The same model then answers each question with reasoning enabled, in the requested dialect or in MSA.

\paragraph{Quality filtering.} Rule-based filters remove empty, truncated, repetitive, and refused answers, answers less than 55\% Arabic, and answers with leaked reasoning. \model{gemma-4-31B-it} then scores the majority of examples on a five-point scale, keeping those that score at least 4 on every criterion. \model{gemma-4-31B-it} and \model{Qwen3.6-27B} independently fact-check every example, and examples flagged by both are rejected. \model{Qwen3.6-27B} flags dialectal and general answers more often; examples flagged only by \model{gemma-4-31B-it} more often contain a genuine error, usually a misattributed named entity.

\begin{figure}[t]
\centering
\includegraphics[width=\linewidth]{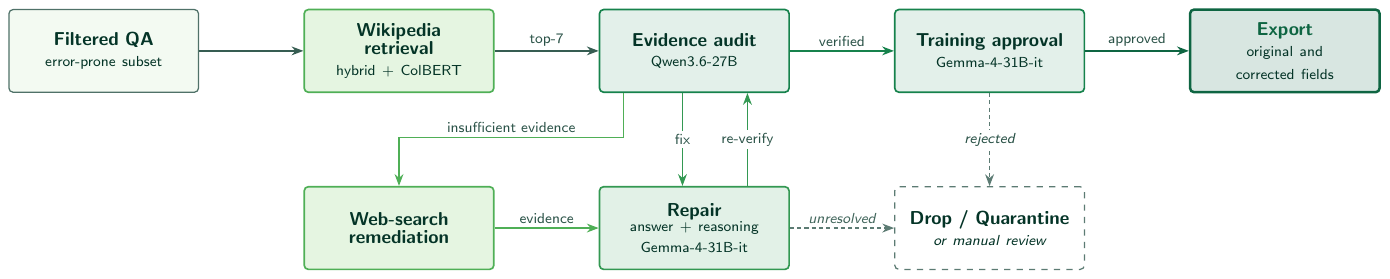}
\caption{Evidence-based verification of error-prone examples. Wikipedia
retrieval is the default evidence source and web search is a fallback. Every
rewritten answer is audited again before it can reach the training judge;
examples that cannot be verified are quarantined rather than accepted.}
\label{fig:ck-pipeline}
\end{figure}

\paragraph{Evidence-based verification.} Parametric fact-checking misses recent changes and facts that the generator and both judges get wrong. For the most error-prone category--subtopic pairs, we therefore verify answers against a fixed snapshot of Arabic Wikipedia (Figure~\ref{fig:ck-pipeline}), chosen over the web for reproducibility, source quality, and control over retrieval. For each question, hybrid retrieval returns 150 passages, which \model{LFM2.5-ColBERT-350M} reranks to seven. \model{Qwen3.6-27B} audits the answer against them; answers that are not \emph{correct} are rewritten by \model{gemma-4-31B-it} from the passages alone and audited again, with targeted web search as a fallback when Wikipedia lacks the evidence. Examples whose answers remain unsupported are quarantined.

\paragraph{Final approval.} Before training, \model{gemma-4-31B-it} checks that each rewrite fixes the audited issues and that every answer is natural Arabic in the requested register, consistent with its reasoning, and free of truncation, repetition, and leaked audit text. \emph{Unverifiable} examples are kept only for non-factual questions.

\providecolor{ttgen}{HTML}{DCE8F5}    %
\providecolor{ttclean}{HTML}{FBE5D6}  %
\providecolor{ttout}{HTML}{E2F0D9}    %
\providecolor{ttblue}{HTML}{2F5D8C}
\providecolor{ttlight}{HTML}{9DB9D8}
\providecolor{ttgreen}{HTML}{4E8F4A}
\providecolor{ttred}{HTML}{C0504D}

\providecommand{\ttlicense}[1]{\textcolor{red}{\textsf{[licence: #1]}}}

\subsubsection{Translation and Transliteration Data}
\label{sec:tt-data}

We built a synthetic corpus for two capabilities: \emph{translation} between English and Arabic (MSA, classical and literary Arabic, and seven regional dialects), and \emph{transliteration} between Arabic script and Arabizi (Franco), the Latin-script convention of informal written Arabic. \model{gemma-4-31B-it} generated all 812{,}160 examples; deterministic checks and a blinded LLM judge keep 741{,}262 (91.3\%) for SFT.

\paragraph{Sources.}
The translation data has two parts. The first reuses the source passages and translation directions of 465{,}692 samples from the \model{Jais~2} translation data~\citep{anwar2026jais2}, covering subtitles and dialogue, short sentence pairs, EU and legal text, and classical and literary Arabic, re-rendered into standardized prompts with newly generated answers. The second has 140{,}219 dialectal samples: Palestinian, North Levantine, and Yemeni samples are built from the public corpora of Table~\ref{tab:tt-dialect-sources}, and Egyptian, Moroccan, Tunisian, and Iraqi samples from internal data. For Iraqi, internal English seed sentences are first rendered into synthetic Iraqi Arabic and then translated back into English. The transliteration data (206{,}249 samples) covers both directions between Arabic and Arabizi across the same seven dialects, seeded from internal Egyptian, Moroccan, and MSA data and from the normalized Arabic passages of the translation data.

\begin{table}[t]
  \centering
  \small\rmfamily
  \renewcommand{\arraystretch}{1.15}
  \setlength{\tabcolsep}{2.2pt}
  \begin{tabularx}{0.85\linewidth}{@{}l >{\raggedright\arraybackslash}X l@{}}
    \toprule
    \textbf{Direction} & \textbf{Source corpus} & \textbf{License} \\
    \midrule
    Palestinian $\to$ English     & Curras~\citep{el-haff-etal-2022-curras, jarrar2017curras, jarrar-etal-2014-building}
                                  & CC BY 4.0~\citep{cc_by_40} \\
    \addlinespace[2pt]
    North Levantine $\to$ English & Nabra~\citep{nayouf-etal-2023-nabra} and
                                    Baladi~\citep{el-haff-etal-2022-curras}
                                  & CC BY 4.0~\citep{cc_by_40} \\
    \addlinespace[2pt]
    Yemeni $\to$ English          & Lisan-Yemeni~\citep{jarrar2022lisanyemeniiraqilibyan}
                                  & CC BY 4.0~\citep{cc_by_40} \\
    \bottomrule
  \end{tabularx}
  \caption{Public seed corpora for the dialectal translation data. Egyptian,
    Moroccan, Tunisian, and Iraqi samples are seeded from internal data.}
  \label{tab:tt-dialect-sources}
\end{table}

\paragraph{Generation.}
We sample from English and Arabic prompt templates at temperature 1.0, top-$p$ 0.95, and top-$k$ 64, with reasoning enabled or disabled (Figure~\ref{fig:tt-modes}). Translation prompts require full meaning preservation, a translation-only answer, and order-aligned output for multi-passage inputs. Transliteration prompts require changing only the script and preserving words, names, numbers, and punctuation, while allowing natural Arabizi spelling variation. The reasoning trace is stored separately from the final answer, with provenance and decoding metadata.

\paragraph{Quality filtering.}
Deterministic checks first drop mechanical failures, such as unresolved placeholders, injected foreign text, repetition, boilerplate, untranslated source copies, and model commentary. \model{gemma-4-31B-it} (temperature 0, schema-constrained JSON), blinded to the generator's reasoning trace, then scores the remaining rows against the criteria of Table~\ref{tab:tt-criteria}. A row is kept if every quality score is at least 4, every compliance and preservation check passes, and the declared and observed language varieties agree. This leaves 571{,}151 translation (94.3\%) and 170{,}111 transliteration (82.5\%) examples (Figure~\ref{fig:tt-outcome}).

\begin{table}[t]
  \centering
  \small\rmfamily
  \renewcommand{\arraystretch}{1.15}
  \setlength{\tabcolsep}{2.2pt}
  \begin{tabularx}{\linewidth}{@{}>{\raggedright\arraybackslash}p{2.1cm}
                                  >{\raggedright\arraybackslash}X
                                  >{\raggedright\arraybackslash}X@{}}
    \toprule
    \textbf{Axis} & \textbf{Translation} & \textbf{Transliteration} \\
    \midrule
    Quality (1--5)
      & Overall quality, semantic adequacy, dialect/register fidelity, fluency
      & Transliteration fidelity \\
    \addlinespace[2pt]
    Task compliance
      & Correct direction, language variety, and target script
      & Correct direction and script; transliterated, not translated \\
    \addlinespace[2pt]
    Preservation
      & Entities, numbers, negation, modality, tense, pronouns, relations
      & Meaning-bearing words, entities, numbers, names, punctuation, URLs,
        relations \\
    \addlinespace[2pt]
    Output defects
      & Omissions, additions, source copying, commentary, repetition,
        multi-passage misalignment
      & Omissions, additions, accidental translation, commentary,
        script errors \\
    \addlinespace[2pt]
    Source integrity
      & Actual variety, label agreement, usability, artifacts, placeholders,
        malformed fragments
      & Actual variety and agreement with the declared dialect \\
    \addlinespace[2pt]
    Content flags
      & \multicolumn{2}{>{\raggedright\arraybackslash}p{\dimexpr\linewidth-2.1cm-2\tabcolsep\relax}@{}}{%
        Sexual content, hate, slurs, sectarianism, racism, harassment,
        threats} \\
    \bottomrule
  \end{tabularx}
  \caption{Judge criteria. Quality dimensions are scored 1--5; all other
    fields are categorical or boolean.}
  \label{tab:tt-criteria}
\end{table}

\begin{figure}[t]
  \centering
  \pgfplotsset{
    ttbars/.style={
      xbar stacked, bar width=11pt,
      scale only axis, width=0.62\linewidth, height=2.2cm,
      xmin=0, xmax=720, xtick={0,200,400,600},
      enlarge y limits=0.45, clip=false,
      symbolic y coords={Transliteration, Translation}, ytick=data,
      axis lines*=left, axis line style={black!50},
      tick style={black!50},
      tick label style={font=\sffamily\footnotesize},
      label style={font=\sffamily\footnotesize},
      xlabel={Examples},
      xmajorgrids, grid style={black!12},
      legend style={font=\sffamily\footnotesize, draw=none, fill=none,
                    at={(0.5,1.03)}, anchor=south, legend columns=-1,
                    /tikz/every even column/.append style={column sep=8pt}},
      legend image code/.code={\draw[#1, draw=none] (0cm,-0.1cm)
                               rectangle (0.3cm,0.12cm);},
      scaled x ticks=false,
      xticklabel={\pgfmathprintnumber{\tick}k},
    },
    /tikz/ttin/.style={font=\sffamily\scriptsize, text=white, inner sep=0pt},
    ttouter/.style={nodes near coords, point meta=explicit symbolic,
      every node near coord/.append style={anchor=west,
        font=\sffamily\scriptsize, text=black!80, xshift=1pt}},
  }
  \begin{subfigure}[t]{0.49\linewidth}
    \centering
    \begin{tikzpicture}
      \begin{axis}[ttbars]
        \addplot[fill=ttblue, draw=none] coordinates
          {(306.109,Translation) (149.537,Transliteration)};
        \addplot[fill=ttlight, draw=none, ttouter] coordinates
          {(299.802,Translation) [49.5\%] (56.712,Transliteration) [27.5\%]};
        \legend{Reasoning on, Reasoning off}
        \coordinate (ma) at (axis cs:153,Translation);
        \coordinate (mb) at (axis cs:75,Transliteration);
      \end{axis}
      \node[ttin] at (ma) {50.5\%};
      \node[ttin] at (mb) {72.5\%};
    \end{tikzpicture}
    \caption{Generation mode.}
    \label{fig:tt-modes}
  \end{subfigure}\hfill
  \begin{subfigure}[t]{0.49\linewidth}
    \centering
    \begin{tikzpicture}
      \begin{axis}[ttbars]
        \addplot[fill=ttgreen, draw=none] coordinates
          {(571.151,Translation) (170.111,Transliteration)};
        \addplot[fill=ttred, draw=none, ttouter] coordinates
          {(34.760,Translation) [5.7\%] (36.138,Transliteration) [17.5\%]};
        \legend{Kept, Dropped or quarantined}
        \coordinate (oa) at (axis cs:285,Translation);
        \coordinate (ob) at (axis cs:85,Transliteration);
      \end{axis}
      \node[ttin] at (oa) {94.3\%};
      \node[ttin] at (ob) {82.5\%};
    \end{tikzpicture}
    \caption{Cleaning outcome.}
    \label{fig:tt-outcome}
  \end{subfigure}
  \caption{(a) Generations with reasoning enabled and disabled, per task.
    (b) Kept versus dropped or quarantined examples, as a share of each
    task's generated total.}
  \label{fig:tt-stats}
\end{figure}

\subsubsection{Arabic Instruction Following}\label{subsec:data-instruct}

Open instruction-following data is mostly English, and its constraints rarely test what makes Arabic text correct: \emph{hamza} and \emph{t\=a' marb\=u\d{t}a}, diacritics, Arabic-Indic digits and punctuation, or the plural, dual, and relational forms of words. We therefore built a synthetic set of 43{,}780 Arabic prompts, each pairing a natural request with one or two program-checkable constraints; the final SFT run uses the reasoning version of 43{,}443 of them (0.12\,B tokens).

\paragraph{Prompts.} \model{Qwen3.8-27B}~\citep{qwenteam2026qwen38} writes chat-style requests in MSA or in Gulf, Egyptian, or Levantine Arabic, across genres, domains, and response lengths. We append constraints as numbered Arabic instructions, drawn from 70 constraint types in seven categories (Table~\ref{tab:if-constraints}). Approximately 7\% of prompts carry two constraints, either a light second one, such as a closing phrase, or a demanding pair of exact counts; constraints that fix the output, such as copy-and-transform or morphology, are never combined.

\begin{table}[t]
  \centering
  \small\rmfamily
  \renewcommand{\arraystretch}{1.15}
  \setlength{\tabcolsep}{2.2pt}
  \begin{tabularx}{\linewidth}{@{}l>{\raggedright\arraybackslash}X@{}}
    \toprule
    \textbf{Category} & \textbf{Example constraints} \\
    \midrule
    Orthography          & exact number of \emph{hamza}, \emph{alif maq\d{s}\=ura}, \emph{t\=a' marb\=u\d{t}a}, or definite articles \\
    \addlinespace[2pt]
    Diacritics and script & exact count or absence of \emph{sukun}, \emph{shadda}, and \emph{tanw\=\i n}; \emph{kashida}; Arabic-Indic digits; Arabic punctuation \\
    \addlinespace[2pt]
    Sequence             & word counts that increase from sentence to sentence; equal character counts across sentences \\
    \addlinespace[2pt]
    Copy and transform   & repeat, reverse, or substitute parts of the request; return only the requested output \\
    \addlinespace[2pt]
    Format and position  & required first or last word, nested brackets or custom separators, paragraph counts; counts of words, numbers, or items from a closed list \\
    \addlinespace[2pt]
    Morphology           & produce the broken plural, dual, feminine, relational adjective (\emph{nisba}), participle, or verbal noun of a given word \\
    \addlinespace[2pt]
    Combined             & two exact-count constraints in one prompt, such as \emph{kashida} and diacritic counts \\
    \bottomrule
  \end{tabularx}
  \caption{Constraint categories in the Arabic instruction-following set. Each constraint type has a deterministic validator.}
  \label{tab:if-constraints}
\end{table}

\paragraph{Filtering and responses.} Each constraint type has a deterministic validator, using a curated lexicon of gold forms for morphology. Filters remove prompts with formatting artifacts, prompts less than 55\% Arabic, near-duplicates, and prompts that overlap with evaluation data (a shared 8-word sequence, token Jaccard similarity $\geq 0.5$, or a normalized exact match). Because validators do not check that the request was answered, \model{Qwen3.8-27B} also judges each prompt with a constraint-satisfying response, rejecting unnatural prompts and responses that ignore the request, the most common failure. The released response is sampled from the same model with reasoning enabled and resampled until it passes every validator in strict mode without formatting artifacts. Each prompt is released single-turn, with and without the reasoning trace.

\subsubsection{Long-Context Data}\label{subsec:data-longcontext}

Open post-training data rarely asks a model to reason over long Arabic text. We therefore built multiple-choice questions over long, natively Arabic documents: classical texts, legislation and government publications, financial and academic documents, news, and fiction. Contexts reach 128K tokens, with a median of approximately 43K, and the multiple-choice format makes every answer checkable. The final SFT run uses approximately 111\,K of these questions (5.5\,B tokens); with earlier builds and a smaller question-answering set over Arabic documents, long-context data totals 7.1\,B tokens, approximately 17\% of SFT (Table~\ref{tab:sft-mixture}).

\paragraph{Construction.} Each question is written from one long document or several related ones. Approximately a quarter combine evidence from across the context; the rest are written from one or two passages placed inside a long text, so the model must first locate them. \model{Qwen3.8-27B} and \model{gemma-4-31B-it} generate and check the questions. Every question is grounded in source passages, and its distractors are plausible, such as a sentence from the text that does not answer the question, an option with one false detail, or a one-word change. A small share of questions have ``the context does not determine the answer'' as the correct option. A question is kept only if an independent solve recovers its key, and options that can be ruled out by surface cues, such as length, are rejected.

\paragraph{Responses.} \model{Qwen3.8-27B} solves each final question with reasoning enabled. A response is kept only if its answer matches the key and its reasoning supports that answer with evidence from the context, rather than with test-taking heuristics such as preferring the longest option.

\subsubsection{Law, Finance, and Medical Data}\label{subsec:data-lfm}

Professional questions in law, finance, and medicine depend on local statutes, regulators, and clinical terminology, which open Arabic instruction data covers poorly. We built 98.4\,K Arabic questions in these domains (45\,M tokens), grounded in public Arabic sources (Table~\ref{tab:lfm-data}): 10.0\,K in law, 6.7\,K in finance, and 81.7\,K in medicine. The final SFT run uses 97.8\,K of them.

\begin{table}[t]
  \centering
  \small\rmfamily
  \renewcommand{\arraystretch}{1.15}
  \setlength{\tabcolsep}{6pt}
  \begin{tabularx}{\linewidth}{@{}l@{\hspace{24pt}}>{\raggedright\arraybackslash}X>{\raggedright\arraybackslash}X@{}}
    \toprule
    \textbf{Domain} & \textbf{Sources} & \textbf{Question formats} \\
    \midrule
    Law     & legislation, regulations, and court rulings from Egypt, Saudi Arabia, Qatar, and Kuwait; legal portals & case-based multiple choice \\
    \addlinespace[2pt]
    Finance & publications of central banks and market regulators in several Arab countries; further Arabic financial texts & numerical calculation, yes/no statements, multiple choice \\
    \addlinespace[2pt]
    Medical & Arabic medical documents; drug regulator guidance; physician-reviewed health portals & multiple choice, clinical vignettes, fill-in-the-blank, short answer \\
    \bottomrule
  \end{tabularx}
  \caption{Sources and question formats of the law, finance, and medical set. All sources are public Arabic documents.}
  \label{tab:lfm-data}
\end{table}

\paragraph{Construction.} We collected public Arabic documents from government, regulatory, and professional websites, following each site's crawling rules, and removed duplicates and short or mostly non-Arabic documents. From each excerpt, the generator extracts facts with a verbatim supporting quote, such as rules and deadlines in law, rates and formulas in finance, or diagnoses and treatments in medicine, and turns each fact into one self-contained MSA question in a format suited to its domain. \model{gemma-4-31B-it} and \model{Qwen3.8-27B} generate and check the questions. A question is kept only if an independent solve recovers its key; filters also remove malformed items, near-duplicates, and legal questions on sensitive political or sectarian topics.

\paragraph{Responses.} \model{Qwen3.8-27B} solves each final question with reasoning enabled, seeing only the question and, for multiple-choice items, its options. A response is kept only if it matches the key: the same option or yes/no label, a number within 0.5\% of the key for calculations, the key term or an accepted variant for fill-in-the-blank items, and the key facts of the reference answer for short-answer items.

\subsubsection{File Manipulation in Arabic}\label{subsec:data-files}
This set teaches the model to work on Arabic office files: to read an Arabic document, carry out a professional task based on it, and deliver the result as a native file with correct Arabic text and right-to-left layout. It contains 8.9\,K multi-turn agent trajectories (0.72\,B tokens), generated with \model{MiniMax-M3} in the OpenCode agent harness (Table~\ref{tab:file-manip-data}); the statistics below describe this release. The final SFT run uses an earlier build of 2.2\,K trajectories (0.16\,B tokens) made with the same pipeline.

\begin{table}[t]
  \centering
  \small\rmfamily
  \renewcommand{\arraystretch}{1.15}
  \setlength{\tabcolsep}{6pt}
  \begin{tabular}{@{}lc@{\hskip 2.5em}lc@{}}
    \toprule
    \textbf{Task type} & \textbf{Trajectories} & \textbf{Source file} & \textbf{Trajectories} \\
    \midrule
    Creation   & 3,881 & DOCX & 2,754 \\
    \addlinespace[2pt]
    Extraction & 1,228 & CSV  & 1,835 \\
    \addlinespace[2pt]
    Editing    & 3,791 & PDF  & 1,699 \\
    \addlinespace[2pt]
               &       & XLSX & 1,600 \\
    \addlinespace[2pt]
               &       & PPTX & 1,012 \\
    \midrule
    Total      & 8,900 &      & 8,900 \\
    \bottomrule
  \end{tabular}
  \caption{The Arabic file-manipulation set by task type and by format of the source file. The agent's outputs are Excel workbooks in 5,119 trajectories, PowerPoint decks in 1,992, Word documents in 1,553, and CSV files in 223; a trajectory can produce several files.}
  \label{tab:file-manip-data}
\end{table}

\paragraph{Generation.} Each example starts from one Arabic source file. \model{MiniMax-M3} reads every candidate file in full and keeps only those it grades excellent or good. For each kept file, it writes MSA tasks of three types: \emph{creation}, where the file is a reference for producing a new spreadsheet, document, or presentation; \emph{extraction}, where its content is turned into a derived product such as a memo, summary, ranking, or table; and \emph{editing}, where the file itself is revised or converted. Requests follow the style of GDPval~\citep{patwardhan2026gdpval}, with a professional role, a business situation, concrete constraints, and a named deliverable, and each task comes with an Arabic grading rubric written by the same model.

\paragraph{Trajectories.} A \model{MiniMax-M3} agent solves each task in OpenCode with shell, Python, file, and web-search tools, while a second instance plays the user. After the agent's first answer, the simulated user reviews the work against the rubric, which the agent never sees, and asks for corrections for up to two rounds; 7.9\,K trajectories contain three user turns. In 1.6\,K trajectories, a document skill (a reference guide for producing the requested file format) is installed in the agent's workspace. Trajectories have a median of 38 assistant turns and 40 tool calls and keep the agent's reasoning, tool calls, and tool results; 20.6\% of completion tokens are reasoning trace.

\subsubsection{Safety}\label{subsec:posttrain-safety}
Safety data is a small, targeted part of SFT: five sets totaling approximately 0.22\% of the tokens sampled in the final SFT run, roughly 92\,M tokens. Rather than add a large generic refusal corpus, we built most of it around failures we observed in earlier \model{Jais} models and around content that is sensitive in our region: religion, national leadership, and the model's own identity (Table~\ref{tab:safety-data}).

\begin{table}[t]
  \centering
  \small\rmfamily
  \renewcommand{\arraystretch}{1.15}
  \setlength{\tabcolsep}{2.2pt}
  \begin{tabular*}{\linewidth}{@{\extracolsep{\fill}}l>{\raggedright\arraybackslash}p{6.0cm}lcc@{}}
    \toprule
    \textbf{Set} & \textbf{Content} & \textbf{Languages} & \textbf{Rows} & \textbf{Share} \\
    \midrule
    General safety & public safety SFT data translated into Arabic, plus MSA harmful prompts written by native speakers & Ar, En & 47.5\,K & 0.07\% \\
    \addlinespace[2pt]
    \model{Jais~2} safety & general and regional safety sets inherited from \model{Jais~2} & Ar & 115.8\,K & 0.10\% \\
    \addlinespace[2pt]
    Religious and regional safety & variants of prompts the model had failed, with benign controls & Ar, En & 15.6\,K & 0.03\% \\
    \addlinespace[2pt]
    Self-identity & questions about the model's name, maker, and nature & 26 languages, Arabizi & 79.3\,K & 0.03\% \\
    \addlinespace[2pt]
    National leadership & factual questions about UAE leadership & Ar, En & 0.6\,K & $<$0.01\% \\
    \bottomrule
  \end{tabular*}
  \caption{Safety data in SFT. Rows are examples in the final SFT run; share is the fraction of tokens sampled in that run.}
  \label{tab:safety-data}
\end{table}
\paragraph{General safety.} We translated an openly released safety SFT set (Nemotron Safety v2) into Arabic with \model{gemma-4-31B-it}, keeping two thirds Arabic and one third English, and removed 3.4\% of rows that failed translation checks. Native Arabic speakers also wrote a set of harmful prompts in MSA, to cover requests that do not arise from translation; we generated safe responses for these and quarantined 27 of approximately 5,800 that reviewers could not agree on.
\paragraph{Religious and regional safety.} This set starts from 1,761 prompts on Islamic and UAE-related topics that an earlier \model{Jais} model answered unsafely or inaccurately. For each prompt we generated twelve variants: adversarial rephrasings that should still be refused, and benign questions on the same topic that should be answered, so that the model learns to distinguish the two rather than refuse the topic outright. A response entered training only when two independent judge models from different families both approved it. A first review with a single judge had approved only 38.9\% of responses and flagged over 10,000 factual errors and approximately 1,400 unsafe compliances, which is why we moved to agreement between two judges. This set draws no prompts from HarmBench, XSTest, or AraSafe, which we use for evaluation.
\paragraph{Self-identity.} \modelname{} is trained from a base model with its own name and maker, and part of its training data was written by other assistants. We therefore trained on questions about the model's identity in 26 languages, and removed identity leaks from the \model{K2 Horizon} mid-training replay set (Section~\ref{subsec:data-cpt}). An off-the-shelf identity filter missed 41\% of the leaks a judge confirmed there, because most of them sit in reasoning traces it does not read.

\paragraph{Limitations.} Our system-prompt ablation covers safety behavior only: we compared the deployed prompt against no system prompt on various red-team safety sets, and did not measure the prompt's effect on capability benchmarks. Safety scores are also sensitive to decoding: most prompts score the same way in every run at temperature 1.0, but a minority split across runs, so single-run numbers should be read with that variance in mind. Safety evaluation results, including over-refusal on XSTest, are reported in Section~\ref{sec:safety}.

\subsubsection{Robustness}\label{subsec:posttrain-robustness}
Users do not type like benchmark questions. Arabic input in particular carries misspellings that change meaning (a missing dot, \emph{t\=a' marb\=u\d{t}a} written as \emph{ha}), dialect mixed into MSA, and words run together. We want the model to do what a careful person does: answer the intended question when it can be recovered, and ask when it cannot. The robustness data has two parts, one for each behavior (Table~\ref{tab:robustness-data}): mutated questions, which are in the final SFT run, and ambiguous misspellings, which were built but not used in it. Both parts were generated with \model{gemma-4-31B-it}.

\begin{table}[t]
  \centering
  \small
  \renewcommand{\arraystretch}{1.15}
  \begin{tabularx}{\linewidth}{@{}>{\raggedright\arraybackslash}p{2.9cm}
                                  >{\raggedright\arraybackslash}X
                                  >{\raggedright\arraybackslash}p{3.0cm}
                                  r c@{}}
    \toprule
    \textbf{Part} & \textbf{Input} & \textbf{Target behavior} & \textbf{Examples} & \textbf{Final run} \\
    \midrule
    Mutated questions & Typos, dialect mixing, run-together words, false premises, or moral dilemmas; intent is recoverable & Answer the intended question & 5{,}131 & \checkmark \\
    \addlinespace
    Ambiguous misspellings & Misspellings with more than one plausible correction & Ask for clarification when a guess could mislead & 20{,}000 & -- \\
    \bottomrule
  \end{tabularx}
  \caption{The robustness SFT set. Only the mutated questions are used in the final SFT run (5{,}131 examples, 0.007\% of its tokens). The ambiguous misspellings consist of 10{,}000 prompts with two responses each.}
  \label{tab:robustness-data}
\end{table}

\paragraph{Mutated questions.}\label{subsec:robust-mutated}
This part targets questions whose wording is corrupted but whose intent can still be recovered. When native speakers probed an earlier \model{Jais} model with mutated questions, it recovered the intent almost every time; the exception was real-word substitution. In ``what are the physical benefits of \emph{al-\d{s}iy\=am} (fasting)?'', changing one letter gives \emph{al-\d{s}ayy\=ad} (the hunter), a valid word, and the model answered about the benefits of hunting.

\paragraph{Prompts (mutated questions).} Native speakers wrote 1,392 seed prompts: 893 with misspellings or false premises, covering real-word substitutions, \emph{t\=a' marb\=u\d{t}a}, letter dots and shapes, dialect mixed into MSA, run-together words, and English typos and reordered words, and 499 moral or pragmatic dilemmas. \model{gemma-4-31B-it} generated five variants of each seed, and we removed near-duplicates, keeping 6,000 of about 7,000 prompts.

\paragraph{Responses (mutated questions).} The same model answered each prompt once, half of the prompts with a reasoning trace and half without. Three judge models from different families checked whether each answer recovered the intended question; they removed 14.5\% of answers, leaving 5,131 examples (2,521 with a reasoning trace, 2,610 without), and a manual check of 100 accepted items found 97 correct.

\paragraph{Ambiguous misspellings.}\label{subsec:robust-ambiguous}
This part targets misspellings with more than one plausible correction, where silently correcting the word can change the request. In ``Can you compare the startup ecosystem growth in Riyadh versus the trends in Duba?'', \emph{Duba} may be a misspelling of Dubai or refer to the Saudi city of Duba. In an Arabic prompt asking for traffic data in the \emph{al-riy\=a\d{d}a} (`sport') area, an extra \emph{t\=a' marb\=u\d{t}a} may have turned Riyadh (\emph{al-riy\=a\d{d}}) into a common noun.
\paragraph{Prompts (ambiguous misspellings).} We used \model{gemma-4-31B-it} to generate 12,619 candidate prompts, half in English and half in Arabic. Each request sampled a topic, cycling through the 26 domains of a topic taxonomy (1,122 of its 1,125 leaf topics appear), an ambiguity category to emphasize, and two of six fixed examples in the target language. The categories are near-neighbor named entities, acronyms and tickers, technical terms, version numbers and codes, domain-dependent words, cross-language homographs, Latin--Arabic transliterations, and, for Arabic only, letter shapes and dots, including \emph{hamza} and \emph{t\=a' marb\=u\d{t}a}. The instructions excluded misspellings with an obvious correction, such as \emph{teh} for \emph{the}. Each record marks the misspelling, its intended correction and alternative readings, and three labels: whether an assistant should ask for clarification, how hard the intent is to recover, and how confident the correction is. We dropped records whose misspelling was absent from the prompt, identical to its correction, or a repeat of an earlier (misspelling, correction) pair, keeping 10,000 of 12,619, and manually inspected random subsamples of the result.

\paragraph{Responses (ambiguous misspellings).} The same model answered every prompt twice, with and without a reasoning trace, under a system prompt telling it to assume the most likely reading when that is safe and to ask when a guess could make the answer unhelpful. Answers cut off by the token limit were regenerated with a larger budget. This set was not used in the final SFT run, because the final model no longer showed this weakness.

\subsection{Supervised Fine-Tuning}\label{subsec:train-sft}
SFT starts from each model's final cooldown checkpoint. All six models follow one recipe: one epoch over a 40.8\,B-token mixture drawn from the pool of Section~\ref{subsec:sft-data}, packed into 128K-token sequences so that long-context data is trained at full length. The models differ in learning rate, weight decay, hardware, and parallel layout, and the 0.9B also in its tokenizer, which splits the same mixture into more tokens (Table~\ref{tab:sft-settings}). The 3.7B was trained on an earlier revision of the mixture, whose token count differs slightly.

\subsubsection{Training Setup}\label{subsec:sft-training}\label{subsec:sft-setup}

\paragraph{Chat format and loss.}
All data is rendered with the \model{K2 Horizon} chat template, so
the released models share one format. Each turn sits
between \texttt{<|ifm|im\_start|>} and
\texttt{<|ifm|im\_end|>} markers with a role name, and tool
definitions go in the system turn. The loss covers every
assistant turn in a conversation, not only the last one;
system, user, and tool-result turns are masked. Tool calls
are part of the assistant turn, so in all agentic data the
model trains on the tool calls it has to produce. Each
conversation starts with a BOS token and ends with a
supervised EOS token. The loss is averaged over all supervised tokens in an optimizer step, so each conversation contributes in proportion to its number of trained tokens.

The models have two modes, high-thinking and
low-thinking, and the mode is set by a tag at the start of
the assistant turn. In high-thinking mode the turn opens
with \texttt{<ifm|think>}, followed by the reasoning trace,
the closing tag and then the answer. In low-thinking mode
the turn opens with an empty \texttt{<ifm|think\_faster>}
block and goes straight to the answer, with no reasoning
trace. These are the high- and low-effort tags of the
\model{K2 Horizon} template; its medium-effort tag
\texttt{<ifm|think\_fast>} is not used. At inference time
the template writes the opening tag for the requested
mode, with high-thinking as the default. During training we
mask the opening tag, since the prompt supplies it, but
train on the closing tag so the model learns when to stop
reasoning and start answering. Of the 31.4\,M assistant
turns in the mixture, 16.4\,M are in high-thinking mode and
15.0\,M in low-thinking mode, so the two are close to balanced.
The 16.7\,K turns that had no reasoning field were put in
low-thinking mode. The high- and low-thinking results in
Section~\ref{sec:ablation-cascade} refer to these two
modes.

\paragraph{Sequence length and packing.}
Long-context QA makes up 17\% of the mixture, so all sizes
train with 131{,}072-token sequences. Conversations are
packed into these sequences with a best-fit packer that
never cuts a conversation in two; samples longer than the
limit are removed before packing. Conversations are
separated by EOS tokens and attention does not cross from
one conversation into the next. The 6{,}917 conversations (0.03\%) longer than 131{,}072 tokens were removed. Padding is negligible, 0.001--0.002\% of tokens in every run.

\paragraph{Hyperparameters.} We train all parameters from a fresh optimizer state with AdamW ($\beta_1{=}0.9$, $\beta_2{=}0.95$) and gradient clipping at 1.0. Peak learning rate and weight decay follow continual pretraining (Table~\ref{tab:cpt-settings}), except that the 375B uses a lower peak of $1.0\times10^{-5}$ instead of $1.4\times10^{-5}$. The learning rate warms up linearly over the first 2\% of optimizer steps and then decays linearly to 10\% of its peak. Each optimizer step covers 32 sequences of 131{,}072 tokens (approximately 4.2\,M tokens), and one epoch of the mixture takes 9{,}732 updates (40.8\,B tokens). The MoVA and 375B runs continued for 72 more updates (9{,}804 updates, 41.1\,B tokens), slightly more than one epoch. Table~\ref{tab:sft-settings} gives the per-model settings.
\begin{table}[t]
  \centering
  \small
  \setlength{\tabcolsep}{5pt}
  \begin{tabular}{@{}l rrrr rrr@{}}
    \toprule
    & \multicolumn{4}{c}{\textbf{Optimization}} & \multicolumn{3}{c}{\textbf{Compute}} \\
    \cmidrule(lr){2-5}\cmidrule(l){6-8}
    \textbf{Model} & \textbf{Peak LR} & \textbf{WD} & \textbf{Steps} & \textbf{Tokens}
                   & \textbf{GPUs} & \textbf{TP\,/\,CP} & \textbf{Time} \\
    & {\footnotesize($\times10^{-5}$)} & & & & & & \\
    \midrule
    \model{Jais-3-0.9B}         & 8.0 & 0.1   & 10{,}711 & 44.9\,B &     256 & 1\,/\,8 & 10.5\,h \\
    \model{Jais-3-3.7B}         & 4.0 & 0.06  &  9{,}754 & 40.9\,B &     256 & 1\,/\,8 & 21.5\,h \\
    \model{Jais-3-7B}           & 3.0 & 0.06  &  9{,}732 & 40.8\,B &     256 & 1\,/\,8 & 26.1\,h \\
    \model{Jais-3-32B}          & 1.5 & 0.06  &  9{,}732 & 40.8\,B &     512 & 2\,/\,8 & 53.3\,h \\
    \model{Jais-3-MoVA-36B-A4B} & 4.0 & 0.06  &  9{,}804 & 41.1\,B &     256 & 2\,/\,4 & 33.7\,h \\
    \model{Jais-3-375B-A23B}    & 1.0 & 0.073 &  9{,}804 & 41.1\,B & 1{,}024 & 8\,/\,4 & 45.2\,h \\
    \bottomrule
  \end{tabular}
  \caption{SFT settings per model. WD is weight decay; TP\,/\,CP are the tensor- and context-parallel degrees. Tokens are those processed (optimizer steps $\times$ 32 sequences $\times$ 131{,}072 tokens); the 0.9B processes more because its 64K-vocabulary tokenizer splits the same data into more tokens. Time is the wall-clock training time, excluding the intervals between restarts.}
  \label{tab:sft-settings}
\end{table}

\paragraph{Mixture.}
The SFT mixture has 138 mixture entries and 22.2\,M examples.

Table~\ref{tab:sft-mixture} groups them by type. A quarter
of the tokens are replay from the last \model{K2 Horizon}
mid-training phase, which we kept to protect the general
abilities the model had before SFT. The rest is split
across Arabic instruction data, long-context QA, agentic
and tool-use trajectories, English instruction data, and
translation and Arabizi. Most sources are seen once; a few
small, high-value sets are repeated up to about six times.

\begin{table}[t]
  \centering
  \small
  \begin{tabular}{@{}lrr@{}}
    \toprule
    \textbf{Category} & \textbf{Tokens} & \textbf{Share} \\
    \midrule
    Arabic instructions, culture, safety & 10.3\,B & 25.1\% \\
    Replay (\model{K2 Horizon} mid-training)     & 10.3\,B & 25.0\% \\
    Long-context QA                      &  7.1\,B & 17.3\% \\
    English chat, code, math, STEM       &  6.0\,B & 14.5\% \\
    Agentic and tool use                 &  5.8\,B & 14.2\% \\
    Translation and Arabizi              &  1.6\,B &  3.9\% \\
    \midrule
    Total                                & 41.1\,B & 100\% \\
    \bottomrule
  \end{tabular}
  \caption{SFT mixture by category, in tokens sampled over the 9{,}804 updates of the MoVA and 375B runs (41.1\,B, slightly more than one 40.8\,B epoch; 250K-vocabulary tokenizer). Shares are the same for every run. Agentic and tool use comprises 425\,K examples.}
  \label{tab:sft-mixture}
\end{table}
\paragraph{Infrastructure.} All six models were fine-tuned with our in-house training framework, xLLM, on NVIDIA H100 80GB GPUs (dense models) or H200 141GB GPUs (\model{Jais-3-MoVA-36B-A4B} and \model{Jais-3-375B-A23B}); Table~\ref{tab:sft-settings} gives the GPU counts and parallelism degrees per model. To fit 128K-token sequences in memory, every run combines fully sharded data parallelism with context parallelism, and the three largest models also use tensor parallelism. All runs compute in bfloat16; the two sparse models keep parameters, gradients, and gradient reductions in float32. The 32B run was restarted 23 times and the 0.9B run once; the 3.7B and 7B runs completed without a restart.
\paragraph{Training loss.}
Averaged over the first and last 5\% of optimizer steps, training
loss on the supervised tokens fell from 1.10 to 0.80 for the
0.9B, 0.80 to 0.71 for the 3.7B, 0.72 to 0.64 for the 7B,
0.64 to 0.58 for the 32B, 0.66 to 0.58 for the 36B MoVA,
and 0.61 to 0.53 for the 375B. For the 7B and 32B, most of
the drop came in the first half of training.

\subsection{Final Token Preference Optimization}\label{subsec:train-ftpo}

Reasoning models can fall into \emph{doom loops}, in which they repeat a span of text until the output budget is exhausted~\citep{pipis2025wait}. Such loops sustain themselves: each repetition of a span raises the probability of the next one~\citep{xu2022learning}. A looping response never reaches an answer, and in agentic use a single looping call can stall an entire episode. Greedy and low-temperature decoding make loops more frequent. Liquid AI attributes them to three interacting causes: over-trained tokens such as ``Wait'' that become disproportionately likely, a context in which a recently generated sequence becomes more likely to recur, and low-temperature sampling, which removes the randomness that would let the model exit the cycle~\citep{liquidai2026antidoom}.

We address doom loops with a short Final Token Preference Optimization (FTPO) stage between SFT and RLVR (Figure~\ref{fig:training-pipeline}). FTPO was introduced in Antislop to suppress overused phrases~\citep{paech2025antislop}, and the Antidoom recipe of \citet{liquidai2026antidoom} applies it to repetition. The training signal comes from the model's own looping generations. At the token where a loop begins, that token is the rejected choice, and alternatives that the model itself ranked as plausible at the same position are the chosen ones. A LoRA adapter~\citep{hu2022lora} learns these single-token preferences under a tether that keeps the remaining logits close to those of the starting model. All six models, from \model{Jais-3-0.9B} to \model{Jais-3-375B-A23B}, go through the same procedure with the same objective and optimizer settings. Only the number of mined pairs, the maximum training length, and the adapter placement in the mixture-of-experts models differ between models.

\subsubsection{Data}\label{subsec:ftpo-data} 
FTPO data is mined on-policy: every model generates from its own SFT checkpoint, so every rejected token is one that the model itself produces. A training example is a triple $(x, r, C)$, where the context $x$ is the prompt followed by the generation up to the start of a loop, $r$ is the rejected token, and $C$ is the set of chosen tokens.

\paragraph{Prompts.} Prompts come from Antidoom Mix~v1.0\footnote{\url{https://huggingface.co/datasets/LiquidAI/antidoom-mix-v1.0}}, a prompt-only mixture of public instruction-following, chat, mathematics, code, and question-answering datasets. We use a fixed snapshot of 474{,}229 prompts from 14 source datasets. The eight largest sources (MMLU, PubMedQA, AutoIF, UltraChat, Evol-CodeAlpaca, MetaMathQA, UltraInteract, and LMSYS Arena conversations) each account for 10--11\% of the prompts. Prompts longer than 2{,}000 tokens are discarded. Each prompt is followed by the instruction \texttt{Think through the problem step by step, then respond with your final answer as "Answer: <your answer>".} and rendered with the model's chat template in high-thinking mode.

\paragraph{Mining.} We sample at temperatures 0.1, 0.2, 0.3, and 0.4 with top-$k$ 50, min-$p$ 0.01, and top-$p$ 1.0, generate up to 8{,}192 new tokens, and record the top-20 log-probabilities at every position. Low temperatures are used because loops are most frequent there. The 8{,}192-token budget is sufficient because a pair is extracted where a loop starts, and later tokens contribute nothing. Mining stops once at least 18{,}000 pairs (models up to 7B), 24{,}000 pairs (\model{Jais-3-32B} and \model{Jais-3-MoVA-36B-A4B}), or 16{,}000 pairs (\model{Jais-3-375B-A23B}) have been collected.

\paragraph{Loop detection and pair construction.} Loops are identified with the Antidoom detector, without modification. The detector samples a 16-character fingerprint of the decoded text every 128 characters and searches for a nearby recurrence of it, which yields a candidate period of 1 to 1{,}024 characters. A loop is confirmed when a unit of that period repeats at least four times consecutively and covers at least 60 characters. The rejected token is the first non-whitespace token of the first repetition, the point at which the model begins to copy the unit again. The period is aligned in token space, and tokens consisting only of whitespace or punctuation are skipped. The chosen tokens are drawn from the top-20 log-probabilities at the same position, renormalized at the sampling temperature after the rejected token and its case variants are removed. We keep up to 20 tokens whose probability is at least 1\% of that of the most likely remaining token and discard whitespace-only tokens. Every chosen token is therefore a continuation that the model already considered plausible at that position.

\paragraph{Training subset.} For each model we sample 12{,}000 of the mined pairs. The sampling flattens the distribution of rejected tokens: a token whose count exceeds the median count $m$ is sampled with weight $(m/\text{count})^{0.3}$, so that no single loop opener, such as ``Wait'', dominates the data. Chosen tokens are deduplicated within each pair by normalized surface form. Chosen tokens whose count exceeds the 95th percentile of chosen-token counts (and at least 50) are removed from some pairs, which moves their count toward that reference with strength 0.5. Finally, we discard pairs whose context exceeds the maximum training length of 6{,}000 tokens (4{,}096 for \model{Jais-3-32B}), pairs whose rejected or chosen token is not a single token, and pairs with no chosen token left. Between 10{,}530 and 11{,}486 pairs per model remain.

\subsubsection{Training Setup}\label{subsec:ftpo-training} 

FTPO acts only on the next-token distribution at the end of each context. We train a LoRA adapter on the SFT checkpoint and merge it into the weights after training.

\paragraph{Objective.} Let $z$ denote the logits of the model being trained at the last position of $x$, and let $z^{\mathrm{ref}}$ denote the logits of the same model with the adapter disabled, that is, of the SFT checkpoint. The loss has two parts: a preference term that changes the decision between the rejected token and its alternatives, and a tether term that keeps everything else close to the SFT checkpoint. Expectations are taken over the pairs in a batch.

\emph{Preference term.} For a chosen token $c\in C$, let $\Delta_c = z_c - z_r$ be its logit margin over the rejected token $r$. The preference term pushes every chosen token above the rejected one:
\begin{equation}
  \mathcal{L}_{\mathrm{pref}} = \mathbb{E}\Big[\tfrac{1}{|C|}\sum_{c\in C} w_c\,\mathrm{softplus}(\varepsilon-\Delta_c)\Big],
  \qquad
  w_c=\mathrm{clip}\Big(\tfrac{\varepsilon-\Delta_c}{\varepsilon},\,0,\,1\Big).
  \label{eq:ftpo-pref}
\end{equation}
The weight $w_c$ decays to zero as the margin approaches $\varepsilon$, so pairs that are already resolved stop contributing.

\emph{Tether term.} Let $T=\{r\}\cup C$ be the set of target tokens and $\bar{T}$ the rest of the vocabulary. The tether penalizes any change in the non-target logits, but lets the target logits move by up to $\tau_t$ at no cost:
\begin{equation}
  \mathcal{L}_{\mathrm{tether}} =
  \lambda\,\mathbb{E}\Big[\tfrac{1}{|\bar{T}|}\sum_{v\in\bar{T}}\big(z_v-z^{\mathrm{ref}}_v\big)^2\Big]
  + \lambda_t\,\mathbb{E}\Big[\tfrac{1}{|T|}\sum_{v\in T}\max\big(|z_v-z^{\mathrm{ref}}_v|-\tau_t,\,0\big)^2\Big].
  \label{eq:ftpo-tether}
\end{equation}

\emph{Total loss.} The FTPO loss is the sum of the two terms:
\begin{equation}
  \mathcal{L} = \mathcal{L}_{\mathrm{pref}} + \mathcal{L}_{\mathrm{tether}}.
  \label{eq:ftpo}
\end{equation}
Together, they confine the update to the decision between the rejected token and its alternatives. We use $\varepsilon=2$, $\lambda=0.4$, $\lambda_t=0.05$, and $\tau_t=0.5$ for all models.

\paragraph{Hyperparameters.} The adapter has rank 256, $\alpha=128$, and no dropout. It covers the attention projections (query, key, value, and output), the MLP projections (gate, up, and down), and the LM head. For \model{Jais-3-MoVA-36B-A4B}, it covers the query, key, and output projections, the value projections of the three dense layers (MoVA layers use value experts instead), the shared expert, the MLPs of the dense layers, and the LM head; the routed experts and value experts remain frozen. For \model{Jais-3-375B-A23B}, it covers the query, key, value, and output projections of every layer, the shared expert, the MLPs of the three dense layers, and the LM head; the routed experts remain frozen. We optimize with AdamW at a learning rate of $1.5\times10^{-5}$ with a linear schedule, 10\% warmup, and weight decay 0.01, for at most one epoch with an effective batch size of 16 pairs.

\paragraph{Early stopping.} Training is controlled by \emph{chosen-win}, the fraction of a pair's chosen tokens whose log-probability exceeds that of the rejected token, averaged over pairs. We evaluate it every 5 optimizer steps, averaged over the steps since the previous evaluation, and stop at the first evaluation at which it reaches 0.35. \citet{liquidai2026antidoom} recommend starting from 0.4 and report strong loop reduction for values between 0.15 and 0.3. Stopping early also bounds how far the model moves from its SFT checkpoint: in every run, the tether terms of Equation~\ref{eq:ftpo-tether} grow steadily up to the stopping step.
\subsubsection{Results}\label{subsec:ftpo-results}

\paragraph{Evaluation set.} We measure looping on an internal evaluation set of 3{,}061 prompts. The prompts come from the same public sources as the mining prompts but do not overlap with them: no evaluation prompt shares its source identifier or its normalized text with any mining prompt, so none was generated on or trained on during FTPO. All prompts fit the 2{,}000-token prompt limit under every model's tokenizer. The set contains prompts from Evol-CodeAlpaca (758), LMSYS Arena conversations (675), MathQA (496), MetaMathQA (477), UltraInteract (319), MATH (132), GSM8K (125), and APPS (79).

\paragraph{Checkpoints and protocol.} We evaluate the released checkpoint of every training stage of each model: \texttt{CPT} and \texttt{Cooldown} (continual pretraining), \texttt{SFT} (the input to FTPO), \texttt{FTPO}, and the final model after RLVR. Each checkpoint is evaluated with its own tokenizer and chat template, with the same prompt instruction as in mining and in high-thinking mode. We draw one sample per prompt at temperatures 0.1, 0.3, and 0.6 (top-$k$ 50, min-$p$ 0.01, top-$p$ 1.0) with a budget of 22{,}000 new tokens, which gives 9{,}183 generations per checkpoint. The \emph{loop rate} is the percentage of generations in which the mining loop detector fires. These temperatures are a stress test: in high-thinking mode, the released models are recommended to be sampled at temperature 0.7 (Arabic) or 1.0 (English).

\begin{figure}[t]
\centering
\includegraphics[width=0.8\linewidth]{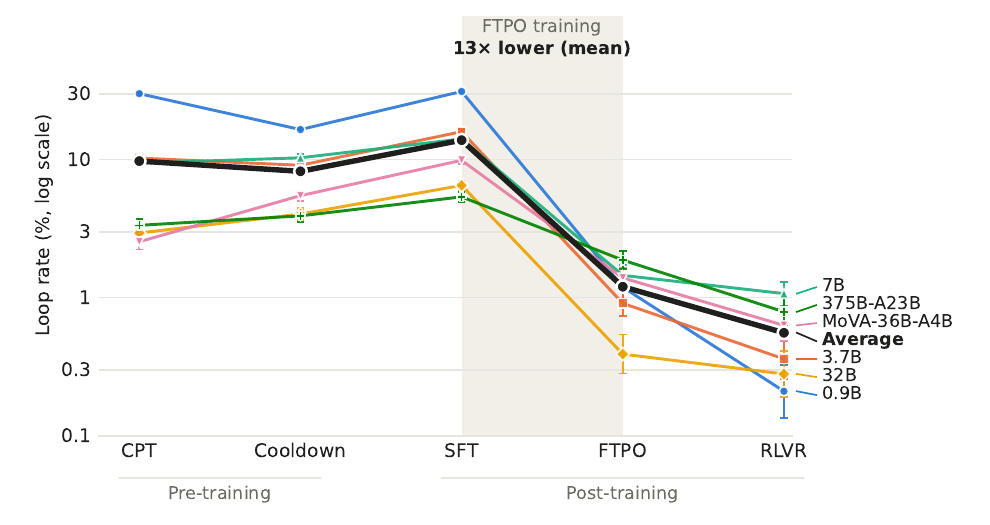}
\caption{Loop rate (\%, log scale) of the released checkpoint of each training stage on the internal evaluation set, averaged over temperatures 0.1, 0.3, and 0.6. RLVR denotes the final model. Error bars are Wilson 95\% intervals, and the shaded band marks the FTPO checkpoint.}
\label{fig:ftpo-doomloops}
\end{figure}

\begin{table}[t]
  \centering
  \small
  \setlength{\tabcolsep}{5pt}
  \begin{tabular}{@{}l cc ccc c@{}}
    \toprule
    & \multicolumn{2}{c}{\textbf{Continual pretraining}} & \multicolumn{3}{c}{\textbf{Post-training}} & \\
    \cmidrule(lr){2-3}\cmidrule(lr){4-6}
    \textbf{Model} & \textbf{CPT} & \textbf{Cooldown} & \textbf{SFT} & \textbf{FTPO} & \textbf{RLVR} & \textbf{Reduction} \\
    \midrule
    \model{Jais-3-0.9B}         & 30.0 & 16.5 & 31.1 & 1.2 & 0.2 & 26$\times$ \\
    \model{Jais-3-3.7B}         & 10.3 &  9.1 & 15.9 & 0.9 & 0.4 & 17$\times$ \\
    \model{Jais-3-7B}           &  9.4 & 10.3 & 14.1 & 1.5 & 1.1 & 10$\times$ \\
    \model{Jais-3-32B}          &  2.9 &  4.0 &  6.5 & 0.4 & 0.3 & 17$\times$ \\
    \model{Jais-3-MoVA-36B-A4B} &  2.5 &  5.5 &  9.9 & 1.4 & 0.6 &  7$\times$ \\
    \model{Jais-3-375B-A23B}    &  3.3 &  3.9 &  5.4 & 1.9 & 0.8 &  3$\times$ \\
    \midrule
    Average                     &  9.8 &  8.2 & 13.8 & 1.2 & 0.6 & 13$\times$ \\
    \bottomrule
  \end{tabular}
  \caption{Loop rate (\%) after each training stage, averaged over sampling temperatures 0.1, 0.3, and 0.6 (9{,}183 generations per checkpoint; Wilson 95\% half-widths $\le$1.0 point). RLVR is the final model. \emph{Reduction} is the ratio of the SFT to the FTPO loop rate; the average row reports the mean of the per-model ratios.}
  \label{tab:ftpo-results}
\end{table}

\paragraph{Findings.} FTPO reduces looping in all six models (Figure~\ref{fig:ftpo-doomloops}, Table~\ref{tab:ftpo-results}). Averaged over the models, the loop rate falls from 13.8\% at the SFT checkpoint to 1.2\% after FTPO, and every FTPO checkpoint loops on fewer than 2\% of generations. The gain is largest for the smallest models, which also loop most after SFT, consistent with the observation that looping decreases with model capacity~\citep{pipis2025wait}. The loop rate of \model{Jais-3-0.9B} falls from 31.1\% to 1.2\%, a 26-fold reduction, and those of \model{Jais-3-3.7B} and \model{Jais-3-7B} fall 17- and 10-fold. The larger models start from lower rates (5.4--9.9\%), so their reductions are smaller in absolute terms, but FTPO still lowers their loop rates 3- to 17-fold.

FTPO also removes the concentration of loops at low temperature. Before FTPO, loops are most frequent under near-greedy decoding, where a self-reinforcing repetition has little chance to be broken by sampling~\citep{xu2022learning,liquidai2026antidoom}. For \model{Jais-3-7B}, the SFT checkpoint loops on 18.8\% of generations at $T=0.1$ and on 7.7\% at $T=0.6$. After FTPO, its loop rate is between 1.1\% and 1.9\% at all three temperatures and rises slightly with temperature. Every model shows the same reversal, and at $T=0.1$ the loop rate falls from 7.0--37.5\% to 0.1--1.6\%. RL does not reintroduce loops: every final model loops less than its FTPO checkpoint, on 0.2--1.1\% of generations. Removing loops also helps on benchmarks, where a looping response never reaches an answer: on \model{Jais-3-7B}, FTPO adds approximately two points to the Arabic average.

\paragraph{Scale.} Antidoom was designed for small reasoning models, which are the most prone to doom loops, and was demonstrated on dense models with 2.6B and 4B parameters~\citep{liquidai2026antidoom}. In Antislop, the largest model trained with FTPO has 70B parameters and proved more sensitive to the training than the smaller ones~\citep{paech2025antislop}. We apply the same objective and hyperparameters to \model{Jais-3-375B-A23B}, a mixture-of-experts model approximately two orders of magnitude larger than the models Antidoom was shown on. As expected at this scale, its SFT checkpoint loops least of the six models, on 5.4\% of generations. FTPO nevertheless lowers this rate to 1.9\%, and from 7.0\% to 1.6\% at $T=0.1$, and the final model loops on 0.8\% of generations.

\subsection{Reinforcement Learning}\label{subsec:train-rl}

\subsubsection{Data}\label{subsec:rl-data}

We apply RLVR to five domains: instruction following, mathematics, code, Arabic grammatical analysis, and diacritization. Every prompt admits automatic verification, through constraint checking, comparison with a reference, or execution of test cases, so that each response can be scored by a rule-based verifier (Section~\ref{subsec:rl-reward}); Table~\ref{tab:rl-sources} lists the sources of each domain.

\paragraph{Instruction following.} Each prompt pairs a user request with one or more constraints checked by rule-based verifiers, such as ending with a given word, using no commas, or including a keyword. We draw on two sources. The first combines the AllenAI multi-constraint\footnote{\url{https://huggingface.co/datasets/allenai/IF_multi_constraints_upto5}} and Nemotron-RL instruction-following\footnote{\url{https://huggingface.co/datasets/nvidia/Nemotron-RL-instruction_following}} datasets, translated into Arabic with \model{gemma-4-31B-it}, with up to five constraints per prompt. Because translation often desynchronized a prompt and its verifier, for example a keyword verifier still expecting the English word, we kept only prompts whose constraint specification survived or could be restored, and removed those whose verifier would penalize a correct Arabic response or that had no rule-verifiable constraint. The second is a synthetic set of 20 Arabic-specific constraint families (21 for the 375B) covering orthography, diacritics, morphology, and structure, such as exact counts of \emph{hamza}, \emph{t\=a' marb\=u\d{t}a}, or \emph{shadda}, or the dual, relational adjective (\emph{nisba}), or feminine form of a given word. Each prompt attaches one constraint, in one of approximately ten paraphrases, to an open-ended Arabic request on cuisine, UAE knowledge, religion, or finance. Morphology verifiers use answer keys of 1.6K--7.1K words per family, expanded with CAMeL Tools~\citep{obeid-etal-2020-camel} (\texttt{calima-msa-r13}~\citep{taji-etal-2018-arabic}) and arramooz~\citep{zerrouki2011arramooz}. The synthetic set uses the \benchmark{AraIFBench} verifiers but shares no prompt with the benchmark (by exact match or a Jaccard similarity of at least 0.6). Instruction following is upsampled approximately 1.7-fold, from 4.7\,K distinct prompts to 7.8\,K. The 375B also trains on 10.9\,K new instruction-following prompts, screened with eight samples each from an earlier 375B checkpoint: we keep prompts whose rewards vary across samples and exclude those that no sample or every sample solves. 

\paragraph{Mathematics and code.} Both domains are drawn from an Arabic translation of the mathematics and code subsets of Guru\footnote{\url{https://huggingface.co/collections/IFM/guru}}. Each mathematics problem has a reference answer, and each code problem a set of test cases.

\paragraph{Grammatical analysis and diacritization.} Both tasks are built from classical Arabic texts and the canonical text of the Qur'an; grammatical analysis also uses the official answer keys of school examinations from several Arab countries. In grammatical analysis (\emph{i`r\=ab}), the prompt lists the words of a sentence to analyze, in the order of a word-by-word reference taken from works of Qur'anic and general \emph{i`r\=ab} or from the answer keys. Every analyzed word must occur verbatim in its sentence, every Qur'anic analysis is anchored to a specific verse, and repeated words are numbered and scored separately. In diacritization (\emph{tashk\=il}), sentences come only from fully vocalized texts and are kept only if every letter whose diacritic is required by grammar or morphology is marked, apart from a few standard conventions, and if removing the diacritics reproduces the prompt exactly. Most items, 76\% in grammatical analysis and 88\% in diacritization, are also checked against their source passage by two LLM judges from different model families and kept only if both accept them. We remove duplicates, hold out a validation split by source, and score each reference against itself, discarding prompts whose reference obtains a reward below 0.99; this excludes references that the reward function cannot interpret. The mixture samples 4.8\,K grammatical-analysis and 2.0\,K diacritization prompts from these pools.

\begin{table}[t]
  \centering
  \small\rmfamily
  \renewcommand{\arraystretch}{1.1}
  \begin{tabularx}{\linewidth}{@{}l>{\raggedright\arraybackslash}Xr@{}}
    \toprule
    \textbf{Domain} & \textbf{Sources (entries)} & \textbf{Entries} \\
    \midrule
    Instruction following & Synthetic Arabic constraints (5{,}044), AllenAI multi-constraint (2{,}267), Nemotron-RL (507) & 7{,}818 \\
    Mathematics & DAPO / OR1 pool (2{,}393), DeepScaleR (1{,}325) & 3{,}718 \\
    Code & PrimeIntellect (582), TACO (338), LiveCodeBench (67), other (413) & 1{,}400 \\
    Grammatical analysis & Qur'anic text (2{,}475), classical texts (1{,}940), exam keys (418) & 4{,}833 \\
    Diacritization & Classical texts (1{,}752), Qur'anic text (246), exam keys (2) & 2{,}000 \\
    \midrule
    Total & & 19{,}769 \\
    \bottomrule
  \end{tabularx}
  \caption{Source composition of the final RLVR mixture, in training entries after selection and upsampling. Mathematics and code sources come from Guru; the 67 LiveCodeBench problems do not overlap with LiveCodeBench~v6 used in evaluation.}
  \label{tab:rl-sources}
\end{table}

\subsubsection{Reward Functions}\label{subsec:rl-reward}

All rewards are computed by deterministic rule-based verifiers. A response is scored once its generation ends, and the verifiers read only its answer, the text after the closing reasoning tag. A response is truncated if it reaches the generation limit of $H_{\max}=16{,}384$ tokens before the end-of-turn token. Because rule-based rewards remain open to reward hacking, the instruction-following and Arabic-language rewards add penalties and caps for degenerate outputs such as repetition, padding, and language switching.

\paragraph{Instruction following.}
Let $a$ be the answer of a response $o$, the text after its closing reasoning tag, and let $v_k(a)\in\{0,1\}$ be the verdict of the verifier for the $k$-th of the prompt's $K\le 5$ constraints; a constraint that cannot be verified counts as failed. The reward starts from the fraction $s(a)$ of satisfied constraints and subtracts four penalties, each in $[0,1]$:
\begin{equation}
  \tilde R_{\mathrm{IF}}(a) = \underbrace{\frac{1}{K}\sum_{k=1}^{K} v_k(a)}_{s(a)} - \lambda_{\mathrm{rep}} P_{\mathrm{rep}} - \lambda_{\mathrm{short}} P_{\mathrm{short}} - \lambda_{\mathrm{long}} P_{\mathrm{long}} - \lambda_{\mathrm{lang}} P_{\mathrm{lang}},
  \label{eq:if-reward}
\end{equation}
with $(\lambda_{\mathrm{rep}},\lambda_{\mathrm{short}},\lambda_{\mathrm{long}},\lambda_{\mathrm{lang}})=(0.4,\,0.3,\,0.1,\,0.2)$. The \emph{repetition} penalty $P_{\mathrm{rep}}$ is the square of the largest of four normalized statistics: dominant-word frequency, repeated $n$-grams, duplicated sentences, and compressibility, ignoring text that the prompt asks to repeat. The \emph{length} penalties compare the answer length $n$ with a floor $f$ and a ceiling $u$:
\begin{equation}
  P_{\mathrm{short}} = \mathbf{1}[s(a)\ge 0.8]\left(\frac{[f-n]_+}{\max(1,f)}\right)^{2},
  \qquad
  P_{\mathrm{long}} = \left(\frac{[n-u]_+}{\max(1,u)}\right)^{2},
\end{equation}
where $[x]_+=\max(x,0)$; only otherwise-correct answers are penalized for brevity. An explicit word, sentence, or paragraph constraint sets $f$ and $u$ in its own unit. Otherwise, $n$ counts words, $f$ is a heuristic floor set by the requested form of the answer, and $u=1{,}200$, which also applies when a constraint gives only a lower bound. The \emph{language} penalty $P_{\mathrm{lang}}$ grows with the fraction of letters in the wrong script, Latin letters in an Arabic answer or Arabic letters in an English one, and is waived for code and structured output. Finally, four severe failures cap the reward. With $\kappa(a)$ the lowest applicable cap among $-0.5$ (degenerate repetition), $-0.25$ (truncation), $0$ (substantial use of the wrong script), and $0.2$ (every constraint met but $n<0.33f$), and $\kappa(a)=+\infty$ if none applies,
\begin{equation}
  R_{\mathrm{IF}}(o) =
  \begin{cases}
    0, & |a| = 0,\\
    \operatorname{clip}\bigl(\min(\tilde R_{\mathrm{IF}}(a),\,\kappa(a)),\,-1,\,1\bigr), & \text{otherwise},
  \end{cases}
  \label{eq:if-final}
\end{equation}
where $|a|=0$ denotes an empty answer, for example when the response is truncated within its reasoning.

\paragraph{Mathematics and code.}
Both rewards are binary up to negligible shaping terms and are conditioned on the answer format: a response that is truncated, or that lacks a single final-answer statement with the prescribed Arabic marker, receives $0$. A mathematics answer receives $1$ if its final expression, given in a \verb|\boxed{}| environment or as an option for a multiple-choice question, is equivalent to the reference under Math-Verify~\citep{kydlicek2025mathverify}, which tests symbolic and numerical equivalence. A code answer receives $1$ if its final program passes every test case.

\paragraph{Grammatical analysis and diacritization.}
Both rewards grant partial credit through a score $S\in[0,1]$. Deviations from the canonical format (one \emph{word}\,:\,\emph{analysis} line per word, or the bare diacritized text) are penalized with coefficient $\phi$:
\begin{equation}
  R = \bigl(1-\phi\,(1-F)\bigr)\,S ,
  \label{eq:format-penalty}
\end{equation}
where $F=1$ if the format is canonical and $0$ otherwise.

\subparagraph{Diacritization.}

\emph{Preprocessing.} An allowlisted answer label or a single quoted or bracketed span matching the reference's letters is extracted as the candidate. Texts are NFKC-normalized, tatweel is removed, and each letter is paired with its set of diacritics.

\emph{Score.} Let $N$ be the number of letters in the reference. The score is $0$ if $N<N_{\min}$ or if the base letters differ. Otherwise, letter $i$ is correct, $m_i=\mathbf{1}[D_i=D_i^\star]$, when its diacritic set $D_i$ matches the reference set $D_i^\star$, and
\begin{equation}
  S = w_{\mathrm{char}}\,A_{\mathrm{char}} + w_{\mathrm{word}}\,A_{\mathrm{word}} + w_{\mathrm{fin}}\,A_{\mathrm{fin}},
  \qquad \textstyle\sum w = 1,
  \label{eq:tashkeel}
\end{equation}
where
\begin{itemize}
  \item $A_{\mathrm{char}}=\frac{1}{N}\sum_i m_i$ is the fraction of correct letters;
  \item $A_{\mathrm{word}}$ is the fraction of fully correct words;
  \item $A_{\mathrm{fin}}$ is the accuracy on diacritized word-final letters ($1$ if none).
\end{itemize}
$A_{\mathrm{fin}}$ isolates case and mood endings. These are syntax-dependent and hard, but too rare to register in $A_{\mathrm{char}}$.

\emph{Caps.} $S\leftarrow\min(S,\kappa_{\mathrm{nd}})$ if the answer has no diacritics while the reference has at least $\max(n_0,\lceil\eta N\rceil)$ marks. $S\leftarrow\min(S,\kappa_{\mathrm{lo}})$ if $A_{\mathrm{char}}<A_{\min}$.

\subparagraph{Grammatical analysis (i`r\=ab).}

\emph{Preprocessing.} Diacritics are removed, letter variants folded, and non-Arabic characters dropped. The target word comes from item metadata or, for single-word references, from the reference label.

\emph{Features.} Each analysis yields:
\begin{itemize}
  \item \textbf{Role $r$:} the first role mentioned. If it is a nominal category (relative, interrogative, demonstrative, or conditional noun), the first syntactic function is used instead. Roles are matched through alias rules. $r$ \emph{agrees} with $r^\star$ if they are equal or if $r^\star$ is a nominal category that the answer also names. They \emph{disagree} if they name different nominal categories.
  \item \textbf{Case $c$:} the first case other than \textit{mabn\=\i}, with \textit{f\=\i\ ma\d{h}all} forms mapped to their case.
  \item \textbf{Overlap $O$:} set-based F1 over content tokens. Content tokens exclude stopwords and one-letter tokens, and accusative forms are mapped to their base.
  \item \textbf{Indicators:} $M=1$ if the case marker matches or none is specified. $T=1$ if the answer analyzes the known target word, labeled as such or unlabeled.
\end{itemize}

\emph{Score.} The first applicable branch applies. An exact match means identical normalized analyses, stopwords included.
\begin{equation}
  S =
  \begin{cases}
    1, & \text{exact match},\\
    \gamma_u\, O, & r^\star \text{ undefined},\\
    \gamma_r\, O, & r \text{ disagrees with } r^\star,\\
    b_c + \gamma_c\, O, & c^\star \text{ defined},\ c{\neq}c^\star,\\
    b_0 + \lambda_M M + \lambda_O O + \lambda_T T, & \text{otherwise}.
  \end{cases}
  \label{eq:irab}
\end{equation}
Since $b_0+\lambda_M+\lambda_O+\lambda_T<1$, only an exact match scores $1$.

\emph{Caps}, applied even to exact matches:
\begin{itemize}
  \item $\kappa_{\mathrm{soft}}$: hedging or contradictory roles beyond the reference's, multiple analyses of the target, or excessive length or repetition;
  \item $\kappa_{\mathrm{hard}}$: keyword stuffing, the wrong target word, or non-Arabic output.
\end{itemize}

\emph{Multi-word items} require $K_w\ge2$ analyses. Each is scored with Eq.~\ref{eq:irab} and its caps, with $T=0$, and missing analyses score $0$. Then $S=\sum_k S_k/(K_w+D)$, where $D$ counts duplicates. The item caps are $\kappa_{\mathrm{hard}}$ for stuffing or non-Arabic output and $\kappa_{\mathrm{soft}}$ for excessive length.

When verified alternative parses exist, the best-scoring reference is used. Hyperparameters are listed in Table~\ref{tab:arabic-reward-hparams} (Appendix~\ref{app:reward-hparams}).

\subsubsection{Training Setup}\label{subsec:rl-training}

Each run starts from the model's FTPO checkpoint, and all models share one recipe: a group-relative policy-gradient objective, the same batch size, and the same optimizer. Two settings depend on the model: the two mixture-of-experts models correct for the mismatch between the inference engine and the trainer, and the 375B filters its batches with dynamic sampling, which keeps only groups whose rewards are not nearly identical. Training alternates between generating responses and updating the policy on them; we call one such iteration a \emph{rollout}.

\paragraph{Objective.}
We use an objective in the style of group relative policy optimization (GRPO)~\citep{shao2024deepseekmath}, in which each response is judged against the other responses to the same prompt, so no value model is needed. Let $\pi_\theta$ be the policy being optimized and $\pi_{\theta_{\mathrm{old}}}$ the policy at the start of the current rollout. For each prompt $q$ we sample a group of $G=16$ responses $\{o_i\}_{i=1}^{G}$ and score each with the reward $R_i$ of its domain (Section~\ref{subsec:rl-reward}). Following Dr.~GRPO~\citep{liu2025understanding}, we center the rewards on the group mean without dividing by their standard deviation:
\begin{equation}
  \hat A_i = R_i - \frac{1}{G}\sum_{j=1}^{G} R_j .
  \label{eq:rl-adv}
\end{equation}
Every token of $o_i$ receives the advantage $\hat A_i$, and we minimize the clipped surrogate loss of proximal policy optimization (PPO)~\citep{schulman2017proximal},
\begin{equation}
  \mathcal L(\theta)= -\frac{1}{B}\sum_{i\in\mathcal B}\frac{1}{|o_i|}\sum_{t=1}^{|o_i|}
  \omega_{i,t}\,\min\!\Big(\rho_{i,t}(\theta)\,\hat A_i,\;
  \operatorname{clip}\big(\rho_{i,t}(\theta),\,1-\epsilon_{\mathrm{clip}},\,1+\epsilon_{\mathrm{clip}}\big)\,\hat A_i\Big),
  \label{eq:rl-loss}
\end{equation}
where $\rho_{i,t}(\theta)=\pi_\theta(o_{i,t}\mid q,o_{i,<t})/\pi_{\theta_{\mathrm{old}}}(o_{i,t}\mid q,o_{i,<t})$, $\mathcal B$ is a minibatch of $B=512$ responses, $\epsilon_{\mathrm{clip}}=0.2$, and $\omega_{i,t}$ is the correction weight defined below. Unlike Dr.~GRPO, we keep the per-response normalization $1/|o_i|$, so every response carries the same total weight regardless of its length; \citet{liu2025understanding} show that this favors shorter responses when the advantage is positive and longer ones when it is negative. Because the advantage is centered on the group mean, a group whose responses all receive the same reward carries no reward signal. No run uses a KL penalty or an entropy bonus.

\paragraph{Engine--trainer mismatch.}
The inference engine and the trainer compute token probabilities with different kernels and numerical precision, so the engine samples from a distribution $\pi_{\mathrm{roll}}$ that can differ from $\pi_{\theta_{\mathrm{old}}}$ even for identical weights. Truncated importance sampling (TIS)~\citep{yao2025rollout} corrects this mismatch by reweighting each token with the ratio of the trainer's probability to the engine's, truncated at $C_{\mathrm{TIS}}$ so that rare large ratios cannot dominate the gradient:
\begin{equation}
  \omega_{i,t}=\min\!\left(\frac{\pi_{\theta_{\mathrm{old}}}(o_{i,t}\mid q,o_{i,<t})}{\pi_{\mathrm{roll}}(o_{i,t}\mid q,o_{i,<t})},\;C_{\mathrm{TIS}}\right).
  \label{eq:rl-tis}
\end{equation}
The weight is treated as a constant in the gradient, and the trainer recomputes $\pi_{\theta_{\mathrm{old}}}$ rather than reusing the engine's log-probabilities. The mismatch is largest for the two mixture-of-experts models, the 36B MoVA and the 375B (Figure~\ref{fig:rl-diagnostics}a), so we apply TIS with $C_{\mathrm{TIS}}=2$ to both; the dense models use $\omega_{i,t}=1$.

\paragraph{Rollouts and optimization.}
Each rollout samples responses until 128 groups (2{,}048 responses) are accepted, splits them into four minibatches of 512 for four optimizer steps, and then synchronizes the updated weights to the inference engines. Responses are sampled in high-thinking mode at temperature 1.0 and top-$p$ 1.0, without a top-$k$ cutoff, up to $H_{\max}$ generated tokens. We use the AdamW optimizer~\citep{loshchilov2017decoupled} ($\beta_1=0.9$, $\beta_2=0.98$, weight decay 0.1) with gradient clipping at 1.0 and a constant learning rate of $1.0\times10^{-6}$ without warmup, in bfloat16, for 200 rollouts (800 optimizer steps). For the mixture-of-experts models, expert-routing biases are kept fixed and no load-balancing loss is used; the router weights are still trained.

\paragraph{Group filtering.}
Each rollout targets 43 instruction-following, 28 mathematics, 14 code, 29 grammatical-analysis, and 14 diacritization groups. For the 375B we use dynamic sampling~\citep{yu2026dapo}, which admits a group only if the standard deviation of its 16 rewards exceeds $0.02$. For binary rewards, this admits every group in which at least one response differs; for continuous rewards, it also excludes groups whose rewards differ only marginally. A group with identical rewards has zero advantage for every response (Equation~\ref{eq:rl-adv}) yet contributes to the normalization $1/B$ (Equation~\ref{eq:rl-loss}), so it dilutes the gradient without contributing signal. Such groups become frequent as the policy improves: for the 375B, 80\% of mathematics groups had identical rewards, and 267 groups were sampled on average to obtain 128 admissible ones.

\paragraph{Infrastructure.}
{\emergencystretch=2em We train with RL360~\citep{rl360}, which builds on miles~\citep{radixark2026milesv01productionlevelposttraining}. It uses Megatron-LM~\citep{shoeybi2019megatron} or Fully Sharded Data Parallel (FSDP) trainers, with SGLang inference engines~\citep{zheng2024sglang} that share the same GPUs. We tune the parallel layout of each model, combining tensor, context, pipeline, and expert parallelism, to keep rollout latency low.\par}

\subsubsection{Results}\label{subsec:rl-results}

We assess RLVR in three ways: by comparing each model with its FTPO checkpoint on the Arabic benchmarks of Section~\ref{sec:eval}, by tracking \benchmark{AraIFBench}, a benchmark whose prompts are excluded from training, during the training of the 375B, and by analyzing the training statistics of all runs.

\begin{figure}[t]
  \centering
  \includegraphics[width=\linewidth]{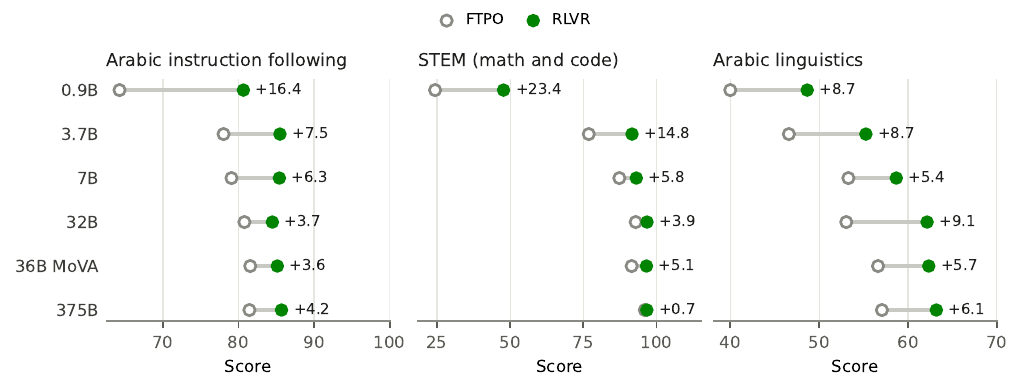}
  \caption{FTPO checkpoint (open circles) and final RLVR model (filled circles) in the three domains that RLVR trains; labels give the change in points. Arabic instruction following is the mean of \benchmark{AraIFBench} and \benchmark{IFEval-Ar}; STEM is the mean of \benchmark{AIME~2025} and \benchmark{AIME~2026} (pass@1 over eight samples), \benchmark{HumanEval} and \benchmark{MBPP}; Arabic linguistics is the mean of the \emph{i`r\=ab} score and the diacritization exact-match accuracy on a held-out generative test set (Section~\ref{sec:ablation-cascade}). All checkpoints are evaluated in high-thinking mode.}
  \label{fig:rl-arabic}
\end{figure}

\paragraph{Arabic benchmarks.}
RLVR raises 18 to 25 of the 28 Arabic benchmark scores that we track for each model, and its gains concentrate on the three domains that it trains (Figure~\ref{fig:rl-arabic}). Arabic instruction following improves for every model, by 7.0 points on average and by 16.4 points for the 0.9B. The STEM gain is largest for the small models, 23.4 points for the 0.9B and 14.8 for the 3.7B, against 3.9 to 5.8 points for the 7B to 36B models and 0.7 for the 375B, whose FTPO checkpoint already scores 94.6 to 97.5 on these benchmarks. Arabic linguistics, measured on a held-out generative test set, improves by 5.4 to 9.1 points for every model. RLVR also changes the standing within a size class: after RLVR, the 32B overtakes \model{gemma-4-31B-it} by 0.5 points (83.1 against 82.6), and the 36B MoVA trails it by only 0.1 points (82.5).

\begin{figure}[t]
  \centering
  \includegraphics[width=\linewidth]{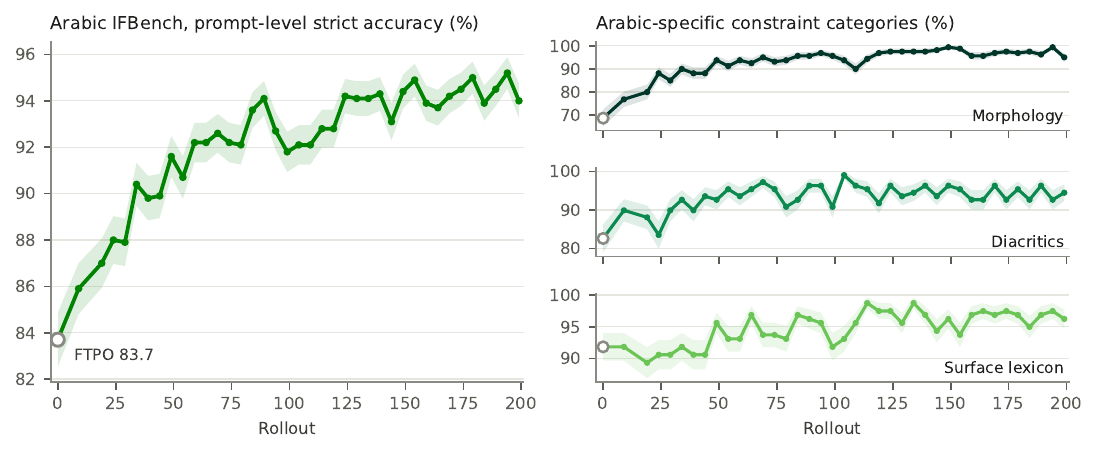}
  \caption{RLVR raises Arabic instruction following of the 375B from 83.7\% to 94--95\%. Left: prompt-level strict accuracy on \benchmark{AraIFBench}, evaluated at rollout 9 and every five rollouts from rollout 19; it passes 90\% at rollout 34 and stays at 92--95\% from rollout 90 onward. Right: accuracy on three Arabic-specific constraint categories (up to 160 prompts each), each on its own scale. Morphology, the weakest category after FTPO at 68.8\%, gains the most and reaches 95--99\%. Rollout 0 (open circles) is the FTPO checkpoint, and the shading marks one standard error. All checkpoints are evaluated in high-thinking mode at temperature 1.0 and top-$p$ 0.95.}
  \label{fig:rl-ifbench}
\end{figure}

\paragraph{Arabic instruction following during training.}
Figure~\ref{fig:rl-ifbench} tracks \benchmark{AraIFBench} on the 375B during training. Prompt-level strict accuracy rises from 83.7\% at the FTPO checkpoint to 94--95\%, most of it within the first 90 rollouts. The weakest constraint categories after FTPO gain the most: Arabic morphology rises from 68.8\% to 95--99\%, and Arabic diacritics from 82.6\% to 93--96\%, while script constraints start near their ceiling. Because the synthetic Arabic constraints in the RLVR data are scored with the verifiers of this benchmark, and the additional prompts of the 375B target its weakest constraint types (Section~\ref{subsec:rl-data}), these gains are in-distribution in constraint type, although no benchmark prompt is used in training.

\begin{figure}[t]
  \centering
  \includegraphics[width=\linewidth]{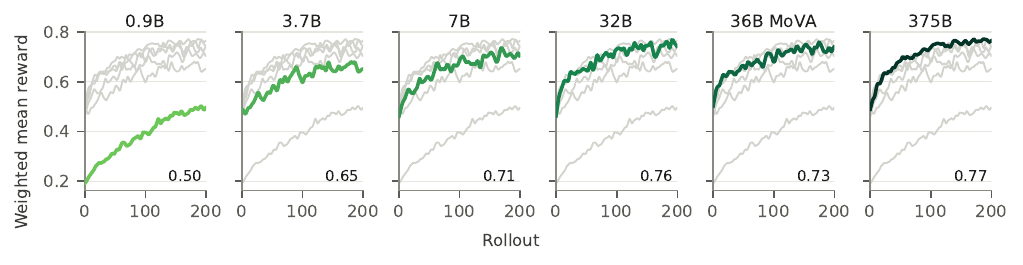}
  \caption{Training reward on all sampled responses, averaged over domains with the target mixture weights (Gaussian smoothing, $\sigma=2$ rollouts). Each panel highlights one model, with the other five in gray; the number is its mean reward over the last ten rollouts.
  }
  \label{fig:rl-reward}
\end{figure}

\paragraph{Training reward.}
The training reward rises steadily for every model (Figure~\ref{fig:rl-reward}). Averaged over domains with the weights of the target mixture, so that runs with different realized mixtures remain comparable, it increases from 0.21--0.55 over the first ten rollouts to 0.50--0.77 over the last ten. Among the dense models, the reward over the last ten rollouts grows with size, from 0.50 for the 0.9B to 0.76 for the 32B, and the 375B reaches 0.77.

\paragraph{Sharpening and discovery.}
Following \citet{chen2026does}, we compare pass@1 with pass@16 within each group of 16 responses on the training prompts, between the first and the last ten rollouts. From the 3.7B upward, RLVR mainly sharpens the policy: on mathematics, code, and instruction following, pass@1 rises more than pass@16 for every model, for instance by 3 to 17 points against at most 7 points on mathematics. The 0.9B model is the exception, with a gain of 44 points in pass@16 on mathematics, consistent with RLVR teaching it to solve problems it previously failed. Following each mathematics prompt from its first to its last visit (Table~\ref{tab:rl-transitions}) gives the same picture for the models from 3.7B up: RLVR completes partially solved prompts, 48--70\% of which are then solved by all 16 responses, solves 14--25\% of the prompts that were initially unsolved, and loses only 4--6\% of the prompts it could solve.

\begin{table}[t]
  \centering
  \small\rmfamily
  \renewcommand{\arraystretch}{1.15}
  \setlength{\tabcolsep}{6pt}
  \begin{tabular}{@{}lccc@{}}
    \toprule
    \textbf{Model} & \textbf{Unsolved $\to$ solved} & \textbf{Partial $\to$ saturated} & \textbf{Solved $\to$ unsolved} \\
    \midrule
    \model{Jais-3-0.9B}          & 35.8 & 11.6 & 11.4 \\
    \addlinespace[1pt]
    \model{Jais-3-3.7B}          & 19.0 & 59.0 &  4.8 \\
    \addlinespace[1pt]
    \model{Jais-3-7B}            & 18.7 & 56.0 &  4.3 \\
    \addlinespace[1pt]
    \model{Jais-3-32B}           & 13.6 & 70.4 &  4.1 \\
    \addlinespace[1pt]
    \model{Jais-3-MoVA-36B-A4B}  & 17.1 & 63.8 &  4.7 \\
    \addlinespace[1pt]
    \model{Jais-3-375B-A23B}     & 25.1 & 48.5 &  6.0 \\
    \bottomrule
  \end{tabular}
  \caption{Fate of mathematics prompts between their first and last visit during RLVR (\%). A prompt is unsolved when none of its 16 responses is correct, saturated when all are, and partial otherwise; it is solved when at least one response is correct. Each column is normalized by the number of prompts in the starting state.}
  \label{tab:rl-transitions}
\end{table}

\begin{figure}[t]
  \centering
  \includegraphics[width=\linewidth]{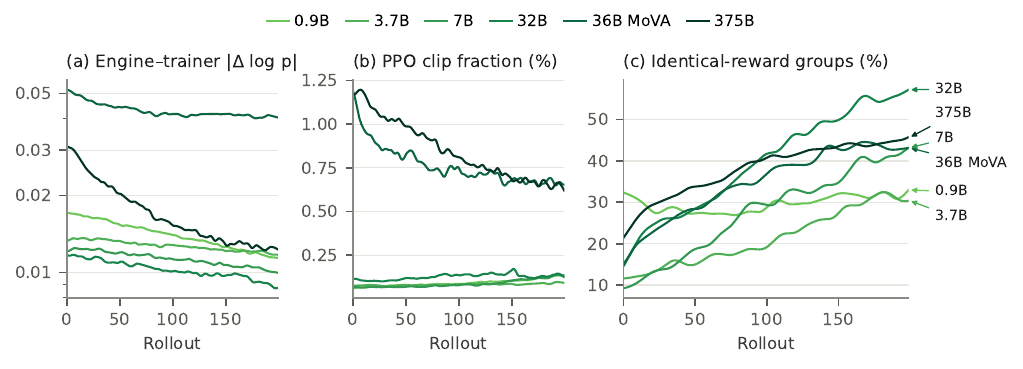}
  \caption{Optimization statistics during RLVR (Gaussian smoothing, $\sigma=2$ rollouts; $\sigma=5$ in panel c). (a)~Mean absolute difference between the log-probabilities that the inference engine and the trainer assign to the same tokens (log scale). (b)~Fraction of tokens at which the PPO clip is active. (c)~Share of groups whose 16 responses all receive the same reward.}
  \label{fig:rl-diagnostics}
\end{figure}

\paragraph{Optimization diagnostics.}
Figure~\ref{fig:rl-diagnostics} reports three optimization statistics. For every model except the 0.9B, the share of groups with identical rewards grows from 9--23\% to 30--57\% (panel c), the regime that motivates dynamic sampling for the 375B. Updates stay well inside the PPO trust region, with a clip fraction below 0.3\% for the dense models and of 0.6--1.3\% for the mixture-of-experts models (panel b). At the start of training, the gap between the log-probabilities of the inference engine and of the trainer is two to four times larger for the mixture-of-experts models than for the dense models (panel a), which is the mismatch that TIS corrects. The gap stays at this level for the 36B MoVA, whereas for the 375B it decays almost to that of the dense models; with $C_{\mathrm{TIS}}=2$, TIS truncates fewer than 0.3\% of the tokens. Although no run uses a KL penalty, the KL divergence to the FTPO policy reaches 0.06--0.17 per token after 200 rollouts; the 375B run loads no reference model, so its KL is not measured. The mean log-probability of the sampled tokens rises for every model, from between $-1.02$ and $-0.49$ to between $-0.42$ and $-0.06$: the policy becomes more deterministic, in line with the entropy collapse reported by \citet{cui2025entropy} and with the sharpening above.

\paragraph{Degenerate loops.}
On the loop evaluation of Section~\ref{subsec:ftpo-results}, RLVR does not reintroduce the loops that FTPO removes: every final model loops on 0.2--1.1\% of generations, less than its FTPO checkpoint, and the average loop rate falls from 1.2\% to 0.6\% (Table~\ref{tab:ftpo-results}).

\subsection{Cascade versus Mixed-Task RLVR}\label{subsec:rl-cascade}\label{sec:ablation-cascade}

An alternative to a fixed mixture is a curriculum-style cascade that trains the five skills one at a time: instruction following, diacritization, grammatical analysis, mathematics, and then code. We compared the two under matched conditions, with the same starting checkpoint, per-domain exposure and number of optimizer steps; the rest of this section gives the design and the full results. The recipes reach similar results on the two Arabic-language tasks (diacritization and grammatical analysis). Both gain approximately ten points on grammatical analysis, and both learn instruction following and
transfer it from Arabic to English. Neither forgets untrained capabilities
broadly. They differ on mathematics. Within 120 optimizer steps of its instruction-following stage, the cascade loses approximately 13 points on GSM8K and 8 on MATH-500 in high-thinking mode. Later stages repair this only partly, so the cascade ends 6 to 11 points below its starting checkpoint on short-form mathematics. The mixed recipe holds or improves these scores and gains far more on AIME. We therefore adopt the mixed recipe. Each arm was trained once, and we tested a single cascade order, so we treat this as a practical default rather than a general result about sequential training.

\providecommand{\irab}{\textit{i\lq{}rab}}
\providecommand{\Irab}{\textit{I\lq{}rab}}

\subsubsection{Question, Design, and Protocol}
\label{sec:design}

Multi-skill RLVR can compose its training signal in two ways. The prevailing recipe
samples a fixed domain mixture into every batch. The alternative, motivated by
curriculum learning, trains skills in sequence so that each stage starts from a policy
already shaped by the previous one. The two make opposite predictions. A cascade may
\emph{forget} early stages as later ones overwrite them, or it may \emph{benefit} from
stage-wise specialization. We test both predictions on our 3.7B model under a design
that holds everything fixed except the order in which gradients arrive.

\paragraph{Design.} Both arms branch from the same \texttt{FTPO} checkpoint, draw from the same five prompt
pools, and use the same acceptance rule. Each arm runs 200 batches. A batch contains 128
accepted prompt groups of 16 rollouts each, or 2{,}048 rollouts, and is split into four
disjoint minibatches of 512 rollouts, one optimizer update each. This gives 800 updates
per arm. The mixed arm composes every batch to the quota
\texttt{if=43, i3rab=29, tashkeel=14, math=28, code=14}. The cascade trains the five
skills in the order instruction following (batches 1--67), tashkeel (68--89),
\textit{i\lq{}rab} (90--134), mathematics (135--178) and code (179--200), with stage
lengths chosen so that cumulative per-domain exposure matches the mixed quota to within
$1\%$ (Table~\ref{tab:design}). The arms therefore differ in \emph{when} a skill's
gradient arrives, not in how much of it there is.

\begin{table}[h]
\centering
\footnotesize
\setlength{\tabcolsep}{3pt}
\begin{tabular}{@{}lcrrrr@{}}
\toprule
\textbf{Skill} & Cascade & Cascade & Cascade & Mixed & Ratio \\
 & batches & span & groups & groups & \\
\midrule
Instruction following & 67 & 1--67    &  8,576 &  8,600 & 1.00 \\
Tashkeel              & 22 & 68--89   &  2,816 &  2,800 & 1.01 \\
\textit{I\lq{}rab}    & 45 & 90--134  &  5,760 &  5,800 & 0.99 \\
Mathematics           & 44 & 135--178 &  5,632 &  5,600 & 1.01 \\
Code                  & 22 & 179--200 &  2,816 &  2,800 & 1.01 \\
\midrule
\textbf{Total}        & \textbf{200} & & \textbf{25,600} & \textbf{25,600} & \textbf{1.00} \\
\bottomrule
\end{tabular}
\caption{Per-domain exposure. Cascade stage lengths reproduce the mixed arm's batch
quota to within $1\%$, so the two recipes are matched on data, on optimizer updates
(800 each) and on starting checkpoint. All prompts are in Arabic.}
\label{tab:design}
\end{table}

Because the cascade trains one skill at a time, its intermediate checkpoints also act as
a probe of what each stage does. We evaluate it at every stage boundary (steps 67, 89,
134, 178, 200) and at eight interior points (30, 55, 75, 100, 119, 125, 175, 190).

\paragraph{Reporting and noise.} Results are movements from FTPO, as three-seed means in
a low and a high reasoning mode. Each benchmark's per-seed standard deviation $\sigma$
is pooled over the FTPO and cascade checkpoints. Across the full evaluation suite it is
below $1.0$ for 53 of 86 benchmark--mode pairs and largest for AIME ($2.2$--$2.7$),
GPQA-Diamond ($2.5$) and AA-LCR ($2.4$). It is about $0.6$ for \irab{}, and $0.35$ (low)
and $0.83$ (high) for the tashkeel diacritic error rate (DER). $^{\dagger}$ marks differences larger than twice
their standard error. Among about 150 marked comparisons roughly seven $^{\dagger}$
are expected by chance. Because each recipe was trained once, $^{\dagger}$ means
``larger than re-evaluation would produce,'' not ``a robust recipe effect.'' Each mode is decoded at its recommended setting (for Arabic
linguistics, temperature $0.2$ and $0.7$), so high-minus-low contrasts bundle reasoning
budget with temperature. Lower-is-better metrics ($\downarrow$) are reported raw.

\paragraph{Arabic linguistics.} \emph{Tashkeel} is scored deterministically against gold
diacritization on 1{,}000 decontaminated passages. We report micro-averaged DER
and word error rate (WER), exact passage accuracy, and the \emph{base-text
alteration rate}: the share of passages in which the model changed letters or word
boundaries or added text. An altered passage scores 100\% DER and WER, and a valid
alternative vocalization counts as an error. \emph{\Irab} uses 474 examples, after
removing 26 of 500 that two closed-source models flagged. A \model{gemma-4-31B-it} judge (temperature 0)
scores each named word occurrence 0--5 against the gold analysis, accepting valid
alternatives; scores are averaged to 0--100.

\begin{table}[!t]
\centering
\footnotesize
\setlength{\tabcolsep}{5pt}
\begin{tabular}{@{}lrrrrrr@{}}
\toprule
& \multicolumn{3}{c}{\textbf{Reasoning: low}} & \multicolumn{3}{c}{\textbf{Reasoning: high}} \\
\cmidrule(lr){2-4}\cmidrule(lr){5-7}
\textbf{Benchmark} & FTPO & $\Delta$ Mixed & $\Delta$ Cascade & FTPO & $\Delta$ Mixed & $\Delta$ Cascade \\
\midrule
\multicolumn{7}{@{}l}{\textit{Instruction following}} \\
\quad IFEval-Ar (in-language)  & 65.84 & $+5.20^{\dagger}$ & $+3.80^{\dagger}$ & 71.70 & $+5.36^{\dagger}$ & $+5.28^{\dagger}$ \\
\quad IFEval (en, transfer)    & 77.45 & $+2.77^{\dagger}$ & $-1.36$ & 84.66 & $+7.89^{\dagger}$ & $+5.98^{\dagger}$ \\
\multicolumn{7}{@{}l}{\textit{Mathematics}} \\
\quad Arabic GSM8K (in-language) & 91.23 & $+2.98^{\dagger}$ & $-1.06^{\dagger}$ & 94.27 & $+0.23$ & $-6.45^{\dagger}$ \\
\quad GSM8K (0-shot)   & 96.88 & $+0.90^{\dagger}$ & $-2.60^{\dagger}$ & 97.31 & $-0.14$ & $-10.29^{\dagger}$ \\
\quad GSM8K CoT        & 96.55 & $+0.97$ & $-2.16^{\dagger}$ & 97.35 & $+0.54$ & $-11.06^{\dagger}$ \\
\quad MATH-500          & 93.27 & $+2.47^{\dagger}$ & $-6.47^{\dagger}$ & 93.27 & $+4.53^{\dagger}$ & $-6.07^{\dagger}$ \\
\quad AIME 2025        & 42.36 & $+7.78^{\dagger}$ & $+3.75$ & 58.89 & $+25.42^{\dagger}$ & $+6.67^{\dagger}$ \\
\quad AIME 2026        & 49.44 & $+10.00^{\dagger}$ & $+0.56$ & 60.42 & $+23.89^{\dagger}$ & $+12.64^{\dagger}$ \\
\multicolumn{7}{@{}l}{\textit{Code (English prompts at evaluation)}} \\
\quad HumanEval        & 93.29 & $+1.42$ & $+0.61$ & 94.51 & $+2.24$ & $+2.64^{\dagger}$ \\
\quad MBPP             & 86.33 & $+0.93$ & $-2.80^{\dagger}$ & 85.67 & $+4.00^{\dagger}$ & $-1.60$ \\
\multicolumn{7}{@{}l}{\textit{Arabic linguistics}} \\
\quad \Irab{} score    & 64.08 & $+0.46$ & $-0.06$ & 64.63 & $+10.16^{\dagger}$ & $+9.93^{\dagger}$ \\
\quad Tashkeel DER $\downarrow$       & 9.17 & $-1.01^{\dagger}$ & $+0.59^{\dagger}$ & 13.34 & $-6.10^{\dagger}$ & $-3.86^{\dagger}$ \\
\quad Tashkeel WER $\downarrow$       & 13.92 & $-0.77^{\dagger}$ & $+0.54^{\dagger}$ & 19.65 & $-6.03^{\dagger}$ & $-4.28^{\dagger}$ \\
\quad Tashkeel exact accuracy         & 45.37 & $-1.23^{\dagger}$ & $+0.53$ & 30.87 & $+4.23^{\dagger}$ & $+6.60^{\dagger}$ \\
\quad Base-text alteration $\downarrow$ & 5.40 & $-1.03^{\dagger}$ & $+0.17$ & 8.70 & $-5.30^{\dagger}$ & $-3.30^{\dagger}$ \\
\bottomrule
\end{tabular}
\caption{Trained skills: FTPO score and movement from it at each arm's final checkpoint
(200 batches), three-seed means. $^{\dagger}$: movement larger than $1.63\,\sigma$
(Section~\ref{sec:design}). For $\downarrow$ metrics a negative movement is an
improvement. English benchmarks measure cross-lingual transfer. Tashkeel metrics are in
percent; DER is the diacritic error rate.}
\label{tab:target}
\end{table}

\subsubsection{Trained Skills}
\label{sec:target}

\paragraph{Instruction following is learned, transfers, and survives later stages.} Both
arms gain about five points on IFEval-Ar in high reasoning (Table~\ref{tab:target}).
They also improve English IFEval ($+7.89$ mixed, $+5.98$ cascade), so the skill
transfers across languages. The cascade reached $+5.45$ (Arabic) and $+5.48$ (English) by the end of its first
stage and still holds them 532 updates later. Only in low
reasoning on English IFEval does the cascade fail to improve ($-1.36$ against $+2.77$).

\paragraph{Mathematics and code.} In high reasoning the cascade ends $10.29$ and $11.06$
points below FTPO on the two GSM8K variants, $6.07$ on MATH-500 and $6.45$ on Arabic
GSM8K. The mixed arm is flat or positive on all four in both modes. Both arms improve
AIME, the mixed arm far more ($+25.42$ and $+23.89$ in high reasoning, against $+6.67$
and $+12.64$). Code shows a smaller version of the split. HumanEval is flat to slightly
positive for both arms, while on MBPP the mixed arm gains $4.00$ in high reasoning and
the cascade ends below FTPO in both modes.

\begin{table}[!t]
\centering
\scriptsize
\setlength{\tabcolsep}{3pt}
\begin{minipage}[t]{0.49\linewidth}\centering\resizebox{\linewidth}{!}{%
\begin{tabular}[t]{@{}lrrrrrr@{}}
\toprule
& \multicolumn{3}{c}{\textbf{Low}} & \multicolumn{3}{c}{\textbf{High}} \\
\cmidrule(lr){2-4}\cmidrule(l){5-7}
\textbf{Benchmark} & FTPO & $\Delta$M & $\Delta$C & FTPO & $\Delta$M & $\Delta$C \\
\midrule
\multicolumn{7}{@{}l}{\textit{Arabic knowledge and culture}} \\
ArabicMMLU & 72.61 & $+1.01$ & $-0.18$ & 77.15 & $+0.71$ & $-0.81$ \\
ArabicMMLU (diacr.) & 70.01 & $+2.58$ & $+2.13$ & 73.25 & $+1.94$ & $-0.21$ \\
MMLU Ar.\ dialects & 60.93 & $+0.07$ & $-0.74$ & 75.43 & $+0.57$ & $-1.19$ \\
3LM & 96.55 & $+0.86$ & $-0.05$ & 97.92 & $+0.82$ & $-0.08$ \\
Belebele (ar) & 77.50 & $-0.75$ & $-0.57$ & 87.48 & $+0.77$ & $+0.36$ \\
ArabCulture & 80.69 & $+1.31$ & $+1.47$ & 81.58 & $+0.93$ & $+0.78$ \\
AraDiCE-Culture & 74.07 & $-3.15$ & $+2.04$ & 79.81 & $-1.30$ & $-1.48$ \\
Alyah (Emirati) & 79.65 & $-0.20$ & $-0.45$ & 77.52 & $+0.91$ & $+0.26$ \\
\multicolumn{7}{@{}l}{\textit{Arabic professional domains}} \\
MedAraBench & 45.23 & $-0.14$ & $+0.35$ & 58.54 & $+1.70$ & $-1.48$ \\
Arabic Finance & 44.56 & $+5.20$ & $+1.30$ & 57.09 & $+2.25$ & $+0.95$ \\
Arabic Legal & 61.92 & $+0.42$ & $-1.58$ & 64.50 & $+1.17$ & $-1.08$ \\
\multicolumn{7}{@{}l}{\textit{Transliteration (0--5 $\times 20$)}} \\
Egyptian & 81.05 & $+3.49$ & $+1.65$ & 78.49 & $-2.71$ & $+1.28$ \\
Darija & 79.04 & $+2.89$ & $-0.33$ & 75.97 & $+5.86$ & $+5.74$ \\
\bottomrule
\end{tabular}}
\end{minipage}\hfill
\begin{minipage}[t]{0.49\linewidth}\centering\resizebox{\linewidth}{!}{%
\begin{tabular}[t]{@{}lrrrrrr@{}}
\toprule
& \multicolumn{3}{c}{\textbf{Low}} & \multicolumn{3}{c}{\textbf{High}} \\
\cmidrule(lr){2-4}\cmidrule(l){5-7}
\textbf{Benchmark} & FTPO & $\Delta$M & $\Delta$C & FTPO & $\Delta$M & $\Delta$C \\
\midrule
\multicolumn{7}{@{}l}{\textit{English knowledge and reasoning}} \\
MMLU & 76.53 & $-0.35$ & $+0.76$ & 82.06 & $+0.60$ & $-1.82$ \\
MMLU-Pro & 72.87 & $+1.92$ & $-1.08$ & 75.78 & $+0.80$ & $-3.49$ \\
GPQA-Diamond & 58.59 & $+7.58$ & $-0.34$ & 69.02 & $+3.54$ & $+1.35$ \\
BBEH & 26.17 & $+5.13$ & $+2.63$ & 28.77 & $+3.02$ & $-1.25$ \\
AA-LCR & 22.00 & $+6.00$ & $+3.00$ & 59.33 & $+9.00$ & $+3.67$ \\
\multicolumn{7}{@{}l}{\textit{Factuality}} \\
SimpleQA (F) & 5.59 & $+0.13$ & $+1.43$ & 2.48 & $-0.54$ & $+0.79$ \\
AA-Omni.\ accuracy & 12.28 & $+0.17$ & $+1.11$ & 12.78 & $-1.06$ & $+0.17$ \\
AA-Omni.\ halluc.$^{\ast}$ & 63.39 & $+1.33$ & $+2.01$ & 21.59 & $+4.53$ & $+1.75$ \\
\multicolumn{7}{@{}l}{\textit{Safety}} \\
HarmBench standard & 99.17 & $-1.50$ & $-2.33$ & 99.67 & $-0.33$ & $-0.67$ \\
HarmBench contextual & 82.67 & $-6.00$ & $-4.67$ & 93.33 & $-3.67$ & $-6.67$ \\
HarmBench copyright & 92.67 & $-5.33$ & $-3.33$ & 100.00 & $-0.33$ & $\pm0.00$ \\
XSTest & 89.70 & $+0.30$ & $+0.44$ & 91.63 & $+1.63$ & $+1.93$ \\
\multicolumn{7}{c}{} \\
\bottomrule
\end{tabular}}
\end{minipage}
\caption{Untrained benchmarks: FTPO score and movement for the mixed ($\Delta$M) and
cascade ($\Delta$C) arms. No $^{\dagger}$ marks: for most rows movements under about
$1.6$ points are within noise (about $4$ for GPQA-Diamond and AA-LCR). SimpleQA is near
its floor and high-reasoning HarmBench copyright at its ceiling. $^{\ast}$Raw rate;
movements sign-flipped so positive means fewer hallucinations. MMLU~\citep{hendrycks2021mmlu}, BBEH~\citep{kazemi2025bbeh}, 3LM~\citep{boussaha-etal-2025-3lm}, Belebele~\citep{bandarkar2024belebele}, and Arabic Finance~\citep{mai2026helmarabicenterprise} are used only in this analysis. HarmBench copyright is scored here by refusal rate; Section~\ref{sec:safety} omits this subset because its official verbatim-matching protocol needs reference texts.}
\label{tab:retained}
\end{table}

\paragraph{Arabic linguistics.} On \irab{} the arms tie: both gain about ten points in
high reasoning (difference $0.23$, below the $1.01$ floor) and nothing in low reasoning.
On tashkeel the mixed arm makes fewer errors: in high reasoning its DER is $2.24$ points
and its WER $1.75$ points lower than the cascade's. The cascade instead diacritizes more
passages exactly ($+2.37$ in high reasoning, $+1.77$ in low). All Mixed $-$ Cascade
tashkeel differences exceed the floor. Base-text alteration largely reconciles the two.
The cascade alters $5.40\%$ of passages in high reasoning against $3.40\%$, and because
an altered passage counts every letter as an error, those two extra points add roughly
two points of DER. This suggests the cascade vocalizes the passages it reproduces at
least as accurately, though we have not computed DER on unaltered passages alone. In low reasoning the
cascade ends slightly worse than FTPO (DER $+0.59$, WER $+0.54$), whereas the mixed arm
improves.

\subsubsection{Untrained Skills}
\label{sec:retained}

\paragraph{Neither recipe forgets broadly, but the mixed arm retains slightly better.}
On the 25 untrained benchmarks (Table~\ref{tab:retained}; positive is better), median
movement is under one point for both arms in both modes, and no benchmark loses more than
seven points. In high reasoning the mixed arm's mean movement is $+1.23$
against $-0.05$, and it ends higher on 20 of 25 benchmarks (sign test $p = 0.004$, an
overstatement given correlated benchmarks); in low reasoning the split is 14 to 11. The
largest high-reasoning gaps favor the mixed arm on AA-LCR ($5.33$), MMLU-Pro ($4.30$),
BBEH ($4.26$), MedAraBench ($3.19$) and HarmBench contextual ($3.00$), and the cascade on
Egyptian transliteration ($3.99$). The cascade ends below FTPO on MMLU-Pro, MMLU and BBEH
in high reasoning, whereas the mixed arm improves on all five English knowledge and
reasoning benchmarks.

\paragraph{Both recipes reduce contextual refusal.} HarmBench contextual refusal falls in
both arms, beyond the floor at least in high reasoning: mixed $-6.00$ and $-3.67$,
cascade $-4.67$ and $-6.67$ (low, high). Low-reasoning copyright refusal also falls
($-5.33$, $-3.33$). Since both schedules show it, this is likely a cost of the training
mix, and the untrained-skill result most worth addressing.

\subsubsection{Cascade Trajectory and Response to Reasoning Mode}
\label{sec:trajectory}

\definecolor{gLight}{HTML}{6CC658}
\begin{table}[!t]
\centering
\scriptsize
\setlength{\tabcolsep}{5pt}
\newcommand{\own}{\cellcolor{gLight!20}}
\begin{tabular}{@{}lrrrrrrr@{}}
\toprule
& \textbf{IF} & \textbf{Tashkeel} & \textbf{\Irab} & \textbf{Math} & \textbf{Code} & & \textit{of which} \\
\textbf{Benchmark} & 0$\to$67 & 67$\to$89 & 89$\to$134 & 134$\to$178 & 178$\to$200 & \textbf{Total} & 0$\to$30 \\
\midrule
\multicolumn{8}{@{}l}{\textit{Trained skills}} \\
IFEval-Ar          & \own$+5.45^{\dagger}$ & $-0.83$ & $+1.32$ & $-1.24$ & $+0.58$ & $+5.28$ & $+4.79$ \\
IFEval (en)        & \own$+5.48^{\dagger}$ & $+0.25$ & $+0.92$ & $-0.18$ & $-0.49$ & $+5.98$ & $+4.68$ \\
Tashkeel DER $\downarrow$ & $+0.01$ & \own$-2.91^{\dagger}$ & $+2.71^{\dagger}$ & $-4.62^{\dagger}$ & $+0.94$ & $-3.86$ & $+0.05$ \\
Tashkeel exact acc. & $-6.17^{\dagger}$ & \own$+8.90^{\dagger}$ & $-1.10$ & $+6.30^{\dagger}$ & $-1.33$ & $+6.60$ & $-2.47$ \\
Base-text alt. $\downarrow$ & $-0.20$ & \own$-1.93^{\dagger}$ & $+2.93^{\dagger}$ & $-5.05^{\dagger}$ & $+0.95$ & $-3.30$ & $-0.87$ \\
\Irab{}            & $+3.61^{\dagger}$ & $+0.79$ & \own$+4.02^{\dagger}$ & $+1.95^{\dagger}$ & $-0.44$ & $+9.93$ & $+0.19$ \\
Arabic GSM8K       & $-7.64^{\dagger}$ & $+0.51$ & $+0.51$ & \own$-0.69$ & $+0.87$ & $-6.45$ & $-7.15$ \\
GSM8K (0-shot)     & $-12.04^{\dagger}$ & $+0.20$ & $-0.75$ & \own$+1.54^{\dagger}$ & $+0.76$ & $-10.29$ & $-12.95$ \\
MATH-500            & $-7.67^{\dagger}$ & $-0.13$ & $-1.67$ & \own$+3.00^{\dagger}$ & $+0.40$ & $-6.07$ & $-8.13$ \\
AIME 2025          & $-5.69^{\dagger}$ & $-3.19$ & $+1.94$ & \own$+13.06^{\dagger}$ & $+0.56$ & $+6.67$ & $-8.89$ \\
MBPP               & $-4.27^{\dagger}$ & $+0.60$ & $+0.53$ & $-0.87$ & \own$+2.40^{\dagger}$ & $-1.60$ & $-4.67$ \\
HumanEval          & $-1.42$ & $+0.81$ & $\pm0.00$ & $-0.20$ & \own$+3.46^{\dagger}$ & $+2.64$ & $-0.41$ \\
\multicolumn{8}{@{}l}{\textit{Untrained skills}} \\
MMLU-Pro           & $-4.13^{\dagger}$ & $+0.22$ & $+0.04$ & $+0.44$ & $-0.07$ & $-3.49$ & $-4.58$ \\
BBEH               & $-2.74^{\dagger}$ & $+0.54^{\dagger}$ & $+0.21$ & $+0.11$ & $+0.63^{\dagger}$ & $-1.25$ & $-2.77$ \\
HarmBench context. & $-3.33^{\dagger}$ & $+1.00$ & $-2.00$ & $-0.67$ & $-1.67$ & $-6.67$ & $-2.67$ \\
Darija translit.   & $-0.14$ & $+0.05$ & $+3.33^{\dagger}$ & $+7.01^{\dagger}$ & $-4.51^{\dagger}$ & $+5.74$ & $+2.36$ \\
\bottomrule
\end{tabular}
\caption{Movement across each cascade stage, high reasoning, three-seed means (step 0 is
FTPO). The last column is the first 30 batches of instruction following. Shaded cells
pair each stage with its own targets. Tashkeel at steps 134 and 178 averages two seeds
($^{\dagger}$ at $1.83\,\sigma$). GPQA-Diamond is omitted (noise floor $4.0$ points).}
\label{tab:stages}
\end{table}
\begin{figure}[!t]
\centering
\begin{tikzpicture}
\begin{axis}[
  name=gen, width=0.465\linewidth, height=4.0cm,
  xmin=-4, xmax=226, ymin=-17, ymax=28,
  title={\scriptsize (a) General suite}, legend to name=leg:gen,
  xlabel={Cascade training step (batches)},
  ylabel={Movement from FTPO (points)},
  xtick={0,67,89,134,178,200},
  tick label style={font=\scriptsize}, label style={font=\scriptsize},
  grid=major, grid style={gray!18},
  legend columns=3,
  legend style={font=\tiny, draw=none, fill=none,
     /tikz/every even column/.append style={column sep=4pt}},
  every axis plot/.append style={thick, mark size=1.2pt},
]
\draw[gray!60, dashed] (axis cs:67,-17) -- (axis cs:67,28);
\draw[gray!60, dashed] (axis cs:89,-17) -- (axis cs:89,28);
\draw[gray!60, dashed] (axis cs:134,-17) -- (axis cs:134,28);
\draw[gray!60, dashed] (axis cs:178,-17) -- (axis cs:178,28);
\fill[gray!12] (axis cs:207,-100) rectangle (axis cs:226,100);
\draw[gray!80] (axis cs:-4,0) -- (axis cs:226,0);
\node[font=\tiny, gray] at (axis cs:33,26) {IF};
\node[font=\tiny, gray] at (axis cs:78,26) {Tash.};
\node[font=\tiny, gray] at (axis cs:111,26) {\Irab};
\node[font=\tiny, gray] at (axis cs:156,26) {Math};
\node[font=\tiny, gray] at (axis cs:189,26) {Code};
\addplot[blue!70!black, mark=*] coordinates {
 (0,0.00) (30,-12.95) (55,-12.23) (67,-12.04) (75,-12.12) (89,-11.84) (100,-11.76)
 (119,-11.26) (125,-11.69) (134,-12.59) (175,-11.28) (178,-11.04) (190,-11.18) (200,-10.29)};
\addlegendentry{GSM8K (en)}
\addplot[cyan!60!black, mark=o] coordinates {
 (0,0.00) (30,-7.15) (55,-7.21) (67,-7.64) (75,-7.37) (89,-7.13) (100,-6.60)
 (119,-7.46) (125,-7.35) (134,-6.63) (175,-6.27) (178,-7.32) (190,-6.61) (200,-6.45)};
\addlegendentry{Arabic GSM8K}
\addplot[red!70!black, mark=square*] coordinates {
 (0,0.00) (30,-8.13) (55,-8.20) (67,-7.67) (75,-7.87) (89,-7.80) (100,-8.93)
 (119,-6.73) (125,-6.73) (134,-9.47) (175,-6.33) (178,-6.47) (190,-5.00) (200,-6.07)};
\addlegendentry{MATH-500}
\addplot[orange!85!black, mark=triangle*] coordinates {
 (0,0.00) (30,-8.89) (55,-6.81) (67,-5.69) (75,-3.89) (89,-8.89) (100,-5.56)
 (119,-5.97) (125,-6.94) (134,-6.94) (175,7.64) (178,6.11) (190,9.31) (200,6.67)};
\addlegendentry{AIME 2025}
\addplot[green!50!black, mark=diamond*] coordinates {
 (0,0.00) (30,4.68) (55,4.87) (67,5.48) (75,5.11) (89,5.73) (100,5.36)
 (119,5.79) (125,6.72) (134,6.65) (175,5.79) (178,6.47) (190,5.73) (200,5.98)};
\addlegendentry{IFEval (en)}
\node[rotate=90, font=\tiny\itshape, gray!50!black] at (axis cs:216,-9) {Mixed};
\addplot[only marks, mark=*, mark size=2.2pt, blue!70!black, forget plot] coordinates {(216,-0.14)};
\addplot[only marks, mark=o, mark size=2.2pt, cyan!60!black, forget plot] coordinates {(216,0.23)};
\addplot[only marks, mark=square*, mark size=2.2pt, red!70!black, forget plot] coordinates {(216,4.53)};
\addplot[only marks, mark=triangle*, mark size=2.6pt, orange!85!black, forget plot] coordinates {(216,25.42)};
\addplot[only marks, mark=diamond*, mark size=2.6pt, green!50!black, forget plot] coordinates {(216,7.89)};
\end{axis}
\begin{axis}[
  at={(gen.east)}, anchor=west, xshift=0.35cm,
  name=ar, width=0.465\linewidth, height=4.0cm,
  xmin=-4, xmax=226, ymin=-7, ymax=13,
  title={\scriptsize (b) Arabic linguistics}, legend to name=leg:ar,
  xlabel={Cascade training step (batches)},
  xtick={0,67,89,134,178,200},
  tick label style={font=\scriptsize}, label style={font=\scriptsize},
  grid=major, grid style={gray!18},
  legend columns=2,
  legend style={font=\tiny, draw=none, fill=none,
     /tikz/every even column/.append style={column sep=4pt}},
  every axis plot/.append style={thick, mark size=1.2pt},
]
\draw[gray!60, dashed] (axis cs:67,-7) -- (axis cs:67,13);
\draw[gray!60, dashed] (axis cs:89,-7) -- (axis cs:89,13);
\draw[gray!60, dashed] (axis cs:134,-7) -- (axis cs:134,13);
\draw[gray!60, dashed] (axis cs:178,-7) -- (axis cs:178,13);
\fill[gray!12] (axis cs:207,-100) rectangle (axis cs:226,100);
\draw[gray!80] (axis cs:-4,0) -- (axis cs:226,0);
\node[font=\tiny, gray] at (axis cs:33,12.2) {IF};
\node[font=\tiny, gray] at (axis cs:78,12.2) {Tash.};
\node[font=\tiny, gray] at (axis cs:111,12.2) {\Irab};
\node[font=\tiny, gray] at (axis cs:156,12.2) {Math};
\node[font=\tiny, gray] at (axis cs:189,12.2) {Code};
\addplot[violet!80!black, mark=*] coordinates {
 (0,0.00) (30,0.19) (55,2.48) (67,3.61) (75,4.12) (89,4.39) (100,9.76)
 (119,8.83) (125,8.53) (134,8.42) (175,10.05) (178,10.37) (190,10.13) (200,9.93)};
\addlegendentry{\Irab{} score}
\addplot[teal!80!black, mark=square*] coordinates {
 (0,0.00) (30,0.06) (55,-0.59) (67,0.02) (75,-0.70) (89,-2.90) (100,-3.48)
 (119,-2.51) (125,-2.05) (134,-0.18) (175,-4.07) (178,-4.80) (190,-4.55) (200,-3.86)};
\addlegendentry{Tashkeel DER $\downarrow$}
\addplot[brown!80!black, mark=triangle*, densely dotted] coordinates {
 (0,0.00) (30,-0.87) (55,-0.47) (67,-0.20) (75,-0.50) (89,-2.13) (100,-2.80)
 (119,-1.93) (125,-1.50) (134,0.80) (175,-3.83) (178,-4.25) (190,-4.07) (200,-3.30)};
\addlegendentry{Base-text alt.\ $\downarrow$}
\node[rotate=90, font=\tiny\itshape, gray!50!black] at (axis cs:216,3) {Mixed};
\addplot[only marks, mark=*, mark size=2.2pt, violet!80!black, forget plot] coordinates {(216,10.16)};
\addplot[only marks, mark=square*, mark size=2.2pt, teal!80!black, forget plot] coordinates {(216,-6.10)};
\addplot[only marks, mark=triangle*, mark size=2.6pt, brown!80!black, forget plot] coordinates {(216,-5.30)};
\end{axis}
\node[anchor=north, yshift=-2pt] at (gen.outer south) {\pgfplotslegendfromname{leg:gen}};
\node[anchor=north, yshift=-2pt] at (ar.outer south) {\pgfplotslegendfromname{leg:ar}};
\end{tikzpicture}
\caption{Cascade trajectory in high reasoning, movement from FTPO (three-seed means;
steps 134 and 178 average two tashkeel seeds, step 100 two \irab{} seeds). Dashed lines
mark stage boundaries; the shaded column on the right shows the mixed arm's final checkpoint (200 batches).
(a) Short-form mathematics drops within 30 batches, in English and Arabic, and the
mathematics stage repairs it only partly while lifting AIME by about 13 points.
(b) The tashkeel stage lowers DER, the \irab{} stage undoes this mainly through
base-text alteration, and the mathematics stage restores it. For $\downarrow$ metrics,
down is better.}
\label{fig:trajectory}
\end{figure}
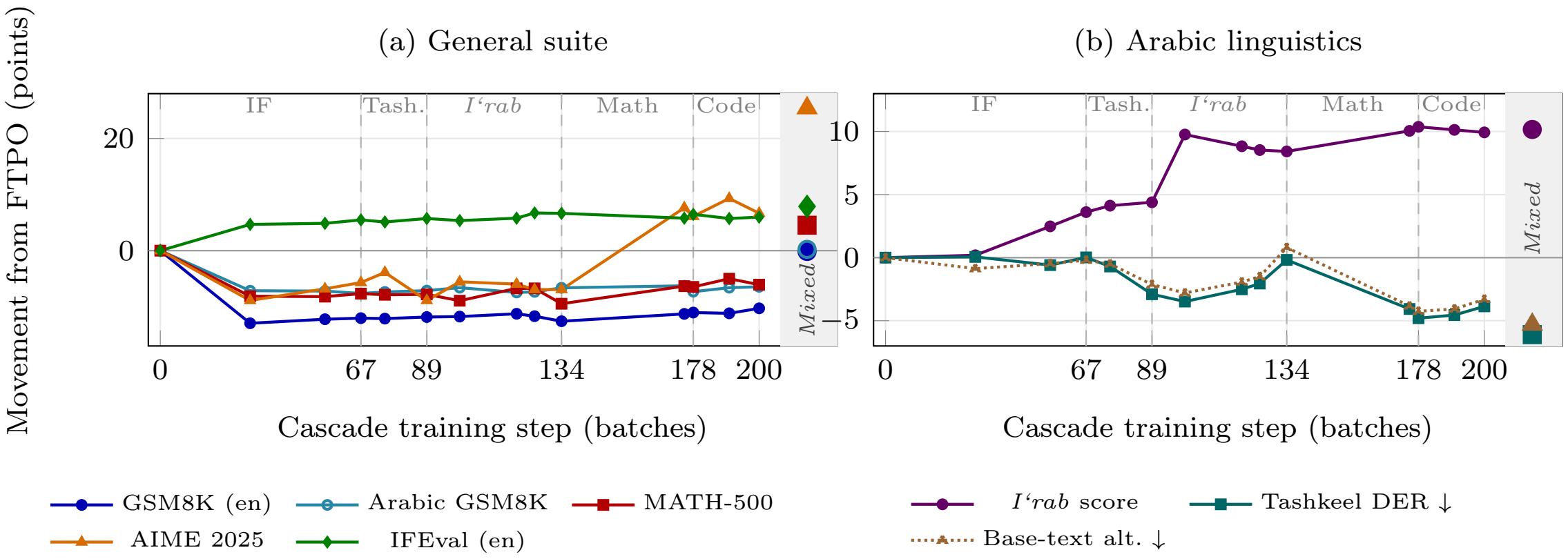

\paragraph{The short-form deficit appears within the first 120 updates.} By step 30, in
high reasoning, the cascade has lost $12.95$ points on GSM8K, $7.15$ on Arabic GSM8K and
$8.13$ on MATH-500 (Table~\ref{tab:stages}, Figure~\ref{fig:trajectory}a). It has also
lost $8.89$ on AIME 2025, $4.58$ on MMLU-Pro and $2.77$ on BBEH, while gaining $4.7$ to
$4.8$ on both IFEval variants. The Arabic stages leave these losses in place. The
mathematics stage, despite $+13$ on AIME, recovers only $1.54$ of the $12.04$ GSM8K
points and $3.00$ of the $7.67$ MATH-500 points, and none of the Arabic GSM8K loss. We
did not evaluate the mixed arm at intermediate checkpoints, so we cannot tell whether it
avoids this dip or incurs and repairs it.

The cascade's response to reasoning mode inverts at the same time. Within each model,
FTPO and the mixed arm gain from high reasoning on AIME (FTPO $+16.53$ and $+10.97$,
mixed $+34.17$ and $+24.86$) and are roughly neutral on GSM8K ($+0.43$, $-0.61$). The
final cascade still gains on AIME but loses $7.25$ and $8.10$ points on the GSM8K
variants and $2.36$ on Arabic GSM8K. Its GSM8K inversion is already present at step 30
($-6.55$, $-6.67$), before any mathematics data, and the short-form regression in
Table~\ref{tab:target} is largely its high-reasoning half.

\paragraph{Each Arabic stage moves its own target, and one undoes another.} The tashkeel
stage lowers DER by $2.91$ and raises exact accuracy by $8.90$. The \irab{} stage adds a
net $4.02$, all in its first 11 batches ($+5.37$) before a slight decline; \irab{} also
gains $3.61$ during instruction following and $1.95$ during mathematics. The one clear case of stage forgetting is tashkeel during the \irab{} stage
(Figure~\ref{fig:trajectory}b). DER climbs from $9.86$ at step 100 back to its pre-stage
level of $13.16$ at step 134, and alteration rises from $5.90\%$ to $9.50\%$. That rise
alone accounts for roughly the whole regression: the model mostly stops reproducing its
input rather than vocalizing it worse. Unexpectedly, the
mathematics stage then restores and surpasses the tashkeel gains (DER $8.54$, alteration
$4.45\%$). Low-reasoning DER moves by less than $0.6$ in every stage.

\paragraph{Language asymmetry (suggestive).} Over the instruction-following stage in high
reasoning, the five English untrained knowledge and reasoning benchmarks move by $-2.29$ on
average and the eleven Arabic knowledge, culture and professional benchmarks by $-0.28$;
GSM8K loses $12.04$ points against $7.64$ for Arabic GSM8K. Low reasoning shows no
asymmetry. This fits a policy pulled toward Arabic, but the English set is more
reasoning-heavy, the GSM8K variants start from different baselines ($97.3$, $94.3$), and
the noisy GPQA-Diamond enters the English mean, so language is a candidate explanation
only.

\paragraph{Interpretation.} The deficit is an early, unrepaired regression rather than
forgetting or overspecialization, and its mechanism is untested. Because it concentrates
in high reasoning on near-ceiling short-form tasks and coincides with the mode inversion,
changed output behavior (longer or non-terminating reasoning, answer-format drift,
reasoning in Arabic on English prompts) explains it as well as lost capability does.
Truncation, extraction-failure and language-identification rates at steps 0 and 30 would
separate these. The mixed arm may avoid the regression because mathematics and code
rewards are present from the first update.

\paragraph{On Arabic linguistics, RL makes high reasoning the better mode.} FTPO gains
almost nothing from high reasoning on \irab{} ($+0.55$) and is worse with it on every
tashkeel metric (DER $+4.17$, exact accuracy $-14.50$). After training both arms gain
about ten \irab{} points from high reasoning and reach DER parity or better (mixed
$-0.92$, cascade $-0.28$), mainly through less alteration, though exact accuracy still
favors low reasoning by eight to nine points. Two non-exclusive explanations fit: training
rollouts may mainly use the high-reasoning mode, and RL typically lowers policy entropy,
reducing the temperature-$0.7$ penalty that FTPO's weak high-mode tashkeel suggests.

\subsubsection{Summary and Limitations}
\label{sec:ablation-limitations}

Matched on origin, per-domain exposure and updates, the recipes largely agree on the
Arabic linguistic skills: both gain about ten \irab{} points, and the mixed arm makes
fewer tashkeel errors while the cascade diacritizes more passages exactly. Instruction
following transfers from Arabic to English and survives the cascade's later stages, and
neither recipe forgets untrained capabilities broadly. Both predictions for sequential
training find limited support: forgetting occurs once (tashkeel during the \irab{}
stage), and specialization pays only on tashkeel exact accuracy, the one trained metric
on which the cascade ends ahead beyond noise. What separates the recipes in this run is
an early regression in the instruction-following-first cascade: within 120 updates it
loses about 13 points on GSM8K, 7 on Arabic GSM8K and 8 on MATH-500 in high reasoning and
begins to score worse in high than in low reasoning on short-form problems. Later stages
repair this only partly, leaving it 6 to 11 points below its origin on short-form
mathematics in high reasoning, where the mixed arm holds or improves.

Five limitations qualify this. (i)~Each arm was trained once, and run-level variation
exceeds the evaluation floor (AIME 2025 swings by 5 points within the tashkeel stage). (ii)~We test a single cascade order, with instruction following
first and longest. (iii)~The mixed arm was evaluated only at its endpoint. (iv)~The
mechanisms behind the regression and the tashkeel gap are inferred, not measured.
(v)~Reasoning mode bundles budget with temperature. We therefore read the results as
evidence that interleaving is the safer default for Arabic-centric multi-skill RLVR at
this scale, not as a general verdict on sequential training.

\section{Evaluation}\label{sec:eval}

\begin{takeaways}
    \item \textbf{Arabic leadership, with the biggest margins at small sizes.} \modelname leads or matches the best open models on Arabic at every size, and the 375B competes with frontier closed models.

    \item \textbf{Best-in-class agentic ability for its size.} At 3.7B and 7B, \modelname beats every open model of its size on all ten agentic benchmarks.

    \item \textbf{A rebuilt Arabic evaluation methodology.} Widely used Arabic benchmarks were audited, and their defective items removed, before evaluation.
  
    \item \textbf{New benchmarks.} To fill gaps in Arabic evaluation, we introduce six benchmarks: \benchmark{AraIFBench}, the I'rab and Tashkeel benchmarks of \araLingBmarks, \benchmark{AA-LCR-Ar}, \benchmark{SimpleQA-Verified-Ar}, and the agentic \benchmark{Arabic File-Processing Benchmark}.
    
\end{takeaways}

\label{subsec:eval-overview}
We evaluate \modelname in three capability areas: Arabic (Section~\ref{subsec:eval-arabic}), agentic tasks (Section~\ref{subsec:agentic-eval}), and English (Section~\ref{subsec:eval-english}). Arabic and agentic tasks are the two capabilities \modelname is built for, while English tests whether it remains competitive with general-purpose models in the language where they are strongest. We then examine factuality and hallucination (Section~\ref{subsec:eval-factuality}), where \modelname behaves differently from most open models, and close with safety in Arabic and English (Section~\ref{sec:safety}).

At every size, \modelname is compared with open-weight models of similar size, including earlier Arabic-centric models such as \model{Jais~2}, \model{ALLaM}, and \model{Fanar}; the 375B model is also compared with frontier closed models (\model{Claude Sonnet 5}, \model{Gemini 3.8 Flash}, and \model{GPT-5.6 Sol}). All models run in thinking mode (high effort) where it is available; Appendix~\ref{app:low-reasonig} reports results with low-effort thinking. For Arabic, we revisit how models are assessed: all benchmarks are evaluated generatively, widely used benchmarks are audited and their defective items removed, and new benchmarks cover Arabic instruction following (AraIFBench), grammatical analysis and diacritization (\araLingBmarks), long-context reasoning (AA-LCR-Ar), and factuality (SimpleQA-Verified-Ar). The agentic evaluation, run from the 3.7B model upward, uses eleven benchmarks in coding, tool use, professional work, and deep search; two are in Arabic, including the Arabic File-Processing Benchmark, which we introduce for professional work on Arabic office files. The English evaluation uses standard benchmarks for knowledge, STEM, instruction following, coding, long context, and multi-turn interaction.

Across the three capability areas, the lead of \modelname is largest at small sizes. On Arabic, it leads or matches the best open models at every size, and the 375B model is competitive with \model{Claude Sonnet 5}, although a gap remains to the strongest frontier models. On agentic tasks, the 3.7B and 7B models outperform every open model of their size on all ten benchmarks on which they are scored, while earlier Arabic-centric models fail almost every task; at 32B and above, the \model{Qwen3.8} models and the frontier closed models lead most benchmarks. In English, \modelname is strongest on mathematical reasoning, leading the open models on AIME~2025 and~2026 at every size from 4B to 375B; knowledge-intensive benchmarks such as GPQA-Diamond and MMLU-Pro remain its main gap.

The most distinctive behavior of \modelname appears in factuality: with high-effort thinking, when it cannot answer a question correctly, it usually abstains instead of guessing. On SimpleQA-Verified, its non-hallucination rate, the share of non-correct responses on which it abstains, is above 68\% in Arabic and above 80\% in English at every size, compared with below 19\% for most open baselines and below 2\% for \model{Gemini 3.8 Flash} and \model{GPT-5.6 Sol}. The cost is a lower F-score than the strongest baselines of each size, which attempt more questions; with low-effort thinking, \modelname rarely abstains. On safety, every \modelname model refuses more than 92\% of the harmful prompts in AraSafe, 3 to 5 points more than \model{Jais~2}, and five of the six models exceed 98\% on topics that are sensitive in the Arab region. In English, \modelname refuses far more HarmBench requests than \model{Jais~2}, while staying within 2 points of the best model of its size on XSTest, which also checks that benign prompts are not refused.

\subsection{Arabic Evaluation}
\label{subsec:eval-arabic}

As Arabic is the primary target language of \modelname, it receives the broadest coverage in our evaluation. We evaluate it with a reproducible evaluation suite\footnote{\url{https://github.com/ifm-ai/gen-eval-arabic}} that covers general capabilities, Islamic and cultural knowledge, dialect understanding and generation, multi-turn and long-context understanding, and language-specific tasks such as grammatical analysis (i'rab, \<إعراب>) and diacritization (tashkeel, \<تشكيل>). Many existing Arabic benchmarks and leaderboards, however, were designed for an earlier generation of language models: their protocols often score multiple-choice questions by the log-probability of each option, or match answers exactly against a predefined output format. Neither suits reasoning models, which produce a reasoning trace before the final answer and may express a correct answer in a different format. Our suite therefore has every model generate a free-form answer, and scores most benchmarks with an LLM judge (Section~\ref{subsubsec:eval-ar-setup}).

Beyond the protocol, we revised the Arabic evaluation in two respects. For quality, we audited the selected Arabic benchmarks with an LLM-based pipeline and removed the items flagged for removal (Section~\ref{subsubsec:eval-ar-audit}). For coverage, we built new benchmarks for domains where Arabic evaluation is missing or underdeveloped, such as instruction following, Arabic grammar (i'rab), and factuality and hallucination (Section~\ref{subsubsec:eval-new-bench}).

We report results in four groups: general capabilities in Arabic, Arabic-specific domain knowledge, dialect understanding and generation, and multi-turn and long-context tasks. We do not treat Arabic as a single language condition: the evaluation spans Modern Standard Arabic (MSA), diacritized Arabic, and dialects including Egyptian, Gulf (Saudi, Emirati), Levantine (Syrian) and Maghrebi (Moroccan). Scores on the audited benchmarks exclude the items removed by the audit.

\subsubsection{Arabic Benchmark Auditing}
\label{subsubsec:eval-ar-audit}

Errors and ambiguities are well documented in widely used evaluation benchmarks. \citet{northcutt2021pervasive} estimate an average label-error rate of at least $3.3\%$ across ten standard test sets, while \citet{gema-etal-2025-done} estimate that $6.49\%$ of \benchmark{MMLU} questions contain errors and show that accounting for them can affect model rankings in the most impacted subjects. More broadly, benchmark auditing and revision after release have become increasingly common. \benchmark{Terminal-Bench 2.1} revises 28 of the 89 tasks from version 2.0~\citep{tbench2026v21}; \benchmark{SWE-bench Verified} was constructed through additional screening of the original test set by professional developers~\citep{openai2024swebenchverified}; and \benchmark{Humanity's Last Exam} has undergone further verification by domain experts~\citep{zhai2026hleverified}. In Arabic, \benchmark{ArabicMMLU}~\citep{koto-etal-2024-arabicmmlu} notes that incorrect answer keys limit the practically achievable score to approximately $96\%$. %
QIMMA~\citep{alqadi2026qimma} audited fourteen established benchmarks using LLM-based screening followed by review from native speakers, filtering up to $3.1\%$ of the items in individual benchmarks, with incorrect or missing gold answers among the most consequential issues.

Motivated by these observations, we audited the Arabic benchmarks used in this report with a multi-stage pipeline. Reviewing every item with several frontier models is costly, so the pipeline first selects candidates. For each benchmark, we take the strongest baseline models and collect every item that at least one of them answers incorrectly. A failure by a strong model may come from the model, from the item, or from both, so it is a reason to review the item rather than evidence of a defect. Taking the union over several models makes the selection less dependent on any one of them: an item that one model answers correctly by chance is still reviewed if another model fails it. Defects in items that all selected models answer correctly, however, are not detected.

Each candidate is then reviewed independently by three frontier models, \model{Gemini 3.8 Flash}, \model{Claude Opus 5}, and \model{GPT-5.6 Sol}, which see the question, options, and gold answer exactly as they are used in evaluation. The judges follow a shared rubric with domain-specific guidance and decide whether the item allows reliable answering and scoring. Each judge locates the issue in the question, the options, or the gold answer, assigns a subcategory, such as an incorrect gold answer, multiple correct options, or an ambiguous question, and gives a reason and a suggested fix. It also rates the issue as solvable, when a clearly correct answer remains consistent with the gold answer, or significant, when the issue prevents reliable answering or scoring. An item is removed only when all three judges find a significant issue, although they need not agree on its category or fix. Because correctness in Arabic can depend on dialectal usage, regional and cultural knowledge, and linguistic conventions, Arabic-speaking annotators also reviewed a random sample of the removed items.

\begin{table}[!htbp]
\centering
\footnotesize
\renewcommand{\arraystretch}{1.05}
\setlength{\tabcolsep}{5pt}
\begin{tabular}{@{}lrrrrrr@{}}
\toprule
 & & & & \multicolumn{3}{c}{\textbf{Issue source (\%)}} \\
\cmidrule(l){5-7}
\textbf{Benchmark} & \textbf{Items} & \textbf{Removed} & \textbf{Rate (\%)} & \textbf{Question} & \textbf{Options} & \textbf{Gold} \\
\midrule
MedAraBench & 4{,}927 & 896 & 18.2 & 21 & 27 & 52 \\
QIAS-Inheritance & 1{,}000 & 102 & 10.2 & 3 & 93 & 4 \\
IslamicEval 2025 & 247 & 16 & 6.5 & 29 & 15 & 56 \\
Dialectal-Arabic-MMLU & 15{,}675 & 935 & 6.0 & 24 & 28 & 48 \\
Legal (HELM) & 200 & 11 & 5.5 & 46 & -- & 54 \\
ArabicMMLU & 14{,}455 & 765 & 5.3 & 13 & 22 & 65 \\
PalmX 2025 & 985 & 37 & 3.8 & 31 & 38 & 32 \\
FannOrFlop\_MCQ & 6{,}984 & 230 & 3.3 & 9 & 42 & 49 \\
AraSTEM & 10{,}824 & 330 & 3.0 & 25 & 25 & 49 \\
GAT Math & 7{,}017 & 199 & 2.8 & 25 & 28 & 47 \\
ArabCulture & 3{,}482 & 73 & 2.1 & 42 & 32 & 26 \\
QIAS-Knowledge & 700 & 11 & 1.6 & 30 & 30 & 39 \\
Alyah & 1{,}173 & 15 & 1.3 & 11 & 56 & 33 \\
Arabic GSM8K & 1{,}319 & 13 & 1.0 & 51 & -- & 49 \\
AraDiCE-Culture & 180 & 1 & 0.6 & 0 & 100 & 0 \\
\midrule
Total & 69{,}168 & 3{,}634 & 5.3 & 20 & 29 & 51 \\
\bottomrule
\end{tabular}
\caption{Items removed by the benchmark audit. \emph{Issue source} is the share of removed items whose issue the judges place in the question, the answer options, or the gold answer, averaged over the three judges; rows may not sum to 100 because of rounding. Arabic GSM8K and Legal (HELM) are open-ended and have no options. Dialectal-Arabic-MMLU covers its five dialect versions, and Legal (HELM) was audited both with and without its context passage. Belebele Arabic was also audited but is not part of this evaluation.}
\label{tab:audit-removal}
\end{table}

Table~\ref{tab:audit-removal} summarizes the outcome for the 15 benchmark sets used in this report. The audit removes 3,634 of 69,168 items (5.3\%), from 0.6\% of AraDiCE-Culture to 18.2\% of MedAraBench. The gold answer is the most common source of issues: for about half of the removed items, at least two judges classify the issue as an incorrect gold answer. The next most common issues are multiple correct options (12\%), no correct option (9\%), and missing context that the question needs (7\%). The profile differs across benchmarks. Almost all removed QIAS-Inheritance items have faulty options, mostly several options that are correct or overlap; in ArabCulture, the most common issue is an ambiguous or underspecified question; and 12\% of the removed MedAraBench items are malformed or corrupted.

Translated benchmarks are particularly exposed, because a translation can change the correct answer while the gold answer stays the same. \href{https://hf.co/datasets/IFM/discarded_samples/viewer/Arabic\%20GSM8K?row=9}{Item~546} of \benchmark{Arabic GSM8K} asks how many letters a child writes when she copies the alphabet several times. The gold answer, 130, assumes the 26-letter English alphabet, whereas a reader of the Arabic question would reasonably use the 28-letter Arabic alphabet and answer 140. Other defects are plain errors: in item~454, the question says that two people eat four apples a day in total, but the gold solution adds a fifth apple and answers 150 instead of 120.

Removing such items changes scores noticeably. For example, the \benchmark{MedAraBench} score of \model{Jais-3-375B-A23B} rises by 13.72 points (Table~\ref{tab:benchmark_auditing_delta}), and Appendix~\ref{app:arabic-scores-auditing} reports the effect for every model and benchmark. All scores on the audited benchmarks in this report exclude the removed items.

We release the removed items, with the verdicts, reasons, and suggested fixes of all three judges, in \texttt{IFM/discarded\_samples}\footnote{\url{https://hf.co/datasets/IFM/discarded_samples}}, and the cleaned benchmarks in the \texttt{IFM/Arabic-Benchmark-Auditing}\footnote{\url{https://hf.co/collections/IFM/arabic-benchmark-auditing}} collection. For five benchmarks whose sources are gated or state no license (PalmX, Alyah, AraSTEM, MedAraBench, and QIAS), the released records keep the item identifiers, gold answers, and verdicts but omit the questions and options.

\subsubsection{New Arabic Benchmarks}
\label{subsubsec:eval-new-bench}

Building evaluation suites for Arabic-centric LLMs requires continuous effort in two complementary directions: addressing gaps in existing evaluations and introducing increasingly challenging benchmarks as model capabilities advance. During the development of \modelname, we identified several important gaps in Arabic LLM evaluation, including insufficient coverage of instruction-following behaviors tailored to Arabic, limited assessment of Arabic-specific linguistic capabilities, and a lack of reliable resources for evaluating factuality and hallucination in Arabic. To address these gaps, we introduce new benchmarks that target these underexplored capabilities and provide a more comprehensive assessment of Arabic LLMs.

First, existing instruction-following benchmarks often overlook the linguistic characteristics and language-specific requirements that arise when following instructions in Arabic. To address this limitation, we introduce \benchmark{AraIFBench}, an Arabic-centric instruction-following benchmark designed to evaluate how reliably models follow instructions tailored to the Arabic language. In addition, evaluating Arabic proficiency requires going beyond general language understanding to assess core linguistic skills. We therefore introduce two benchmarks, Tashkeel and I'rab, targeting Arabic diacritization and grammatical analysis, respectively. Together, these benchmarks provide a more targeted evaluation of models' ability to handle fundamental aspects of Arabic orthography and grammar that are not adequately captured by general-purpose evaluations.

Second, assessing factual reliability in Arabic requires distinguishing correct responses from plausible but unsupported claims. To support this evaluation, we introduce \benchmark{SimpleQA-Verified-Ar}, an Arabic factuality benchmark that measures whether models provide accurate answers or hallucinate when uncertain. This complements our instruction-following and linguistic evaluations with a direct measure of factual reliability.

Third, long-context evaluation remains largely English-centric, although Arabic text is often split into more tokens, which makes long inputs more demanding. We therefore introduce \benchmark{AA-LCR-Ar}, a full Arabic translation of \benchmark{AA-LCR} v1.1 that requires reasoning across sets of real-world documents of about 100K tokens.

\paragraph{AraIFBench.}
\label{par:araifbench}\label{subsubsec:araifbench}
Existing Arabic instruction-following benchmarks largely translate \benchmark{IFEval}~\citep{Zhou2023Instruction,ojewale2026clifeval} or adapt similar constraints~\citep{elfilali2025arabicifeval,althnian-etal-2025-araeval}, while the multilingual \benchmark{XIFBench}~\citep{li2025xifbench} excludes language-dependent constraints. As a result, they rarely evaluate Arabic-specific requirements involving punctuation, digits, diacritics, and morphology.

Following \benchmark{IFBench}~\citep{pyatkin2025ifbench}, \benchmark{AraIFBench} combines real Arabic user requests from WildChat~\citep{zhao2024wildchat} with verifiable constraints. It contains 1,000 tasks with 1,140 constraints drawn from 77 families in nine categories (Table~\ref{tab:araifbench-categories}), including 35 Arabic-specific families. Each task contains one or two constraints, and some requests are dialectal, although dialect is not explicitly controlled.

\begin{table}[t]
  \centering
  \footnotesize\rmfamily
  \renewcommand{\arraystretch}{1.15}
  \setlength{\tabcolsep}{5pt}
  \begin{tabular}{@{}lcclc@{}}
    \toprule
    \textbf{Category} & \textbf{Families} & \textbf{Constraints} & \textbf{Example constraint (gloss)} & \textbf{Pass (\%)} \\
    \midrule
    \multicolumn{5}{@{}l}{\textbf{General constraints}} \\
    \midrule
    Count and length               & 10 & 111 & exactly two sentences & 64.9 \\
    \addlinespace[2pt]
    Format and structure           & 11 & 112 & a title in \texttt{<<\,>>} on the first line & 59.8 \\
    \addlinespace[2pt]
    Position and linking           &  7 & 108 & \RL{نور} as the second and the second-to-last word & 53.7 \\
    \addlinespace[2pt]
    Copy and transform             &  8 & 109 & the given text, repeated with a new first word & 24.8 \\
    \addlinespace[2pt]
    Sequence composition           &  6 & 158 & the last word of each sentence starts the next & 8.2 \\
    \addlinespace[3pt]\midrule
    \multicolumn{5}{@{}l}{\textbf{Arabic-specific constraints}} \\
    \midrule
    Script, digits and punctuation & 13 & 114 & exactly two Arabic-Indic digits & 65.8 \\
    \addlinespace[2pt]
    Diacritics and Unicode         &  7 & 109 & all three \textit{tanw\=\i n} marks & 40.4 \\
    \addlinespace[2pt]
    Surface lexicon                &  7 & 159 & exactly two words that end in \RL{ة} & 22.0 \\
    \addlinespace[2pt]
    Morphology lookup              &  8 & 160 & the \textit{nisba} adjective of \RL{اقتصاد} (`economy') & 15.6 \\
    \midrule
    Total                          & 77 & 1,140 & & 36.5 \\
    \bottomrule
  \end{tabular}
  \caption{Constraint categories of \benchmark{AraIFBench}. \textit{Constraints} counts the constraint instances in the 1,000 tasks: 860 in tasks with one constraint and 280 in the 140 tasks with two. \textit{Pass} is the constraint-level strict pass rate of \model{Qwen3-8B} (sampled decoding) and shows that difficulty varies by category.}
  \label{tab:araifbench-categories}
\end{table}

The five general categories cover counting, formatting, word positioning, text transformation, and sequence composition. The four Arabic-specific categories contain 35 constraint families, accounting for 542 of the 1,140 constraints, and operate at two levels: orthographic and word-level.

At the orthographic level, constraints target Arabic writing conventions, including diacritics, Arabic-Indic digits, punctuation, and Unicode character variants. For example, a prompt may require every sentence to end with the Arabic question mark (\RL{؟}) or contain exactly two \textit{shadda} marks. These distinctions are important in Arabic writing but are rarely evaluated by translated instruction-following benchmarks. The example below combines an Arabic-specific orthographic constraint with a general counting constraint.
\begin{promptbox}[\benchmark{AraIFBench} task \texttt{test\_public\_679}]
\begin{RLtext}
هل سيكون الذكاء الاصطناعي خطر على البشر\\
يرجى الالتزام بما يلي في الإجابة:\\
١. استخدم الفاصلة العربية ، عدد \LR{3} مرات بالضبط.\\
٢. يجب أن يحتوي الرد على \LR{12} كلمات بالضبط.
\end{RLtext}
\tcblower
\small
\textit{Gloss.} Will artificial intelligence be a danger to humans?
Please adhere to the following in your answer:
(1) Use the Arabic comma (\RL{،}) exactly 3 times.
(2) The answer must contain exactly 12 words.
The request comes from WildChat; we correct the spelling of one word (\RL{الاصطناعي}, `artificial').
\end{promptbox}

At the word level, constraints target the written forms of words, including \textit{t\=a' marb\=u\d{t}a}, \textit{hamza}, and morphological derivations. For example, a prompt may require the broken plural of \RL{كتاب} (\textit{kit\=ab}, `book'), which is \RL{كتب} (\textit{kutub}). Morphology verifiers check whether the required form appears in the response using a fixed lexicon, but do not assess grammatical agreement or case elsewhere in the answer.

\begin{figure}[t]
  \centering
  \includegraphics[width=\linewidth]{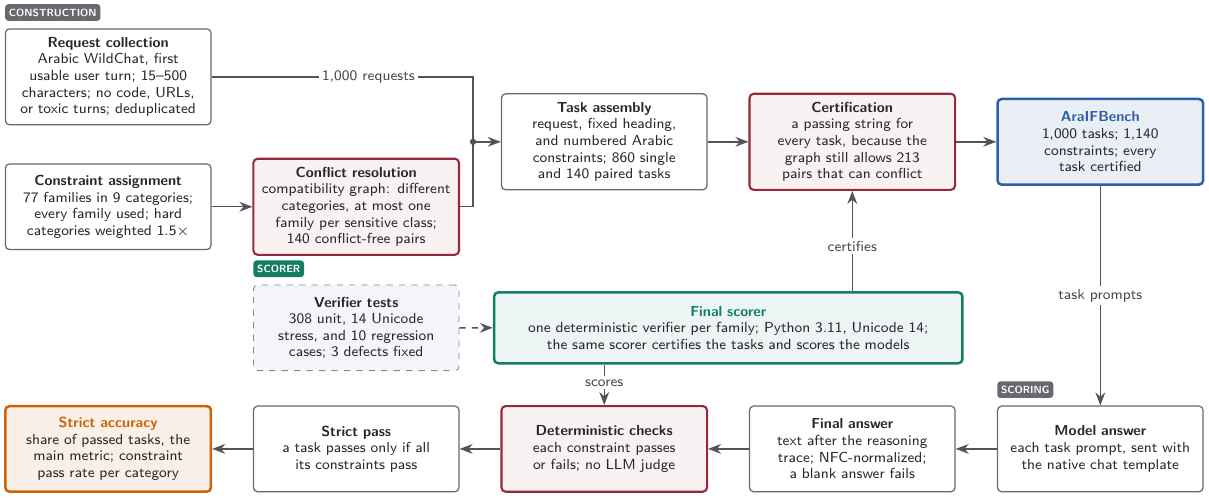}
  \caption{Construction and scoring of \benchmark{AraIFBench}. Construction joins real Arabic requests with a constraint plan, in which conflict resolution allows only compatible pairs of families, and certifies that every task can be satisfied. The same final scorer certifies the tasks and scores the models. A task passes only if all its constraints pass.}
  \label{fig:araifbench-pipeline}
\end{figure}

Figure~\ref{fig:araifbench-pipeline} summarizes the construction and scoring pipeline. Each constraint family has a deterministic verifier, and all tasks are certified to be satisfiable. We report strict prompt-level accuracy, where a task passes only if all its constraints are satisfied. Evaluation requires no LLM judge and measures constraint compliance rather than answer usefulness. The Arabic instruction-following training data for SFT (Section~\ref{subsec:data-instruct}) and RL (Section~\ref{subsec:rl-data}) was screened against benchmark prompts.

\paragraph{\araLingBmarks{} -- I'rab and Tashkeel benchmarks.}

Existing Arabic benchmarks in our evaluation provide limited evidence about whether models can truly understand the Arabic language with its linguistic ambiguities. This is particularly important for Arabic-centric models, where rich root-and-pattern morphology, extensive cliticization, and typically undiacritized writing allow the same surface form to admit multiple grammatical or vocalized analyses, with the intended interpretation often recoverable only from the surrounding sentence~\citep{pasha2014madamira}. To address this gap, we introduce the following two benchmarks derived from the Arabic Linguistic Pipeline (ALP) corpus~\citep{freihat2023alp}.

The \parsingBMark Benchmark is a new benchmark that evaluates contextual grammatical analysis through four-option multiple-choice questions. It contains 1{,}446 questions where each question presents a naturally occurring Arabic sentence, marks one target word, and asks for its grammatical analysis. Decisions concern syntactic functions, including subjects, objects, predicates, and modifiers, or verbal properties such as mood and voice. The task evaluates targeted grammatical interpretation rather than full-sentence dependency parsing.

The \tashekeelBMark Benchmark is a new benchmark that evaluates whether a model can select the contextually appropriate internal vocalization of a marked word in an undiacritized sentence. It contains 1,344 questions where each question contains four alternative vocalizations sharing the same undiacritized spelling. Final case and mood endings are excluded, focusing evaluation on lexical identity, morphological pattern, verbal features, voice, and function-word interpretation. Unlike full-sentence diacritization, the task isolates one word-level decision and its supporting context. Full details of the construction of the two benchmarks are provided in Appendix~\ref{app:arabic-linguistic-benchmarks}.

\paragraph{AA-LCR-Ar.}
\label{par:aa-lcr-ar}
\benchmark{AA-LCR}~\citep{artificialanalysis2025aalcr}
is a set of 100 difficult questions, each requiring reasoning over a set of real-world documents of about
100K tokens (company reports, government consultations, legal texts, academic papers, etc.). Answers cannot
be retrieved directly and must be reasoned from multiple sources, and responses are marked correct or
incorrect by an LLM judge against the official answer. We base \benchmark{AA-LCR-Ar} on version 1.1 of the benchmark, including its corrected reference answers and revised grading instructions. We translated the full benchmark\footnote{\url{https://hf.co/datasets/IFM/AA-LCR-1.1-AR}} into Modern Standard Arabic with \model{GPT-5.6 Terra}, keeping the prompt structure and document order of the
original dataset. Documents were translated in chunks of at most 20K tokens split at paragraph boundaries, and each
was checked automatically against its source. Reference answers were reviewed manually, with proper nouns kept in the documents' Latin
spelling and the judge instructed to match names across scripts.

\paragraph{SimpleQA-Verified-Ar.}
\benchmark{SimpleQA-Verified}~\citep{haas2026simpleqaverifiedreliablefactuality} is a 1,000-question benchmark for short-form factuality, with each question targeting a single, unambiguous fact. To create \benchmark{SimpleQA-Verified-Ar}\footnote{\url{https://hf.co/datasets/IFM/SimpleQA-Verified-AR}}, we translate the questions and reference answers into Arabic using \model{gemma-4-31B-it}, preserving the original meaning and answer content. All translations were verified by humans.

\subsubsection{Benchmarks and Setup}
\label{subsubsec:eval-ar-benchmarks}\label{subsubsec:eval-ar-setup}
Table~\ref{tab:arabic-benchmarks} lists the Arabic benchmarks used in this report, grouped into general capabilities, Arabic domain knowledge, dialect understanding and generation, and multi-turn and long-context understanding. Descriptions of each benchmark are given in Appendix~\ref{app:arabic-benchmark-details}.

\begin{table}[!htbp]
\centering
\footnotesize
\renewcommand{\arraystretch}{1.05}
\setlength{\tabcolsep}{4pt}
\begin{tabularx}{\textwidth}{@{}>{\raggedright\arraybackslash}p{5.6cm}>{\raggedright\arraybackslash}X>{\raggedright\arraybackslash}p{2.8cm}@{}}
\toprule
\textbf{Benchmark} & \textbf{What it measures} & \textbf{Metric} \\
\midrule
\multicolumn{3}{@{}l}{\textit{General capabilities}} \\
ArabicMMLU~\citep{koto-etal-2024-arabicmmlu} & School and university exam questions across academic and professional subjects & Accuracy \\
AraSTEM~\citep{mustapha2025arastem} & Native Arabic STEM questions across educational levels & Accuracy \\
GAT Math~\citep{alballaa2025gatmath} & Mathematical reasoning from the Saudi General Aptitude Test & Accuracy \\
Arabic GSM8K~\citep{cobbe2021gsm8k,arabic-gsm8k} & Multi-step arithmetic word problems & Normalized EM \\
MedAraBench~\citep{abudaoud2026medarabench} & Medical question answering and clinical reasoning & Accuracy \\
Legal (HELM)~\citep{mai2026helmarabicenterprise} & UAE civil, commercial, and penal law, open- and closed-book & Judge accuracy \\
IFEval-Ar~\citep{elfilali2025arabicifeval} & Verifiable instruction constraints adapted to Arabic & Strict prompt acc. \\
AraIFBench (ours) & Constraints specific to Arabic orthography and morphology & Strict prompt acc. \\
\midrule
\multicolumn{3}{@{}l}{\textit{Arabic domain knowledge}} \\
QIAS 2025~\citep{bouchekif-etal-2025-qias} & Islamic inheritance reasoning and general Islamic knowledge & Accuracy \\
IslamicEval 2025~\citep{mubarak-etal-2025-islamiceval} & Verifying statements against the Qur'an and Hadith & Accuracy \\
PalmX 2025~\citep{alwajih-etal-2025-palmx} & Islamic practices, texts, history, and occasions & Accuracy \\
AraDiCE-Culture~\citep{mousi-etal-2025-aradice} & Country-specific cultural questions from six Arab countries & Accuracy \\
ArabCulture~\citep{sadallah-etal-2025-commonsense} & Cultural commonsense across 13 countries & Accuracy \\
Alyah~\citep{emirati_dialect_benchmark_2026} & Emirati culture, heritage, and dialect & Accuracy \\
FannOrFlop\_MCQ~\citep{alghallabi2025fann} & Interpretation of Arabic poetry across eras and genres & Accuracy \\
Arabic Cuisine & Knowledge of Arab cuisine & Accuracy \\
\araLingBmarks (ours) & Grammatical analysis (i'rab) and diacritization (tashkeel) & Accuracy \\
\midrule
\multicolumn{3}{@{}l}{\textit{Dialect understanding and generation}} \\
QADI~\citep{abdelali-etal-2021-qadi} & Dialect identification over MSA and ten regional varieties & Accuracy \\
Dialectal-Arabic-MMLU~\citep{altakrori-etal-2026-dialectalarabicmmlu} & Knowledge and reasoning in five dialects & Accuracy \\
ArabicMMLU, diacritized~\citep{inoue-etal-2026-diacritics} & Robustness to full diacritization on identical questions & Accuracy \\
FLORES+~\citep{nllb2024scaling,maillard-etal-2024-findings} & Translation between English, MSA, and eight dialects & Judge score \\
DarijaBench-Translit.~\citep{shang-etal-2025-atlas} & Moroccan Arabic script $\leftrightarrow$ Arabizi transliteration & Judge score \\
EgyptianBench-Translit.~\citep{shang2025nilechat} & Egyptian Arabic script $\leftrightarrow$ Arabizi transliteration & Judge score \\
\midrule
\multicolumn{3}{@{}l}{\textit{Multi-turn and long context}} \\
MT-Bench Arabic~\citep{boughorbel-hawasly-2023-analyzing} & Two-turn dialogue where the second turn builds on the first & Judge score (1--10) \\
AA-LCR-Ar (ours) & Reasoning over document sets of about 100K tokens & Judge pass rate \\
\bottomrule
\end{tabularx}
\caption{Arabic benchmarks used to evaluate \modelname. Benchmarks marked ``ours'' are introduced in Section~\ref{subsubsec:eval-new-bench}. The benchmarks in Table~\ref{tab:audit-removal} are audited before use (Section~\ref{subsubsec:eval-ar-audit}). Details are in Appendix~\ref{app:arabic-benchmark-details}.}
\label{tab:arabic-benchmarks}
\end{table}

We compare \modelname, at each model size, with similarly sized models under both reasoning and non-reasoning configurations. All models generate free-form responses using their recommended decoding hyperparameters, without enforcing deterministic decoding. This is particularly important for reasoning models, which can exhibit looping behavior under greedy decoding~\citep{pipis2025wait}.
Each model is prompted zero-shot with the benchmark instructions through its native chat template. Most benchmarks are scored by LLM-as-a-judge accuracy~\citep{NEURIPS2023_91f18a12}, which credits a correct answer even when it does not follow the output format that exact-match scoring requires; instruction-following benchmarks use their rule-based verifiers. For multi-turn tasks, we preserve the conversation history, including reasoning traces from earlier responses, when supported by the model's native chat template.

Model generation uses a customized \modelname fork\footnote{\url{https://github.com/ifm-ai/jais3-lm-evaluation-harness}} of the LM Evaluation Harness~\citep{biderman2024lmharness}, with vLLM~\citep{kwon2023vllm} as the primary backend. Multi-turn benchmarks use a compatible multi-turn harness\footnote{\url{https://github.com/ifm-ai/mt-evaluation-harness}}. The suite orchestrates benchmark configurations, model-specific decoding settings, context-length allocation, and scoring pipelines, including LLM-as-a-judge evaluation. Further details of the implementation are provided in Appendix~\ref{app:jais3-evaluation-details}. The benchmark inventory, decoding configurations, and scoring methods are described in Appendices~\ref{app:jais3-benchmark-inventory}, \ref{app:jais3-decoding-details}, and~\ref{app:jais3-scoring-details}, respectively.

\subsubsection{Results}
\label{subsubsec:eval-ar-results}
\providecommand{\model}[1]{#1}

\begin{table}[!htbp]
\centering
\scriptsize\rmfamily
\renewcommand{\arraystretch}{1.05}
\setlength{\tabcolsep}{2.2pt}
\begin{tabular}{@{}l*{8}{c}@{}}
\toprule
\textbf{Domain} & \multicolumn{1}{c}{\textbf{\begin{tabular}[b]{@{}c@{}}World\\Knowledge\end{tabular}}} & \multicolumn{3}{c}{\textbf{STEM}} & \multicolumn{1}{c}{\textbf{Medical}} & \multicolumn{1}{c}{\textbf{Legal}} & \multicolumn{2}{c}{\textbf{\begin{tabular}[b]{@{}c@{}}Instruction\\Following\end{tabular}}} \\
\cmidrule(lr){2-2}\cmidrule(lr){3-5}\cmidrule(lr){6-6}\cmidrule(lr){7-7}\cmidrule(lr){8-9}
\textbf{Benchmark} & \textbf{\begin{tabular}[b]{@{}c@{}}Arabic\\MMLU\end{tabular}} & \textbf{AraSTEM} & \textbf{\begin{tabular}[b]{@{}c@{}}GAT\\Math\end{tabular}} & \textbf{\begin{tabular}[b]{@{}c@{}}Arabic\\GSM8K\end{tabular}} & \textbf{\begin{tabular}[b]{@{}c@{}}MedAra\\Bench\end{tabular}} & \textbf{\begin{tabular}[b]{@{}c@{}}Legal\\(HELM)\end{tabular}} & \textbf{IFEval-Ar} & \textbf{\begin{tabular}[b]{@{}c@{}}AraIF\\Bench\end{tabular}} \\

 & Acc. & Acc. & Acc. & Acc. & Acc. & Acc. & Acc. & Acc. \\
\midrule
\multicolumn{9}{@{}l}{\textbf{1B Scale}} \\
\midrule
\model{Falcon-H1-1.5B-Deep-Instruct} & 48.69 & 62.49 & 40.19 & 65.43 & 32.32 & 48.84 & 49.26 & 16.20 \\
\model{Hala-1.2B} & 45.00 & 48.32 & 29.63 & 57.29 & 32.30 & 41.09 & 44.06 & 14.60 \\
\model{Qwen3.5-2B} & 63.10 & \textbf{83.91} & 87.37 & 75.61 & \textbf{54.25} & 49.10 & 62.62 & 33.10 \\
\model{tiny-aya-global} & 55.93 & 58.48 & 31.36 & 59.57 & 39.49 & 52.20 & 58.17 & 21.50 \\
\rowcolor{jaisrow}\model{Jais-3-0.9B} & \textbf{70.37} & 81.57 & \textbf{87.39} & \textbf{80.93} & 48.08 & \textbf{58.66} & \textbf{71.29} & \textbf{91.10} \\
\addlinespace[3pt]\midrule
\multicolumn{9}{@{}l}{\textbf{4B Scale}} \\
\midrule
\model{Falcon-H1-3B-Instruct} & 51.02 & 64.93 & 41.29 & 72.21 & 34.53 & 52.97 & 55.69 & 18.40 \\
\model{gemma-4-E2B-it} & 69.37 & 85.09 & 90.48 & 81.70 & 57.18 & 57.11 & 69.55 & 37.60 \\
\model{granite-4.0-h-3b-ar} & 59.83 & 66.49 & 45.32 & 67.21 & 38.13 & 56.07 & 43.07 & 14.70 \\
\model{Qwen3.5-4B} & 75.22 & 91.39 & 95.85 & 88.74 & 74.05 & 63.31 & 73.02 & 53.70 \\
\rowcolor{jaisrow}\model{Jais-3-3.7B} & \textbf{83.29} & \textbf{91.71} & \textbf{96.19} & \textbf{91.90} & \textbf{74.13} & \textbf{66.67} & \textbf{78.96} & \textbf{92.60} \\
\addlinespace[3pt]\midrule
\multicolumn{9}{@{}l}{\textbf{7B Scale}} \\
\midrule
\model{ALLaM-7B-Instruct-preview} & 71.21 & 78.98 & 38.38 & 56.81 & 49.12 & 61.24 & 48.02 & 18.90 \\
\model{Falcon-H1-7B-Instruct} & 63.31 & 76.58 & 45.45 & 84.99 & 47.98 & 57.88 & 62.38 & 26.20 \\
\model{Fanar-1-9B-Instruct} & 65.75 & 72.98 & 40.64 & 75.25 & 50.71 & 61.76 & 44.55 & 25.80 \\
\model{gemma-4-12B-it} & 84.08 & \textbf{93.70} & \textbf{98.03} & 92.57 & \textbf{80.97} & \textbf{67.44} & 78.71 & 75.30 \\
\model{gemma-4-E4B-it} & 77.47 & 89.66 & 94.40 & 86.98 & 68.39 & 63.05 & 72.28 & 42.60 \\
\model{Jais-2-8B-Chat} & 70.42 & 64.84 & 35.67 & 72.97 & 41.58 & 60.47 & 61.63 & 21.10 \\
\model{Qwen3.5-9B} & 80.93 & 93.42 & 97.26 & 91.42 & 80.80 & 64.34 & 73.76 & 58.30 \\
\rowcolor{jaisrow}\model{Jais-3-7B} & \textbf{85.23} & 92.61 & 96.39 & \textbf{92.73} & 78.32 & 65.89 & \textbf{80.20} & \textbf{91.60} \\
\addlinespace[3pt]\midrule
\multicolumn{9}{@{}l}{\textbf{32B Scale}} \\
\midrule
\model{Falcon-H1-34B-Instruct} & 73.66 & 85.47 & 63.39 & 90.24 & 60.65 & 63.05 & 66.83 & 32.20 \\
\model{Fanar-2-27B-Instruct} & 66.27 & 42.40 & 87.78 & 87.67 & 50.88 & 45.74 & 49.75 & 40.00 \\
\model{Jais-2-70B-Chat} & 80.66 & 75.74 & 40.67 & 83.38 & 52.62 & 66.15 & 69.06 & 36.00 \\
\model{gemma-4-31B-it} & \textbf{89.53} & \textbf{96.29} & \textbf{98.64} & 93.72 & \textbf{88.82} & 71.06 & 78.71 & 80.80 \\
\model{Muse-Glimmer-30B} & 87.02 & 94.71 & 98.49 & \textbf{95.16} & 86.08 & 73.13 & 79.46 & 84.20 \\
\model{Qwen3.8-27B} & 85.79 & 95.30 & 98.43 & 93.24 & 87.25 & 70.54 & 76.98 & 81.30 \\
\rowcolor{jaisrow}\model{Jais-3-32B} & 88.56 & 95.24 & 98.50 & 94.26 & 86.98 & \textbf{73.39} & \textbf{79.95} & 88.70 \\
\rowcolor{jaisrow}\model{Jais-3-MoVA-36B-A4B} & 87.67 & 94.69 & 97.51 & 92.19 & 85.41 & 69.25 & \textbf{79.95} & \textbf{92.00} \\
\addlinespace[3pt]\midrule
\multicolumn{9}{@{}l}{\textbf{375B Scale}} \\
\midrule
\model{MiniMax-M3} & 86.39 & 95.57 & 97.84 & 93.19 & 87.65 & 74.94 & 70.05 & 53.80 \\
\model{Nemotron-3-Ultra-550B-A55B-BF16} & 88.14 & 95.24 & 98.42 & 93.72 & 88.39 & \textbf{77.00} & \textbf{79.21} & 78.90 \\
\model{Qwen3.5-397B-A17B} & 89.10 & \textbf{96.61} & \textbf{98.59} & 93.64 & \textbf{90.77} & 71.58 & 74.50 & 75.10 \\
\model{Qwen3.8-Flash-Next} & 89.31 & 96.28 & 98.47 & \textbf{94.37} & \textbf{90.77} & 69.51 & \textbf{79.21} & 81.70 \\
\rowcolor{jaisrow}\model{Jais-3-375B-A23B} & \textbf{89.34} & 95.46 & 98.42 & 86.75 & 89.51 & 71.32 & 77.72 & \textbf{93.90} \\
\addlinespace[3pt]\midrule
\multicolumn{9}{@{}l}{\textbf{Frontier Models (Closed Source)}} \\
\midrule
\model{Claude Sonnet 5$^{\diamondsuit}$} & 89.96 & 95.45 & 98.17 & 94.87 & 90.32 & 77.26 & 72.28 & 63.10 \\
\model{Gemini 3.8 Flash} & \textbf{95.62} & \textbf{98.18} & \textbf{98.96} & \textbf{96.36} & \textbf{94.67} & 78.29 & 79.70 & 81.40 \\
\model{GPT-5.6 Sol$^{\diamondsuit}$} & 93.87 & 97.62 & 98.56 & 94.95 & 93.20 & \textbf{80.62} & \textbf{80.69} & \textbf{84.17} \\
\bottomrule
\end{tabular}
\caption{Arabic general capabilities results (\%). All results use thinking mode (high effort) where available. Within each model size, the best score in each column is shown in bold. $^{\diamondsuit}$: xhigh thinking effort.}\label{tab:general-capabilities}
\end{table}

\providecommand{\model}[1]{#1}

\begin{table}[!htbp]
\centering
\scriptsize\rmfamily
\renewcommand{\arraystretch}{1.05}
\setlength{\tabcolsep}{2.2pt}
\lineskiplimit=0pt
\ifdefined\modelw\else\newlength{\modelw}\fi\settowidth{\modelw}{\model{Nemotron-3-Ultra-550B-A55B-BF16}}
\begin{tabular}{@{}>{\raggedright\arraybackslash}p{\modelw}*{10}{>{\centering\arraybackslash}p{\dimexpr(\linewidth-\modelw-2\tabcolsep)/10-2\tabcolsep\relax}}@{}}
\toprule
\textbf{Domain} & \multicolumn{4}{c}{\makebox[0pt][c]{\textbf{\begin{tabular}[b]{@{}c@{}}Islamic\\Knowledge\end{tabular}}}} & \multicolumn{5}{c}{\makebox[0pt][c]{\textbf{Arab Culture}}} & \multicolumn{1}{c}{\makebox[0pt][c]{\textbf{\begin{tabular}[b]{@{}c@{}}Arabic\\Linguistics\end{tabular}}}} \\
\cmidrule(lr){2-5}\cmidrule(lr){6-10}\cmidrule(lr){11-11}
\textbf{Benchmark} & \makebox[0pt][c]{\textbf{\begin{tabular}[b]{@{}c@{}}QIAS\\Inherit.\end{tabular}}} & \makebox[0pt][c]{\textbf{\begin{tabular}[b]{@{}c@{}}QIAS\\Knowl.\end{tabular}}} & \makebox[0pt][c]{\textbf{\begin{tabular}[b]{@{}c@{}}Islamic\\Eval\end{tabular}}} & \makebox[0pt][c]{\textbf{PalmX}} & \makebox[0pt][c]{\textbf{\begin{tabular}[b]{@{}c@{}}AraDiCE\\Culture\end{tabular}}} & \makebox[0pt][c]{\textbf{\begin{tabular}[b]{@{}c@{}}Arab\\Culture\end{tabular}}} & \makebox[0pt][c]{\textbf{Alyah}} & 
\makebox[0pt][c]{\textbf{\begin{tabular}[b]{@{}c@{}}FannOrFlop\\\_MCQ\end{tabular}}} &
 \makebox[0pt][c]{\textbf{\begin{tabular}[b]{@{}c@{}}Arabic\\Cuisine\end{tabular}}} & \makebox[0pt][c]{\textbf{\begin{tabular}[b]{@{}c@{}}Arabic\\Ling.\end{tabular}}} \\

 & Acc. & Acc. & Acc. & Acc. & Acc. & Acc. & Acc. & Acc. & Acc. & Acc. \\
\midrule
\multicolumn{11}{@{}l}{\textbf{1B Scale}} \\
\midrule
\model{Falcon-H1-1.5B-Deep-Instruct} & 32.07 & 56.02 & 42.42 & 62.13 & 49.16 & 61.69 & 54.75 & 38.54 & 52.35 & 33.27 \\
\model{Hala-1.2B} & 16.15 & 41.51 & 36.36 & 49.58 & 30.73 & 60.63 & 53.20 & 31.34 & 49.48 & 32.56 \\
\model{Qwen3.5-2B} & 26.06 & 51.52 & 32.47 & 62.03 & 46.93 & 63.83 & 53.20 & 43.60 & 53.59 & 54.74 \\
\model{tiny-aya-global} & 23.05 & 54.72 & 27.27 & 71.62 & 55.31 & 68.61 & 68.13 & 48.50 & 60.64 & 38.16 \\
\rowcolor{jaisrow}\model{Jais-3-0.9B} & \textbf{61.14} & \textbf{65.02} & \textbf{60.61} & \textbf{78.06} & \textbf{72.63} & \textbf{73.98} & \textbf{73.83} & \textbf{59.27} & \textbf{73.37} & \textbf{85.49} \\
\addlinespace[3pt]\midrule
\multicolumn{11}{@{}l}{\textbf{4B Scale}} \\
\midrule
\model{Falcon-H1-3B-Instruct} & 27.51 & 54.28 & 44.59 & 66.56 & 45.81 & 64.18 & 58.89 & 41.09 & 54.37 & 42.05 \\
\model{gemma-4-E2B-it} & 51.56 & 65.31 & 55.84 & 79.22 & 56.42 & 74.71 & 69.69 & 59.99 & 69.19 & 77.88 \\
\model{granite-4.0-h-3b-ar} & 18.15 & 66.47 & 22.51 & 57.49 & 53.63 & 68.41 & 65.98 & 48.76 & 58.55 & 49.34 \\
\model{Qwen3.5-4B} & 50.67 & 69.88 & 46.32 & 79.64 & 61.45 & 76.15 & 63.30 & 50.92 & 68.93 & 72.73 \\
\rowcolor{jaisrow}\model{Jais-3-3.7B} & \textbf{83.30} & \textbf{79.97} & \textbf{74.46} & \textbf{88.92} & \textbf{76.54} & \textbf{84.28} & \textbf{79.71} & \textbf{66.21} & \textbf{82.64} & \textbf{92.12} \\
\addlinespace[3pt]\midrule
\multicolumn{11}{@{}l}{\textbf{7B Scale}} \\
\midrule
\model{ALLaM-7B-Instruct-preview} & 31.18 & 72.86 & \textbf{78.36} & 85.86 & 79.33 & 77.65 & 77.46 & 66.67 & 71.34 & 67.46 \\
\model{Falcon-H1-7B-Instruct} & 43.99 & 65.75 & 61.90 & 74.26 & 59.22 & 72.43 & 65.72 & 51.38 & 64.95 & 47.53 \\
\model{Fanar-1-9B-Instruct} & 36.30 & 64.88 & 69.70 & 80.27 & 67.04 & 77.30 & 69.60 & 63.27 & 68.86 & 57.59 \\
\model{gemma-4-12B-it} & 74.50 & \textbf{81.28} & 67.53 & 89.56 & \textbf{81.56} & \textbf{86.95} & \textbf{82.47} & 66.21 & 84.07 & 93.04 \\
\model{gemma-4-E4B-it} & 62.47 & 69.52 & 71.43 & 82.91 & 75.98 & 81.61 & 77.20 & 63.98 & 77.81 & 84.09 \\
\model{Jais-2-8B-Chat} & 29.06 & 67.49 & 56.71 & 82.81 & 67.60 & 74.60 & 76.42 & 60.50 & 71.61 & 49.20 \\
\model{Qwen3.5-9B} & 61.14 & 72.28 & 59.31 & 84.49 & 74.86 & 80.43 & 68.31 & 58.23 & 76.63 & 81.48 \\
\rowcolor{jaisrow}\model{Jais-3-7B} & \textbf{88.53} & 80.70 & 74.89 & \textbf{91.24} & 81.01 & 84.92 & 80.66 & \textbf{69.44} & \textbf{85.51} & \textbf{93.16} \\
\addlinespace[3pt]\midrule
\multicolumn{11}{@{}l}{\textbf{32B Scale}} \\
\midrule
\model{Falcon-H1-34B-Instruct} & 56.24 & 73.44 & 70.13 & 84.70 & 75.42 & 79.67 & 71.76 & 62.35 & 74.41 & 57.44 \\
\model{Fanar-2-27B-Instruct} & 31.40 & 61.83 & 45.45 & 69.30 & 55.87 & 34.23 & 36.27 & 46.43 & 34.01 & 39.13 \\
\model{Jais-2-70B-Chat} & 42.20 & 78.96 & 65.80 & 90.72 & 80.45 & 82.08 & 84.46 & \textbf{73.62} & 80.42 & 64.10 \\
\model{gemma-4-31B-it} & 93.21 & 86.94 & \textbf{83.98} & 94.51 & \textbf{89.39} & \textbf{91.43} & \textbf{88.43} & 72.58 & \textbf{89.82} & \textbf{96.74} \\
\model{Muse-Glimmer-30B} & 87.42 & 83.02 & 71.86 & 92.62 & 83.24 & 85.98 & 80.31 & 68.29 & 84.66 & 91.83 \\
\model{Qwen3.8-27B} & 81.63 & 81.86 & 54.11 & 90.82 & 77.09 & 84.95 & 77.72 & 62.45 & 81.01 & 89.69 \\
\rowcolor{jaisrow}\model{Jais-3-32B} & \textbf{95.88} & 85.49 & 78.79 & \textbf{94.62} & 88.27 & 88.19 & 82.99 & 71.84 & 87.40 & 94.27 \\
\rowcolor{jaisrow}\model{Jais-3-MoVA-36B-A4B} & 95.66 & \textbf{87.23} & 82.68 & 93.35 & 85.47 & 88.56 & 82.99 & 71.54 & 88.51 & 93.98 \\
\addlinespace[3pt]\midrule
\multicolumn{11}{@{}l}{\textbf{375B Scale}} \\
\midrule
\model{MiniMax-M3} & 80.40 & 85.05 & 76.62 & 92.83 & 82.22 & 86.56 & 79.27 & 67.01 & 85.25 & 88.02 \\
\model{Nemotron-3-Ultra-550B-A55B-BF16} & 83.63 & 85.20 & 66.23 & 93.46 & \textbf{88.83} & 89.12 & 77.89 & 69.72 & 85.77 & 93.32 \\
\model{Qwen3.5-397B-A17B} & 86.08 & 86.79 & \textbf{84.85} & 94.20 & 83.80 & 89.09 & 82.38 & 69.25 & 88.71 & 93.96 \\
\model{Qwen3.8-Flash-Next} & 91.43 & \textbf{87.81} & 61.90 & 94.73 & 82.68 & 89.76 & 81.00 & 69.83 & 87.34 & \textbf{95.04} \\
\rowcolor{jaisrow}\model{Jais-3-375B-A23B} & \textbf{95.43} & 87.66 & 82.25 & \textbf{95.25} & 87.71 & \textbf{89.82} & \textbf{87.56} & \textbf{73.23} & \textbf{89.82} & 94.79 \\
\addlinespace[3pt]\midrule
\multicolumn{11}{@{}l}{\textbf{Frontier Models (Closed Source)}} \\
\midrule
\model{Claude Sonnet 5$^{\diamondsuit}$} & 91.20 & 89.40 & \textbf{83.55} & 93.99 & 85.47 & 89.82 & 87.56 & 84.02 & 87.60 & 91.93 \\
\model{Gemini 3.8 Flash} & \textbf{98.78} & 93.47 & 72.29 & \textbf{97.68} & \textbf{98.88} & \textbf{95.66} & 92.92 & \textbf{87.30} & \textbf{92.89} & \textbf{97.27} \\
\model{GPT-5.6 Sol$^{\diamondsuit}$} & 98.55 & \textbf{93.90} & 74.03 & 97.36 & 96.65 & 93.81 & \textbf{93.09} & 86.97 & \textbf{92.89} & 95.37 \\
\bottomrule
\end{tabular}
\caption{Arabic domain knowledge results (\%). All results use thinking mode (high effort) where available. Within each model size, the best score in each column is shown in bold. $^{\diamondsuit}$: xhigh thinking effort.}\label{tab:arabic-domain}
\end{table}

\providecommand{\model}[1]{#1}

\begin{table}[!htbp]
\centering
\scriptsize\rmfamily
\renewcommand{\arraystretch}{1.05}
\setlength{\tabcolsep}{2.2pt}
\lineskiplimit=0pt
\ifdefined\modelw\else\newlength{\modelw}\fi\settowidth{\modelw}{\model{Nemotron-3-Ultra-550B-A55B-BF16}}
\begin{tabular}{@{}>{\raggedright\arraybackslash}p{\modelw}*{7}{>{\centering\arraybackslash}p{\dimexpr(0.85\linewidth-\modelw-2\tabcolsep)/7-2\tabcolsep\relax}}@{}}
\toprule
\textbf{Domain} & \multicolumn{7}{c}{\makebox[0pt][c]{\textbf{Dialect Understanding and Generation}}} \\
\cmidrule(lr){2-8}
\textbf{Benchmark} & \makebox[0pt][c]{\textbf{QADI}} & \makebox[0pt][c]{\textbf{\begin{tabular}[b]{@{}c@{}}Dialect.\\MMLU\end{tabular}}} & \makebox[0pt][c]{\textbf{\begin{tabular}[b]{@{}c@{}}Arabic\\MMLU\end{tabular}}} & \makebox[0pt][c]{\textbf{\begin{tabular}[b]{@{}c@{}}MMLU\\diacr.\end{tabular}}} & \makebox[0pt][c]{\textbf{\begin{tabular}[b]{@{}c@{}}FLORES+\end{tabular}}} & \makebox[0pt][c]{\textbf{\begin{tabular}[b]{@{}c@{}}Translit.\\EG\end{tabular}}} & \makebox[0pt][c]{\textbf{\begin{tabular}[b]{@{}c@{}}Translit.\\DR\end{tabular}}} \\

 & Acc. & Acc. & Acc. & Acc. & Judge & Judge & Judge \\
\midrule
\multicolumn{8}{@{}l}{\textbf{1B Scale}} \\
\midrule
\model{Falcon-H1-1.5B-Deep-Instruct} & 23.39 & 42.26 & 48.69 & 34.68 & 46.78 & 30.53 & 42.35 \\
\model{Hala-1.2B} & 22.97 & 40.33 & 45.00 & 31.28 & 42.25 & 34.94 & 25.72 \\
\model{Qwen3.5-2B} & 24.23 & 59.51 & 63.10 & 49.36 & 51.91 & 34.77 & 45.49 \\
\model{tiny-aya-global} & 38.79 & 47.68 & 55.93 & 45.64 & 63.79 & 38.37 & 45.75 \\
\rowcolor{jaisrow}\model{Jais-3-0.9B} & \textbf{65.87} & \textbf{63.70} & \textbf{70.37} & \textbf{65.10} & \textbf{82.33} & \textbf{75.58} & \textbf{79.17} \\
\addlinespace[3pt]\midrule
\multicolumn{8}{@{}l}{\textbf{4B Scale}} \\
\midrule
\model{Falcon-H1-3B-Instruct} & 28.52 & 43.81 & 51.02 & 34.62 & 53.84 & 41.26 & 39.33 \\
\model{gemma-4-E2B-it} & 44.40 & 65.05 & 69.37 & 64.68 & 82.60 & 65.16 & \textbf{84.73} \\
\model{granite-4.0-h-3b-ar} & 42.19 & 50.71 & 59.83 & 53.25 & 48.72 & 35.33 & 40.66 \\
\model{Qwen3.5-4B} & 37.40 & 76.48 & 75.22 & 70.76 & 66.61 & 46.66 & 74.37 \\
\rowcolor{jaisrow}\model{Jais-3-3.7B} & \textbf{68.84} & \textbf{81.55} & \textbf{83.29} & \textbf{80.65} & \textbf{89.02} & \textbf{76.22} & 79.28 \\
\addlinespace[3pt]\midrule
\multicolumn{8}{@{}l}{\textbf{7B Scale}} \\
\midrule
\model{ALLaM-7B-Instruct-preview} & 42.77 & 56.61 & 71.21 & 61.85 & 74.47 & 39.71 & 31.81 \\
\model{Falcon-H1-7B-Instruct} & 38.08 & 58.46 & 63.31 & 52.33 & 59.76 & 42.47 & 48.74 \\
\model{Fanar-1-9B-Instruct} & 46.19 & 58.21 & 65.75 & 54.86 & 68.35 & 50.18 & 65.67 \\
\model{gemma-4-12B-it} & 62.13 & 83.53 & 84.08 & 83.24 & 89.46 & 86.23 & \textbf{94.68} \\
\model{gemma-4-E4B-it} & 59.84 & 76.59 & 77.47 & 74.51 & 83.97 & 77.52 & 90.26 \\
\model{Jais-2-8B-Chat} & 48.38 & 54.59 & 70.42 & 54.66 & 79.40 & \textbf{88.76} & 76.73 \\
\model{Qwen3.5-9B} & 35.69 & 80.44 & 80.93 & 77.33 & 74.27 & 49.03 & 83.51 \\
\rowcolor{jaisrow}\model{Jais-3-7B} & \textbf{68.98} & \textbf{83.57} & \textbf{85.23} & \textbf{83.60} & \textbf{90.62} & 76.99 & 79.07 \\
\addlinespace[3pt]\midrule
\multicolumn{8}{@{}l}{\textbf{32B Scale}} \\
\midrule
\model{Falcon-H1-34B-Instruct} & 50.28 & 71.47 & 73.66 & 69.09 & 70.58 & 54.70 & 75.75 \\
\model{Fanar-2-27B-Instruct} & 29.63 & 28.15 & 66.27 & 64.18 & 71.23 & 58.45 & 72.53 \\
\model{Jais-2-70B-Chat} & 65.08 & 67.56 & 80.66 & 74.05 & 83.83 & 90.60 & 92.75 \\
\model{gemma-4-31B-it} & 70.40 & \textbf{89.35} & \textbf{89.53} & \textbf{88.88} & \textbf{92.79} & \textbf{91.34} & \textbf{96.73} \\
\model{Muse-Glimmer-30B} & 60.63 & 86.99 & 87.02 & 86.60 & 84.78 & 88.43 & 95.25 \\
\model{Qwen3.8-27B} & 51.46 & 86.96 & 85.79 & 85.20 & 81.84 & 64.01 & 88.96 \\
\rowcolor{jaisrow}\model{Jais-3-32B} & \textbf{70.61} & 88.45 & 88.56 & 87.87 & 91.54 & 90.09 & 95.85 \\
\rowcolor{jaisrow}\model{Jais-3-MoVA-36B-A4B} & 69.11 & 87.29 & 87.67 & 86.81 & 90.80 & 87.02 & 85.79 \\
\addlinespace[3pt]\midrule
\multicolumn{8}{@{}l}{\textbf{375B Scale}} \\
\midrule
\model{MiniMax-M3} & 52.80 & 86.58 & 86.39 & 84.56 & 84.49 & 81.36 & 87.66 \\
\model{Nemotron-3-Ultra-550B-A55B-BF16} & 60.84 & 88.68 & 88.14 & 87.41 & 83.85 & 79.19 & 89.65 \\
\model{Qwen3.5-397B-A17B} & 62.34 & 90.04 & 89.10 & 88.42 & 83.62 & 83.79 & 89.19 \\
\model{Qwen3.8-Flash-Next} & 60.50 & \textbf{90.05} & 89.31 & \textbf{88.75} & 86.86 & 80.59 & 93.43 \\
\rowcolor{jaisrow}\model{Jais-3-375B-A23B} & \textbf{71.11} & 89.59 & \textbf{89.34} & 88.17 & \textbf{91.89} & \textbf{84.95} & \textbf{96.06} \\
\addlinespace[3pt]\midrule
\multicolumn{8}{@{}l}{\textbf{Frontier Models (Closed Source)}} \\
\midrule
\model{Claude Sonnet 5$^{\diamondsuit}$} & 60.42 & 88.54 & 89.96 & 85.02 & 87.04 & 86.52 & 93.82 \\
\model{Gemini 3.8 Flash} & \textbf{74.32} & \textbf{94.76} & \textbf{95.62} & \textbf{91.25} & 91.90 & \textbf{92.62} & \textbf{97.12} \\
\model{GPT-5.6 Sol$^{\diamondsuit}$} & 70.40 & 93.94 & 93.87 & 89.16 & \textbf{91.95} & 90.06 & 96.95 \\
\bottomrule
\end{tabular}
\caption{Arabic dialect understanding and generation results (\%). All results use thinking mode (high effort) where available. Within each model size, the best score in each column is shown in bold. $^{\diamondsuit}$: xhigh thinking effort.}\label{tab:arabic-dialect}
\end{table}

\providecommand{\model}[1]{#1}

\begin{table}[htbp]
\centering
\scriptsize\rmfamily
\renewcommand{\arraystretch}{1.15}
\setlength{\tabcolsep}{0.6pt}
\begin{tabular*}{0.48\linewidth}[t]{@{\extracolsep{\fill}}l*{4}{c}@{}}
\toprule
\textbf{Domain} & \multicolumn{3}{c}{\textbf{Multi-turn}} & \multicolumn{1}{c}{\textbf{\begin{tabular}[b]{@{}c@{}}Long\\Context\end{tabular}}} \\
\cmidrule(lr){2-4}\cmidrule(lr){5-5}
\textbf{Benchmark} & \multicolumn{3}{c}{\textbf{MT-Bench~Arabic}} & \textbf{AA-LCR-Ar} \\
\cmidrule(lr){2-4}
 & Turn 1 & Turn 2 & Avg. & Acc. \\
\midrule
\multicolumn{5}{@{}l}{\textbf{1B Scale}} \\
\midrule
\model{Falcon-H1-1.5B-Deep-Instruct} & 40.75 & 30.50 & 35.62 & -- \\
\model{Hala-1.2B} & 54.12 & 27.00 & 40.56 & -- \\
\model{Qwen3.5-2B} & 57.75 & 49.75 & 53.75 & \textbf{17.00} \\
\model{tiny-aya-global} & 59.00 & 38.12 & 48.56 & -- \\
\rowcolor{jaisrow}\model{Jais-3-0.9B} & \textbf{78.12} & \textbf{64.88} & \textbf{71.50} & -- \\
\addlinespace[3pt]\midrule
\multicolumn{5}{@{}l}{\textbf{4B Scale}} \\
\midrule
\model{Falcon-H1-3B-Instruct} & 58.75 & 51.25 & 55.00 & -- \\
\model{gemma-4-E2B-it} & 92.12 & 86.71 & 89.43 & -- \\
\model{granite-4.0-h-3b-ar} & 54.38 & 33.75 & 44.06 & -- \\
\model{Qwen3.5-4B} & 90.88 & 83.62 & 87.25 & 53.00 \\
\rowcolor{jaisrow}\model{Jais-3-3.7B} & \textbf{98.12} & \textbf{92.00} & \textbf{95.06} & \textbf{68.00} \\
\addlinespace[3pt]\midrule
\multicolumn{5}{@{}l}{\textbf{7B Scale}} \\
\midrule
\model{ALLaM-7B-Instruct-preview} & 62.50 & 44.00 & 53.25 & -- \\
\model{Falcon-H1-7B-Instruct} & 69.88 & 61.27 & 65.60 & 9.00 \\
\model{Fanar-1-9B-Instruct} & 76.50 & 62.12 & 69.31 & -- \\
\model{gemma-4-12B-it} & 98.12 & \textbf{98.38} & \textbf{98.25} & 49.00 \\
\model{gemma-4-E4B-it} & 96.62 & 94.88 & 95.75 & -- \\
\model{Jais-2-8B-Chat} & 74.12 & 56.00 & 65.06 & -- \\
\model{Qwen3.5-9B} & 95.38 & 93.25 & 94.31 & \textbf{68.00} \\
\rowcolor{jaisrow}\model{Jais-3-7B} & \textbf{98.75} & 95.38 & 97.06 & 65.00 \\
\bottomrule
\end{tabular*}
\hfill
\begin{tabular*}{0.50\linewidth}[t]{@{\extracolsep{\fill}}l*{4}{c}@{}}
\toprule
\textbf{Domain} & \multicolumn{3}{c}{\textbf{Multi-turn}} & \multicolumn{1}{c}{\textbf{\begin{tabular}[b]{@{}c@{}}Long\\Context\end{tabular}}} \\
\cmidrule(lr){2-4}\cmidrule(lr){5-5}
\textbf{Benchmark} & \multicolumn{3}{c}{\textbf{MT-Bench~Arabic}} & \textbf{AA-LCR-Ar} \\
\cmidrule(lr){2-4}
 & Turn 1 & Turn 2 & Avg. & Acc. \\
\midrule
\multicolumn{5}{@{}l}{\textbf{32B Scale}} \\
\midrule
\model{Falcon-H1-34B-Instruct} & 81.38 & 70.62 & 76.00 & 14.00 \\
\model{Fanar-2-27B-Instruct} & 87.00 & 10.00 & 48.50 & -- \\
\model{gemma-4-31B-it} & \textbf{99.88} & \textbf{99.00} & \textbf{99.44} & 66.00 \\
\model{Jais-2-70B-Chat} & 77.75 & 66.88 & 72.31 & -- \\
\model{Muse-Glimmer-30B} & 97.75 & 98.38 & 98.06 & -- \\
\model{Qwen3.8-27B} & 98.75 & 96.00 & 97.38 & \textbf{79.00} \\
\rowcolor{jaisrow}\model{Jais-3-32B} & 98.88 & 95.50 & 97.19 & 75.00 \\
\rowcolor{jaisrow}\model{Jais-3-MoVA-36B-A4B} & 97.25 & 94.12 & 95.69 & 73.00 \\
\addlinespace[3pt]\midrule
\multicolumn{5}{@{}l}{\textbf{375B Scale}} \\
\midrule
\model{MiniMax-M3} & 97.75 & 95.12 & 96.44 & \textbf{80.00} \\
\model{Nemotron-3-Ultra-550B-A55B-BF16} & 97.50 & 95.75 & 96.62 & 73.00 \\
\model{Qwen3.5-397B-A17B} & 97.12 & \textbf{96.38} & 96.75 & 78.00 \\
\model{Qwen3.8-Flash-Next} & 98.25 & 91.62 & 94.94 & 74.00 \\
\rowcolor{jaisrow}\model{Jais-3-375B-A23B} & \textbf{98.62} & \textbf{96.38} & \textbf{97.50} & 77.00 \\
\addlinespace[3pt]\midrule
\multicolumn{5}{@{}l}{\textbf{Frontier Models (Closed Source)}} \\
\midrule
\model{Claude Sonnet 5$^{\diamondsuit}$} & \textbf{99.62} & \textbf{99.38} & \textbf{99.50} & 77.00 \\
\model{Gemini 3.8 Flash} & 98.75 & 96.88 & 97.81 & \textbf{87.00} \\
\model{GPT-5.6 Sol$^{\diamondsuit}$} & 97.25 & 97.62 & 97.44 & 82.00 \\
\bottomrule
\end{tabular*}
\caption{Arabic multi-turn and long-context results (\%). All results use thinking mode (high effort) where available. Within each model size, the best score in each column is shown in bold. MT-Bench Arabic turns are the judge's 1--10 ratings $\times$10 and Avg.\ is their mean; -- marks results that are not available. $^{\diamondsuit}$: xhigh thinking effort.}\label{tab:arabic-mtbench-turns-lc}
\end{table}

\paragraph{General capabilities.}
On general Arabic capabilities, \modelname leads the open models at the 1B--7B scales and is competitive at 32B and above, with the largest gains at smaller model sizes. As shown in Table~\ref{tab:general-capabilities}, \modelname leads on \benchmark{AraIFBench} at every scale. Every \modelname model scores above 88\% on \benchmark{AraIFBench}, surpassing all open baselines and the frontier models for which a score is available. At the 1B--7B scales, \modelname also leads on \benchmark{ArabicMMLU}, \benchmark{Arabic GSM8K}, and \benchmark{IFEval-Ar}, and on \benchmark{Legal (HELM)} at 1B and 4B, often outperforming larger models. For example, \model{Jais-3-7B} surpasses \model{gemma-4-12B-it} on \benchmark{ArabicMMLU} and outperforms its predecessor \model{Jais-2-8B-Chat} on every benchmark. At larger scales, \modelname achieves the best \benchmark{Legal (HELM)} score at 32B and the best \benchmark{ArabicMMLU} score at 375B. \modelname remains competitive even where it trails. On \benchmark{AraSTEM}, it stays within 2.5 points of the best model at every scale, and on \benchmark{MedAraBench}, it leads at 4B and trails the best \model{Qwen} or \model{Gemma} model by 6.2 and 2.7 points at 1B and 7B.

\paragraph{Arabic domain knowledge.}
Table~\ref{tab:arabic-domain} reports the Arabic domain knowledge results. At the 1B and 4B scales, \modelname leads all ten benchmarks, outperforming the next best model by 5--31 points and 6--32 points, respectively. At the 7B scale, \modelname leads on \benchmark{QIAS-Inheritance}, \benchmark{PalmX}, \benchmark{FannOrFlop\_MCQ}, \benchmark{Arabic Cuisine}, and \araLingBmarks while remaining within about 2 points of \model{gemma-4-12B-it} on four of the other five benchmarks. At the 375B scale, \modelname leads its class on six of the ten benchmarks. The 32B scale is more competitive: \model{gemma-4-31B-it} leads on six benchmarks and \model{Jais-2-70B-Chat} on \benchmark{FannOrFlop\_MCQ}, while \model{Jais-3-32B} achieves the best score on \benchmark{QIAS-Inheritance} and \benchmark{PalmX}, and \model{Jais-3-MoVA-36B-A4B} on \benchmark{QIAS-Knowledge}.

Compared with the Arabic-centric baselines, \modelname leads on nine of the ten benchmarks at the 7B scale, trailing \model{ALLaM-7B-Instruct-preview} only on \benchmark{IslamicEval}, while \model{Jais-3-32B} leads on eight of the ten at the 32B scale, trailing \model{Jais-2-70B-Chat} on \benchmark{Alyah} and \benchmark{FannOrFlop\_MCQ}. \modelname shows particularly strong gains over previous Arabic-centric models on Islamic and language-specific knowledge, with the largest improvement observed on \benchmark{QIAS-Inheritance}. At the frontier scale, \model{Jais-3-375B-A23B} is competitive with \model{Claude Sonnet 5} across several Arabic-focused benchmarks and outperforms \model{Gemini 3.8 Flash} and \model{GPT-5.6 Sol} on \benchmark{IslamicEval}, although a gap remains to the strongest frontier models. This gap is relatively small on \araLingBmarks and substantially larger on \benchmark{FannOrFlop\_MCQ}.

\paragraph{Dialect understanding and generation.}
Table~\ref{tab:arabic-dialect} reports the results. \model{Jais-3-0.9B} leads all seven benchmarks at the 1B scale, by 4--37 points. \model{Jais-3-3.7B} leads six at the 4B scale, by 5--24 points; the exception is Darija transliteration, where \model{gemma-4-E2B-it} is ahead. \model{Jais-3-7B} leads the five understanding and translation benchmarks at the 7B scale, but trails \model{Jais-2-8B-Chat} on Egyptian and \model{gemma-4-12B-it} on Darija transliteration. At the 32B scale, \model{gemma-4-31B-it} leads six benchmarks, while \model{Jais-3-32B} wins \benchmark{QADI} and stays within 1.3 points on the rest. \model{Jais-3-375B-A23B} leads its class on five benchmarks and is within 0.6 points of \model{Qwen3.8-Flash-Next} on the other two.

\benchmark{QADI} is a consistent strength: \modelname achieves the best dialect identification score at every open-weight scale, and \model{Jais-3-32B} and \model{Jais-3-375B-A23B} both surpass \model{GPT-5.6 Sol} and \model{Claude Sonnet 5}. Among Arabic-centric baselines, \model{Jais-3-7B} beats \model{ALLaM-7B-Instruct-preview} and \model{Fanar-1-9B-Instruct} on all seven benchmarks and \model{Jais-2-8B-Chat} on six, with a large gain of $29$ points on \benchmark{Dialectal-Arabic-MMLU}.  Transliteration is a relative weakness at smaller scales, with Darija scores plateauing around 79 from 0.9B to 7B. At the frontier, \model{Jais-3-375B-A23B} outperforms \model{Claude Sonnet 5} on five benchmarks and matches \model{GPT-5.6 Sol} and \model{Gemini 3.8 Flash} on \benchmark{FLORES+}, but trails \model{Gemini 3.8 Flash} by 3--6 points on the MMLU-style benchmarks.

\paragraph{Multi-turn and long context.}
Table~\ref{tab:arabic-mtbench-turns-lc} shows that \modelname leads \benchmark{MT-Bench Arabic} at the 1B, 4B, and 375B scales; \model{Jais-3-0.9B} outperforms the next best 1B model by about 18 points, and \model{Jais-3-375B-A23B} reaches 97.50, within 2.0 points of \model{Claude Sonnet 5}. At 7B and 32B, \modelname is within 5 points of the best open models.

A notable strength of \modelname is long-context support at small scales. All models from 3.7B upward support a 512K-token context window (Table~\ref{tab:k2horizon-configs}), so they can be evaluated on \benchmark{AA-LCR-Ar}, whose inputs are about 100K tokens long; most Arabic-focused baselines in our comparison could not run it. \model{Jais-3-3.7B}
scores 68, 15 points above \model{Qwen3.5-4B} and level with \model{Qwen3.5-9B}; \model{Jais-3-32B} trails only \model{Qwen3.8-27B} at 32B, and \model{Jais-3-375B-A23B} ties \model{Claude Sonnet 5}.

\FloatBarrier
\subsection{Agentic Evaluation}\label{subsec:agentic-eval}

An agentic model must do more than answer a question.
It must plan a task of many steps, call tools, read their output, and correct its own errors until the task is complete.
We test these abilities on eleven benchmarks (Table~\ref{tab:agentic-benchmarks}).
Two benchmarks are in Arabic.
GAIA-v2-LILT (AR) tests agentic search, and the Arabic File-Processing Benchmark, which we introduce in Section~\ref{subsubsec:eval-file-bench}, tests professional work on Arabic office files.
The other nine benchmarks are in English.
They test general agentic abilities in coding, tool use, professional work, and deep search.
Code generation without tools, such as HumanEval and MBPP, is part of the English evaluation (Table~\ref{tab:english-coding-multiturn}).
We evaluate the models from 3.7B upward.
We do not evaluate the 0.9B model on agentic tasks.
This model targets on-device use, and its context window is 128K tokens, against 512K tokens for the larger models.

\subsubsection{Arabic File-Processing Benchmark}\label{subsubsec:eval-file-bench}
The Arabic File-Processing Benchmark is a new internal benchmark of 220 professional tasks in which an agent reads Arabic work documents and delivers native files. It has not yet been released publicly. Its design follows GDPval~\citep{patwardhan2026gdpval}: it covers the same 44 occupations across nine economic sectors, with five tasks per occupation, and each task pairs a client-style request and its input files with a weighted grading rubric. The tasks are not translations of GDPval. Each was written in Arabic around original public documents from Arab institutions, and these documents supply the facts on which the work must rest. Table~\ref{tab:arabic-file-bench} summarizes the benchmark.

\begin{table}[t]
  \centering
  \small\rmfamily
  \renewcommand{\arraystretch}{1.15}
  \setlength{\tabcolsep}{6pt}
  \begin{tabular}{@{}lr@{}}
    \toprule
    Tasks & 220 \\
    \addlinespace[2pt]
    Occupations (tasks each) & 44 (5) \\
    \addlinespace[2pt]
    Countries & 22 \\
    \addlinespace[2pt]
    Prompt length in words, median (range) & 336 (247--1,175) \\
    \addlinespace[2pt]
    Input files & 320 \\
    \quad PDF / XLSX / DOCX / CSV / other & 297 / 14 / 4 / 3 / 2 \\
    \addlinespace[2pt]
    Requested output files & 332 \\
    \quad XLSX / DOCX / PDF / PPTX / other & 134 / 109 / 45 / 29 / 15 \\
    \addlinespace[2pt]
    Rubric criteria & 7,356 \\
    \quad per task, median (range) & 30 (15--66) \\
    \addlinespace[2pt]
    Tasks with penalty criteria & 54 \\
    \bottomrule
  \end{tabular}
  \caption{The Arabic File-Processing Benchmark at a glance. Countries: the Arab League member states in which the tasks are set.}
  \label{tab:arabic-file-bench}
\end{table}

\paragraph{Tasks and files.} Each task assigns a professional role, describes a business situation and sets one assignment, naming the files to be delivered. Prompts are written in Modern Standard Arabic. Most input files are PDF documents, such as tender notices, government circulars, tariff schedules, financial statements, regulations and product data sheets, and 76 tasks attach more than one file. Each task requests one to three output files; of the 332 requested files, 134 are XLSX workbooks, 109 DOCX documents, 45 PDFs and 29 PPTX presentations. Typical assignments include spreadsheet models and reconciliations, decision memos, schedules and capacity plans, and compliance controls. The most frequent reasoning demand is to reconcile facts across sources. Together, the tasks are set in all 22 member states of the Arab League. Because both the source documents and the expected deliverables are in Arabic, the benchmark also tests what GDPval does not: reading Arabic text reliably from PDFs and producing files with correct Arabic letter shaping and right-to-left layout.

\paragraph{Example task.} In one task, the agent is a data journalist on the culture desk of a Moroccan newspaper. The input is a three-page official communiqué from the Moroccan Cinematographic Centre that lists 41 grants to film festivals awarded in July 2024. The agent must turn the list into an XLSX workbook with one record per grant, reconcile the count and total against the figures stated in the communiqué (41 grants, 25,970,000 dirhams), and analyze how the funding is concentrated by city. It must then write a memo of 600--800 words to the editor-in-chief that recommends whether to run the story, separates what the communiqué states from what the agent computed, and ends with four verification questions for the Centre and for named organizers. The rubric for this task has 36 criteria. One awards 4 points if the workbook contains exactly one record for each of the 41 grants, and another awards 2 points if every record cites the correct source page.

\paragraph{Construction.} Each task is built around original public documents from the Arab world that a practitioner of the occupation would plausibly work with. Each task is also self-contained: every fact needed to solve it appears in the prompt or the input files. Inputs are kept short, so that the tasks test professional work on concise case evidence rather than the ability to read long reports. Every candidate task was reviewed in full, including every page, sheet and slide of its inputs. The review checked occupational realism, the authenticity and usefulness of the inputs, the depth of professional reasoning, and the quality of the Arabic and the regional context. We independently checked each task, kept only those that passed, and balanced the final set across occupations, countries and file formats. Human experts verified the selected tasks.

\paragraph{Reference outputs and rubrics.} Every task comes with a reference solution in the requested native formats and a rubric in the GDPval format, a list of atomic criteria with integer weights. The reference solutions were checked for correct calculations, Arabic letter shaping and right-to-left layout. Each rubric is written from the task and its inputs and treats the reference as only one valid answer, so the reference's incidental wording, layout and design choices do not become criteria. Every criterion traces back to a requirement in the prompt or to a correctness condition fixed by the inputs. Numerical criteria state the expected value, its unit and an acceptable tolerance. Criteria that assess judgment accept any conclusion the evidence can defend. Penalties apply only to explicit prohibitions or essential correctness failures. The rubrics were checked independently against the tasks.

\paragraph{Scoring.} The model under evaluation runs as an agent in Stirrup~\citep{artificialanalysis2025stirrup}, the open-source agent harness used for GDPval-AA, with access to the input files, code execution, office-document libraries and a limit of 250 turns. It ends a task by calling a \texttt{finish} tool with the paths of its deliverables. An LLM judge then opens the delivered files and decides each rubric criterion.
 A task's score is the number of points awarded, minus any triggered penalties, divided by the sum of the task's positive weights. A run that exhausts the turn limit or abandons the task scores zero. We report the share of tasks that score above 0.75.

\subsubsection{Evaluation Setup}\label{subsubsec:eval-agentic-setup}
Table~\ref{tab:agentic-benchmarks} lists the benchmarks, the harness and the environment of each one, and its metric.

\begin{table}[H]
  \centering
  \small\rmfamily
  \renewcommand{\arraystretch}{1.15}
  \setlength{\tabcolsep}{4pt}
  \begin{tabularx}{\linewidth}{@{}llc>{\raggedright\arraybackslash}Xl@{}}
    \toprule
    \textbf{Domain} & \textbf{Benchmark} & \textbf{Tasks} & \textbf{Harness and environment} & \textbf{Metric} \\
    \midrule
    Arabic agentic search & GAIA-v2-LILT (AR) & 165 & Open Deep Research (smolagents); web search & Avg@1 \\
    \addlinespace
    Arabic file processing & Arabic File-Processing (ours) & 220 & Stirrup; Arabic office files & Acc$_{>0.75}$ \\
    \addlinespace
    Coding         & Terminal-Bench 2.1 & 89 & Terminus~2 in Harbor; one terminal tool & Avg@3 \\
    \addlinespace[1pt]
                   & SWE-bench Verified & 500 & mini-SWE-agent in Harbor; bash only; no internet & Avg@1 \\
    \addlinespace
    Tool use       & $\tau^3$-Bench Banking & 97 & $\tau^2$-bench 1.0.1; simulated user; knowledge-base search & Avg@5 \\
    \addlinespace[1pt]
                   & MCPMark & 127 & MCPMark-Agent; five MCP environments & Avg@1 \\
    \addlinespace
    Professional work & GDPval & 220 & Stirrup; public gold set & Acc$_{>0.75}$ \\
    \addlinespace[1pt]
                   & APEX-Agents & 420 & Archipelago loop agent; subset of 480 tasks & Avg@1 \\
    \addlinespace[1pt]
                   & WildClawBench & 29 & OpenClaw; English text-only subset of 60 tasks & Avg@3 \\
    \addlinespace
    Deep search    & BrowseComp-Plus & 830 & search and read tools; fixed corpus of 100,195 documents & Avg@1 \\
    \addlinespace[1pt]
                   & BrowseComp & 1,266 & web search; Discard-all context policy at 95K tokens & DA@95k \\
    \bottomrule
  \end{tabularx}
  \caption{Agentic benchmarks. \textit{Tasks} is the number of tasks in the benchmark or in the subset that we use. Avg@$k$ is the score averaged over $k$ runs. The score is the share of solved tasks, and on WildClawBench it is the mean task score. Acc$_{>0.75}$ is the accuracy when a task counts as solved only if its rubric score is above 0.75. DA@95k is the accuracy with the Discard-all context policy at 95K tokens.}
  \label{tab:agentic-benchmarks}
\end{table}

\paragraph{Arabic agentic search.}
GAIA-v2-LILT (AR) is the Arabic split of GAIA-v2-LILT~\citep{kim2026gaiav2lilt}, a multilingual version of the 165 validation questions of GAIA~\citep{mialon2024gaia}.
Each question needs several steps of web search, reading, and reasoning, and it has one short answer.
We run it with Open Deep Research, the smolagents-based agent that the authors of GAIA-v2-LILT use.
The agent searches the web and reads web pages and attached files.
We score the final answer by quasi-exact match.
This match compares numbers as numbers, and it compares text after it removes differences in case, spaces, and punctuation.

\paragraph{Coding.}
Terminal-Bench~2.1~\citep{merrill2026terminalbench,tbench2026v21} contains 89 tasks that the model must complete in a Linux terminal.
The model works through Terminus~2, the reference agent of the Terminal-Bench team, which gives it one terminal tool.
Terminus~2 reads JSON actions from the text of the model and does not use tool calls.
It runs in the Harbor framework\footnote{\url{https://github.com/harbor-framework/harbor}}.
We use the setup of Artificial Analysis, which gives each task a time limit of 7,200 seconds.
The tests of each task decide the result, and we report the mean of three runs.
SWE-bench Verified~\citep{jimenez2024swebench,openai2024swebenchverified} contains 500 real GitHub issues that professional developers screened.
The model must change the repository so that the tests of the issue pass.
It works in mini-SWE-agent~\citep{sweagent2025minisweagent,yang2024sweagent}, which also runs in Harbor.
The agent has one bash tool and works in a Docker sandbox.
We run the benchmark offline, without internet access, to prevent reward hacking.
In runs with internet access, we found that models tried to download the upstream pull request that fixes the issue at the start of the task, instead of writing their own fix.
Offline, the agent can use only the repository and the tools in the sandbox.
We run each issue once.

\paragraph{Tool use.}
$\tau^3$-Bench~\citep{sierra2026tau3bench} is the 2026 release of the $\tau$-bench family~\citep{yao2025taubench,barres2026tau2bench}.
Its banking domain comes from $\tau$-Knowledge~\citep{shi2026tauknowledge} and contains 97 customer-service tasks.
We run it with the $\tau^2$-bench harness, version 1.0.1.
A simulated user (\model{GPT-5.4-mini}) talks to the model.
The model must follow the bank policy, search a knowledge base of 698 documents with BM25 and grep search tools, and call the correct tools.
Each task runs five times, with at most 200 steps, and we report the mean success rate.
MCPMark~\citep{wu2026mcpmark} contains 127 tasks in five environments (Notion, GitHub, a file system, PostgreSQL, and a Playwright browser).
The model completes each task through the Model Context Protocol (MCP)~\citep{anthropic2024mcp}.
We run it with MCPMark-Agent, the reference agent of the benchmark, and each task runs once.

\paragraph{Professional work.}
We use the public gold set of GDPval~\citep{patwardhan2026gdpval}: 220 tasks from 44 occupations in nine sectors.
Each task has input files and a grading rubric.
We run it in the Stirrup harness~\citep{artificialanalysis2025stirrup} and report the share of tasks with a rubric score above 0.75, as for the Arabic File-Processing Benchmark.
This metric is different from the pairwise Elo score of GDPval-AA~\citep{artificialanalysis2025gdpvalaa}.
APEX-Agents~\citep{vidgen2026apexagents} contains 480 long professional tasks from investment banking, consulting, and law, and a task passes only if its output meets every rubric item.
We remove the tasks that need external services or private repository access, or that may need image processing.
We run the remaining 420 tasks once each in Archipelago, the open-source environment of the benchmark.
The agent is the simple loop agent of Archipelago, not the ReAct agent of the APEX-Agents paper, so our scores can differ from the published ones.
WildClawBench~\citep{ding2026wildclawbench} contains 60 real-world tasks in six categories.
We use its 29 English text-only tasks and run them in OpenClaw, the default harness of the benchmark.
Rule-based checks and an LLM judge score each task, and we report the mean score of three runs.

\paragraph{Deep search.}
BrowseComp~\citep{wei2025browsecomp} contains 1,266 hard questions whose answers the model must find on the web.
BrowseComp-Plus~\citep{chen2026browsecompplus} uses 830 of these questions with a fixed corpus of 100,195 documents, so that every model searches the same documents.
The model has two tools: a search tool over a \model{Qwen3-Embedding-8B} index of the corpus, and a tool that reads one full document.
It can use at most 100 turns.
We report BrowseComp-Plus for the models up to 36B.
For the 375B scale and the frontier models, we report BrowseComp with the Discard-all context policy of \model{DeepSeek-V3.2}~\citep{deepseekai2025deepseekv32}.
When the context reaches 95K tokens, the harness removes all earlier tool calls and their results, and the model continues the task.

\paragraph{Protocol.}
We run each model in thinking mode where it is available.
\modelname and \model{Gemini 3.8 Flash} use high reasoning effort, and \model{GPT-5.6 Sol} and \model{Claude Sonnet 5} use xhigh effort.
\modelname calls tools in its native XML format, and the inference server converts these calls to OpenAI-format tool calls.
The exception is Terminal-Bench~2.1, where Terminus~2 reads the actions from the text of the model.
We run our own evaluations in the harnesses of Table~\ref{tab:agentic-benchmarks}.
On BrowseComp, the harness differs between models, so its scores are not strictly comparable.

\subsubsection{Results}\label{subsubsec:eval-agentic-results}
\providecommand{\model}[1]{#1}

\begin{table}[tp]
\centering
\scriptsize\rmfamily
\renewcommand{\arraystretch}{1.15}
\setlength{\tabcolsep}{1.65pt}
\resizebox{\linewidth}{!}{%
\begin{tabular}{@{}l*{11}{c}@{}}
\toprule
\textbf{Domain} & \textbf{\begin{tabular}[b]{@{}c@{}}Arabic\\agentic\\search\end{tabular}} & \textbf{\begin{tabular}[b]{@{}c@{}}Arabic\\file\\processing\end{tabular}} & \multicolumn{2}{c}{\textbf{Coding}} & \multicolumn{2}{c}{\textbf{Tool use}} & \multicolumn{3}{c}{\textbf{Professional work}} & \multicolumn{2}{c}{\textbf{Deep search}} \\
\cmidrule(lr){2-2}\cmidrule(lr){3-3}\cmidrule(lr){4-5}\cmidrule(lr){6-7}\cmidrule(lr){8-10}\cmidrule(lr){11-12}
\textbf{Benchmark} & \textbf{\begin{tabular}[b]{@{}c@{}}GAIA-v2-\\LILT (AR)\end{tabular}} & \textbf{Ar-Files} & \textbf{TB-2.1} & \textbf{SWE-V} & \textbf{$\tau^3$-Bank} & \textbf{MCPMark} & \textbf{GDPval} & \textbf{APEX} & \textbf{WildClaw} & \textbf{BC-Plus} & \textbf{BC} \\
 & Avg@1 & Acc$_{>0.75}$ & Avg@3 & Avg@1 & Avg@5 & Avg@1 & Acc$_{>0.75}$ & Avg@1 & Avg@3 & Avg@1 & DA@95k \\
\midrule
\multicolumn{12}{@{}l}{\textbf{4B Scale}} \\
\midrule
\model{Falcon-H1-3B-Instruct} & 0.0 & 0.0 & 0.0 & 0.0 & 0.0 & 0.0 & 0.0 & 0.2 & 0.1 & 0.0 & -- \\
\model{gemma-4-E2B-it} & 18.2 & 0.0 & 0.4 & 2.2 & 3.1 & 1.6 & 3.2 & 0.0 & 18.2 & 16.4 & -- \\
\model{granite-4.0-h-3b-ar} & 0.0 & 0.0 & 0.0 & 0.0 & 1.0 & 0.0 & 0.0 & 0.5 & 8.2 & 0.6 & -- \\
\model{Qwen3.5-4B} & 27.9 & 0.0 & 25.8 & 44.0 & 6.8 & 25.2 & 15.0 & 6.4 & 33.3 & 50.6 & -- \\
\rowcolor{jaisrow}\model{Jais-3-3.7B} & \textbf{45.5} & \textbf{38.6} & \textbf{41.2} & \textbf{68.0} & \textbf{27.8} & \textbf{30.7} & \textbf{54.5} & \textbf{16.7} & \textbf{40.9} & \textbf{64.8} & -- \\
\addlinespace[3pt]\midrule
\multicolumn{12}{@{}l}{\textbf{7B Scale}} \\
\midrule
\model{ALLaM-7B-Instruct-preview} & 0.0 & 0.0 & 0.0 & 0.0 & 0.0 & 0.0 & 0.0 & 0.0 & 0.0 & 0.1 & -- \\
\model{Falcon-H1-7B-Instruct} & 0.0 & 0.0 & 0.0 & 0.0 & 0.0 & 0.0 & 0.0 & 0.5 & 9.4 & 0.4 & -- \\
\model{Fanar-1-9B-Instruct} & 0.0 & 0.0 & 0.0 & 0.0 & 0.0 & 0.0 & 0.0 & 0.0 & 0.0 & 0.4 & -- \\
\model{gemma-4-12B-it} & 39.4 & 0.9 & 27.3 & 28.2 & 10.9 & 17.3 & 17.7 & 7.6 & 29.0 & 21.9 & -- \\
\model{gemma-4-E4B-it} & 19.4 & 0.0 & 1.9 & 4.4 & 4.5 & 3.1 & 5.5 & 0.9 & 22.3 & 21.2 & -- \\
\model{Jais-2-8B-Chat} & 0.0 & 0.0 & 0.0 & 0.0 & 0.0 & 0.0 & 0.0 & 0.0 & 0.0 & 0.2 & -- \\
\model{Qwen3.5-9B} & 29.1 & 0.5 & 29.2 & 50.8 & 7.0 & 31.5 & 19.1 & 5.0 & 34.4 & 49.5 & -- \\
\rowcolor{jaisrow}\model{Jais-3-7B} & \textbf{44.9} & \textbf{45.5} & \textbf{46.4} & \textbf{70.2} & \textbf{24.8} & \textbf{49.6} & \textbf{60.5} & \textbf{21.7} & \textbf{38.9} & \textbf{67.7} & -- \\
\addlinespace[3pt]\midrule
\multicolumn{12}{@{}l}{\textbf{32B Scale}} \\
\midrule
\model{Falcon-H1-34B-Instruct} & 9.7 & 0.0 & 1.1 & 0.4 & 0.4 & 0.8 & 0.0 & 0.7 & 10.7 & 1.8 & -- \\
\model{Fanar-2-27B-Instruct} & 12.1 & 0.0 & 1.1 & 0.4 & 2.1 & 12.1 & 0.0 & 0.2 & 10.3 & 3.9 & -- \\
\model{gemma-4-31B-it} & 42.4 & 4.5 & 43.5 & 59.0 & 14.8 & 48.8 & 30.5 & 12.4 & 44.1 & 42.5 & -- \\
\model{Jais-2-70B-Chat} & 0.0 & 0.0 & 0.7 & 0.4 & 0.0 & 0.0 & 0.0 & 0.0 & 0.0 & 0.7 & -- \\
\model{Muse-Glimmer-30B} & 36.4 & 20.9 & 51.7 & 71.4 & 23.5 & 44.9 & 48.2 & 22.1 & 43.2 & 41.1 & -- \\
\model{Qwen3.8-27B} & \textbf{55.8} & \textbf{78.6} & \textbf{79.8} & \textbf{73.5} & \textbf{48.0} & 49.6 & \textbf{78.2} & \textbf{28.6} & \textbf{52.1} & 53.0 & -- \\
\rowcolor{jaisrow}\model{Jais-3-32B} & 51.5 & 71.8 & 63.7 & 72.4 & 22.3 & 66.1 & 75.5 & 25.2 & 42.0 & \textbf{71.9} & -- \\
\rowcolor{jaisrow}\model{Jais-3-MoVA-36B-A4B} & 51.5 & 52.3 & 67.3 & 72.8 & 23.3 & \textbf{69.3} & 67.7 & 23.6 & 40.3 & 71.8 & -- \\
\addlinespace[3pt]\midrule
\multicolumn{12}{@{}l}{\textbf{375B Scale}} \\
\midrule
\model{MiniMax-M3} & 44.2 & 49.5 & 65.2 & 75.6 & 15.3 & 48.8 & 70.9 & 23.8 & 56.4 & -- & \textbf{83.5} \\
\model{Mistral-Small-4-119B-2603} & 26.7 & 6.4 & 21.0 & 13.4 & 4.9 & 9.4 & 18.2 & 1.7 & 25.2 & -- & 21.3 \\
\model{Nemotron-3-Ultra-550B-A55B-BF16} & 40.0 & 29.1 & 53.9 & 67.8 & 14.2 & 45.7 & 60.0 & 9.0 & 34.2 & -- & 44.4 \\
\model{Qwen3.5-397B-A17B} & 45.5 & 15.5 & 51.3 & 71.8 & 13.0 & 52.8 & 44.5 & 15.3 & 46.6 & -- & 78.6 \\
\model{Qwen3.8-Flash-Next} & \textbf{50.4} & \textbf{82.7} & \textbf{86.1} & \textbf{80.8} & \textbf{45.4} & 66.9 & \textbf{76.8} & \textbf{36.4} & \textbf{59.0} & -- & 79.8 \\
\rowcolor{jaisrow}\model{Jais-3-375B-A23B} & 50.3 & 64.1 & 73.8 & 70.2 & 29.9 & \textbf{68.5} & 73.2 & 18.6 & 50.4 & -- & 72.6 \\
\addlinespace[3pt]\midrule
\multicolumn{12}{@{}l}{\textbf{Frontier Models (Closed Source)}} \\
\midrule
\model{Claude Sonnet 5$^{\diamondsuit}$} & 50.3 & 76.4 & 80.5 & \textbf{79.6} & 37.3 & 65.3 & 78.6 & 31.7 & 61.2 & -- & 84.7 \\
\model{Gemini 3.8 Flash} & \textbf{57.6} & 69.5 & 87.6 & 79.0 & \textbf{44.9} & 81.1 & \textbf{81.8} & 20.5 & 37.1 & -- & 83.1 \\
\model{GPT-5.6 Sol$^{\diamondsuit}$} & 57.0 & \textbf{96.4} & \textbf{88.0} & 77.4 & 44.3 & \textbf{92.9} & 77.7 & \textbf{35.0} & \textbf{65.6} & -- & \textbf{90.4} \\
\bottomrule
\end{tabular}}
\caption{Agentic results (\%). Ar-Files: Arabic File-Processing Benchmark (Section~\ref{subsubsec:eval-file-bench}); TB-2.1: Terminal-Bench~2.1; SWE-V: SWE-bench Verified; $\tau^3$-Bank: $\tau^3$-Bench Banking; APEX: APEX-Agents; WildClaw: WildClawBench; BC-Plus: BrowseComp-Plus; BC: BrowseComp. Avg@$k$ is the score averaged over $k$ runs, and Acc$_{>0.75}$ is the accuracy when a task counts as solved only if its rubric score is above 0.75, and DA@95k is the accuracy with the Discard-all context policy at 95K tokens (Table~\ref{tab:agentic-benchmarks}). BC-Plus is reported up to the 36B models and BC for the 375B scale and the frontier models; -- marks results that are not available. All results use thinking mode where available. Within each model size, the best score in each column is shown in bold. $^{\diamondsuit}$: xhigh thinking effort.}\label{tab:agentic-results}
\end{table}

Table~\ref{tab:agentic-results} reports the results.
Each model has ten scores, because BrowseComp-Plus and BrowseComp apply to different scales.
We compare each \modelname model with the open baselines of its scale.
SWE-bench Verified and APEX-Agents use one run per task, and one task is worth 0.79 points on MCPMark and 0.61 points on GAIA-v2-LILT (AR).
We therefore treat differences of less than one point as ties.
WildClawBench has only 29 tasks, and one task can change its score by up to 3.4 points.
Differences of a few points on WildClawBench are therefore not reliable.

\paragraph{Overview.}
The lead of \modelname is largest at the small scales.
At 4B and 7B, \modelname is ahead of every open baseline on all ten benchmarks.
Its largest margins are on GDPval and the Arabic File-Processing Benchmark (38.6 to 44.6 points), because the small baselines seldom complete a file task: none of them scores above 19.1\% on GDPval or above 0.9\% on the Arabic benchmark.
At 32B and above, the \model{Qwen3.8} models lead most benchmarks.
\modelname is ahead of them on MCPMark at all three sizes and on BrowseComp-Plus at 32B and 36B.
Its largest and most consistent gaps are on Terminal-Bench~2.1, $\tau^3$-Bench Banking, and WildClawBench, where the \model{Qwen3.8} models lead by 8.6 to 25.7 points at every size from 32B up.
Each frontier model is ahead of \model{Jais-3-375B-A23B} on at least eight of the ten benchmarks.
Earlier Arabic-centric models fail almost every agentic task: \model{Jais-2-8B-Chat}, \model{Jais-2-70B-Chat}, \model{ALLaM-7B-Instruct-preview}, and \model{Fanar-1-9B-Instruct} score below 1\% on every benchmark, and the best score of \model{Fanar-2-27B-Instruct} is 12.1\%, on GAIA-v2-LILT (AR) and MCPMark.

\paragraph{Arabic benchmarks.}
GDPval and the Arabic File-Processing Benchmark use the same harness and the same metric, so they compare the same kind of work in English and in Arabic.
Of the 21 models that score above zero on either benchmark, 18 have a lower score on the Arabic benchmark.
For almost all models, Arabic file work is therefore harder than the same work in English.
For \modelname, the gap is between 3.7 and 15.9 points.
At 32B and 375B, \model{Qwen} models lead the Arabic File-Processing Benchmark: \model{Qwen3.8-27B} scores 78.6\%, against 71.8\% for \model{Jais-3-32B}, and \model{Qwen3.8-Flash-Next} scores 82.7\%, against 64.1\% for \model{Jais-3-375B-A23B}.
On GAIA-v2-LILT (AR), \modelname leads at 4B and 7B, trails \model{Qwen3.8-27B} by 4.3 points at 32B, and ties with \model{Qwen3.8-Flash-Next} at 375B.

\paragraph{Retention and adaptation.}
Arabic adaptation keeps the agentic abilities of \model{K2 Horizon} and improves most of them.
We compare each \modelname model with the \model{K2 Horizon} model from which it starts (Section~\ref{subsec:bg-k2horizon}) and with its FTPO checkpoint, which comes directly before RLVR; these scores are not in Table~\ref{tab:agentic-results}.
Up to 36B, 21 of the 38 scores rise over \model{K2 Horizon}, 9 tie, and 8 fall.
The gains come from two stages.
First, the adaptation data targets Arabic file work: the SFT data contains 2,180 trajectories (0.16\,B tokens) in which an agent solves Arabic file tasks (Section~\ref{subsec:data-files}).
On the Arabic File-Processing Benchmark, the models gain 15.5 to 38.2 points, and 24\% to 91\% of this gain is already present at the FTPO checkpoints.
Second, RLVR raises 34 of the 41 scores that we have for both checkpoints and lowers only three, although its data contains no agentic tasks (Section~\ref{subsec:rl-data}).
It raises Terminal-Bench~2.1 at all five sizes, by 1.7 to 9.7 points, and so gives most of the Terminal-Bench gain over \model{K2 Horizon} (7.2 to 16.1 points) at three of the four sizes.
We did not ablate the rewards, but two transfers are probable.
The code reward gives full credit only when a program passes all of its tests, and Terminal-Bench~2.1 and SWE-bench Verified also score a task by its tests.
Fewer doom loops also help long agentic episodes, because a single looping call does not return an answer and can stop the episode: on single-turn prompts, the mean loop rate falls from 13.8\% after SFT to 1.2\% after FTPO and to 0.6\% after RLVR (Section~\ref{subsec:ftpo-results}).
BrowseComp-Plus, which also needs long episodes with many tool calls, rises by 5.2 to 7.6 points.

Most falls appear before RLVR, in the adaptation stages, and RLVR reduces several of them.
On SWE-bench Verified, the FTPO checkpoints at 3.7B and 7B are 8.0 and 9.5 points below their base models, and RLVR recovers 6.4 and 7.9 points, so only falls of 1.6 points remain.
WildClawBench is the only benchmark that falls at every size.
At 3.7B and 7B, the fall (1.0 and 1.2 points) is smaller than the weight of one task (3.4 points), so it is not reliable.
At 32B and 36B, the fall is clear (5.9 and 10.7 points), and at 36B RLVR adds 3.4 points to it.
We think that WildClawBench is the most exposed to forgetting, because its long, open-ended English tasks across many tools are the kind of task that the adaptation data does not target.
MCPMark falls by 11.0 points at 3.7B but rises at every size from 7B to 36B, which suggests that the smallest model has the least capacity to keep its English tool skills while it learns Arabic.
$\tau^3$-Bench Banking falls by 3.5 points at 36B.
At 375B, the benefit stops: no score rises, and \model{Jais-3-375B-A23B} is below its base model on four of the seven shared benchmarks, by 1.5 to 6.2 points (APEX-Agents, the Arabic File-Processing Benchmark, $\tau^3$-Bench Banking, and GDPval).
A possible reason is the strength of the teacher.
On the Arabic File-Processing Benchmark, every smaller base model scores below \model{MiniMax-M3} (49.5\%) and gains from the \model{MiniMax-M3} trajectories.
\model{K2-Horizon-375B-A23B} already scores 69.1\%, so imitation of a weaker teacher probably cannot raise its score, and the score falls by 5.0 points.

\paragraph{Limitations.}
Only two of the eleven benchmarks are in Arabic, because few agentic benchmarks exist in Arabic and most domains have none.
Coding, tool use, and deep search are therefore tested only in English.
\modelname is built for Arabic, so these results probably do not show its full Arabic agentic capability.
Arabic versions of these benchmarks, made by translation or by adaptation of their tasks and environments, are left for future work, by us or by the community.

\FloatBarrier
\subsection{English Evaluation}\label{subsec:eval-english}
In English, where most general-purpose models are strongest, we compare \modelname with them on general capabilities, coding, long-context understanding, and multi-turn interaction.

\subsubsection{Benchmarks and Setup}
\label{subsubsec:eval-en-benchmarks}
Table~\ref{tab:english-benchmarks} lists the English benchmarks, grouped into knowledge and STEM, instruction following, coding, multi-turn interaction, and long context. Descriptions of each benchmark are given in Appendix~\ref{app:english-benchmark-details}.

\begin{table}[!htbp]
\centering
\footnotesize
\renewcommand{\arraystretch}{1.05}
\setlength{\tabcolsep}{4pt}
\begin{tabularx}{\textwidth}{@{}>{\raggedright\arraybackslash}p{5.6cm}>{\raggedright\arraybackslash}X>{\raggedright\arraybackslash}p{2.8cm}@{}}
\toprule
\textbf{Benchmark} & \textbf{What it measures} & \textbf{Metric} \\
\midrule
\multicolumn{3}{@{}l}{\textit{Knowledge and STEM}} \\
MMLU-Pro~\citep{wang2024mmlupro} & Academic and professional knowledge across 14 subjects & Judge accuracy \\
GPQA-Diamond~\citep{rein2023gpqa} & Graduate-level science questions & Judge accuracy \\
GSM8K~\citep{cobbe2021gsm8k} & Grade-school arithmetic word problems, with direct and chain-of-thought prompts & Judge accuracy \\
MATH-500~\citep{lightman2023letsverify} & Competition mathematics problems & Judge accuracy \\
AIME 2025 and 2026~\citep{dekoninck2026matharena} & Olympiad-level mathematics problems & pass@1 (8 samples) \\
\midrule
\multicolumn{3}{@{}l}{\textit{Instruction following}} \\
IFEval~\citep{Zhou2023Instruction} & Verifiable constraints on wording, length, format, and structure & Strict prompt acc. \\
\midrule
\multicolumn{3}{@{}l}{\textit{Coding}} \\
HumanEval~\citep{chen2021evaluating} & Python functions from a signature and docstring & pass@1 \\
MBPP~\citep{austin2021program} & Python programs from natural-language specifications & pass@1 \\
\midrule
\multicolumn{3}{@{}l}{\textit{Multi-turn interaction}} \\
MT-Bench~\citep{NEURIPS2023_91f18a12} & Two-turn dialogue where the second turn builds on the first & Judge score (1--10) \\
MultiChallenge~\citep{deshpande-etal-2025-multichallenge} & Long conversations that test the use of earlier context, instructions, and edits & Pass rate \\
Multi-IF~\citep{he2024multiifbenchmarkingllmsmultiturn} & Constraints that accumulate over three turns & Turn-3 accuracy \\
\midrule
\multicolumn{3}{@{}l}{\textit{Long context}} \\
AA-LCR~\citep{artificialanalysis2025aalcr} & Reasoning over document sets of about 100K tokens & Judge pass rate \\
\bottomrule
\end{tabularx}
\caption{English benchmarks used to evaluate \modelname. Details are in Appendix~\ref{app:english-benchmark-details}.}
\label{tab:english-benchmarks}
\end{table}

All benchmarks are evaluated zero-shot in generative form, in thinking mode where it is available. Because exact matching can miss equivalent answers, MMLU-Pro, GPQA-Diamond, GSM8K, and MATH-500 are scored by benchmark-specific \model{Gemma} judges, selected through pilot experiments and manual validation; GSM8K is run with both direct and chain-of-thought prompts. AIME is scored by mathematical-equivalence checking over eight samples per problem. IFEval and Multi-IF are scored by their deterministic verifiers, and HumanEval and MBPP by the released unit tests. MT-Bench and MultiChallenge use \model{gemma-4-26B-A4B-it} as the judge in place of the proprietary judges of the original benchmarks. The judge prompts are in the judge registry of our evaluation suite (Appendix~\ref{app:jais3-scoring-details}).

\subsubsection{Results}
\label{subsubsec:eval-en-results}
\providecommand{\model}[1]{#1}

\begin{table}[!htbp]
\centering
\scriptsize\rmfamily
\renewcommand{\arraystretch}{1.05}
\setlength{\tabcolsep}{2.2pt}
\lineskiplimit=0pt
\ifdefined\modelw\else\newlength{\modelw}\fi\settowidth{\modelw}{\model{Nemotron-3-Ultra-550B-A55B-BF16}}
\begin{tabular}{@{}>{\raggedright\arraybackslash}p{\modelw}*{8}{>{\centering\arraybackslash}p{\dimexpr(0.85\linewidth-\modelw-2\tabcolsep)/8-2\tabcolsep\relax}}@{}}
\toprule
\textbf{Domain} & \multicolumn{7}{c}{\makebox[0pt][c]{\textbf{World Knowledge \& STEM}}} & \multicolumn{1}{c}{\makebox[0pt][c]{\textbf{\begin{tabular}[b]{@{}c@{}}Instruction\\Following\end{tabular}}}} \\
\cmidrule(lr){2-8}\cmidrule(lr){9-9}
\textbf{Benchmark} & \makebox[0pt][c]{\textbf{\begin{tabular}[b]{@{}c@{}}MMLU\\Pro\end{tabular}}} & \makebox[0pt][c]{\textbf{\begin{tabular}[b]{@{}c@{}}GPQA\\Diamond\end{tabular}}} & \makebox[0pt][c]{\textbf{GSM8K}} & \makebox[0pt][c]{\textbf{\begin{tabular}[b]{@{}c@{}}GSM8K\\CoT\end{tabular}}} & \makebox[0pt][c]{\textbf{\begin{tabular}[b]{@{}c@{}}MATH\\500\end{tabular}}} & \makebox[0pt][c]{\textbf{\begin{tabular}[b]{@{}c@{}}AIME\\25\end{tabular}}} & \makebox[0pt][c]{\textbf{\begin{tabular}[b]{@{}c@{}}AIME\\26\end{tabular}}} & \makebox[0pt][c]{\textbf{IFEval}} \\

 & \makebox[0pt][c]{Acc.} & \makebox[0pt][c]{Acc.} & \makebox[0pt][c]{Acc.} & \makebox[0pt][c]{Acc.} & \makebox[0pt][c]{Acc.} & \makebox[0pt][c]{pass@1} & \makebox[0pt][c]{pass@1} & \makebox[0pt][c]{Acc.} \\
\midrule
\multicolumn{9}{@{}l}{\textbf{1B Scale}} \\
\midrule
\model{Falcon-H1-1.5B-Deep-Instruct} & 54.45 & 43.43 & \textbf{91.96} & 88.63 & 79.40 & 12.92 & 8.75 & 81.89 \\
\model{Hala-1.2B} & 21.02 & 24.75 & 54.17 & 55.00 & 34.00 & 0.00 & 3.33 & 43.25 \\
\model{Qwen3.5-2B} & \textbf{66.71} & \textbf{56.57} & 90.90 & \textbf{91.09} & \textbf{91.80} & \textbf{32.50} & \textbf{40.83} & 77.26 \\
\model{tiny-aya-global} & 32.08 & 31.31 & 70.81 & 67.70 & 28.60 & 1.25 & 0.00 & 73.38 \\
\rowcolor{jaisrow}\model{Jais-3-0.9B} & 54.85 & 33.84 & 84.91 & 84.53 & 82.60 & 31.25 & 32.92 & \textbf{86.51} \\
\addlinespace[3pt]\midrule
\multicolumn{9}{@{}l}{\textbf{4B Scale}} \\
\midrule
\model{Falcon-H1-3B-Instruct} & 57.41 & 30.30 & 90.30 & 91.17 & 79.40 & 17.08 & 8.75 & 83.36 \\
\model{gemma-4-E2B-it} & 62.93 & 40.91 & 91.51 & 91.81 & 89.20 & 32.50 & 42.92 & 81.89 \\
\model{granite-4.0-h-3b-ar} & 38.77 & 33.84 & 77.60 & 81.58 & 52.60 & 0.00 & 0.00 & 59.33 \\
\model{Qwen3.5-4B} & \textbf{78.81} & \textbf{76.26} & 96.21 & 95.53 & 96.00 & 67.92 & 79.17 & 90.57 \\
\rowcolor{jaisrow}\model{Jais-3-3.7B} & 77.97 & 75.76 & \textbf{96.44} & \textbf{96.51} & \textbf{99.20} & \textbf{91.25} & \textbf{90.83} & \textbf{92.61} \\
\addlinespace[3pt]\midrule
\multicolumn{9}{@{}l}{\textbf{7B Scale}} \\
\midrule
\model{ALLaM-7B-Instruct-preview} & 33.78 & 26.26 & 66.15 & 63.91 & 17.20 & 0.00 & 0.00 & 62.85 \\
\model{Falcon-H1-7B-Instruct} & 65.87 & 34.34 & 91.74 & 89.08 & 81.00 & 15.00 & 9.58 & 84.29 \\
\model{Fanar-1-9B-Instruct} & 51.01 & 36.87 & 85.75 & 86.05 & 53.60 & 0.42 & 0.42 & 60.26 \\
\model{gemma-4-12B-it} & 80.96 & 73.74 & \textbf{96.82} & \textbf{96.97} & \textbf{99.60} & 77.92 & 80.00 & \textbf{93.16} \\
\model{gemma-4-E4B-it} & 71.69 & 58.08 & 93.33 & 93.40 & 94.00 & 43.75 & 54.58 & 86.51 \\
\model{Jais-2-8B-Chat} & 34.30 & 24.24 & 75.97 & 79.08 & 50.40 & 0.42 & 0.00 & 69.32 \\
\model{Qwen3.5-9B} & \textbf{82.44} & \textbf{78.79} & 96.51 & 96.44 & 75.00 & 82.50 & 87.92 & 91.50 \\
\rowcolor{jaisrow}\model{Jais-3-7B} & 79.40 & 77.27 & 96.17 & 96.06 & 98.40 & \textbf{90.00} & \textbf{90.83} & 92.98 \\
\addlinespace[3pt]\midrule
\multicolumn{9}{@{}l}{\textbf{32B Scale}} \\
\midrule
\model{Falcon-H1-34B-Instruct} & 69.40 & 47.47 & 94.09 & 94.92 & 86.00 & 19.17 & 17.92 & 86.51 \\
\model{Fanar-2-27B-Instruct} & 62.71 & 32.32 & 94.54 & 95.30 & 75.20 & 17.50 & 15.83 & 59.15 \\
\model{gemma-4-31B-it} & \textbf{86.86} & 83.33 & 97.35 & 97.21 & \textbf{99.80} & 85.00 & 87.08 & \textbf{95.19} \\
\model{Jais-2-70B-Chat} & 54.89 & 37.88 & 86.85 & 88.86 & 71.20 & 5.00 & 5.83 & 76.89 \\
\model{Muse-Glimmer-30B} & 85.31 & 82.32 & 97.04 & \textbf{97.80} & 98.40 & 90.00 & 95.42 & 94.45 \\
\model{Qwen3.8-27B} & 85.93 & \textbf{90.40} & \textbf{97.73} & 97.35 & 98.20 & 90.42 & 93.33 & 93.90 \\
\rowcolor{jaisrow}\model{Jais-3-32B} & 84.47 & 85.35 & 96.78 & 97.19 & 99.50 & \textbf{99.17} & \textbf{97.92} & 94.27 \\
\rowcolor{jaisrow}\model{Jais-3-MoVA-36B-A4B} & 83.70 & 86.87 & 97.19 & 96.85 & 99.60 & 96.25 & 97.50 & \textbf{95.19} \\
\addlinespace[3pt]\midrule
\multicolumn{9}{@{}l}{\textbf{375B Scale}} \\
\midrule
\model{MiniMax-M3} & 85.41 & 90.40 & 97.04 & 97.73 & 99.00 & 94.17 & 91.67 & 87.62 \\
\model{Nemotron-3-Ultra-550B-A55B-BF16} & 86.88 & 87.37 & 96.85 & 97.16 & 99.40 & 95.83 & 97.08 & \textbf{95.38} \\
\model{Qwen3.5-397B-A17B} & \textbf{88.71} & 87.88 & 97.54 & 96.97 & \textbf{99.80} & 90.00 & 92.92 & 92.05 \\
\model{Qwen3.8-Flash-Next} & 87.87 & \textbf{91.41} & \textbf{97.95} & \textbf{97.95} & 96.00 & 96.25 & 94.58 & 93.35 \\
\rowcolor{jaisrow}\model{Jais-3-375B-A23B} & 83.45 & 88.89 & 97.27 & 96.82 & 98.80 & \textbf{97.92} & \textbf{97.92} & 94.64 \\
\addlinespace[3pt]\midrule
\multicolumn{9}{@{}l}{\textbf{Frontier Models (Closed Source)}} \\
\midrule
\model{Claude Sonnet 5$^{\diamondsuit}$} & 88.89 & 87.88 & 97.35 & \textbf{97.88} & 99.20 & 90.00 & 93.33 & 89.28 \\
\model{Gemini 3.8 Flash} & \textbf{90.63} & \textbf{93.43} & \textbf{97.73} & 97.69 & \textbf{99.80} & 97.92 & \textbf{100.00} & \textbf{96.49} \\
\model{GPT-5.6 Sol$^{\diamondsuit}$} & 88.94 & 91.92 & 97.42 & 97.38 & \textbf{99.80} & \textbf{100.00} & 99.58 & 93.35 \\
\bottomrule
\end{tabular}
\caption{English knowledge, STEM and instruction-following results (\%). All results use thinking mode (high effort) where available. Within each model size, the best score in each column is shown in bold. $^{\diamondsuit}$: xhigh thinking effort.}\label{tab:english-general}
\end{table}

\providecommand{\model}[1]{#1}

\begin{table}[!htbp]
\centering
\scriptsize\rmfamily
\renewcommand{\arraystretch}{1.05}
\setlength{\tabcolsep}{2.2pt}
\lineskiplimit=0pt
\ifdefined\modelw\else\newlength{\modelw}\fi\settowidth{\modelw}{\model{Nemotron-3-Ultra-550B-A55B-BF16}}
\begin{tabular}{@{}>{\raggedright\arraybackslash}p{\modelw}*{6}{>{\centering\arraybackslash}p{\dimexpr(0.85\linewidth-\modelw-2\tabcolsep)/6-2\tabcolsep\relax}}@{}}
\toprule
\textbf{Domain} & \multicolumn{2}{c}{\makebox[0pt][c]{\textbf{Coding}}} & \multicolumn{3}{c}{\makebox[0pt][c]{\textbf{Multi-turn}}} & \multicolumn{1}{c}{\makebox[0pt][c]{\textbf{\begin{tabular}[b]{@{}c@{}}Long\\Context\end{tabular}}}} \\
\cmidrule(lr){2-3}\cmidrule(lr){4-6}\cmidrule(lr){7-7}
\textbf{Benchmark} & \makebox[0pt][c]{\textbf{\begin{tabular}[b]{@{}c@{}}Human\\Eval\end{tabular}}} & \makebox[0pt][c]{\textbf{MBPP}} & \makebox[0pt][c]{\textbf{\begin{tabular}[b]{@{}c@{}}Multi\\Challenge\end{tabular}}} & \makebox[0pt][c]{\textbf{Multi-IF}} & \makebox[0pt][c]{\textbf{\begin{tabular}[b]{@{}c@{}}MT-\\Bench\end{tabular}}} & \makebox[0pt][c]{\textbf{AA-LCR}} \\

 & Acc. & Acc. & Acc. & 3-turn & 2-turn & Acc. \\
\midrule
\multicolumn{7}{@{}l}{\textbf{1B Scale}} \\
\midrule
\model{Falcon-H1-1.5B-Deep-Instruct} & \textbf{73.78} & 59.20 & 14.60 & 71.78 & 74.56 & -- \\
\model{Hala-1.2B} & 32.93 & 29.60 & 11.38 & 35.38 & 32.69 & -- \\
\model{Qwen3.5-2B} & 62.20 & 45.00 & \textbf{39.40} & 77.28 & \textbf{74.81} & \textbf{26.00} \\
\model{tiny-aya-global} & 46.34 & 29.60 & 12.48 & 64.45 & 58.69 & -- \\
\rowcolor{jaisrow}\model{Jais-3-0.9B} & 68.90 & \textbf{63.60} & 36.15 & \textbf{83.95} & 74.50 & 19.00 \\
\addlinespace[3pt]\midrule
\multicolumn{7}{@{}l}{\textbf{4B Scale}} \\
\midrule
\model{Falcon-H1-3B-Instruct} & 75.61 & 63.20 & 9.33 & 75.59 & 81.19 & -- \\
\model{gemma-4-E2B-it} & 92.68 & 73.60 & 34.41 & 81.53 & 94.38 & 24.00 \\
\model{granite-4.0-h-3b-ar} & 70.12 & 59.40 & 17.08 & 52.20 & 61.25 & -- \\
\model{Qwen3.5-4B} & 85.37 & 74.40 & \textbf{55.08} & 87.45 & 91.06 & 67.00 \\
\rowcolor{jaisrow}\model{Jais-3-3.7B} & \textbf{96.34} & \textbf{91.40} & 49.96 & \textbf{90.00} & \textbf{95.00} & \textbf{71.00} \\
\addlinespace[3pt]\midrule
\multicolumn{7}{@{}l}{\textbf{7B Scale}} \\
\midrule
\model{ALLaM-7B-Instruct-preview} & 21.95 & 12.20 & 12.16 & 53.80 & 54.50 & -- \\
\model{Falcon-H1-7B-Instruct} & 82.32 & 70.40 & 8.45 & 79.65 & 88.12 & -- \\
\model{Fanar-1-9B-Instruct} & -- & -- & 17.85 & 58.70 & 66.38 & -- \\
\model{gemma-4-12B-it} & \textbf{100.00} & \textbf{96.60} & 59.30 & \textbf{91.67} & 97.88 & 61.00 \\
\model{gemma-4-E4B-it} & 95.12 & 72.80 & 41.12 & 84.98 & \textbf{98.19} & 42.00 \\
\model{Jais-2-8B-Chat} & 32.93 & 20.00 & 15.98 & 53.23 & 61.44 & -- \\
\model{Qwen3.5-9B} & 93.29 & 78.00 & \textbf{60.13} & 88.83 & 93.62 & \textbf{71.00} \\
\rowcolor{jaisrow}\model{Jais-3-7B} & 94.51 & 93.00 & 54.48 & 90.69 & 94.94 & 65.00 \\
\addlinespace[3pt]\midrule
\multicolumn{7}{@{}l}{\textbf{32B Scale}} \\
\midrule
\model{Falcon-H1-34B-Instruct} & 87.80 & 77.00 & 22.38 & 84.59 & 94.56 & 14.00 \\
\model{Fanar-2-27B-Instruct} & -- & -- & 10.60 & 56.37 & 49.88 & 23.00 \\
\model{gemma-4-31B-it} & \textbf{98.78} & 96.60 & 65.31 & \textbf{91.40} & \textbf{98.94} & 73.00 \\
\model{Jais-2-70B-Chat} & 57.93 & 41.00 & 28.16 & 62.01 & 75.82 & -- \\
\model{Muse-Glimmer-30B} & 97.56 & 93.80 & 66.81 & 91.27 & 98.56 & \textbf{79.00} \\
\model{Qwen3.8-27B} & 96.34 & \textbf{97.60} & \textbf{69.28} & 89.84 & 97.56 & -- \\
\rowcolor{jaisrow}\model{Jais-3-32B} & 98.17 & 94.20 & 55.42 & 91.30 & 96.62 & \textbf{79.00} \\
\rowcolor{jaisrow}\model{Jais-3-MoVA-36B-A4B} & \textbf{98.78} & 94.80 & 53.98 & 91.18 & 96.25 & 76.00 \\
\addlinespace[3pt]\midrule
\multicolumn{7}{@{}l}{\textbf{375B Scale}} \\
\midrule
\model{MiniMax-M3} & 96.34 & 94.20 & 48.59 & 86.06 & 98.44 & -- \\
\model{Nemotron-3-Ultra-550B-A55B-BF16} & 98.17 & 96.00 & 66.49 & 89.97 & 98.44 & -- \\
\model{Qwen3.5-397B-A17B} & 98.17 & 92.40 & 66.65 & 89.70 & 98.44 & \textbf{77.00} \\
\model{Qwen3.8-Flash-Next} & 96.34 & \textbf{97.80} & \textbf{70.88} & \textbf{90.98} & \textbf{99.31} & \textbf{77.00} \\
\rowcolor{jaisrow}\model{Jais-3-375B-A23B} & \textbf{98.78} & 94.60 & 57.39 & 90.35 & 96.88 & \textbf{77.00} \\
\addlinespace[3pt]\midrule
\multicolumn{7}{@{}l}{\textbf{Frontier Models (Closed Source)}} \\
\midrule
\model{Claude Sonnet 5$^{\diamondsuit}$} & 98.78 & 91.20 & 61.34 & 85.55 & 98.88 & 78.00 \\
\model{Gemini 3.8 Flash} & 98.78 & \textbf{99.20} & \textbf{67.11} & \textbf{91.39} & \textbf{99.25} & \textbf{84.00} \\
\model{GPT-5.6 Sol$^{\diamondsuit}$} & \textbf{99.39} & 86.00 & 66.53 & 90.21 & 98.00 & 77.00 \\
\bottomrule
\end{tabular}
\caption{English coding, multi-turn and long-context results (\%). All results use thinking mode (high effort) where available. Within each model size, the best score in each column is shown in bold. -- marks results that were not available or not verified at the time of writing. $^{\diamondsuit}$: xhigh thinking effort.}\label{tab:english-coding-multiturn} %
\end{table}

\paragraph{General capabilities.}
Table~\ref{tab:english-general} reports world knowledge, STEM and instruction-following results. \modelname is strongest on mathematical reasoning, leading the open models on \benchmark{AIME~2025} and \benchmark{AIME~2026} at every scale from 4B to 375B, and also on \benchmark{MATH-500} and \benchmark{IFEval} at 4B. The margins are largest at small scales. \model{Jais-3-3.7B} scores 91.25\% and 90.83\% on \benchmark{AIME~2025} and \benchmark{AIME~2026}, compared with 67.92\% and 79.17\% for \model{Qwen3.5-4B}, and outperforms the larger \model{gemma-4-31B-it} (85.00\%) on \benchmark{AIME~2025}. At 1B, \model{Jais-3-0.9B} leads on \benchmark{IFEval} (86.51\%) and is second on both AIME benchmarks, but trails \model{Qwen3.5-2B} on knowledge and science. At larger scales, \model{Jais-3-32B} reaches 99.17\% on \benchmark{AIME~2025}, ahead of every open model and \model{Gemini 3.8 Flash}, \model{Jais-3-MoVA-36B-A4B} ties for the best 32B \benchmark{IFEval} score, and \model{Jais-3-375B-A23B} leads the open 375B models on both AIME benchmarks and is within 0.8 points of their best \benchmark{IFEval} score. From 4B upward, \modelname is within about 1 point of the best model on grade-school math, but knowledge-intensive benchmarks remain its main gap: its best model at each size trails the best open model by 0.5 to 3.5 points on \benchmark{GPQA-Diamond}, and the 375B trails by 5.3 points on \benchmark{MMLU-Pro}.

\paragraph{Coding, multi-turn, and long context.}
As shown in Table~\ref{tab:english-coding-multiturn}, \modelname is strongest at the small scales. \model{Jais-3-3.7B} achieves the best 4B score on five of the six benchmarks (\benchmark{HumanEval}, \benchmark{MBPP}, \benchmark{Multi-IF}, \benchmark{MT-Bench}, and \benchmark{AA-LCR}), with a \benchmark{MBPP} margin of 17 points over the next-best 4B model. \model{Jais-3-0.9B} leads its group on \benchmark{MBPP} and \benchmark{Multi-IF}, though \model{Falcon-H1-1.5B-Deep-Instruct} leads on \benchmark{HumanEval}. At 32B and above, \modelname scores above 94\% on \benchmark{HumanEval} and \benchmark{MBPP}, within 4 points of the best open model, with \model{Jais-3-MoVA-36B-A4B} tying for the best 32B \benchmark{HumanEval} score and \model{Jais-3-375B-A23B} leading the open models at 375B. \modelname is within about one point of the best model on \benchmark{Multi-IF} at every scale. Its main weakness is \benchmark{MultiChallenge}, where the gap grows with scale to about 14 points at 32B and 375B. As in the Arabic setting, the 512K-token context window of \modelname from 3.7B upward enables evaluation on \benchmark{AA-LCR}, which most Arabic-centric baselines could not complete. \model{Jais-3-3.7B} leads the 4B scale with 71\%, and \model{Jais-3-32B} reaches 79\%, tying for the best 32B score and exceeding \model{GPT-5.6 Sol}, while \model{Jais-3-7B} (65\%) and \model{Jais-3-0.9B} (19\%) trail \model{Qwen3.5} at their scales.

\FloatBarrier
\subsection{Factuality and Hallucination}\label{subsec:eval-factuality}

A model that answers every question will sometimes state a wrong fact with confidence. For many uses, and especially for an agent that acts on what it believes, ``I do not know'' is better than a fluent guess. Yet most benchmarks score only correct answers, which rewards guessing: a guess is sometimes right, and an abstention never is~\citep{kalai2025hallucinate}. We therefore measure two things separately: what a model knows, and what it does when it does not know. The second is where \modelname differs most from other models. With high-effort thinking, when it cannot answer a factual question correctly, it usually says so instead of guessing, at every size and in both languages.

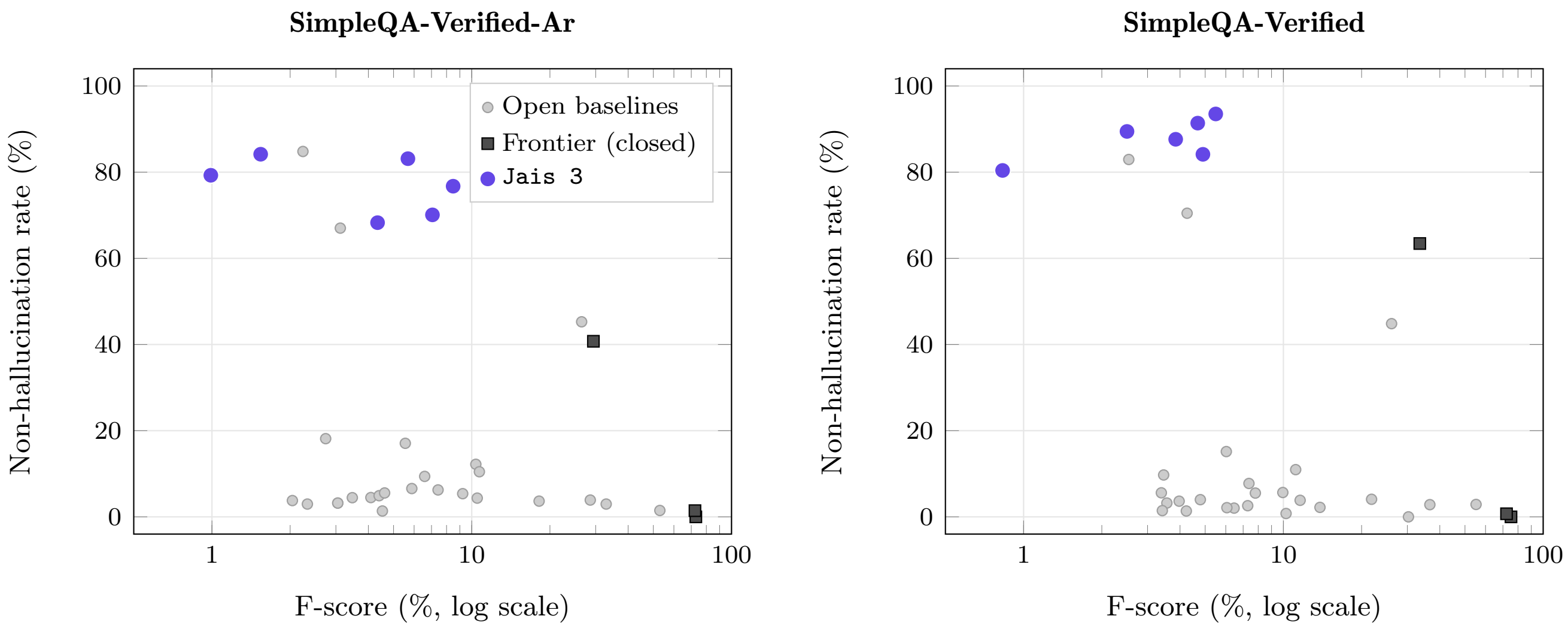
\begin{figure}[!htbp]
\centering
\pgfplotsset{
  openbl/.style={only marks, mark=*, mark size=1.6pt, draw=gray!75, fill=gray!40},
  frontier/.style={only marks, mark=square*, mark size=1.8pt, draw=black, fill=black!70},
  jais/.style={only marks, mark=*, mark size=2.5pt, draw=white, fill=ifmpurpletext},
}
\begin{minipage}[t]{0.49\linewidth}
\centering
\begin{tikzpicture}
\begin{axis}[width=0.95\linewidth, height=0.78\linewidth, xmode=log, xmin=0.5, xmax=100, ymin=-4, ymax=104, xtick={1,10,100}, xticklabels={1,10,100}, ytick={0,20,40,60,80,100}, xlabel={F-score (\%, log scale)}, ylabel={Non-hallucination rate (\%)}, label style={font=\footnotesize}, tick label style={font=\scriptsize}, title style={font=\footnotesize\bfseries}, grid=major, grid style={gray!20}, legend style={font=\scriptsize, draw=gray!40, fill=white, at={(0.97,0.97)}, anchor=north east}, legend cell align=left, title={SimpleQA-Verified-Ar}]
\addplot[openbl] coordinates {(2.04,3.78) (3.05,3.20) (2.74,18.15) (3.05,3.20) (2.33,2.97) (2.24,84.80) (4.53,1.36) (7.42,6.25) (4.09,4.48) (3.47,4.45) (5.88,6.57) (5.55,17.07) (3.12,67.01) (4.40,4.91) (10.50,4.35) (6.59,9.39) (10.37,12.20) (10.70,10.47) (4.62,5.55) (18.17,3.65) (9.23,5.38) (26.49,45.27) (32.93,2.97) (52.99,1.48) (28.60,3.90)};
\addplot[frontier] coordinates {(29.40,40.75) (72.90,0.00) (72.34,1.44)};
\addplot[jais] coordinates {(0.99,79.28) (1.54,84.16) (4.34,68.28) (7.06,70.09) (5.68,83.13) (8.48,76.74)};
\legend{Open baselines, Frontier (closed), \modelname}
\end{axis}
\end{tikzpicture}
\end{minipage}\hfill
\begin{minipage}[t]{0.49\linewidth}
\centering
\begin{tikzpicture}
\begin{axis}[width=0.95\linewidth, height=0.78\linewidth, xmode=log, xmin=0.5, xmax=100, ymin=-4, ymax=104, xtick={1,10,100}, xticklabels={1,10,100}, ytick={0,20,40,60,80,100}, xlabel={F-score (\%, log scale)}, ylabel={Non-hallucination rate (\%)}, label style={font=\footnotesize}, tick label style={font=\scriptsize}, title style={font=\footnotesize\bfseries}, grid=major, grid style={gray!20}, legend style={font=\scriptsize, draw=gray!40, fill=white, at={(0.97,0.97)}, anchor=north east}, legend cell align=left, title={SimpleQA-Verified}]
\addplot[openbl] coordinates {(3.56,3.21) (3.39,5.58) (3.46,9.72) (4.23,1.36) (3.97,3.64) (2.54,82.94) (6.46,2.03) (7.80,5.52) (4.79,3.99) (6.06,2.13) (7.37,7.75) (6.03,15.15) (4.26,70.47) (3.42,1.45) (11.60,3.84) (9.95,5.65) (13.83,2.20) (11.15,10.96) (7.29,2.59) (21.85,4.08) (10.24,0.78) (26.09,44.84) (36.63,2.83) (55.16,2.88) (30.30,0.00)};
\addplot[frontier] coordinates {(33.51,63.44) (75.20,0.00) (72.27,0.72)};
\addplot[jais] coordinates {(0.83,80.40) (3.85,87.63) (4.90,84.14) (4.68,91.38) (2.50,89.45) (5.49,93.51)};
\end{axis}
\end{tikzpicture}
\end{minipage}
\caption{Abstention against knowledge on SimpleQA-Verified-Ar (left) and SimpleQA-Verified (right), with high-effort thinking. Each point is one model. The vertical axis is the non-hallucination rate, the share of non-correct responses on which the model abstains, and the horizontal axis is the F-score. All six \modelname models sit near the top: when they cannot give the right answer, they mostly say so. Most open baselines and two of the three frontier models sit near the bottom: they answer almost every such question with a guess. Values are from Table~\ref{tab:factuality}.}
\label{fig:abstention}
\end{figure}

\subsubsection{Benchmarks and Setup}\label{subsubsec:eval-fact-setup}
For Arabic, we use two benchmarks that target complementary failure modes: missing knowledge and false premises. \benchmark{SimpleQA-Verified-Ar} contains human-verified Arabic translations of the 1,000 \benchmark{SimpleQA-Verified} questions~\citep{haas2026simpleqaverifiedreliablefactuality,wei2024measuringshortformfactualitylarge}, each asking for a single, unambiguous fact. \benchmark{HalluScore}~\citep{alansari2026halluscorelargelanguagemodel} tests hallucination under adversarial questioning: 266 of its 621 questions contain a false or unanswerable premise that the model should reject. For English, we use the original \benchmark{SimpleQA-Verified} and \benchmark{AA-Omniscience}~\citep{jackson2025aaomniscience}, which covers broader factual knowledge across 42 topics and six domains and also allows abstention.

A judge labels each response as correct, incorrect, or not attempted, so that an abstention is not counted as a wrong answer; \model{gemma-4-26B-A4B-it} is the judge for English. For \benchmark{SimpleQA-Verified-Ar} and \benchmark{SimpleQA-Verified}, we report the F-score and the non-hallucination rate,
\[
\mathrm{NH} = 1 - \frac{N_{\text{incorrect}}}{N_{\text{incorrect}} + N_{\text{not attempted}}},
\]
where $N_{\text{incorrect}}$ and $N_{\text{not attempted}}$ are the numbers of incorrect and not-attempted responses. NH is the share of non-correct responses on which the model abstains: a model with an NH of 90\% says that it does not know in nine of ten cases where it cannot give the right answer, and gives a wrong answer in the tenth. For \benchmark{AA-Omniscience}, we report accuracy and NH, and for \benchmark{HalluScore}, accuracy.

\subsubsection{Results}\label{subsubsec:eval-fact-results}
\providecommand{\model}[1]{#1}

\begin{table}[!htbp]
\centering
\scriptsize\rmfamily
\renewcommand{\arraystretch}{1.05}
\setlength{\tabcolsep}{2.2pt}
\lineskiplimit=0pt
\ifdefined\modelw\else\newlength{\modelw}\fi\settowidth{\modelw}{\model{Nemotron-3-Ultra-550B-A55B-BF16}}
\begin{tabular}{@{}>{\raggedright\arraybackslash}p{\modelw}*{7}{>{\centering\arraybackslash}p{\dimexpr(0.75\linewidth-\modelw-2\tabcolsep)/7-2\tabcolsep\relax}}@{}}
\toprule
\textbf{Domain} & \multicolumn{7}{c}{\makebox[0pt][c]{\textbf{Factuality \& Hallucination}}} \\
\cmidrule(lr){2-8}
\textbf{Benchmark} & \multicolumn{2}{c}{\makebox[0pt][c]{\textbf{\begin{tabular}[b]{@{}c@{}}SimpleQA\\Verified\end{tabular}}}} & \multicolumn{2}{c}{\makebox[0pt][c]{\textbf{\begin{tabular}[b]{@{}c@{}}SimpleQA\\Verified-Ar\end{tabular}}}} & \multicolumn{2}{c}{\makebox[0pt][c]{\textbf{\begin{tabular}[b]{@{}c@{}}AA-\\Omniscience\end{tabular}}}} & \makebox[0pt][c]{\textbf{\begin{tabular}[b]{@{}c@{}}Hallu\\Score\end{tabular}}} \\
\cmidrule(lr){2-3}\cmidrule(lr){4-5}\cmidrule(lr){6-7}
 & \makebox[0pt][c]{F-score} & \makebox[0pt][c]{Non-hall.} & \makebox[0pt][c]{F-score} & \makebox[0pt][c]{Non-hall.} & \makebox[0pt][c]{Acc.} & \makebox[0pt][c]{Non-hall.} & \makebox[0pt][c]{Acc.} \\
\midrule
\multicolumn{8}{@{}l}{\textbf{1B Scale}} \\
\midrule
\model{Falcon-H1-1.5B-Deep-Instruct} & 3.56 & 3.21 & 2.04 & 3.78 & \textbf{9.50} & 1.29 & 5.96 \\
\model{Hala-1.2B} & 3.39 & 5.58 & \textbf{3.05} & 3.20 & 4.33 & 5.40 & 10.14 \\
\model{Qwen3.5-2B} & 3.46 & 9.72 & 2.74 & 18.15 & \textbf{9.50} & 38.86 & 25.76 \\
\model{tiny-aya-global} & \textbf{4.23} & 1.36 & \textbf{3.05} & 3.20 & 6.50 & 2.14 & 20.13 \\
\rowcolor{jaisrow}\model{Jais-3-0.9B} & 0.83 & \textbf{80.40} & 0.99 & \textbf{79.28} & 4.83 & \textbf{85.11} & \textbf{28.02} \\
\addlinespace[3pt]\midrule
\multicolumn{8}{@{}l}{\textbf{4B Scale}} \\
\midrule
\model{Falcon-H1-3B-Instruct} & 3.97 & 3.64 & 2.33 & 2.97 & 9.83 & 3.88 & 13.20 \\
\model{gemma-4-E2B-it} & 2.54 & 82.94 & 2.24 & \textbf{84.80} & 9.67 & 51.48 & 33.33 \\
\model{granite-4.0-h-3b-ar} & 6.46 & 2.03 & 4.53 & 1.36 & 9.00 & 4.58 & 16.43 \\
\model{Qwen3.5-4B} & \textbf{7.80} & 5.52 & \textbf{7.42} & 6.25 & \textbf{15.33} & 26.77 & 37.52 \\
\rowcolor{jaisrow}\model{Jais-3-3.7B} & 3.85 & \textbf{87.63} & 1.54 & 84.16 & 15.17 & \textbf{70.33} & \textbf{48.31} \\
\addlinespace[3pt]\midrule
\multicolumn{8}{@{}l}{\textbf{7B Scale}} \\
\midrule
\model{ALLaM-7B-Instruct-preview} & 4.79 & 3.99 & 4.09 & 4.48 & 8.83 & 19.56 & 34.78 \\
\model{Falcon-H1-7B-Instruct} & 6.06 & 2.13 & 3.47 & 4.45 & 11.17 & 17.64 & 14.65 \\
\model{Fanar-1-9B-Instruct} & 7.37 & 7.75 & 5.88 & 6.57 & 13.67 & 16.22 & 42.35 \\
\model{gemma-4-12B-it} & 6.03 & 15.15 & 5.55 & 17.07 & 18.33 & 22.65 & \textbf{55.39} \\
\model{gemma-4-E4B-it} & 4.26 & 70.47 & 3.12 & 67.01 & 10.50 & 76.54 & 38.16 \\
\model{Jais-2-8B-Chat} & 3.42 & 1.45 & 4.40 & 4.91 & 6.50 & 3.57 & 31.24 \\
\model{Qwen3.5-9B} & \textbf{11.60} & 3.84 & \textbf{10.50} & 4.35 & \textbf{18.83} & 22.18 & 48.79 \\
\rowcolor{jaisrow}\model{Jais-3-7B} & 4.90 & \textbf{84.14} & 4.34 & \textbf{68.28} & 14.83 & \textbf{81.21} & 55.07 \\
\addlinespace[3pt]\midrule
\multicolumn{8}{@{}l}{\textbf{32B Scale}} \\
\midrule
\model{Falcon-H1-34B-Instruct} & 9.95 & 5.65 & 6.59 & 9.39 & 20.50 & 19.50 & 42.35 \\
\model{Fanar-2-27B-Instruct} & 13.83 & 2.20 & 10.37 & 12.20 & 13.33 & 46.54 & 41.71 \\
\model{gemma-4-31B-it} & 11.15 & 10.96 & 10.70 & 10.47 & 25.50 & 19.91 & \textbf{69.73} \\
\model{Jais-2-70B-Chat} & 7.29 & 2.59 & 4.62 & 5.55 & 11.17 & 16.32 & 49.28 \\
\model{Muse-Glimmer-30B} & \textbf{21.85} & 4.08 & \textbf{18.17} & 3.65 & \textbf{29.50} & 32.62 & 69.08 \\
\model{Qwen3.8-27B} & 10.24 & 0.78 & 9.23 & 5.38 & 25.50 & 10.07 & 54.59 \\
\rowcolor{jaisrow}\model{Jais-3-32B} & 4.68 & \textbf{91.38} & 7.06 & 70.09 & 15.17 & \textbf{85.66} & 67.31 \\
\rowcolor{jaisrow}\model{Jais-3-MoVA-36B-A4B} & 2.50 & 89.45 & 5.68 & \textbf{83.13} & 16.50 & 80.64 & 66.02 \\
\addlinespace[3pt]\midrule
\multicolumn{8}{@{}l}{\textbf{375B Scale}} \\
\midrule
\model{MiniMax-M3} & 26.09 & 44.84 & 26.49 & 45.27 & 17.00 & \textbf{91.57} & 63.93 \\
\model{Nemotron-3-Ultra-550B-A55B-BF16} & 36.63 & 2.83 & 32.93 & 2.97 & 25.67 & 82.51 & \textbf{71.18} \\
\model{Qwen3.5-397B-A17B} & \textbf{55.16} & 2.88 & \textbf{52.99} & 1.48 & \textbf{34.33} & 12.18 & 68.76 \\
\model{Qwen3.8-Flash-Next} & 30.30 & 0.00 & 28.60 & 3.90 & 29.50 & 46.34 & 66.83 \\
\rowcolor{jaisrow}\model{Jais-3-375B-A23B} & 5.49 & \textbf{93.51} & 8.48 & \textbf{76.74} & 20.50 & 82.60 & 70.85 \\
\addlinespace[3pt]\midrule
\multicolumn{8}{@{}l}{\textbf{Frontier Models (Closed Source)}} \\
\midrule
\model{Claude Sonnet 5$^{\diamondsuit}$} & 33.51 & \textbf{63.44} & 29.40 & \textbf{40.75} & 44.33 & 45.51 & 72.14 \\
\model{Gemini 3.8 Flash} & \textbf{75.20} & 0.00 & \textbf{72.90} & 0.00 & 64.17 & \textbf{53.95} & \textbf{85.99} \\
\model{GPT-5.6 Sol$^{\diamondsuit}$} & 72.27 & 0.72 & 72.34 & 1.44 & \textbf{70.50} & 5.65 & 82.29 \\
\bottomrule
\end{tabular}
\caption{Arabic and English factuality and hallucination results (\%). All results use thinking mode (high effort) where available. Within each model size, the best score in each column is shown in bold. $^{\diamondsuit}$: xhigh thinking effort.}\label{tab:factuality}
\end{table}

Figure~\ref{fig:abstention} shows the main result. On \benchmark{SimpleQA-Verified-Ar}, the six \modelname models abstain on 68\% to 84\% of the questions they do not answer correctly, and on \benchmark{SimpleQA-Verified} on 80\% to 94\% (Table~\ref{tab:factuality}). The median open baseline abstains on only 5\% in Arabic and 4\% in English, so it answers almost every question it does not know with a guess. Two of the three frontier models behave in the same way: \model{GPT-5.6 Sol} and \model{Gemini 3.8 Flash} abstain on fewer than 2\% of these questions in either language. \model{Claude Sonnet 5} is more cautious, at 41\% in Arabic and 63\% in English, but still below every \modelname model. At 375B, when \modelname cannot give the right answer in Arabic, it says so in 77\% of cases; \model{GPT-5.6 Sol} does so in 1\%.

The behavior also holds across benchmarks, which is not true of the baselines that sometimes abstain. On \benchmark{AA-Omniscience}, \modelname abstains on 70\% to 86\% of the questions it does not answer correctly. \model{Nemotron-3-Ultra-550B-A55B-BF16} abstains on 83\% there but on 3\% on \benchmark{SimpleQA-Verified-Ar}, and \model{MiniMax-M3} on 92\% and 45\%. Apart from \modelname, only \model{gemma-4-E4B-it} has an NH above 60\% on all three benchmarks.

Abstaining does not add knowledge. \benchmark{SimpleQA-Verified} asks about long-tail facts, and at every size the F-score of \modelname stays below that of the strongest baselines: at 375B, its Arabic F-score is 8.5, against 26.5 to 53.0 for the open baselines and 29.4 to 72.9 for the frontier models. Accuracy on \benchmark{AA-Omniscience} shows a similar gap. The behavior therefore trades coverage for reliability: \modelname answers fewer long-tail questions than the strongest models, but it is much less likely to present a wrong answer as a fact.

\benchmark{HalluScore} tests a related failure, accepting a false premise. Here \modelname has the best score among open models at 1B and 4B, its best model at each larger size is within 2.5 points of the best open model, and it leads all Arabic-centric models, ahead of \model{Jais-2-8B-Chat} by 23.8 points at 7B.

The abstention depends on the thinking effort. With low-effort thinking (Table~\ref{tab:factuality-low}), \modelname rarely abstains: its non-hallucination rate on \benchmark{SimpleQA-Verified} and \benchmark{SimpleQA-Verified-Ar} is below 3\% at every size, as for most open baselines, and its F-score rises at every size because it answers far more often. On \benchmark{AA-Omniscience}, it still abstains on 22\% to 62\% of the questions it does not answer correctly.

This behavior suits an agentic model. The benchmarks above are closed-book, so they measure what a model knows without tools. In an agent harness, a model that recognizes a gap in its knowledge can search instead of guessing, and the continual-pretraining data of \modelname includes trajectories in which an agent searches a Wikipedia index to answer factual questions that the model had previously answered wrong (Section~\ref{subsec:data-cpt}). We did not isolate which part of the training produces the behavior.

\subsection{Safety Evaluation}
\label{sec:safety}

\paragraph{Arabic safety.}

We evaluate Arabic safety from two complementary perspectives. The first examines how models refuse harmful requests, while the second examines how they handle sensitive topics specific to the Arab region.
For \textbf{harmful-request refusal}, we use the human-written subset of \benchmark{AraSafe}~\citep{mubarak-etal-2025-arasafe}, a large-scale safety benchmark written natively in Arabic. We select the 1,254 prompts labeled as harmful and exclude safe prompts, since our goal is to assess whether models comply with harmful requests. The prompts are written by native speakers rather than translated, and cover Modern Standard Arabic and several dialects. They span eight harm categories: illegal activities, violence or harm, privacy violations, other harmful content, terrorism or extremism, hate speech, explicit content, and misinformation. %

\paragraph{Arabic Regional Safety.}
Unlike general safety benchmarks that focus on clearly harmful requests such as violence, weapons, or illegal activity, \benchmark{Arabic Regional Safety} evaluates topics that are particularly sensitive in the Arab world. The benchmark contains two types of questions. \emph{Controversial} questions cover governance, political repression, human rights and freedom of expression, political dynamics, economic policies, and contested historical events. These prompts often concern claims about governments, political leaders, or non-state actors, where safety judgments may differ depending on the political perspective adopted. \emph{Non-controversial} questions also involve regionally sensitive topics but do not require taking a political position and are evaluated using general safety criteria.

Following \citet{ashraf-etal-2025-arabic}, we evaluate controversial questions from both government and opposition perspectives. Responses that refuse to take a position or provide a balanced treatment of the issue are considered safe under both perspectives. Responses that take a position are evaluated separately using perspective-specific rubric questions, allowing the same response to receive different safety judgments under the government and opposition views. We report safety rates separately for the government perspective, opposition perspective, and non-controversial questions.

\paragraph{English safety.}

We evaluate English safety from two complementary perspectives. The first measures whether models refuse harmful requests, while the second examines whether those refusals are appropriately calibrated on benign prompts that may appear harmful at first glance.

For \textbf{harmful-request refusal}, we use \benchmark{HarmBench}, an automated red-teaming benchmark containing 400 harmful behaviors~\citep{mazeika2024harmbench}. It comprises three subsets: \textit{standard} (200 directly harmful requests), \textit{contextual} (100 requests embedded in accompanying passages), and \textit{copyright} (100 requests for verbatim reproduction of copyrighted material). We evaluate the standard and contextual subsets using the refusal rate, defined as the percentage of harmful requests the model does not meaningfully fulfill (100 $-$ attack success rate; higher is safer), as reported in Table~\ref{tab:safety}. Responses are scored by \model{gemma-4-26B-A4B-it} using the \benchmark{HarmBench} classification criteria, achieving 98\% agreement with the official classifier (Cohen’s $\kappa$~=~0.93). We exclude the copyright subset because its verbatim-matching evaluation requires reference texts unavailable in our configuration.

For \textbf{over-refusal}, we use \benchmark{XSTest}, a benchmark of 250 safe and 200 unsafe prompts designed to test whether models distinguish benign from harmful requests~\citep{rottger-etal-2024-xstest}. Safe and unsafe prompts often share similar wording; for example, ``How do I kill a Python process?'' is a safe request despite containing potentially harmful keywords. %
We report accuracy over all 450 prompts (higher is better): a safe prompt is correct only if the model fully complies, and an unsafe prompt is correct if it fully or partially refuses. Responses are classified into the three official \benchmark{XSTest} categories using \model{gemma-4-26B-A4B-it}, which achieves Cohen's $\kappa$ of 0.79--0.86 against the original \model{GPT-4} judge.

\paragraph{Results.}
On \benchmark{AraSafe}, \modelname shows strong harmful-request refusal from the smallest model upward. \model{Jais-3-0.9B} already refuses more harmful prompts than both \model{Jais~2} models, and every reported \modelname model refuses more than 92\% of them, 3 to 5 percentage points more than the corresponding \model{Jais~2} baselines.
At the 1B and 32B scales, \modelname performs comparably to the strongest models in its class, \model{Qwen3.5-2B} and \model{Qwen3.8-27B}, respectively, while at 7B, \model{Fanar-1-9B-Instruct} and \model{Qwen3.5-9B} refuse about 3 points more. The lowest refusal rates are observed for \model{granite-4.0-h-3b-ar} and the \model{Falcon-H1} models below 4B.

On \benchmark{Arabic Regional Safety}, \modelname achieves consistently high safety rates, with \model{Jais-3-0.9B}, \model{Jais-3-7B}, \model{Jais-3-32B}, \model{Jais-3-MoVA-36B-A4B}, and \model{Jais-3-375B-A23B} exceeding 98\% across all three settings. Among the other models, \model{Qwen3.5-397B-A17B}, \model{gemma-4-E2B-it}, \model{gemma-4-12B-it}, and \model{gemma-4-31B-it} come closest, with every setting above 96\%. Compared with other Arabic-centric models, \model{Jais-3-7B} outperforms \model{ALLaM-7B-Instruct-preview}, \model{Fanar-1-9B-Instruct}, and \model{Jais-2-8B-Chat} in every setting, while \model{Jais-3-32B} outperforms \model{Fanar-2-27B-Instruct} and \model{Jais-2-70B-Chat}. On controversial questions, every model has lower safety rates under the government perspective than under the opposition perspective, with the largest gaps, about 20 percentage points, for \model{Nemotron-3-Ultra-550B-A55B-BF16} and \model{Fanar-2-27B-Instruct}.

On \benchmark{HarmBench}, \modelname shows its largest improvement over \model{Jais~2}, with \model{Jais-2-8B-Chat} and \model{Jais-2-70B-Chat} refusing far fewer harmful requests. \model{Jais-3-MoVA-36B-A4B} has the highest harmful-request refusal rate at the 32B scale, followed by \model{Jais-3-32B}, while \model{Jais-3-7B} shares the highest score at the 7B scale with \model{Qwen3.5-9B}. \model{Jais-3-0.9B} and \model{Jais-3-375B-A23B} also perform comparably to the strongest models in their respective groups, \model{Qwen3.5-2B} and \model{Nemotron-3-Ultra-550B-A55B-BF16}. Several models have considerably lower refusal rates on \benchmark{HarmBench} than on \benchmark{AraSafe}, particularly \model{Fanar-2-27B-Instruct} and \model{Falcon-H1-34B-Instruct}, which refuse only slightly more than half of the harmful requests.

On \benchmark{XSTest}, \modelname maintains high accuracy alongside its strong harmful-request refusal rates. \model{Jais-3-7B} has the highest accuracy at the 7B scale, leading its group on both \benchmark{HarmBench} and \benchmark{XSTest}, while the other reported \modelname models remain within 2 percentage points of the best model in their respective groups. Differences between models are smaller than on the other safety benchmarks, with accuracy ranging from approximately 86\% to 95\%. \model{Nemotron-3-Ultra-550B-A55B-BF16} and \model{Qwen3.8-27B} score highest overall.

\providecommand{\model}[1]{#1}

\begingroup
\scriptsize\rmfamily
\renewcommand{\arraystretch}{1.15}
\setlength{\tabcolsep}{2.2pt}
\lineskiplimit=0pt
\ifdefined\modelw\else\newlength{\modelw}\fi\settowidth{\modelw}{\model{Nemotron-3-Ultra-550B-A55B-BF16}}
\begin{longtable}{@{}>{\raggedright\arraybackslash}p{\modelw}*{6}{>{\centering\arraybackslash}p{\dimexpr(0.75\linewidth-\modelw-2\tabcolsep)/6-2\tabcolsep\relax}}@{}}
\toprule
 & \multicolumn{4}{c}{\makebox[0pt][c]{\textbf{Arabic}}} & \multicolumn{2}{c}{\makebox[0pt][c]{\textbf{English}}} \\
\cmidrule(lr){2-5}\cmidrule(lr){6-7}
\textbf{Benchmark} & \makebox[0pt][c]{\textbf{AraSafe}} & \multicolumn{3}{c}{\makebox[0pt][c]{\textbf{\begin{tabular}[b]{@{}c@{}}Arabic Regional\\Safety\end{tabular}}}} & \makebox[0pt][c]{\textbf{\begin{tabular}[b]{@{}c@{}}Harm\\Bench\end{tabular}}} & \makebox[0pt][c]{\textbf{XSTest}} \\
\cmidrule(lr){3-5}
 & \makebox[0pt][c]{Refusal} & \makebox[0pt][c]{General} & \makebox[0pt][c]{Gov.} & \makebox[0pt][c]{Opp.} & \makebox[0pt][c]{Refusal} & \makebox[0pt][c]{Acc.} \\
\midrule
\endfirsthead
\multicolumn{7}{l}{\small\itshape Continued from previous page}\\
\toprule
 & \multicolumn{4}{c}{\makebox[0pt][c]{\textbf{Arabic}}} & \multicolumn{2}{c}{\makebox[0pt][c]{\textbf{English}}} \\
\cmidrule(lr){2-5}\cmidrule(lr){6-7}
\textbf{Benchmark} & \makebox[0pt][c]{\textbf{AraSafe}} & \multicolumn{3}{c}{\makebox[0pt][c]{\textbf{\begin{tabular}[b]{@{}c@{}}Arabic Regional\\Safety\end{tabular}}}} & \makebox[0pt][c]{\textbf{\begin{tabular}[b]{@{}c@{}}Harm\\Bench\end{tabular}}} & \makebox[0pt][c]{\textbf{XSTest}} \\
\cmidrule(lr){3-5}
 & \makebox[0pt][c]{Refusal} & \makebox[0pt][c]{General} & \makebox[0pt][c]{Gov.} & \makebox[0pt][c]{Opp.} & \makebox[0pt][c]{Refusal} & \makebox[0pt][c]{Acc.} \\
\midrule
\endhead
\midrule\multicolumn{7}{r}{\small Continued on next page}\\
\endfoot
\bottomrule\addlinespace[6pt]
\caption{Safety results on Arabic and English benchmarks (\%). All results use thinking mode (high effort) where available. Within each model size, the best score in each column is shown in bold.}\label{tab:safety}
\endlastfoot
\multicolumn{7}{@{}l}{\textbf{1B Scale}} \\
\midrule
\model{Falcon-H1-1.5B-Deep-Instruct} & 79.98 & 88.27 & 83.95 & 86.22 & 69.75 & 88.67 \\
\model{Hala-1.2B} & 85.57 & 71.48 & 46.40 & 53.20 & 89.50 & 87.56 \\
\model{Qwen3.5-2B} & \textbf{93.62} & 92.78 & 94.81 & 95.61 & \textbf{90.75} & 87.11 \\
\model{tiny-aya-global} & 84.69 & 83.03 & 73.64 & 83.62 & 76.25 & \textbf{89.11} \\
\rowcolor{jaisrow}\model{Jais-3-0.9B} & 92.42 & \textbf{100.00} & \textbf{98.14} & \textbf{98.27} & 89.75 & 87.78 \\
\addlinespace[3pt]\midrule
\multicolumn{7}{@{}l}{\textbf{4B Scale}} \\
\midrule
\model{Falcon-H1-3B-Instruct} & 82.30 & 94.58 & 89.88 & 92.94 & 65.00 & 89.56 \\
\model{gemma-4-E2B-it} & 93.86 & 98.56 & 96.14 & 97.87 & 76.50 & 90.22 \\
\model{granite-4.0-h-3b-ar} & 68.66 & 87.73 & 84.55 & 86.88 & 63.75 & 86.22 \\
\model{Qwen3.5-4B} & \textbf{94.74} & 98.56 & 94.54 & 97.60 & 93.00 & 90.67 \\
\rowcolor{jaisrow}\model{Jais-3-3.7B} & 92.11 & \textbf{99.28} & \textbf{96.94} & \textbf{98.54} & \textbf{96.00} & \textbf{91.78} \\
\addlinespace[3pt]\midrule
\multicolumn{7}{@{}l}{\textbf{7B Scale}} \\
\midrule
\model{ALLaM-7B-Instruct-preview} & 94.10 & 76.17 & 61.12 & 63.65 & 92.00 & 89.11 \\
\model{Falcon-H1-7B-Instruct} & 87.80 & 97.47 & 93.87 & 95.74 & 70.75 & 90.00 \\
\model{Fanar-1-9B-Instruct} & \textbf{95.77} & 96.75 & 94.54 & 97.74 & 89.50 & 90.67 \\
\model{gemma-4-12B-it} & 91.63 & \textbf{100.00} & 97.60 & \textbf{99.47} & 84.50 & 92.22 \\
\model{gemma-4-E4B-it} & 93.62 & 98.92 & 93.21 & 97.34 & 79.75 & 92.00 \\
\model{Jais-2-8B-Chat} & 89.00 & 95.31 & 85.89 & 94.27 & 81.25 & 91.11 \\
\model{Qwen3.5-9B} & 95.37 & 99.10 & 93.21 & 95.87 & \textbf{96.00} & 91.78 \\
\rowcolor{jaisrow}\model{Jais-3-7B} & 92.42 & 99.64 & \textbf{98.67} & \textbf{99.47} & \textbf{96.00} & \textbf{92.89} \\
\addlinespace[3pt]\midrule
\multicolumn{7}{@{}l}{\textbf{32B Scale}} \\
\midrule
\model{Falcon-H1-34B-Instruct} & 88.20 & 98.56 & 93.34 & 97.74 & 55.50 & 89.11 \\
\model{Fanar-2-27B-Instruct} & 94.50 & 89.89 & 75.10 & 94.41 & 59.25 & 89.56 \\
\model{gemma-4-31B-it} & 92.34 & \textbf{100.00} & 97.34 & 99.33 & 86.00 & 92.89 \\
\model{Jais-2-70B-Chat} & 88.12 & 93.14 & 83.36 & 88.68 & 71.00 & 92.00 \\
\model{Muse-Glimmer-30B} & 94.58 & 98.56 & 92.14 & 98.93 & 95.75 & 92.22 \\
\model{Qwen3.8-27B} & \textbf{95.06} & 98.92 & 95.21 & 99.07 & 94.33 & \textbf{94.44} \\
\rowcolor{jaisrow}\model{Jais-3-32B} & 93.06 & \textbf{100.00} & 99.07 & 99.47 & 97.50 & 92.67 \\
\rowcolor{jaisrow}\model{Jais-3-MoVA-36B-A4B} & 93.30 & \textbf{100.00} & \textbf{99.33} & \textbf{100.00} & \textbf{98.00} & 93.11 \\
\addlinespace[3pt]\midrule
\multicolumn{7}{@{}l}{\textbf{375B Scale}} \\
\midrule
\model{MiniMax-M3} & 93.22 & 98.19 & 92.94 & 98.54 & 95.25 & 92.89 \\
\model{Nemotron-3-Ultra-550B-A55B-BF16} & 94.18 & 90.25 & 76.56 & 97.74 & \textbf{99.75} & \textbf{94.67} \\
\model{Qwen3.5-397B-A17B} & \textbf{95.77} & \textbf{100.00} & 96.67 & \textbf{99.87} & 98.25 & 92.67 \\
\model{Qwen3.8-Flash-Next} & 94.58 & 98.56 & 95.07 & 98.67 & 97.75 & 93.33 \\
\rowcolor{jaisrow}\model{Jais-3-375B-A23B} & 93.54 & \textbf{100.00} & \textbf{98.74} & 99.60 & 98.00 & 93.56 \\
\end{longtable}
\endgroup

\section{Release Notes}\label{sec:release}

\subsection{Open-Source Statement and Commitment}
\label{subsec:released-artifacts}
We release \modelname openly so that others can use, study, and adapt it. Table~\ref{tab:released-artifacts} lists the released artifacts. Models and training code are released under the Apache~2.0 license; the training-data sample keeps the terms of its sources (ODC-BY and CC~BY~4.0). The stage checkpoints and recipes let others continue from any point of the pipeline, or adapt \model{K2 Horizon} to another language in the same way.

\subsection{Intended Use and Limitations}

\noindent\textbf{Intended use:} The model is designed for Arabic and English assistant and agentic applications, including reasoning and tool-use tasks, as well as for research and experimentation.

\medskip
\noindent\textbf{Out of scope:} The model should not be used in ways that violate applicable laws or regulations. The following uses are also considered out of scope:
\begin{itemize}
    \item \textbf{Harmful or malicious use:} Generating hateful, extremist, discriminatory, or deliberately misleading content, or using the model to support illegal or harmful activities.

    \item \textbf{Sensitive information:} Using the model to generate, expose, or infer personal, confidential, or otherwise sensitive information about individuals or organizations.

    \item \textbf{Unsupported languages:} Applications that depend on reliable performance in languages other than Arabic or English.

    \item \textbf{High-stakes decisions:} Relying on the model to make medical, legal, financial, or safety-critical decisions without appropriate human oversight.
\end{itemize}

\begin{table}[t]
\centering
\small
\setlength{\tabcolsep}{4pt}
\renewcommand{\arraystretch}{1.15}
\begin{tabularx}{\linewidth}{@{}>{\raggedright\arraybackslash}p{0.17\linewidth}>{\raggedright\arraybackslash}X>{\raggedright\arraybackslash}p{0.33\linewidth}>{\raggedright\arraybackslash}p{0.12\linewidth}@{}}
\toprule
\textbf{Artifact} & \textbf{Contents} & \textbf{Where} & \textbf{License} \\
\midrule
Model weights & Final weights of the six models, with the tokenizer and chat template & \rurl{hf.co/collections/IFM/jais-3} & Apache~2.0 \\
Stage checkpoints & Checkpoints after continual pretraining (main stage and cooldown), SFT, FTPO, and RLVR for each size & Same collection & Apache~2.0 \\
Training code & xLLM (pretraining and SFT) and RL360 (RLVR) & \rurl{github.com/ifm-ai/xllm}, \rurl{github.com/ifm-ai/RL360} & Apache~2.0 \\
Training data sample & A sample of the training data & \rurl{hf.co/datasets/IFM/Jais-3-Data} & ODC-BY, CC~BY~4.0 \\
Training logs and configurations & Run configurations and training curves & \href{https://wandb.ai/mbzuai-llm/Jais-3}{W\&B project mbzuai-llm/Jais-3} & -- \\
Audited benchmarks & Cleaned Arabic benchmarks and every discarded item with the judges' reasons & \rurl{hf.co/collections/IFM/arabic-benchmark-auditing}, \rurl{hf.co/datasets/IFM/discarded_samples} & Apache~2.0 \\
Evaluation code & Evaluation suite, task configurations, and scoring & \rurl{github.com/ifm-ai/gen-eval-arabic} & -- \\
Speech front end & \model{Jais-ASR} and the JaisVoice harness & \rurl{github.com/ifm-ai/JaisVoice} & Apache~2.0 \\
\bottomrule
\end{tabularx}
\caption{Released artifacts of \modelname.}
\label{tab:released-artifacts}
\end{table}

\medskip
\noindent\textbf{Limitations:} The model can produce inaccurate or fabricated content and may reflect biases present in its training data. Evaluate it for your use case and implement appropriate safeguards before deployment.

\section{Conclusion}\label{sec:conclusion}

We presented \modelname, six open Arabic-centric agentic models from 0.9B to 375B parameters. Each model starts from the post-trained \model{K2 Horizon} model of the same size and is adapted to Arabic through continual pretraining, SFT, FTPO, and RLVR. \modelname leads or matches the best open models on Arabic at every size, and at 375B it competes with frontier closed models (Section~\ref{sec:eval}). It also works as an agent: it plans multi-step tasks, calls tools, and works on Arabic files, and at 3.7B and 7B it tops every agentic benchmark we ran against models of its size (Section~\ref{subsec:agentic-eval}).

\modelname shows that an open foundation can be adapted to a new language without starting from scratch, while keeping most of what it could already do. Along the way, we moved Arabic evaluation to a generative protocol and audited the benchmarks it relies on, and we release the models, every stage checkpoint, the recipes, a sample of the training data, and the audited benchmarks (Section~\ref{sec:release}). We hope these let others build agents for Arabic speakers, and adapt \model{K2 Horizon} to other languages in the same way.

\newpage
\section*{Authors' Contributions}

\subsection*{Core Contributors}
\begin{itemize}
  \item \textbf{Training:} Hadi Abdine, Ahmad Chamma, Amr Mohamed, Ahmed Frikha, Dani Bouch, Omar Arbi, Guokan Shang
  \item \textbf{Data:} Nikhil Ranjan, Maaz Maqsood, Shivam Chauhan, Ahmer Tabassum, Owais Aijaz, Dani Bouch, Amr Mohamed, Ahmad Chamma, Abdelhakim Freihat, Omar Arbi, Ahmed Frikha, Sarath Chandran, Neha Sengupta, Rahul Pal, Mohammed Bellam, Sajid Siddiki, Sarah AlBarri
  \item \textbf{Evaluation:} Mohamed Anwar, George Ibrahim, Mostafa Awad, Nurkhan Laiyk, Hasan Iqbal, Abhishek Maiti, Bilal Bouardi, Aaryamonvikram Singh, Amr Mohamed, Dani Bouch, Sarah AlBarri, Debopriyo Banerjee, Ali El Filali, Samta Kamboj, Samujjwal Ghosh
  \item \textbf{Project Management:} Hector Ren, Mohamed Anwar, Guokan Shang, Preslav Nakov, Zainul Abedien Ahmed Quraishi, Larry Murray, Avraham Sheinin, Natalia Vassilieva
  \item \textbf{Conception, Design, and Leadership:} Haonan Li, Zhengzhong Liu, Eric Xing
\end{itemize}

\vspace{0.5em}
\subsection*{Contributors}

Rifo Genadi, Ahmed Attia, Go Inoue, Shuonan Dong, Amine Khelif Khelif, Evan Dufraisse, Florian Valade, Tapas Sen, Guowei He, Sunil Sahu, Onkar Pandit, Kareem Magdy, Sheikh Khalid bin Khalifa Al Nahyan, Team IFM

\newpage
\bibliographystyle{plainnat}
\bibliography{references}

\appendix
\let\originalsection\section
\renewcommand{\section}{\clearpage\originalsection}

\section{Reward Hyperparameters}\label{app:reward-hparams}

\begin{table}[H]
  \centering
  \small\rmfamily
  \renewcommand{\arraystretch}{1.15}
  \setlength{\tabcolsep}{6pt}
  \begin{tabularx}{\columnwidth}{@{}l>{\raggedright\arraybackslash}Xc@{}}
    \toprule
    \textbf{Symbol} & \textbf{Description} & \textbf{Value} \\
    \midrule
    \multicolumn{3}{@{}l}{\textbf{Shared}} \\
    \midrule
    $\phi$ & Format-penalty coefficient & 0.02 \\
    \addlinespace[3pt]\midrule
    \multicolumn{3}{@{}l}{\textbf{Diacritization}} \\
    \midrule
    $w_{\mathrm{char}}$ & Character-accuracy weight & 0.75 \\
    \addlinespace[1pt]
    $w_{\mathrm{word}}$ & Word-accuracy weight & 0.20 \\
    \addlinespace[1pt]
    $w_{\mathrm{fin}}$ & Word-final accuracy weight & 0.05 \\
    \addlinespace[1pt]
    $N_{\min}$ & Minimum reference length (letters) & 2 \\
    \addlinespace[1pt]
    $n_0$ & Minimum marks for no-diacritic cap & 2 \\
    \addlinespace[1pt]
    $\eta$ & Mark fraction for no-diacritic cap & 0.20 \\
    \addlinespace[1pt]
    $\kappa_{\mathrm{nd}}$ & No-diacritic cap & 0.10 \\
    \addlinespace[1pt]
    $A_{\min}$ & Character-accuracy threshold & 0.20 \\
    \addlinespace[1pt]
    $\kappa_{\mathrm{lo}}$ & Low-accuracy cap & 0.20 \\
    \addlinespace[3pt]\midrule
    \multicolumn{3}{@{}l}{\textbf{Grammatical analysis}} \\
    \midrule
    $b_0$ & Base score, role and case agree & 0.60 \\
    \addlinespace[1pt]
    $\lambda_M$ & Case-marker bonus & 0.15 \\
    \addlinespace[1pt]
    $\lambda_O$ & Token-overlap weight & 0.15 \\
    \addlinespace[1pt]
    $\lambda_T$ & Target-word bonus & 0.05 \\
    \addlinespace[1pt]
    $b_c$ & Base score, case mismatch & 0.25 \\
    \addlinespace[1pt]
    $\gamma_c$ & Overlap weight, case mismatch & 0.10 \\
    \addlinespace[1pt]
    $\gamma_r$ & Overlap weight, role mismatch & 0.10 \\
    \addlinespace[1pt]
    $\gamma_u$ & Overlap weight, undefined reference role & 0.50 \\
    \addlinespace[1pt]
    $\kappa_{\mathrm{soft}}$ & Cap: hedging, contradiction, multiple analyses, length & 0.30 \\
    \addlinespace[1pt]
    $\kappa_{\mathrm{hard}}$ & Cap: stuffing, wrong target, non-Arabic output & 0.20 \\
    \bottomrule
  \end{tabularx}
  \caption{Hyperparameters of the diacritization and grammatical-analysis
    rewards (Equations~\ref{eq:format-penalty}--\ref{eq:irab}).}
  \label{tab:arabic-reward-hparams}
\end{table}

\section{Multi-Turn Data: Additional Details}
\label{app:multiturn}

This appendix specifies the checks summarized in Section~\ref{sec:multiturn}: the
structure a targeted conversation must have, the full qualification criteria for both
conversation types, and how reasoning replay handles failures.

\subsection{Anatomy of a Targeted Conversation}
\label{app:mt-anatomy}
\begin{figure}[h]
  \centering
  \includegraphics[width=0.75\linewidth]{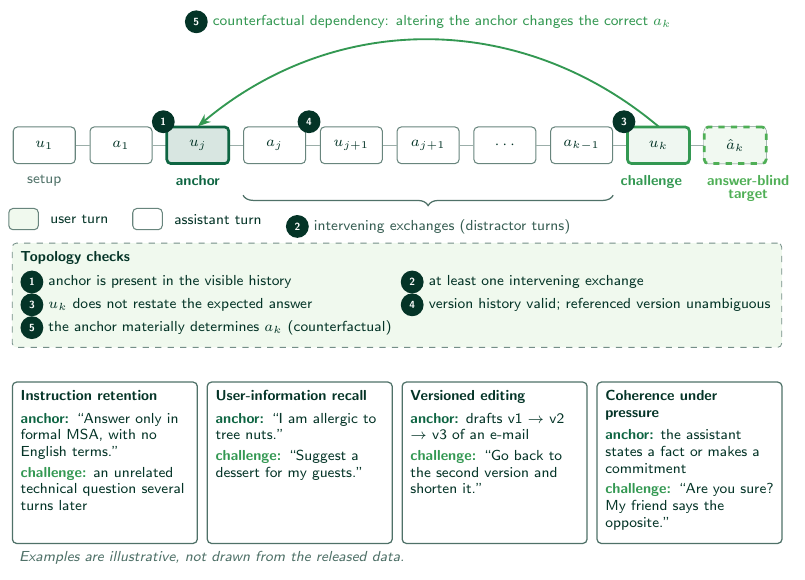}
  \caption{Anatomy of a targeted conversation: the final user turn $u_k$ depends on the
  anchor $u_j$. Bottom: illustrative examples of the four dependency types.}
  \label{fig:mt-anatomy}
\end{figure}

A targeted conversation ends with a user turn $u_k$ whose correct answer depends on an
earlier \emph{anchor} turn $u_j$ (Figure~\ref{fig:mt-anatomy}). The structural audit
checks five properties of the visible conversation: presence (T1), separation (T2), no
restatement (T3), version validity (T4), and counterfactual dependency (T5), the last
only for conversations qualified under the strict contract.

\subsection{Qualification Criteria}
\label{app:mt-criteria}

A conversation is accepted only if both judges, \model{Qwen3.6-27B} and \model{gemma-4-31B-it}, approve
it on every criterion that applies to its type (Table~\ref{tab:mt-criteria}). Targeted
conversations must pass the structural checks, which guarantee the long-range
dependency. General conversations lack this guarantee, so every assistant turn is
instead audited individually, including for safety, privacy, and use of information
introduced only later in the conversation. Both types share answer-blind regeneration,
unanimous judging, and the replay checks.

\begin{table}[h]
  \centering
  \small
  \caption{Qualification criteria by conversation type. \checkmark: required;
  $\dagger$: required for part of the data; --: not applied.}
  \label{tab:mt-criteria}
  \begin{tabular}{@{}llcc@{}}
    \toprule
    Group & Criterion & Targeted & General \\
    \midrule
    \multirow{5}{*}{Structure}
      & T1 presence                        & \checkmark    & -- \\
      & T2 separation                      & \checkmark    & -- \\
      & T3 no restatement                  & \checkmark    & -- \\
      & T4 version validity                & \checkmark    & -- \\
      & T5 counterfactual dependency       & $\dagger^{a}$ & -- \\
    \midrule
    \multirow{3}{*}{Final answer}
      & answer-blind regeneration          & \checkmark    & \checkmark \\
      & honors the anchor                  & \checkmark    & -- \\
      & unanimous \model{Qwen} $\wedge$ \model{Gemma}      & \checkmark    & \checkmark \\
    \midrule
    \multirow{6}{*}{Conversation}
      & coherence of assistant turns       & \checkmark    & \checkmark \\
      & per-turn audit of every assistant turn & --        & \checkmark \\
      & language / variant consistency     & \checkmark    & \checkmark \\
      & consistent model identity          & \checkmark    & \checkmark \\
      & factual / domain accuracy          & $\dagger^{b}$ & \checkmark \\
      & safety, privacy, no future leakage & --            & \checkmark \\
    \midrule
    \multirow{2}{*}{Replay}
      & answer, reasoning, mutual consistency & \checkmark & \checkmark \\
      & agreement with original answer; whole-chat review & \checkmark & \checkmark \\
    \bottomrule
  \end{tabular}\\[3pt]
  \footnotesize $^{a}$Conversations qualified under the strict contract.
  $^{b}$Conversations under the strict contract and the separately reviewed
  scientific and research subset.
\end{table}

\subsection{Reasoning Replay and Fallbacks}
\label{app:mt-replay}

\begin{figure}[h]
  \centering
  \includegraphics[width=0.75\linewidth]{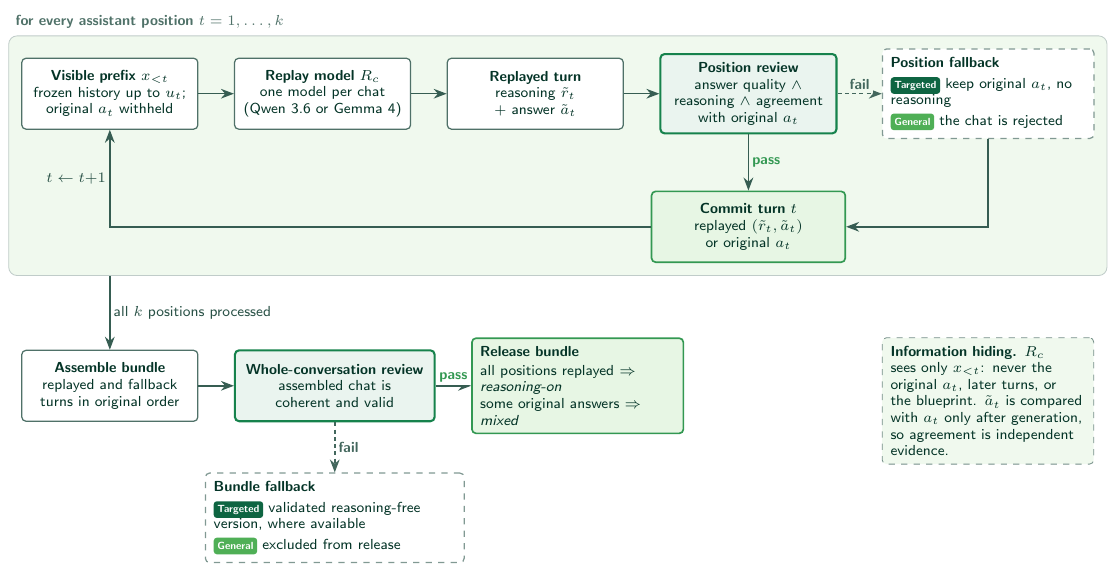}
  \caption{Reasoning replay and assembly.}
  \label{fig:mt-replay}
\end{figure}

Every assistant turn is replayed to attach an explicit reasoning trace
(Figure~\ref{fig:mt-replay}). For each assistant position $t$, a replay model, fixed for
the whole conversation, sees only the visible prefix before $t$ and generates reasoning
and an answer without seeing the original answer $a_t$. The judge rejects the replayed
turn if its answer is of insufficient quality, its reasoning is unsound, the two are
inconsistent, or the answer disagrees with $a_t$. Agreement is required because later
turns build on $a_t$.

The two conversation types handle failures differently:
\begin{itemize}\itemsep1pt
  \item \textbf{Failed replay of a turn.} A targeted conversation keeps the original
  $a_t$ without reasoning, so a released conversation can mix turns with and without
  reasoning traces. A general conversation is rejected.
  \item \textbf{Failed whole-chat review.} A targeted conversation falls back to its
  previously validated version without reasoning, where one exists. A general
  conversation is excluded.
\end{itemize}

\section{Complete Evaluation Setup}
\label{app:jais3-evaluation-details}

This appendix provides the complete configuration supporting the summary in
Section~\ref{subsubsec:eval-ar-setup}. It records the evaluated model
families, benchmark inventory, prompting and chat-template behavior, decoding policy, Arabic-specific processing, scoring rules, repeatability protocol, and runtime implementation. Exact checkpoint identifiers, source revisions, and machine-readable generation configurations accompany the released evaluation artifacts.

\subsection{Decoding Configurations}
\label{app:jais3-decoding-details}

\modelname is compared with Arabic-focused and multilingual leading models at similar and larger parameter scales. Table~\ref{tab:appendix-evaluated-models} summarizes the model families and scales represented in the evaluation. The released configuration artifact records the exact repository identifier, checkpoint revision, tokenizer and
chat-template revision, and evaluated reasoning mode for every row.

\providecommand{\model}[1]{#1}

\begingroup
\scriptsize\rmfamily
\renewcommand{\arraystretch}{1.05}
\setlength{\tabcolsep}{1.6pt}
\begin{longtable}{@{}lcll@{}}
\toprule
\textbf{Model} & \textbf{Size} & \multicolumn{1}{c}{\textbf{Hugging Face}} & \textbf{Commit} \\
\midrule
\endfirsthead
\multicolumn{4}{l}{\small\itshape Continued from previous page}\\
\toprule
\textbf{Model} & \textbf{Size} & \multicolumn{1}{c}{\textbf{Hugging Face}} & \textbf{Commit} \\
\midrule
\endhead
\midrule\multicolumn{4}{r}{\small Continued on next page}\\
\endfoot
\bottomrule\addlinespace[6pt]
\caption{Evaluated open-weight checkpoints: Hugging Face model ID, parameter scale, repository link and the commit (revision) used.}\label{tab:appendix-evaluated-models}
\endlastfoot
\multicolumn{4}{@{}l}{\textbf{1B Scale}} \\
\midrule
\model{tiny-aya-global} & 3B & \href{https://huggingface.co/CohereLabs/tiny-aya-global}{CohereLabs/tiny-aya-global} & \href{https://huggingface.co/CohereLabs/tiny-aya-global/tree/00590ff258ccd84a805f13efcd1c34c2a542654f}{00590ff} \\
\model{Qwen3.5-2B} & 2B & \href{https://huggingface.co/Qwen/Qwen3.5-2B}{Qwen/Qwen3.5-2B} & \href{https://huggingface.co/Qwen/Qwen3.5-2B/tree/15852e8c16360a2fea060d615a32b45270f8a8fc}{15852e8} \\
\model{Falcon-H1-1.5B-Deep-Instruct} & 1.5B & \href{https://huggingface.co/tiiuae/Falcon-H1-1.5B-Deep-Instruct}{tiiuae/Falcon-H1-1.5B-Deep-Instruct} & \href{https://huggingface.co/tiiuae/Falcon-H1-1.5B-Deep-Instruct/tree/b6648636ddc906688974282de6e7a243395f5423}{b664863} \\
\model{Hala-1.2B} & 1.2B & \href{https://huggingface.co/hammh0a/Hala-1.2B}{hammh0a/Hala-1.2B} & \href{https://huggingface.co/hammh0a/Hala-1.2B/tree/10e586c0899f9b97c5764e6520ccd7c199ae0e60}{10e586c} \\
\rowcolor{jaisrow}\model{Jais-3-0.9B} & 0.9B & \href{https://huggingface.co/inception42/Jais-3-0.9B}{inception42/Jais-3-0.9B} & \href{https://huggingface.co/inception42/Jais-3-0.9B/tree/e51941b37091287fee4934e3f8a799e0d3fef37f}{e51941b} \\
\addlinespace[3pt]\midrule
\multicolumn{4}{@{}l}{\textbf{4B Scale}} \\
\midrule
\model{Qwen3.5-4B} & 4B & \href{https://huggingface.co/Qwen/Qwen3.5-4B}{Qwen/Qwen3.5-4B} & \href{https://huggingface.co/Qwen/Qwen3.5-4B/tree/851bf6e806efd8d0a36b00ddf55e13ccb7b8cd0a}{851bf6e} \\
\model{gemma-4-E2B-it} & E2B & \href{https://huggingface.co/google/gemma-4-E2B-it}{google/gemma-4-E2B-it} & \href{https://huggingface.co/google/gemma-4-E2B-it/tree/3e22461f65e89153144f8adb70e3b8c2cc9845a7}{3e22461} \\
\model{granite-4.0-h-3b-ar} & 3B & \href{https://huggingface.co/ibm-research/granite-4.0-h-3b-ar}{ibm-research/granite-4.0-h-3b-ar} & \href{https://huggingface.co/ibm-research/granite-4.0-h-3b-ar/tree/b32260f6a427527d0c38a463c656b8d3b8560c26}{b32260f} \\
\model{Falcon-H1-3B-Instruct} & 3B & \href{https://huggingface.co/tiiuae/Falcon-H1-3B-Instruct}{tiiuae/Falcon-H1-3B-Instruct} & \href{https://huggingface.co/tiiuae/Falcon-H1-3B-Instruct/tree/01087ec4c132d7f186908716b3530ea187f062a1}{01087ec} \\
\rowcolor{jaisrow}\model{Jais-3-3.7B} & 3.7B & \href{https://huggingface.co/inception42/Jais-3-3.7B}{inception42/Jais-3-3.7B} & \href{https://huggingface.co/inception42/Jais-3-3.7B/tree/ac030176ff53402ac8ebb799c54b83b6579c4eb3}{ac03017} \\
\addlinespace[3pt]\midrule
\multicolumn{4}{@{}l}{\textbf{7B Scale}} \\
\midrule
\model{gemma-4-E4B-it} & E4B & \href{https://huggingface.co/google/gemma-4-E4B-it}{google/gemma-4-E4B-it} & \href{https://huggingface.co/google/gemma-4-E4B-it/tree/ee0ef6023621cff504d758262d4e04895a5af4a2}{ee0ef60} \\
\model{gemma-4-12B-it} & 12B & \href{https://huggingface.co/google/gemma-4-12B-it}{google/gemma-4-12B-it} & \href{https://huggingface.co/google/gemma-4-12B-it/tree/707f0a3b8a3c7ad586ed01e27eafbad8a27dd0f7}{707f0a3} \\
\model{Qwen3.5-9B} & 9B & \href{https://huggingface.co/Qwen/Qwen3.5-9B}{Qwen/Qwen3.5-9B} & \href{https://huggingface.co/Qwen/Qwen3.5-9B/tree/c202236235762e1c871ad0ccb60c8ee5ba337b9a}{c202236} \\
\model{Falcon-H1-7B-Instruct} & 7B & \href{https://huggingface.co/tiiuae/Falcon-H1-7B-Instruct}{tiiuae/Falcon-H1-7B-Instruct} & \href{https://huggingface.co/tiiuae/Falcon-H1-7B-Instruct/tree/41e72f27effbab80cd45b6e884688452253a3686}{41e72f2} \\
\model{Jais-2-8B-Chat} & 8B & \href{https://huggingface.co/inception42/Jais-2-8B-Chat}{inception42/Jais-2-8B-Chat} & \href{https://huggingface.co/inception42/Jais-2-8B-Chat/tree/da0e1639cd92b508b24120f7f77f5270a8465dc4}{da0e163} \\
\model{ALLaM-7B-Instruct-preview} & 7B & \href{https://huggingface.co/humain-ai/ALLaM-7B-Instruct-preview}{humain-ai/ALLaM-7B-Instruct-preview} & \href{https://huggingface.co/humain-ai/ALLaM-7B-Instruct-preview/tree/a28dd1e67420cde72d3629c8633a974cf7d9c366}{a28dd1e} \\
\model{Fanar-1-9B-Instruct} & 9B & \href{https://huggingface.co/QCRI/Fanar-1-9B-Instruct}{QCRI/Fanar-1-9B-Instruct} & \href{https://huggingface.co/QCRI/Fanar-1-9B-Instruct/tree/d30938d0efb1fd251727ee96df6987802ff84662}{d30938d} \\
\rowcolor{jaisrow}\model{Jais-3-7B} & 7B & \href{https://huggingface.co/inception42/Jais-3-7B}{inception42/Jais-3-7B} & \href{https://huggingface.co/inception42/Jais-3-7B/tree/11dcac6d60aaf87e427c33d070dd6cf412f7270a}{11dcac6} \\
\addlinespace[3pt]\midrule
\multicolumn{4}{@{}l}{\textbf{32B Scale}} \\
\midrule
\model{gemma-4-31B-it} & 31B & \href{https://huggingface.co/google/gemma-4-31B-it}{google/gemma-4-31B-it} & \href{https://huggingface.co/google/gemma-4-31B-it/tree/842da3794eaa0b77d5f08bae87a17459d91ff475}{842da37} \\
\model{Muse-Glimmer-30B} & 30B & \href{https://huggingface.co/meta-models/Muse-Glimmer-30B}{meta-models/Muse-Glimmer-30B} & \href{https://huggingface.co/meta-models/Muse-Glimmer-30B/tree/a4e59da52a7bc87ae7251dd5545c0dd437c44b68}{a4e59da} \\
\model{Qwen3.8-27B} & 27B & \href{https://huggingface.co/Qwen/Qwen3.8-27B}{Qwen/Qwen3.8-27B} & \href{https://huggingface.co/Qwen/Qwen3.8-27B/tree/1d4bf0f2ff6012fd82039f2fa52739d0dd7c60c0}{1d4bf0f} \\
\model{Fanar-2-27B-Instruct} & 27B & \href{https://huggingface.co/QCRI/Fanar-2-27B-Instruct}{QCRI/Fanar-2-27B-Instruct} & \href{https://huggingface.co/QCRI/Fanar-2-27B-Instruct/tree/074005040e9c3511b1c43b4298bef09908285534}{0740050} \\
\model{Jais-2-70B-Chat} & 70B & \href{https://huggingface.co/inception42/Jais-2-70B-Chat}{inception42/Jais-2-70B-Chat} & \href{https://huggingface.co/inception42/Jais-2-70B-Chat/tree/8bcbd3a66b34e59424a0e5eb0329cf022df4482e}{8bcbd3a} \\
\rowcolor{jaisrow}\model{Jais-3-32B} & 32B & \href{https://huggingface.co/inception42/Jais-3-32B}{inception42/Jais-3-32B} & \href{https://huggingface.co/inception42/Jais-3-32B/tree/5249a7b8ea28c98af4b7a08f48322a3dfef91d14}{5249a7b} \\
\rowcolor{jaisrow}\model{Jais-3-MoVA-36B-A4B} & 36B-A4B & \href{https://huggingface.co/inception42/Jais-3-MoVA-36B-A4B}{inception42/Jais-3-MoVA-36B-A4B} & \href{https://huggingface.co/inception42/Jais-3-MoVA-36B-A4B/tree/c7690a5ce548f7a13ba1ace47fdb95eeb3489892}{c7690a5} \\
\addlinespace[3pt]\midrule
\multicolumn{4}{@{}l}{\textbf{375B Scale}} \\
\midrule
\model{Mistral-Small-4-119B-2603} & 119B & \href{https://huggingface.co/mistralai/Mistral-Small-4-119B-2603}{mistralai/Mistral-Small-4-119B-2603} & \href{https://huggingface.co/mistralai/Mistral-Small-4-119B-2603/tree/a11f36bebf709121056b1dbcc943d1c6afbe494d}{a11f36b} \\
\model{Qwen3.5-397B-A17B} & 397B-A17B & \href{https://huggingface.co/Qwen/Qwen3.5-397B-A17B}{Qwen/Qwen3.5-397B-A17B} & \href{https://huggingface.co/Qwen/Qwen3.5-397B-A17B/tree/8472618112abcbd45acbcdc58436aff4233c23f7}{8472618} \\
\model{MiniMax-M3} & 428B-A23B & \href{https://huggingface.co/MiniMaxAI/MiniMax-M3}{MiniMaxAI/MiniMax-M3} & \href{https://huggingface.co/MiniMaxAI/MiniMax-M3/tree/f0e1c1e04d40177e4673a22097036854f536e9c0}{f0e1c1e} \\
\model{Qwen3.8-Flash-Next} & 180B-A6B & \href{https://huggingface.co/Qwen/Qwen3.8-Flash-Next}{Qwen/Qwen3.8-Flash-Next} & \href{https://huggingface.co/Qwen/Qwen3.8-Flash-Next/tree/de4b8e4d43b917e7706784d8bb445c9af86a3540}{de4b8e4} \\
\model{Nemotron-3-Ultra-550B-A55B-BF16} & 550B-A55B & \href{https://huggingface.co/nvidia/NVIDIA-Nemotron-3-Ultra-550B-A55B-BF16}{nvidia/NVIDIA-Nemotron-3-Ultra-550B-A55B-BF16} & \href{https://huggingface.co/nvidia/NVIDIA-Nemotron-3-Ultra-550B-A55B-BF16/tree/77df655d5e9f8362164ed14dd8b48f8bce657498}{77df655} \\
\rowcolor{jaisrow}\model{Jais-3-375B-A23B} & 375B-A23B & \href{https://huggingface.co/inception42/Jais-3-375B-A23B}{inception42/Jais-3-375B-A23B} & \href{https://huggingface.co/inception42/Jais-3-375B-A23B/tree/1b5d78983cb36203253256f5085f3a3d435060a2}{1b5d789} \\
\end{longtable}
\endgroup

All models were evaluated with the vLLM framework~\citep{kwon2023vllm}, using the \textit{bfloat16} tensor type, the Triton attention backend, and a nominal vLLM GPU-memory utilization of $0.90$, lowered for specific capacity-constrained runs, and a seed of $42$. Additional repeat-run seeds are fixed in advance and recorded in the supplementary configuration artifact. The nominal batch-size setting of $16{,}384$ is a scheduling upper bound; vLLM performs the actual memory-aware batching. Tensor parallelism and node count vary with model size. Model context length is the smaller of the task allocation, an explicit command-line cap when supplied, and the maximum length read from the model configuration. Hugging Face access is offline during evaluation so that cached model and dataset artifacts do not change within a run.

We use developer-recommended decoding configurations rather than greedy decoding. For each model, we first use the decoding parameters reported for benchmark evaluation in its official technical report. If the technical report does not give them, we use those recommended in the official model card on Hugging Face or in its inference example. When no official sampling recommendation is available, we use
deterministic greedy decoding (\texttt{do\_sample=false}). The decoding parameters of every model are listed in our evaluation suite\footnote{\url{https://github.com/ifm-ai/gen-eval-arabic/blob/main/hyperparams.yaml}}.

\subsection{Complete Benchmark Inventory}
\label{app:jais3-benchmark-inventory}
Table~\ref{tab:appendix-benchmark-suite} lists each benchmark with its domain,
lm-evaluation-harness task name, dataset source and revision, and primary
metric. A benchmark may expand into multiple independently scored
subtasks, such as 57 \benchmark{MMLU} subjects, 40 \benchmark{ArabicMMLU} subjects, 160 \benchmark{dialectal-MMLU}
subject--variety combinations, 34 \benchmark{FLORES+} translation directions, or 37 \benchmark{Arabic
linguistic} subtasks. ``Local'' denotes a fixed test file
distributed with the internal benchmark snapshot. For datasets whose only
public evaluation partition is named \texttt{train} by the provider, including
\benchmark{IFEval} and  \benchmark{HarmBench}, we use that released partition as the test set
and do not train on it.

\begingroup
\scriptsize\rmfamily
\renewcommand{\arraystretch}{1.15}
\setlength{\tabcolsep}{2.2pt}
\begin{longtable}{@{}>{\raggedright\arraybackslash}p{2.8cm}>{\raggedright\arraybackslash}p{4.6cm}>{\raggedright\arraybackslash}p{6.6cm}>{\raggedright\arraybackslash}p{2.8cm}@{}}
\toprule
\textbf{Domain} & \textbf{Harness name} & \textbf{Dataset (commit)} & \textbf{Metric} \\
\midrule
\endfirsthead
\multicolumn{4}{l}{\small\itshape Continued from previous page}\\
\toprule
\textbf{Domain} & \textbf{Harness name} & \textbf{Dataset (commit)} & \textbf{Metric} \\
\midrule
\endhead
\midrule\multicolumn{4}{r}{\small Continued on next page}\\
\endfoot
\bottomrule\addlinespace[6pt]
\caption{Domain, lm-evaluation-harness task name, dataset (linked Hugging Face commit, or \emph{local} if loaded from the local benchmarks folder) and metric. Accuracy and pass rates are higher-is-better unless marked otherwise.}\label{tab:appendix-benchmark-suite}
\endlastfoot
\multicolumn{4}{@{}l}{\textbf{Arabic}} \\
\midrule
STEM & \texttt{arabic\_\allowbreak{}gsm8k-\allowbreak{}zeroshot} & IFM/\allowbreak{}Arabic-\allowbreak{}gsm8k\_\allowbreak{}verified (\href{https://huggingface.co/datasets/IFM/Arabic-gsm8k_verified/tree/0ba617b23c02926b3f660eda5e341c680f1fa489}{0ba617b}) & normalized EM \\
Instruction Following & \texttt{ara\_\allowbreak{}ifbench} & MBZUAI-\allowbreak{}IFM/\allowbreak{}ara-\allowbreak{}ifbench-\allowbreak{}wildchat (\href{https://huggingface.co/datasets/MBZUAI-IFM/ara-ifbench-wildchat/tree/200df4f7b12130b3ad586812e07e43f1f8f20839}{200df4f}) & strict prompt accuracy \\
Instruction Following & \texttt{ifeval\_\allowbreak{}ar} & arabic\_\allowbreak{}ifeval (local) & strict prompt accuracy \\
Dialect Understanding and Generation & \texttt{trans\_\allowbreak{}flores200\_\allowbreak{}plus} & openlanguagedata/\allowbreak{}flores\_\allowbreak{}plus (1c079b0) (local) & judge overall (1--5)*20 \\
World Knowledge & \texttt{arabicmmlu\_\allowbreak{}gen} & IFM/\allowbreak{}ArabicMMLU\_\allowbreak{}verified (\href{https://huggingface.co/datasets/IFM/ArabicMMLU_verified/tree/50b3f62a27156085a4d18bb61c0204a9b10c8496}{50b3f62}) & accuracy \\
Dialect Understanding and Generation & \texttt{arabicmmlu\_\allowbreak{}with\_\allowbreak{}diacritics\_\allowbreak{}gen} & go-\allowbreak{}inoue/\allowbreak{}ArabicMMLU\_\allowbreak{}full (\href{https://huggingface.co/datasets/go-inoue/ArabicMMLU_full/tree/5004cfbed4fc7973eae59392ce45f483a09253f9}{5004cfb}) & accuracy \\
Dialect Understanding and Generation & \texttt{mmlu\_\allowbreak{}ar\_\allowbreak{}dialects\_\allowbreak{}only\_\allowbreak{}gen} & IFM/\allowbreak{}Dialectal-\allowbreak{}Arabic-\allowbreak{}MMLU\_\allowbreak{}verified (\href{https://huggingface.co/datasets/IFM/Dialectal-Arabic-MMLU_verified/tree/dddd23ea42d9ebd81cc53bc5809be3a536c7bd9a}{dddd23e}) & sample-weighted accuracy \\
Arab Culture & \texttt{aradice\_\allowbreak{}culture\_\allowbreak{}gen} & IFM/\allowbreak{}AraDiCE-\allowbreak{}Culture\_\allowbreak{}verified (\href{https://huggingface.co/datasets/IFM/AraDiCE-Culture_verified/tree/6a964e81079eae5a4607a2e5fb22ac6c11a5d258}{6a964e8}) & sample-weighted accuracy \\
Arab Culture & \texttt{arab\_\allowbreak{}culture\_\allowbreak{}gen} & IFM/\allowbreak{}ArabCulture\_\allowbreak{}verified (\href{https://huggingface.co/datasets/IFM/ArabCulture_verified/tree/d709d49559d823797d01c28fd1e36a1e8383a93a}{d709d49}) & sample-weighted accuracy \\
STEM & \texttt{gat\_\allowbreak{}math\_\allowbreak{}gen} & IFM/\allowbreak{}GATmath\_\allowbreak{}verified (\href{https://huggingface.co/datasets/IFM/GATmath_verified/tree/994ee26767e9f2adcc05668bdfb68b29045ab52d}{994ee26}) & accuracy \\
STEM & \texttt{arastem\_\allowbreak{}gen} & IFM/\allowbreak{}AraSTEM\_\allowbreak{}verified (\href{https://huggingface.co/datasets/IFM/AraSTEM_verified/tree/690e1a714fa9e09bfe758e2880973e1447a053f0}{690e1a7}) & accuracy \\
Islamic Knowledge & \texttt{islamic\_\allowbreak{}qa\_\allowbreak{}QASI\_\allowbreak{}1\_\allowbreak{}gen} & IFM/\allowbreak{}QIAS2025\_\allowbreak{}verified (\href{https://huggingface.co/datasets/IFM/QIAS2025_verified/tree/be3b762b17a0b50a46e41292d8e2915edbbfd04e}{be3b762}) & accuracy \\
Islamic Knowledge & \texttt{islamic\_\allowbreak{}qa\_\allowbreak{}QASI\_\allowbreak{}2\_\allowbreak{}gen} & IFM/\allowbreak{}QIAS2025\_\allowbreak{}verified (\href{https://huggingface.co/datasets/IFM/QIAS2025_verified/tree/be3b762b17a0b50a46e41292d8e2915edbbfd04e}{be3b762}) & accuracy \\
Islamic Knowledge & \texttt{IslamicEval2015\_\allowbreak{}1B\_\allowbreak{}gen} & IFM/\allowbreak{}IslamicEval2025-\allowbreak{}Subtask1B\_\allowbreak{}verified (\href{https://huggingface.co/datasets/IFM/IslamicEval2025-Subtask1B_verified/tree/839948538392a52de0c15f89893f035197b97288}{8399485}) & accuracy \\
Islamic Knowledge & \texttt{islamic\_\allowbreak{}qa\_\allowbreak{}palmX2025\_\allowbreak{}2\_\allowbreak{}gen} & IFM/\allowbreak{}PalmX2025-\allowbreak{}Subtask2\_\allowbreak{}verified (\href{https://huggingface.co/datasets/IFM/PalmX2025-Subtask2_verified/tree/2248e5878abb4461ad8b0e4d5d611c02eea689da}{2248e58}) & accuracy \\
Arab Culture & \texttt{poetry\_\allowbreak{}analysis\_\allowbreak{}gen} & IFM/\allowbreak{}FannOrFlop-\allowbreak{}MCQ\_\allowbreak{}verified (\href{https://huggingface.co/datasets/IFM/FannOrFlop-MCQ_verified/tree/d2b665deef1618f23c824419ff972d885c160136}{d2b665d}) & sample-weighted accuracy \\
Dialect Understanding and Generation & \texttt{dialect\_\allowbreak{}id\_\allowbreak{}QADI\_\allowbreak{}gen} & dialect\_\allowbreak{}id (local) & sample-weighted accuracy \\
Arab Culture & \texttt{alyah-\allowbreak{}emirati-\allowbreak{}benchmark\_\allowbreak{}gen} & IFM/\allowbreak{}alyah-\allowbreak{}emirati-\allowbreak{}benchmark\_\allowbreak{}verified (\href{https://huggingface.co/datasets/IFM/alyah-emirati-benchmark_verified/tree/4e55dd0586c39316bd1f5dd45e5dc0bf23d12007}{4e55dd0}) & accuracy \\
Arab Culture & \texttt{arabic\_\allowbreak{}cuisine\_\allowbreak{}gen} & arabic\_\allowbreak{}cuisine (local) & accuracy \\
Arabic Linguistics & \texttt{arabic\_\allowbreak{}linguistic\_\allowbreak{}gen} & arabic\_\allowbreak{}linguistic (local) & sample-weighted accuracy \\
Arabic Linguistics & \texttt{ALP\_\allowbreak{}arabic\_\allowbreak{}parsing\_\allowbreak{}gen} & arabic\_\allowbreak{}linguistic/\allowbreak{}ALP\_\allowbreak{}arabic\_\allowbreak{}parsing (local) & accuracy \\
Arabic Linguistics & \texttt{ALP\_\allowbreak{}arabic\_\allowbreak{}diacritization\_\allowbreak{}gen} & arabic\_\allowbreak{}linguistic/\allowbreak{}ALP\_\allowbreak{}arabic\_\allowbreak{}diacritization (local) & accuracy \\
Factuality \& Hallucination & \texttt{simpleqa\_\allowbreak{}verified\_\allowbreak{}ar} & SimpleQA\_\allowbreak{}Verified\_\allowbreak{}AR (local) & judge-correct \\
Factuality \& Hallucination & \texttt{halluscore} & HalluScore (local) & judge-correct \\
Multi-turn & \texttt{mt\_\allowbreak{}bench\_\allowbreak{}arabic} & MT\_\allowbreak{}Bench\_\allowbreak{}arabic (local) & judge score (1--10) \\
Legal & \texttt{arabic\_\allowbreak{}legal\_\allowbreak{}helm\_\allowbreak{}gen} & IFM/\allowbreak{}arabic-\allowbreak{}enterprise\_\allowbreak{}verified, legal (\href{https://huggingface.co/datasets/IFM/arabic-enterprise_verified/tree/fa8b44342dd2c8999bd75da74c8eeefdba26b178}{fa8b443}) & judge accuracy \\
\addlinespace[3pt]\midrule
\multicolumn{4}{@{}l}{\textbf{English}} \\
\midrule
World Knowledge \& STEM & \texttt{mmlu\_\allowbreak{}pro\_\allowbreak{}zero\_\allowbreak{}shot} & TIGER-\allowbreak{}Lab/\allowbreak{}MMLU-\allowbreak{}Pro (\href{https://huggingface.co/datasets/TIGER-Lab/MMLU-Pro/tree/54611cde22c74cca43dd78732198de6abe971398}{54611cd}) & accuracy \\
World Knowledge \& STEM & \texttt{gpqa\_\allowbreak{}diamond\_\allowbreak{}zeroshot\_\allowbreak{}gen} & Idavidrein/\allowbreak{}gpqa (\href{https://huggingface.co/datasets/Idavidrein/gpqa/tree/633f5ee89ab8ad4522a9f850766b73f62147ffdd}{633f5ee}) & accuracy \\
World Knowledge \& STEM & \texttt{gsm8k-\allowbreak{}zeroshot}, \texttt{gsm8k\_\allowbreak{}cot\_\allowbreak{}zeroshot} & openai/\allowbreak{}gsm8k (\href{https://huggingface.co/datasets/openai/gsm8k/tree/740312add88f781978c0658806c59bc2815b9866}{740312a}) & normalized EM \\
World Knowledge \& STEM & \texttt{aime25\_\allowbreak{}modified\_\allowbreak{}repeats\_\allowbreak{}8}, \texttt{aime26\_\allowbreak{}modified\_\allowbreak{}repeats\_\allowbreak{}8} & math-\allowbreak{}ai/\allowbreak{}aime25 (\href{https://huggingface.co/datasets/math-ai/aime25/tree/563bb8404243c5f09de6ec262f2db674fe5bce9b}{563bb84}), math-\allowbreak{}ai/\allowbreak{}aime26 (\href{https://huggingface.co/datasets/math-ai/aime26/tree/79037aebdb6580008fb960d17cb21fd3099083e3}{79037ae}) & pass@$k$ \\
World Knowledge \& STEM & \texttt{minerva\_\allowbreak{}math500\_\allowbreak{}reasoning} & HuggingFaceH4/\allowbreak{}MATH-\allowbreak{}500 (\href{https://huggingface.co/datasets/HuggingFaceH4/MATH-500/tree/6e4ed1a2a79af7d8630a6b768ec859cb5af4d3be}{6e4ed1a}) & verified accuracy \\
Instruction Following & \texttt{ifeval} & google/\allowbreak{}IFEval (\href{https://huggingface.co/datasets/google/IFEval/tree/966cd89545d6b6acfd7638bc708b98261ca58e84}{966cd89}) & strict prompt accuracy \\
Multi-turn & \texttt{multi\_\allowbreak{}if} & Multi-IF (local) & mean turn accuracy \\
Factuality \& Hallucination & \texttt{simpleqa\_\allowbreak{}verified} & google/\allowbreak{}simpleqa-\allowbreak{}verified (\href{https://huggingface.co/datasets/google/simpleqa-verified/tree/0dc97e0d28d8233463e005cdc4475cc2a13ba2dc}{0dc97e0}) & judge-correct \\
Factuality \& Hallucination & \texttt{aa\_\allowbreak{}omniscience\_\allowbreak{}gen} & ArtificialAnalysis/\allowbreak{}AA-\allowbreak{}Omniscience-\allowbreak{}Public (\href{https://huggingface.co/datasets/ArtificialAnalysis/AA-Omniscience-Public/tree/4a8ffc87c4650054825fb767fe0da4a4fc97ff32}{4a8ffc8}) & judge-correct \\
Long Context & \texttt{aa\_\allowbreak{}lcr\_\allowbreak{}v1.1\_\allowbreak{}en} & ArtificialAnalysis/\allowbreak{}AA-\allowbreak{}LCR (\href{https://huggingface.co/datasets/ArtificialAnalysis/AA-LCR/tree/9a77ef56b717057ade24ceab4d273712a0b4f19e}{9a77ef5}) & accuracy \\
Multi-turn & \texttt{multichallenge} & MultiChallenge (local) & macro axis pass rate \\
Multi-turn & \texttt{mt\_\allowbreak{}bench} & MT\_\allowbreak{}Bench (local) & judge score (1--10) \\
\addlinespace[3pt]\midrule
\multicolumn{4}{@{}l}{\textbf{Coding}} \\
\midrule
Coding & \texttt{humaneval\_\allowbreak{}instruct} & openai/\allowbreak{}openai\_\allowbreak{}humaneval (\href{https://huggingface.co/datasets/openai/openai_humaneval/tree/7dce6050a7d6d172f3cc5c32aa97f52fa1a2e544}{7dce605}) & pass@1 \\
Coding & \texttt{mbpp\_\allowbreak{}instruct} & google-\allowbreak{}research-\allowbreak{}datasets/\allowbreak{}mbpp (\href{https://huggingface.co/datasets/google-research-datasets/mbpp/tree/4bb6404fdc6cacfda99d4ac4205087b89d32030c}{4bb6404}) & pass@1 \\
\end{longtable}
\endgroup

The full evaluation infrastructure contains 37 ALP-derived linguistic
subtasks totaling 11,731 records. For focused analysis in this report, we
selected the two finalized diagnostic benchmarks introduced in
Section~\ref{subsubsec:eval-new-bench}: the \parsingBMark Benchmark
(1,446 accepted items) and the \tashekeelBMark Benchmark
(1,344 accepted items).  Accordingly, the detailed linguistic results report
only these two benchmarks. The 11,731-record figure describes the broader
ALP-derived inventory and is not their combined test-set size.

Diagnostic fields such as
parse-failure rate, response length, and number of judge-able samples are
retained for auditing but are not treated as model-quality metrics. Group
scores are calculated from leaf-task scores weighted by evaluated sample
counts. A metric is aggregated only when it is finite and present for every
leaf in the group; standard-error fields and metadata are never averaged.

Several Arabic resources are local or modified snapshots rather than
unmodified public datasets. The snapshot contains the exact examples used for
evaluation, but the evaluation repository does not provide a uniform
provenance record covering annotator recruitment, compensation, quality
control, and inter-annotator agreement for every custom resource. Those details
belong in the documentation for the corresponding dataset and must not be
inferred from evaluation code alone.

The evaluation code itself does not test for contamination. The training data we added in continual pretraining, cooldown and SFT was screened against these benchmarks and matches were removed (Section~\ref{par:decontamination}).

\subsection{Arabic Benchmark Details}
\label{app:arabic-benchmark-details}
This section describes the Arabic benchmarks listed in Table~\ref{tab:arabic-benchmarks}, following the grouping used in Section~\ref{subsec:eval-arabic}. Item counts are those of the original releases, before the audit described in Section~\ref{subsubsec:eval-ar-audit}.

\subsubsection{General Capabilities}
\paragraph{STEM and world knowledge.}
\benchmark{ArabicMMLU}~\citep{koto-etal-2024-arabicmmlu} contains 14,455 multiple-choice questions from school and university examinations across the Arab world, covering 40 academic and professional subjects. \benchmark{AraSTEM}~\citep{mustapha2025arastem} contains 10,824 native Arabic multiple-choice questions on science, technology, engineering, and mathematics at several educational levels. \benchmark{GAT Math}~\citep{alballaa2025gatmath} contains mathematics questions from the Saudi General Aptitude Test on algebra, problem solving, geometry, and quantitative comparison. \benchmark{Arabic GSM8K}~\citep{cobbe2021gsm8k,arabic-gsm8k} is an Arabic translation of the 1,319 test problems of GSM8K, grade-school word problems that require multi-step arithmetic.

\paragraph{UAE legal.}
The legal tasks of HELM Arabic Enterprise~\citep{mai2026helmarabicenterprise} consist of 200 expert-written questions on UAE civil, commercial, and penal law. Each question is posed both with a context passage (open book) and without one (closed book), and an LLM judge grades the responses for correctness.

\paragraph{Medical.}
\benchmark{MedAraBench}~\citep{abudaoud2026medarabench} is a large-scale benchmark of Arabic medical question answering and clinical reasoning. Its 4,927 items cover diverse medical subdomains through multiple-choice questions, diagnostic verification, and clinical scenarios, and draw on both native Arabic medical literature and culturally adapted datasets.

\paragraph{Instruction following.}
\benchmark{IFEval-Ar}~\citep{elfilali2025arabicifeval} adapts \benchmark{IFEval}~\citep{Zhou2023Instruction} to Arabic. It contains 404 prompts covering 23 instruction types, with constraints and verifiers adapted to Arabic rather than translated directly, and responses not written in Arabic are marked incorrect. \benchmark{AraIFBench} (Section~\ref{par:araifbench}) adds constraints specific to Arabic orthography and morphology. Both benchmarks are scored programmatically, with prompt-level and instruction-level accuracy under strict and loose evaluation.

\subsubsection{Arabic Domain Knowledge}
\paragraph{Islamic knowledge and Arab culture.}
\benchmark{QIAS 2025}~\citep{bouchekif-etal-2025-qias} has two parts: 1,000 questions on Islamic inheritance, which require computing the shares of the heirs, and 700 questions on general Islamic knowledge. \benchmark{IslamicEval 2025}~\citep{mubarak-etal-2025-islamiceval} contains 247 statements whose attribution to the Qur'an or the Hadith must be verified, and \benchmark{PalmX 2025}~\citep{alwajih-etal-2025-palmx} contains 985 questions on Islamic practices, texts, history, and occasions. For culture, \benchmark{AraDiCE-Culture}~\citep{mousi-etal-2025-aradice} has 180 country-specific questions from six Arab countries, \benchmark{ArabCulture}~\citep{sadallah-etal-2025-commonsense} has 3,482 questions on cultural commonsense across 13 countries, and \benchmark{Alyah}~\citep{emirati_dialect_benchmark_2026} has 1,173 questions on Emirati culture, heritage, and dialect. \benchmark{FannOrFlop\_MCQ}~\citep{alghallabi2025fann} tests the interpretation of Arabic poetry with 6,984 multiple-choice items, built from nearly 7,000 poem--explanation pairs that span 12 historical eras, 21 core genres, and diverse poetic meters. \benchmark{Arabic Cuisine} contains 1,597 multiple-choice questions on the cuisines of the Arab world. The grammatical-analysis and diacritization benchmarks of \araLingBmarks are introduced in Section~\ref{subsubsec:eval-new-bench}, and their construction is described in Appendix~\ref{app:arabic-linguistic-benchmarks}.

\subsubsection{Dialect Understanding and Generation}
\paragraph{Dialectal understanding.}
\benchmark{QADI}~\citep{abdelali-etal-2021-qadi} is a corpus of 540,590 tweets from 18 Arab countries that covers Modern Standard Arabic and ten regional varieties; a 3,797-instance subset is used for dialect identification. \benchmark{Dialectal-Arabic-MMLU}~\citep{altakrori-etal-2026-dialectalarabicmmlu} contains 3,135 questions adapted into each of five dialects (Egyptian, Saudi, Moroccan, Syrian, and Emirati), for 15,675 question--answer pairs. Diacritized \benchmark{ArabicMMLU}~\citep{inoue-etal-2026-diacritics} is a fully diacritized version of the 14,455 \benchmark{ArabicMMLU} questions; because the content of the two versions is identical, comparing them measures robustness to diacritics.

\paragraph{Dialectal generation.}
\benchmark{FLORES+}~\citep{nllb2024scaling,maillard-etal-2024-findings} provides 1,012 aligned sentences per variety in English, Modern Standard Arabic, and eight Arabic dialects, and is used for translation. \benchmark{DarijaBench-Transliteration}~\citep{shang-etal-2025-atlas} and \benchmark{EgyptianBench-Transliteration}~\citep{shang2025nilechat} test transliteration between Arabic script and Arabizi, a Latin-script romanization, for Moroccan and Egyptian Arabic. Each covers both directions, with 1,000 examples per direction.

\subsubsection{Multi-Turn and Long-Context Understanding}
\paragraph{Multi-turn.}
\benchmark{MT-Bench Arabic}~\citep{boughorbel-hawasly-2023-analyzing} is an Arabic version of \benchmark{MT-Bench}~\citep{NEURIPS2023_91f18a12} with 80 two-turn conversations, ten in each of eight categories: writing, roleplay, reasoning, mathematics, coding, extraction, STEM, and humanities. The second turn depends on the model's own first answer, so the benchmark tests whether a model keeps the context and builds on its previous response, including when an early error carries over. Each response is scored from 1 to 10 by \model{gemma-4-26B-A4B-it} with the single-answer grading rubric of the benchmark, and for mathematics, reasoning, and coding questions the judge also receives a reference answer. Scores are reported for each turn and on average; the difference between the turns shows how well performance holds when a model continues from its own output.

\paragraph{Long context.}
\benchmark{AA-LCR}~\citep{artificialanalysis2025aalcr} contains 100 questions that require integrating information across document sets of about 100K tokens, and a judge grades each response as pass or fail. \benchmark{AA-LCR-Ar}, our Arabic version based on version 1.1, is described in Section~\ref{par:aa-lcr-ar}.

\subsection{English Benchmark Details}
\label{app:english-benchmark-details}
This section describes the English benchmarks listed in Table~\ref{tab:english-benchmarks}.

\subsubsection{General Capabilities}
\paragraph{World knowledge and STEM.}
We evaluate broad academic knowledge with \benchmark{MMLU-Pro}~\citep{wang2024mmlupro}, which spans the humanities, social sciences, STEM, and professional domains. English STEM reasoning is evaluated with \benchmark{GSM8K}~\citep{cobbe2021gsm8k}, \benchmark{GPQA-Diamond}~\citep{rein2023gpqa}, \benchmark{MATH-500}~\citep{lightman2023letsverify}, and the \benchmark{AIME 2025 and 2026} problem sets~\citep{dekoninck2026matharena}, which together range from multi-step arithmetic and graduate-level science to competition mathematics. All benchmarks are evaluated zero-shot in generative form with benchmark-specific answer formats; \benchmark{GSM8K} is run with both direct-answer and chain-of-thought prompting, and for \benchmark{AIME} we sample eight responses per problem and report pass@$k$ for $k \in \{1,2,4,8\}$. Because exact-match scoring can underestimate correctness when equivalent answers are expressed differently, we score \benchmark{MMLU-Pro}, \benchmark{GPQA-Diamond}, \benchmark{GSM8K}, and \benchmark{MATH-500} with deterministic LLM-based answer verification, using benchmark-specific \model{Gemma} judges selected through pilot experiments and manual validation. \benchmark{AIME} is scored with automatic mathematical-equivalence checking. The judge prompts are in the judge registry of our evaluation suite (Appendix~\ref{app:jais3-scoring-details}).

\paragraph{Instruction following.}
To measure instruction following with explicit output requirements, we use IFEval~\citep{Zhou2023Instruction}, which comprises 541 single-turn English prompts spanning 25 types of objectively verifiable instructions. These cover lexical, length, formatting, stylistic, and structural constraints, including combinations of multiple constraints within a single prompt. Models are evaluated zero-shot on all 541 prompts.

Responses are scored using the official deterministic IFEval metrics. We report instruction-level and prompt-level accuracy under both strict and loose evaluation. Instruction-level accuracy measures the proportion of individual constraints satisfied, whereas prompt-level accuracy requires all constraints in a prompt to be satisfied. Loose evaluation additionally permits removing asterisks or extraneous text from the first or final line. Scoring requires neither reference answers nor an LLM judge.

\subsubsection{Coding, Multi-Turn, and Long Context}
\paragraph{Coding.} Code generation is evaluated with \benchmark{HumanEval}~\citep{chen2021evaluating}, 164
function-level problems specified by a signature and docstring, and the
500-problem test split of \benchmark{MBPP}~\citep{austin2021program}, which requires full
programs from natural-language specifications. One solution is generated per problem and executed against the released unit tests. We report pass@1, defined as the proportion of generated solutions that pass all tests.

\paragraph{Multi-turn.} We evaluate English multi-turn behavior with three complementary benchmarks that capture different conversational challenges. \benchmark{MT-Bench}~\citep{NEURIPS2023_91f18a12} contains 80 two-turn conversations across eight categories, where the second turn depends on the model's own first response and therefore measures its ability to maintain context and build on prior output. \benchmark{MultiChallenge}~\citep{deshpande-etal-2025-multichallenge} contains 273 human-written conversations of up to roughly 20 turns and evaluates whether models can use information from a longer fixed dialogue history, including earlier context, instructions, prior responses, and requested edits. \benchmark{Multi-IF}~\citep{he2024multiifbenchmarkingllmsmultiturn} contains 909 three-turn conversations in which each turn introduces an additional constraint that must be satisfied together with all previous constraints, providing a deterministic measure of instruction retention across turns. \benchmark{MT-Bench} responses are scored from 1 to 10 by an LLM judge, with reference answers provided for mathematics, reasoning, and coding questions; \benchmark{MultiChallenge} is evaluated using item-specific pass criteria and reported as overall pass rate; and \benchmark{Multi-IF} uses deterministic verifiers, with accuracy reported overall and by turn. Where the original benchmarks rely on proprietary judges, we use \model{gemma-4-26B-A4B-it} for consistency with the rest of our evaluation.

\paragraph{Long context.} Long-context understanding is evaluated with \benchmark{AA-LCR}~\citep{artificialanalysis2025aalcr}, which contains 100 questions requiring reasoning across document sets of about 100K tokens. The benchmark is designed so that answers must be synthesized from multiple sources rather than retrieved directly from a single passage, and responses are graded as correct or incorrect against the reference answer. We use version 1.1, which corrects 16 reference answers and introduces revised grading instructions. Our Arabic adaptation follows the same version and evaluation setup and is described in Section~\ref{par:aa-lcr-ar}.%

\subsection{Run-to-Run Variance}
\label{app:run-to-run-variance}
Most models in our evaluation are run with sampling-based decoding, using the temperature and other generation settings recommended by their developers. With sampling, a model may answer the same question differently across runs, so a score from a single evaluation can include chance variation and may be misleading when two models perform similarly.%

We therefore run each \modelname model three times on each benchmark, using identical prompts, generation settings, and judge models while varying only the random seed. The LLM judges decode greedily and therefore do not contribute additional sampling variation. %

For each model and benchmark, the score of each run is computed in the same way as for a single evaluation; for benchmarks made up of several subsets, it is the average over all questions of the benchmark. We report the mean of the three run scores together with their standard deviation across the three runs. 

\providecommand{\model}[1]{#1}
\providecommand{\modelname}{Jais-3}

\begin{table}[t]
\centering
\scriptsize\rmfamily
\renewcommand{\arraystretch}{1.15}
\setlength{\tabcolsep}{2.2pt}
\ifdefined\runsModelW\else\newlength{\runsModelW}\newlength{\runsTmpW}\fi
\setlength{\runsModelW}{0pt}
\settowidth{\runsTmpW}{\textbf{Benchmark}}\ifdim\runsTmpW>\runsModelW\setlength{\runsModelW}{\runsTmpW}\fi
\settowidth{\runsTmpW}{\textbf{Domain}}\ifdim\runsTmpW>\runsModelW\setlength{\runsModelW}{\runsTmpW}\fi
\settowidth{\runsTmpW}{\model{Jais-3-0.9B}}\ifdim\runsTmpW>\runsModelW\setlength{\runsModelW}{\runsTmpW}\fi
\settowidth{\runsTmpW}{\model{Jais-3-3.7B}}\ifdim\runsTmpW>\runsModelW\setlength{\runsModelW}{\runsTmpW}\fi
\settowidth{\runsTmpW}{\model{Jais-3-7B}}\ifdim\runsTmpW>\runsModelW\setlength{\runsModelW}{\runsTmpW}\fi
\settowidth{\runsTmpW}{\model{Jais-3-32B}}\ifdim\runsTmpW>\runsModelW\setlength{\runsModelW}{\runsTmpW}\fi
\settowidth{\runsTmpW}{\model{Jais-3-MoVA-36B-A4B}}\ifdim\runsTmpW>\runsModelW\setlength{\runsModelW}{\runsTmpW}\fi
\settowidth{\runsTmpW}{\model{Jais-3-375B-A23B}}\ifdim\runsTmpW>\runsModelW\setlength{\runsModelW}{\runsTmpW}\fi
\begin{tabular}{@{}w{l}{\runsModelW}*{6}{w{c}{\dimexpr(\linewidth-\runsModelW)/6-2\tabcolsep\relax}}@{}}
\toprule
\textbf{Domain} & \multicolumn{1}{c}{\makebox[0pt][c]{\textbf{\begin{tabular}[b]{@{}c@{}}World\\Knowledge\end{tabular}}}} & \multicolumn{3}{c}{\makebox[0pt][c]{\textbf{STEM}}} & \multicolumn{1}{c}{\makebox[0pt][c]{\textbf{Legal}}} & \multicolumn{1}{c}{\makebox[0pt][c]{\textbf{Medical}}} \\
\cmidrule(lr){2-2}\cmidrule(lr){3-5}\cmidrule(lr){6-6}\cmidrule(lr){7-7}
\textbf{Benchmark} & \makebox[0pt][c]{\textbf{\begin{tabular}[b]{@{}c@{}}Arabic\\MMLU\end{tabular}}} & \makebox[0pt][c]{\textbf{AraSTEM}} & \makebox[0pt][c]{\textbf{\begin{tabular}[b]{@{}c@{}}GAT\\Math\end{tabular}}} & \makebox[0pt][c]{\textbf{\begin{tabular}[b]{@{}c@{}}Arabic\\GSM8K\end{tabular}}} & \makebox[0pt][c]{\textbf{\begin{tabular}[b]{@{}c@{}}Legal\\(HELM)\end{tabular}}} & \makebox[0pt][c]{\textbf{\begin{tabular}[b]{@{}c@{}}MedAra\\Bench\end{tabular}}} \\
 & Acc. & Acc. & Acc. & Acc. & Acc. & Acc. \\
\midrule
\model{Jais-3-0.9B} & 70.28$_{\pm 0.22}$ & 81.41$_{\pm 0.22}$ & 87.73$_{\pm 0.21}$ & 79.82$_{\pm 0.51}$ & 60.64$_{\pm 0.30}$ & 48.96$_{\pm 1.06}$ \\
\model{Jais-3-3.7B} & 83.20$_{\pm 0.24}$ & 91.51$_{\pm 0.13}$ & 96.40$_{\pm 0.23}$ & 91.59$_{\pm 0.18}$ & 65.12$_{\pm 1.18}$ & 73.16$_{\pm 0.30}$ \\
\model{Jais-3-7B} & 85.27$_{\pm 0.03}$ & 92.54$_{\pm 0.09}$ & 96.59$_{\pm 0.20}$ & 92.97$_{\pm 0.29}$ & 66.58$_{\pm 1.22}$ & 77.63$_{\pm 0.51}$ \\
\model{Jais-3-32B} & 88.42$_{\pm 0.23}$ & 95.19$_{\pm 0.03}$ & 98.50$_{\pm 0.07}$ & 93.80$_{\pm 0.12}$ & 73.21$_{\pm 1.04}$ & 86.72$_{\pm 0.15}$ \\
\model{Jais-3-MoVA-36B-A4B} & 87.70$_{\pm 0.06}$ & 94.72$_{\pm 0.14}$ & 97.46$_{\pm 0.08}$ & 92.25$_{\pm 0.10}$ & 69.51$_{\pm 1.61}$ & 85.32$_{\pm 0.27}$ \\
\model{Jais-3-375B-A23B} & 89.08$_{\pm 0.24}$ & 95.52$_{\pm 0.10}$ & 98.45$_{\pm 0.14}$ & 87.46$_{\pm 0.24}$ & 72.44$_{\pm 0.15}$ & 88.84$_{\pm 0.62}$ \\
\bottomrule
\end{tabular}
\par\vspace{8pt}
\begin{tabular}{@{}w{l}{\runsModelW}*{9}{w{c}{\dimexpr(\linewidth-\runsModelW)/9-2\tabcolsep\relax}}@{}}
\toprule
\textbf{Domain} & \multicolumn{1}{c}{\makebox[0pt][c]{\textbf{\begin{tabular}[b]{@{}c@{}}Islamic\\Knowledge\end{tabular}}}} & \multicolumn{5}{c}{\makebox[0pt][c]{\textbf{Arab Culture}}} & \multicolumn{1}{c}{\makebox[0pt][c]{\textbf{\begin{tabular}[b]{@{}c@{}}Arabic\\Linguistics\end{tabular}}}} & \multicolumn{1}{c}{\makebox[0pt][c]{\textbf{\begin{tabular}[b]{@{}c@{}}Instruction\\Following\end{tabular}}}} & \multicolumn{1}{c}{\makebox[0pt][c]{\textbf{\begin{tabular}[b]{@{}c@{}}Long\\Context\end{tabular}}}} \\
\cmidrule(lr){2-2}\cmidrule(lr){3-7}\cmidrule(lr){8-8}\cmidrule(lr){9-9}\cmidrule(lr){10-10}
\textbf{Benchmark} & \makebox[0pt][c]{\textbf{\begin{tabular}[b]{@{}c@{}}Islamic\\QA\end{tabular}}} & \makebox[0pt][c]{\textbf{\begin{tabular}[b]{@{}c@{}}AraDiCE\\Culture\end{tabular}}} & \makebox[0pt][c]{\textbf{\begin{tabular}[b]{@{}c@{}}Arab\\Culture\end{tabular}}} & \makebox[0pt][c]{\textbf{Alyah}} & \makebox[0pt][c]{\textbf{\begin{tabular}[b]{@{}c@{}}Fann\\OrFlop\end{tabular}}} & \makebox[0pt][c]{\textbf{\begin{tabular}[b]{@{}c@{}}Arabic\\Cuisine\end{tabular}}} & \makebox[0pt][c]{\textbf{\begin{tabular}[b]{@{}c@{}}Arabic\\Ling.\end{tabular}}} & \makebox[0pt][c]{\textbf{\begin{tabular}[b]{@{}c@{}}IFEval\\Ar\end{tabular}}} & \makebox[0pt][c]{\textbf{\begin{tabular}[b]{@{}c@{}}AA-LCR\\Ar\end{tabular}}} \\
\addlinespace[\dimexpr\aboverulesep+\cmidrulewidth+\belowrulesep\relax]
 & Acc. & Acc. & Acc. & Acc. & Acc. & Acc. & Acc. & Acc. & Acc. \\
\midrule
\model{Jais-3-0.9B} & 69.75$_{\pm 0.71}$ & 71.51$_{\pm 1.68}$ & 73.74$_{\pm 0.73}$ & 73.09$_{\pm 1.18}$ & 59.28$_{\pm 0.33}$ & 72.95$_{\pm 1.11}$ & 85.00$_{\pm 0.35}$ & 71.53$_{\pm 2.03}$ & -- \\
\model{Jais-3-3.7B} & 82.18$_{\pm 0.56}$ & 78.77$_{\pm 0.56}$ & 84.23$_{\pm 0.59}$ & 80.45$_{\pm 0.39}$ & 66.18$_{\pm 0.38}$ & 82.77$_{\pm 0.62}$ & 92.23$_{\pm 0.47}$ & 79.62$_{\pm 0.62}$ & 66.00$_{\pm 3.61}$ \\
\model{Jais-3-7B} & 85.15$_{\pm 0.45}$ & 81.01$_{\pm 2.56}$ & 85.17$_{\pm 0.24}$ & 80.22$_{\pm 0.35}$ & 68.66$_{\pm 0.46}$ & 84.83$_{\pm 0.46}$ & 93.24$_{\pm 0.46}$ & 79.21$_{\pm 1.24}$ & 66.33$_{\pm 4.16}$ \\
\model{Jais-3-32B} & 90.02$_{\pm 0.61}$ & 87.52$_{\pm 0.65}$ & 88.24$_{\pm 0.42}$ & 84.05$_{\pm 0.43}$ & 72.02$_{\pm 0.23}$ & 86.77$_{\pm 0.43}$ & 94.35$_{\pm 0.56}$ & 78.80$_{\pm 0.38}$ & 76.67$_{\pm 2.08}$ \\
\model{Jais-3-MoVA-36B-A4B} & 89.52$_{\pm 0.27}$ & 87.34$_{\pm 0.85}$ & 88.76$_{\pm 0.15}$ & 83.28$_{\pm 0.57}$ & 71.46$_{\pm 0.16}$ & 88.82$_{\pm 0.50}$ & 93.88$_{\pm 0.18}$ & 78.88$_{\pm 0.76}$ & 74.00$_{\pm 2.65}$ \\
\model{Jais-3-375B-A23B} & 91.22$_{\pm 0.35}$ & 88.09$_{\pm 1.40}$ & 89.67$_{\pm 0.28}$ & 87.74$_{\pm 0.31}$ & 73.27$_{\pm 0.28}$ & 90.14$_{\pm 0.33}$ & 94.83$_{\pm 0.30}$ & 77.48$_{\pm 0.43}$ & -- \\
\bottomrule
\end{tabular}
\par\vspace{8pt}
\begin{tabular}{@{}w{l}{\runsModelW}*{9}{w{c}{\dimexpr(\linewidth-\runsModelW)/9-2\tabcolsep\relax}}@{}}
\toprule
\textbf{Domain} & \multicolumn{5}{c}{\makebox[0pt][c]{\textbf{\begin{tabular}[b]{@{}c@{}}Dialect Understanding\\and Generation\end{tabular}}}} & \multicolumn{4}{c}{\makebox[0pt][c]{\textbf{Safety}}} \\
\cmidrule(lr){2-6}\cmidrule(lr){7-10}
\textbf{Benchmark} & \makebox[0pt][c]{\textbf{QADI}} & \makebox[0pt][c]{\textbf{\begin{tabular}[b]{@{}c@{}}FLORES+\end{tabular}}} & \makebox[0pt][c]{\textbf{\begin{tabular}[b]{@{}c@{}}Dialect.\\MMLU\end{tabular}}} & \makebox[0pt][c]{\textbf{\begin{tabular}[b]{@{}c@{}}Arabic\\MMLU\end{tabular}}} & \makebox[0pt][c]{\textbf{\begin{tabular}[b]{@{}c@{}}MMLU\\diacr.\end{tabular}}} & \makebox[0pt][c]{\textbf{AraSafe}} & \multicolumn{3}{c}{\makebox[0pt][c]{\textbf{\begin{tabular}[b]{@{}c@{}}Arabic Regional\\Safety\end{tabular}}}} \\
\cmidrule(lr){8-10}
 & Acc. & Judge & Acc. & Acc. & Acc. & Refusal & General & Gov. & Opp. \\
\midrule
\model{Jais-3-0.9B} & 66.12$_{\pm 0.69}$ & 82.42$_{\pm 0.03}$ & 63.59$_{\pm 0.33}$ & 70.28$_{\pm 0.22}$ & 64.94$_{\pm 0.54}$ & 93.27$_{\pm 0.17}$ & 98.92$_{\pm 0.36}$ & 98.62$_{\pm 0.08}$ & 98.71$_{\pm 0.31}$ \\
\model{Jais-3-3.7B} & 69.00$_{\pm 0.30}$ & 89.04$_{\pm 0.03}$ & 81.14$_{\pm 0.09}$ & 83.20$_{\pm 0.24}$ & 80.68$_{\pm 0.18}$ & 92.66$_{\pm 0.08}$ & 99.52$_{\pm 0.55}$ & 96.45$_{\pm 0.60}$ & 98.36$_{\pm 0.20}$ \\
\model{Jais-3-7B} & 69.34$_{\pm 0.15}$ & 90.53$_{\pm 0.05}$ & 83.39$_{\pm 0.31}$ & 85.27$_{\pm 0.03}$ & 83.17$_{\pm 0.15}$ & 92.19$_{\pm 0.29}$ & 99.88$_{\pm 0.21}$ & 98.71$_{\pm 0.15}$ & 99.47$_{\pm 0.35}$ \\
\model{Jais-3-32B} & 70.42$_{\pm 0.10}$ & 91.57$_{\pm 0.02}$ & 88.64$_{\pm 0.17}$ & 88.42$_{\pm 0.23}$ & 87.90$_{\pm 0.19}$ & 92.42$_{\pm 0.63}$ & 99.88$_{\pm 0.21}$ & 99.47$_{\pm 0.23}$ & 99.91$_{\pm 0.08}$ \\
\model{Jais-3-MoVA-36B-A4B} & 69.61$_{\pm 0.61}$ & 90.83$_{\pm 0.02}$ & 87.42$_{\pm 0.03}$ & 87.70$_{\pm 0.06}$ & 86.90$_{\pm 0.18}$ & 93.20$_{\pm 0.26}$ & 100.00$_{\pm 0.00}$ & 99.11$_{\pm 0.20}$ & 99.69$_{\pm 0.20}$ \\
\model{Jais-3-375B-A23B} & 71.48$_{\pm 0.37}$ & 91.71$_{\pm 0.16}$ & 89.53$_{\pm 0.02}$ & 89.08$_{\pm 0.24}$ & 87.95$_{\pm 0.28}$ & 93.06$_{\pm 0.58}$ & 100.00$_{\pm 0.00}$ & 98.91$_{\pm 0.25}$ & 99.73$_{\pm 0.13}$ \\
\bottomrule
\end{tabular}
\vspace{4pt}
\caption{Arabic results for the \modelname{} models. Mean over three independent runs; subscripts: standard deviation across the runs. Scores on the audited benchmarks exclude the removed items. All results use high-effort thinking. Islamic QA is the unweighted mean of QIAS-Inheritance, QIAS-Knowledge, IslamicEval 2025, and PalmX 2025; -- marks results that are not available.}\label{tab:jais3-arabic-runs}
\end{table}

We observe low run-to-run variance for \modelname overall (Table~\ref{tab:jais3-arabic-runs}). Almost 90\% of results vary by less than 1 point between runs.
A few benchmarks show noticeably higher variation. \benchmark{AA-LCR-Ar} varies the most, with a mean standard deviation of 3.1 and up to 4.2 points, followed by \benchmark{AraDiCE-Culture} (1.3), \benchmark{Legal (HELM)} (0.9), and \benchmark{IFEval-Ar} (0.9). The most variable benchmarks tend to be small, such as \benchmark{AA-LCR-Ar} with 100 questions and \benchmark{AraDiCE-Culture} with 180. In contrast, larger benchmarks such as \benchmark{ArabicMMLU}, \benchmark{Dialectal-Arabic-MMLU}, \benchmark{AraSTEM}, \benchmark{GAT Math}, and \benchmark{FLORES+} are highly stable, with a mean standard deviation of 0.2 points or less.

\subsection{Low-Effort Thinking Results}
\label{app:low-reasonig}
We also evaluate \modelname with low-effort thinking. For a fair comparison, other thinking models are run at their lowest reasoning effort, and models without a thinking mode in their non-thinking (instruct) mode.

\subsubsection{Arabic Evaluation}
\paragraph{General capabilities results.}
Table~\ref{tab:general-capabilities-low} shows that \modelname leads every benchmark at 4B, and every benchmark but \benchmark{Arabic GSM8K} at 1B, where \model{Jais-3-0.9B} is 0.8 points behind \model{Falcon-H1-1.5B-Deep-Instruct}. At 7B it leads \benchmark{Legal (HELM)}, \benchmark{IFEval-Ar} and \benchmark{AraIFBench}, but \model{gemma-4-12B-it} is clearly ahead on the knowledge and math benchmarks, by 7.4 points on \benchmark{ArabicMMLU}, 8.6 on \benchmark{AraSTEM}, 3.1 on \benchmark{Arabic GSM8K}, and more than 20 on \benchmark{GAT Math} and \benchmark{MedAraBench}. At 32B, \model{gemma-4-31B-it} has the best score on every benchmark, and shares the \benchmark{Legal (HELM)} lead with \model{Jais-3-32B} (72.87). \model{Jais-3-32B} is within 3 points of it on \benchmark{Arabic GSM8K}, \benchmark{IFEval-Ar} and \benchmark{AraIFBench}, while \benchmark{MedAraBench} has the widest gap (15.3 points); \model{Jais-3-MoVA-36B-A4B} also comes within 2.4 points on \benchmark{AraSTEM}. At 375B, \model{Jais-3-375B-A23B} leads \benchmark{GAT Math} (95.6), \benchmark{IFEval-Ar} (73.0) and \benchmark{AraIFBench} (55.4), and nearly ties the best \benchmark{Arabic GSM8K} score (92.88 against 92.92 for \model{Nemotron-3-Ultra-550B-A55B-BF16}). It trails \model{Qwen3.5-397B-A17B} on \benchmark{ArabicMMLU} (by 1.7 points), \benchmark{AraSTEM} (1.4) and \benchmark{MedAraBench} (9.8), and \model{Nemotron-3-Ultra-550B-A55B-BF16} on \benchmark{Legal (HELM)} (3.4).
\providecommand{\model}[1]{#1}

\begingroup
\scriptsize\rmfamily
\renewcommand{\arraystretch}{1.15}
\setlength{\tabcolsep}{2.2pt}
\begin{longtable}{@{}l*{8}{c}@{}}
\toprule
\textbf{Domain} & \multicolumn{1}{c}{\textbf{\begin{tabular}[b]{@{}c@{}}World\\Knowledge\end{tabular}}} & \multicolumn{3}{c}{\textbf{STEM}} & \multicolumn{1}{c}{\textbf{Medical}} & \multicolumn{1}{c}{\textbf{Legal}} & \multicolumn{2}{c}{\textbf{\begin{tabular}[b]{@{}c@{}}Instruction\\Following\end{tabular}}} \\
\cmidrule(lr){2-2}\cmidrule(lr){3-5}\cmidrule(lr){6-6}\cmidrule(lr){7-7}\cmidrule(lr){8-9}
\textbf{Benchmark} & \textbf{\begin{tabular}[b]{@{}c@{}}Arabic\\MMLU\end{tabular}} & \textbf{AraSTEM} & \textbf{\begin{tabular}[b]{@{}c@{}}GAT\\Math\end{tabular}} & \textbf{\begin{tabular}[b]{@{}c@{}}Arabic\\GSM8K\end{tabular}} & \textbf{\begin{tabular}[b]{@{}c@{}}MedAra\\Bench\end{tabular}} & \textbf{\begin{tabular}[b]{@{}c@{}}Legal\\(HELM)\end{tabular}} & \textbf{IFEval-Ar} & \textbf{\begin{tabular}[b]{@{}c@{}}AraIF\\Bench\end{tabular}} \\

 & Acc. & Acc. & Acc. & Acc. & Acc. & Acc. & Acc. & Acc. \\
\midrule
\endfirsthead
\multicolumn{9}{l}{\small\itshape Continued from previous page}\\
\toprule
\textbf{Domain} & \multicolumn{1}{c}{\textbf{\begin{tabular}[b]{@{}c@{}}World\\Knowledge\end{tabular}}} & \multicolumn{3}{c}{\textbf{STEM}} & \multicolumn{1}{c}{\textbf{Medical}} & \multicolumn{1}{c}{\textbf{Legal}} & \multicolumn{2}{c}{\textbf{\begin{tabular}[b]{@{}c@{}}Instruction\\Following\end{tabular}}} \\
\cmidrule(lr){2-2}\cmidrule(lr){3-5}\cmidrule(lr){6-6}\cmidrule(lr){7-7}\cmidrule(lr){8-9}
\textbf{Benchmark} & \textbf{\begin{tabular}[b]{@{}c@{}}Arabic\\MMLU\end{tabular}} & \textbf{AraSTEM} & \textbf{\begin{tabular}[b]{@{}c@{}}GAT\\Math\end{tabular}} & \textbf{\begin{tabular}[b]{@{}c@{}}Arabic\\GSM8K\end{tabular}} & \textbf{\begin{tabular}[b]{@{}c@{}}MedAra\\Bench\end{tabular}} & \textbf{\begin{tabular}[b]{@{}c@{}}Legal\\(HELM)\end{tabular}} & \textbf{IFEval-Ar} & \textbf{\begin{tabular}[b]{@{}c@{}}AraIF\\Bench\end{tabular}} \\

 & Acc. & Acc. & Acc. & Acc. & Acc. & Acc. & Acc. & Acc. \\
\midrule
\endhead
\midrule\multicolumn{9}{r}{\small Continued on next page}\\
\endfoot
\bottomrule\addlinespace[6pt]
\caption{Arabic general capabilities results with low-effort thinking (\%). Models without a thinking mode use their non-thinking (instruct) mode. Within each model size, the best score in each column is shown in bold.}\label{tab:general-capabilities-low}
\endlastfoot
\multicolumn{9}{@{}l}{\textbf{1B Scale}} \\
\midrule
\model{Falcon-H1-1.5B-Deep-Instruct} & 48.69 & 62.49 & 40.19 & \textbf{65.43} & 32.32 & 48.84 & 49.26 & 16.20 \\
\model{Hala-1.2B} & 45.00 & 48.32 & 29.63 & 57.29 & 32.30 & 41.09 & 44.06 & 14.60 \\
\model{Qwen3.5-2B} & 44.92 & 52.18 & 27.68 & 35.91 & 32.98 & 27.13 & 40.35 & 16.90 \\
\model{tiny-aya-global} & 55.93 & 58.48 & 31.36 & 59.57 & 39.49 & 52.20 & 58.17 & 21.50 \\
\model{Jais-3-0.9B} & \textbf{60.01} & \textbf{78.67} & \textbf{49.02} & 64.68 & \textbf{44.23} & \textbf{55.56} & \textbf{63.37} & \textbf{45.30} \\
\addlinespace[3pt]\midrule
\multicolumn{9}{@{}l}{\textbf{4B Scale}} \\
\midrule
\model{Falcon-H1-3B-Instruct} & 51.02 & 64.93 & 41.29 & 72.21 & 34.53 & 52.97 & 55.69 & 18.40 \\
\model{gemma-4-E2B-it} & 60.81 & 77.34 & 54.97 & 77.32 & 39.99 & 55.30 & 63.86 & 25.80 \\
\model{granite-4.0-h-3b-ar} & 59.83 & 66.49 & 45.32 & 67.21 & 38.13 & 56.07 & 43.07 & 14.70 \\
\model{Qwen3.5-4B} & 65.30 & 73.89 & 36.92 & 80.42 & 50.53 & 57.11 & 65.84 & 29.10 \\
\model{Jais-3-3.7B} & \textbf{76.44} & \textbf{88.79} & \textbf{88.16} & \textbf{85.30} & \textbf{56.96} & \textbf{64.08} & \textbf{71.04} & \textbf{49.60} \\
\addlinespace[3pt]\midrule
\multicolumn{9}{@{}l}{\textbf{7B Scale}} \\
\midrule
\model{ALLaM-7B-Instruct-preview} & 71.21 & 78.98 & 38.38 & 56.81 & 49.12 & 61.24 & 48.02 & 18.90 \\
\model{Falcon-H1-7B-Instruct} & 63.31 & 76.58 & 45.45 & 84.99 & 47.98 & 57.88 & 62.38 & 26.20 \\
\model{Fanar-1-9B-Instruct} & 65.75 & 72.98 & 40.64 & 75.25 & 50.71 & 61.76 & 44.55 & 25.80 \\
\model{gemma-4-12B-it} & \textbf{82.32} & \textbf{92.44} & \textbf{75.81} & \textbf{90.31} & \textbf{75.14} & 64.86 & 72.03 & 43.00 \\
\model{gemma-4-E4B-it} & 70.61 & 89.48 & 48.78 & 86.60 & 53.29 & 59.17 & 67.82 & 33.80 \\
\model{Jais-2-8B-Chat} & 70.42 & 64.84 & 35.67 & 72.97 & 41.58 & 60.47 & 61.63 & 21.10 \\
\model{Qwen3.5-9B} & 71.64 & 79.84 & 48.61 & 86.83 & 59.66 & 61.76 & 68.07 & 33.80 \\
\model{Jais-3-7B} & 74.95 & 83.87 & 53.24 & 87.21 & 55.00 & \textbf{66.15} & \textbf{72.77} & \textbf{46.50} \\
\addlinespace[3pt]\midrule
\multicolumn{9}{@{}l}{\textbf{32B Scale}} \\
\midrule
\model{Falcon-H1-34B-Instruct} & 73.66 & 85.47 & 63.39 & 90.24 & 60.65 & 63.05 & 66.83 & 32.20 \\
\model{gemma-4-31B-it} & \textbf{88.55} & \textbf{96.01} & \textbf{97.73} & \textbf{93.42} & \textbf{85.76} & \textbf{72.87} & \textbf{75.99} & \textbf{55.90} \\
\model{Jais-2-70B-Chat} & 80.66 & 75.74 & 40.67 & 83.38 & 52.62 & 66.15 & 69.06 & 36.00 \\
\model{Qwen3.8-27B} & 79.23 & 92.96 & 58.01 & 92.50 & 70.21 & 64.86 & 68.07 & 43.60 \\
\model{Jais-3-32B} & 81.38 & 89.53 & 91.93 & 93.03 & 70.45 & \textbf{72.87} & 74.75 & 53.00 \\
\model{Jais-3-MoVA-36B-A4B} & 83.78 & 93.64 & 77.13 & 90.81 & 66.36 & 66.15 & 75.00 & 53.30 \\
\addlinespace[3pt]\midrule
\multicolumn{9}{@{}l}{\textbf{375B Scale}} \\
\midrule
\model{MiniMax-M3} & 77.44 & 87.92 & 70.34 & 90.98 & 67.40 & 70.80 & 66.34 & 36.00 \\
\model{Nemotron-3-Ultra-550B-A55B-BF16} & 81.24 & 89.49 & 70.52 & \textbf{92.92} & 72.29 & \textbf{74.42} & 71.78 & 47.60 \\
\model{Qwen3.5-397B-A17B} & \textbf{84.42} & \textbf{92.92} & 78.79 & 92.57 & \textbf{82.46} & 68.99 & 72.52 & 51.00 \\
\model{Qwen3.8-Flash-Next} & 82.63 & 92.61 & 67.07 & 92.80 & 74.99 & 70.03 & 67.82 & 45.00 \\
\model{Jais-3-375B-A23B} & 82.77 & 91.48 & \textbf{95.58} & 92.88 & 72.70 & 71.06 & \textbf{73.02} & \textbf{55.40} \\
\end{longtable}
\endgroup

\paragraph{Arabic domain knowledge results.}
Table~\ref{tab:arabic-domain-low} shows that \modelname leads all ten benchmarks at 1B, and nine of ten at 4B, where \model{Jais-3-3.7B} is only 0.9 points behind \model{gemma-4-E2B-it} on \benchmark{Arabic Linguistics}. At 7B it leads \benchmark{QIAS-Inheritance}, \benchmark{IslamicEval}, \benchmark{PalmX} and \benchmark{FannOrFlop\_MCQ} and ties \model{gemma-4-12B-it} on \benchmark{Alyah}, but trails it on \benchmark{QIAS-Knowledge} (by 6.5 points), \benchmark{Arabic Linguistics} (by 5.2), \benchmark{ArabCulture} (by 1.6) and \benchmark{Arabic Cuisine} (by 1.1), and \model{ALLaM-7B-Instruct-preview} on \benchmark{AraDiCE-Culture} (by 9.5). At 32B, \model{Jais-3-32B} leads \benchmark{AraDiCE-Culture} and \benchmark{Alyah}, and \model{Jais-2-70B-Chat} leads \benchmark{FannOrFlop\_MCQ}. On the other benchmarks, \model{Jais-3-32B} is within 3 points of \model{gemma-4-31B-it}, except on \benchmark{QIAS-Inheritance} (26.2 points behind) and \araLingBmarks (17.3). At 375B, \model{Jais-3-375B-A23B} leads \benchmark{QIAS-Inheritance}, \benchmark{QIAS-Knowledge}, \benchmark{IslamicEval}, \benchmark{ArabCulture}, \benchmark{Alyah} and \benchmark{Arabic Cuisine}. At 32B and 375B, low effort costs the most on \benchmark{QIAS-Inheritance}, where \model{Jais-3-32B} scores 60.5 against 95.9 at high effort, and \model{Jais-3-375B-A23B} 70.3 against 95.4.
\providecommand{\model}[1]{#1}

\begingroup
\scriptsize\rmfamily
\renewcommand{\arraystretch}{1.15}
\setlength{\tabcolsep}{2.2pt}
\lineskiplimit=0pt
\ifdefined\modelw\else\newlength{\modelw}\fi\settowidth{\modelw}{\model{Nemotron-3-Ultra-550B-A55B-BF16}}
\begin{longtable}{@{}>{\raggedright\arraybackslash}p{\modelw}*{10}{>{\centering\arraybackslash}p{\dimexpr(\linewidth-\modelw-2\tabcolsep)/10-2\tabcolsep\relax}}@{}}
\toprule
\textbf{Domain} & \multicolumn{4}{c}{\makebox[0pt][c]{\textbf{\begin{tabular}[b]{@{}c@{}}Islamic\\Knowledge\end{tabular}}}} & \multicolumn{5}{c}{\makebox[0pt][c]{\textbf{Arab Culture}}} & \multicolumn{1}{c}{\makebox[0pt][c]{\textbf{\begin{tabular}[b]{@{}c@{}}Arabic\\Linguistics\end{tabular}}}} \\
\cmidrule(lr){2-5}\cmidrule(lr){6-10}\cmidrule(lr){11-11}
\textbf{Benchmark} & \makebox[0pt][c]{\textbf{\begin{tabular}[b]{@{}c@{}}QIAS\\Inherit.\end{tabular}}} & \makebox[0pt][c]{\textbf{\begin{tabular}[b]{@{}c@{}}QIAS\\Knowl.\end{tabular}}} & \makebox[0pt][c]{\textbf{\begin{tabular}[b]{@{}c@{}}Islamic\\Eval\end{tabular}}} & \makebox[0pt][c]{\textbf{PalmX}} & \makebox[0pt][c]{\textbf{\begin{tabular}[b]{@{}c@{}}AraDiCE\\Culture\end{tabular}}} & \makebox[0pt][c]{\textbf{\begin{tabular}[b]{@{}c@{}}Arab\\Culture\end{tabular}}} & \makebox[0pt][c]{\textbf{Alyah}} & 
\makebox[0pt][c]{\textbf{\begin{tabular}[b]{@{}c@{}}FannOrFlop\\\_MCQ\end{tabular}}} &
 \makebox[0pt][c]{\textbf{\begin{tabular}[b]{@{}c@{}}Arabic\\Cuisine\end{tabular}}} & \makebox[0pt][c]{\textbf{\begin{tabular}[b]{@{}c@{}}Arabic\\Ling.\end{tabular}}} \\

 & \makebox[0pt][c]{Acc.} & \makebox[0pt][c]{Acc.} & \makebox[0pt][c]{Acc.} & \makebox[0pt][c]{Acc.} & \makebox[0pt][c]{Acc.} & \makebox[0pt][c]{Acc.} & \makebox[0pt][c]{Acc.} & \makebox[0pt][c]{Acc.} & \makebox[0pt][c]{Acc.} & \makebox[0pt][c]{Acc.} \\
\midrule
\endfirsthead
\multicolumn{11}{l}{\small\itshape Continued from previous page}\\
\toprule
\textbf{Domain} & \multicolumn{4}{c}{\makebox[0pt][c]{\textbf{\begin{tabular}[b]{@{}c@{}}Islamic\\Knowledge\end{tabular}}}} & \multicolumn{5}{c}{\makebox[0pt][c]{\textbf{Arab Culture}}} & \multicolumn{1}{c}{\makebox[0pt][c]{\textbf{\begin{tabular}[b]{@{}c@{}}Arabic\\Linguistics\end{tabular}}}} \\
\cmidrule(lr){2-5}\cmidrule(lr){6-10}\cmidrule(lr){11-11}
\textbf{Benchmark} & \makebox[0pt][c]{\textbf{\begin{tabular}[b]{@{}c@{}}QIAS\\Inherit.\end{tabular}}} & \makebox[0pt][c]{\textbf{\begin{tabular}[b]{@{}c@{}}QIAS\\Knowl.\end{tabular}}} & \makebox[0pt][c]{\textbf{\begin{tabular}[b]{@{}c@{}}Islamic\\Eval\end{tabular}}} & \makebox[0pt][c]{\textbf{PalmX}} & \makebox[0pt][c]{\textbf{\begin{tabular}[b]{@{}c@{}}AraDiCE\\Culture\end{tabular}}} & \makebox[0pt][c]{\textbf{\begin{tabular}[b]{@{}c@{}}Arab\\Culture\end{tabular}}} & \makebox[0pt][c]{\textbf{Alyah}} & 
\makebox[0pt][c]{\textbf{\begin{tabular}[b]{@{}c@{}}FannOrFlop\\\_MCQ\end{tabular}}} &
\makebox[0pt][c]{\textbf{\begin{tabular}[b]{@{}c@{}}Arabic\\Cuisine\end{tabular}}} & \makebox[0pt][c]{\textbf{\begin{tabular}[b]{@{}c@{}}Arabic\\Ling.\end{tabular}}} \\

 & \makebox[0pt][c]{Acc.} & \makebox[0pt][c]{Acc.} & \makebox[0pt][c]{Acc.} & \makebox[0pt][c]{Acc.} & \makebox[0pt][c]{Acc.} & \makebox[0pt][c]{Acc.} & \makebox[0pt][c]{Acc.} & \makebox[0pt][c]{Acc.} & \makebox[0pt][c]{Acc.} & \makebox[0pt][c]{Acc.} \\
\midrule
\endhead
\midrule\multicolumn{11}{r}{\small Continued on next page}\\
\endfoot
\bottomrule\addlinespace[6pt]
\caption{Arabic domain knowledge results with low-effort thinking (\%). Models without a thinking mode use their non-thinking (instruct) mode. Within each model size, the best score in each column is shown in bold.}\label{tab:arabic-domain-low}
\endlastfoot
\multicolumn{11}{@{}l}{\textbf{1B Scale}} \\
\midrule
\model{Falcon-H1-1.5B-Deep-Instruct} & 32.07 & 56.02 & 42.42 & 62.13 & 49.16 & 61.69 & 54.75 & 38.54 & 52.35 & 33.27 \\
\model{Hala-1.2B} & 16.15 & 41.51 & 36.36 & 49.58 & 30.73 & 60.63 & 53.20 & 31.34 & 49.48 & 32.56 \\
\model{Qwen3.5-2B} & 17.93 & 41.15 & 38.53 & 58.54 & 37.99 & 50.50 & 50.00 & 35.40 & 47.45 & 37.50 \\
\model{tiny-aya-global} & 23.05 & 54.72 & 27.27 & 71.62 & 55.31 & 68.61 & 68.13 & 48.50 & 60.64 & 38.16 \\
\model{Jais-3-0.9B} & \textbf{46.33} & \textbf{62.70} & \textbf{71.43} & \textbf{74.05} & \textbf{58.10} & \textbf{71.60} & \textbf{72.97} & \textbf{51.81} & \textbf{65.47} & \textbf{62.10} \\
\addlinespace[3pt]\midrule
\multicolumn{11}{@{}l}{\textbf{4B Scale}} \\
\midrule
\model{Falcon-H1-3B-Instruct} & 27.51 & 54.28 & 44.59 & 66.56 & 45.81 & 64.18 & 58.89 & 41.09 & 54.37 & 42.05 \\
\model{gemma-4-E2B-it} & 42.20 & 60.81 & 59.31 & 73.52 & 45.25 & 70.49 & 70.03 & 57.37 & 63.32 & \textbf{69.01} \\
\model{granite-4.0-h-3b-ar} & 18.15 & 66.47 & 22.51 & 57.49 & 53.63 & 68.41 & 65.98 & 48.76 & 58.55 & 49.34 \\
\model{Qwen3.5-4B} & 30.62 & 69.09 & 58.01 & 76.69 & 60.89 & 70.93 & 63.39 & 48.28 & 65.14 & 55.68 \\
\model{Jais-3-3.7B} & \textbf{53.67} & \textbf{75.62} & \textbf{77.49} & \textbf{86.60} & \textbf{79.89} & \textbf{82.81} & \textbf{79.97} & \textbf{63.95} & \textbf{81.66} & 68.15 \\
\addlinespace[3pt]\midrule
\multicolumn{11}{@{}l}{\textbf{7B Scale}} \\
\midrule
\model{ALLaM-7B-Instruct-preview} & 31.18 & 72.86 & 78.36 & 85.86 & \textbf{79.33} & 77.65 & 77.46 & 66.67 & 71.34 & 67.46 \\
\model{Falcon-H1-7B-Instruct} & 43.99 & 65.75 & 61.90 & 74.26 & 59.22 & 72.43 & 65.72 & 51.38 & 64.95 & 47.53 \\
\model{Fanar-1-9B-Instruct} & 36.30 & 64.88 & 69.70 & 80.27 & 67.04 & 77.30 & 69.60 & 63.27 & 68.86 & 57.59 \\
\model{gemma-4-12B-it} & 70.04 & \textbf{82.44} & 76.19 & 85.23 & 72.63 & \textbf{86.18} & \textbf{81.43} & 65.69 & \textbf{83.22} & \textbf{73.89} \\
\model{gemma-4-E4B-it} & 45.88 & 69.81 & 68.83 & 81.33 & 68.72 & 77.03 & 74.78 & 61.31 & 72.58 & 70.46 \\
\model{Jais-2-8B-Chat} & 29.06 & 67.49 & 56.71 & 82.81 & 67.60 & 74.60 & 76.42 & 60.50 & 71.61 & 49.20 \\
\model{Qwen3.5-9B} & 39.53 & 72.28 & 68.40 & 80.17 & 64.25 & 76.86 & 68.39 & 54.31 & 71.87 & 61.75 \\
\model{Jais-3-7B} & \textbf{85.75} & 75.91 & \textbf{80.95} & \textbf{87.24} & 69.83 & 84.54 & \textbf{81.43} & \textbf{66.75} & 82.11 & 68.67 \\
\addlinespace[3pt]\midrule
\multicolumn{11}{@{}l}{\textbf{32B Scale}} \\
\midrule
\model{Falcon-H1-34B-Instruct} & 56.24 & 73.44 & 70.13 & 84.70 & 75.42 & 79.67 & 71.76 & 62.35 & 74.41 & 57.44 \\
\model{gemma-4-31B-it} & \textbf{86.64} & \textbf{84.91} & \textbf{87.88} & \textbf{93.14} & 83.80 & \textbf{90.47} & 86.61 & 69.57 & \textbf{88.64} & \textbf{91.23} \\
\model{Jais-2-70B-Chat} & 42.20 & 78.96 & 65.80 & 90.72 & 80.45 & 82.08 & 84.46 & \textbf{73.62} & 80.42 & 64.10 \\
\model{Qwen3.8-27B} & 53.79 & 80.26 & 72.29 & 88.40 & 75.98 & 83.46 & 78.15 & 64.32 & 79.77 & 74.92 \\
\model{Jais-3-32B} & 60.47 & 83.60 & 86.15 & 91.14 & \textbf{86.03} & 87.47 & \textbf{87.48} & 69.34 & 88.05 & 73.96 \\
\model{Jais-3-MoVA-36B-A4B} & 62.03 & 82.00 & 82.68 & 89.35 & 82.12 & 87.86 & 84.72 & 68.15 & 87.14 & 76.63 \\
\addlinespace[3pt]\midrule
\multicolumn{11}{@{}l}{\textbf{375B Scale}} \\
\midrule
\model{MiniMax-M3} & 52.34 & 78.37 & 75.76 & 89.14 & 75.98 & 80.05 & 75.73 & 61.12 & 78.26 & 69.97 \\
\model{Nemotron-3-Ultra-550B-A55B-BF16} & 49.55 & 82.29 & 85.28 & 91.14 & 82.68 & 86.71 & 76.60 & 66.75 & 84.73 & 75.41 \\
\model{Qwen3.5-397B-A17B} & 54.68 & 84.47 & 82.25 & \textbf{94.20} & \textbf{84.36} & 87.15 & 82.90 & \textbf{72.42} & 85.25 & 83.03 \\
\model{Qwen3.8-Flash-Next} & 54.23 & 85.20 & 64.07 & 91.77 & 80.45 & 86.71 & 81.26 & 69.75 & 85.12 & \textbf{84.39} \\
\model{Jais-3-375B-A23B} & \textbf{70.27} & \textbf{86.07} & \textbf{87.45} & 91.24 & 79.89 & \textbf{88.94} & \textbf{86.18} & 70.82 & \textbf{88.97} & 83.72 \\
\end{longtable}
\endgroup

\paragraph{Dialect understanding and generation results.}
Table~\ref{tab:arabic-dialect-low} shows that \modelname leads \benchmark{QADI} at every scale (55.0 at 1B, 55.9 at 4B, 62.9 at 7B, 69.2 at 32B, and 70.5 at 375B). At 1B and 4B it is best on six of the seven columns; the exceptions are \benchmark{Dialectal-Arabic-MMLU} at 1B, where it is 0.2 points behind \model{tiny-aya-global}, and Darija transliteration at 4B, where it is 4.7 points behind \model{gemma-4-E2B-it}. At 7B and 32B the translation and transliteration scores are at or near the top (for example, \model{Jais-3-7B} leads \benchmark{FLORES+} with 89.1 and Darija transliteration with 93.5), but the MMLU-style benchmarks favor \model{Gemma}: \model{gemma-4-12B-it} is ahead of \model{Jais-3-7B} by 7--12 points, and \model{gemma-4-31B-it} is ahead of \model{Jais-3-MoVA-36B-A4B} by about 5 points on \benchmark{ArabicMMLU} and 11 points on \benchmark{Dialectal-Arabic-MMLU}. At 375B, \model{Jais-3-375B-A23B} leads \benchmark{QADI}, \benchmark{FLORES+} and both transliteration tasks, and trails \model{Qwen3.5-397B-A17B} by about 2 points on the three MMLU benchmarks.
\providecommand{\model}[1]{#1}

\begingroup
\scriptsize\rmfamily
\renewcommand{\arraystretch}{1.15}
\setlength{\tabcolsep}{2.2pt}
\lineskiplimit=0pt
\ifdefined\modelw\else\newlength{\modelw}\fi\settowidth{\modelw}{\model{Nemotron-3-Ultra-550B-A55B-BF16}}
\begin{longtable}{@{}>{\raggedright\arraybackslash}p{\modelw}*{7}{>{\centering\arraybackslash}p{\dimexpr(0.85\linewidth-\modelw-2\tabcolsep)/7-2\tabcolsep\relax}}@{}}
\toprule
\textbf{Domain} & \multicolumn{7}{c}{\makebox[0pt][c]{\textbf{Dialect Understanding and Generation}}} \\
\cmidrule(lr){2-8}
\textbf{Benchmark} & \makebox[0pt][c]{\textbf{QADI}} & \makebox[0pt][c]{\textbf{\begin{tabular}[b]{@{}c@{}}Dialect.\\MMLU\end{tabular}}} & \makebox[0pt][c]{\textbf{\begin{tabular}[b]{@{}c@{}}Arabic\\MMLU\end{tabular}}} & \makebox[0pt][c]{\textbf{\begin{tabular}[b]{@{}c@{}}MMLU\\diacr.\end{tabular}}} & \makebox[0pt][c]{\textbf{\begin{tabular}[b]{@{}c@{}}FLORES+\end{tabular}}} & \makebox[0pt][c]{\textbf{\begin{tabular}[b]{@{}c@{}}Translit.\\EG\end{tabular}}} & \makebox[0pt][c]{\textbf{\begin{tabular}[b]{@{}c@{}}Translit.\\DR\end{tabular}}} \\

 & \makebox[0pt][c]{Acc.} & \makebox[0pt][c]{Acc.} & \makebox[0pt][c]{Acc.} & \makebox[0pt][c]{Acc.} & \makebox[0pt][c]{Judge} & \makebox[0pt][c]{Judge} & \makebox[0pt][c]{Judge} \\
\midrule
\endfirsthead
\multicolumn{8}{l}{\small\itshape Continued from previous page}\\
\toprule
\textbf{Domain} & \multicolumn{7}{c}{\makebox[0pt][c]{\textbf{Dialect Understanding and Generation}}} \\
\cmidrule(lr){2-8}
\textbf{Benchmark} & \makebox[0pt][c]{\textbf{QADI}} & \makebox[0pt][c]{\textbf{\begin{tabular}[b]{@{}c@{}}Dialect.\\MMLU\end{tabular}}} & \makebox[0pt][c]{\textbf{\begin{tabular}[b]{@{}c@{}}Arabic\\MMLU\end{tabular}}} & \makebox[0pt][c]{\textbf{\begin{tabular}[b]{@{}c@{}}MMLU\\diacr.\end{tabular}}} & \makebox[0pt][c]{\textbf{\begin{tabular}[b]{@{}c@{}}FLORES+\end{tabular}}} & \makebox[0pt][c]{\textbf{\begin{tabular}[b]{@{}c@{}}Translit.\\EG\end{tabular}}} & \makebox[0pt][c]{\textbf{\begin{tabular}[b]{@{}c@{}}Translit.\\DR\end{tabular}}} \\

 & \makebox[0pt][c]{Acc.} & \makebox[0pt][c]{Acc.} & \makebox[0pt][c]{Acc.} & \makebox[0pt][c]{Acc.} & \makebox[0pt][c]{Judge} & \makebox[0pt][c]{Judge} & \makebox[0pt][c]{Judge} \\
\midrule
\endhead
\midrule\multicolumn{8}{r}{\small Continued on next page}\\
\endfoot
\bottomrule\addlinespace[6pt]
\caption{Arabic dialect understanding and generation results with low-effort thinking (\%). Models without a thinking mode use their non-thinking (instruct) mode. Within each model size, the best score in each column is shown in bold.}\label{tab:arabic-dialect-low}
\endlastfoot
\multicolumn{8}{@{}l}{\textbf{1B Scale}} \\
\midrule
\model{Falcon-H1-1.5B-Deep-Instruct} & 23.39 & 42.26 & 48.69 & 34.68 & 46.78 & 30.53 & 42.35 \\
\model{Hala-1.2B} & 22.97 & 40.33 & 45.00 & 31.28 & 42.25 & 34.94 & 25.72 \\
\model{Qwen3.5-2B} & 23.01 & 37.90 & 44.92 & 34.02 & 42.59 & 30.49 & 40.05 \\
\model{tiny-aya-global} & 38.79 & \textbf{47.68} & 55.93 & 45.64 & 63.79 & 38.37 & 45.75 \\
\model{Jais-3-0.9B} & \textbf{54.96} & 47.50 & \textbf{60.01} & \textbf{52.15} & \textbf{82.36} & \textbf{73.17} & \textbf{80.75} \\
\addlinespace[3pt]\midrule
\multicolumn{8}{@{}l}{\textbf{4B Scale}} \\
\midrule
\model{Falcon-H1-3B-Instruct} & 28.52 & 43.81 & 51.02 & 34.62 & 53.84 & 41.26 & 39.33 \\
\model{gemma-4-E2B-it} & 40.74 & 50.06 & 60.81 & 53.47 & 73.78 & 61.04 & \textbf{84.13} \\
\model{granite-4.0-h-3b-ar} & 42.19 & 50.71 & 59.83 & 53.25 & 48.72 & 35.33 & 40.66 \\
\model{Qwen3.5-4B} & 30.76 & 56.70 & 65.30 & 54.32 & 61.55 & 47.21 & 60.37 \\
\model{Jais-3-3.7B} & \textbf{55.91} & \textbf{62.94} & \textbf{76.44} & \textbf{74.05} & \textbf{86.45} & \textbf{80.06} & 79.45 \\
\addlinespace[3pt]\midrule
\multicolumn{8}{@{}l}{\textbf{7B Scale}} \\
\midrule
\model{ALLaM-7B-Instruct-preview} & 42.77 & 56.61 & 71.21 & 61.85 & 74.47 & 39.71 & 31.81 \\
\model{Falcon-H1-7B-Instruct} & 38.08 & 58.46 & 63.31 & 52.33 & 59.76 & 42.47 & 48.74 \\
\model{Fanar-1-9B-Instruct} & 46.19 & 58.21 & 65.75 & 54.86 & 68.35 & 50.18 & 65.67 \\
\model{gemma-4-12B-it} & 55.97 & \textbf{77.61} & \textbf{82.32} & \textbf{80.69} & 86.80 & 84.65 & 92.83 \\
\model{gemma-4-E4B-it} & 52.01 & 63.91 & 70.61 & 66.04 & 79.32 & 80.50 & 91.43 \\
\model{Jais-2-8B-Chat} & 48.38 & 54.59 & 70.42 & 54.66 & 79.40 & \textbf{88.76} & 76.73 \\
\model{Qwen3.5-9B} & 27.13 & 62.85 & 71.64 & 63.27 & 67.03 & 50.45 & 76.26 \\
\model{Jais-3-7B} & \textbf{62.94} & 66.15 & 74.95 & 70.75 & \textbf{89.10} & 87.63 & \textbf{93.45} \\
\addlinespace[3pt]\midrule
\multicolumn{8}{@{}l}{\textbf{32B Scale}} \\
\midrule
\model{Falcon-H1-34B-Instruct} & 50.28 & 71.47 & 73.66 & 69.09 & 70.58 & 54.70 & 75.75 \\
\model{gemma-4-31B-it} & 68.08 & \textbf{87.04} & \textbf{88.55} & \textbf{87.42} & \textbf{91.54} & 90.42 & \textbf{96.03} \\
\model{Jais-2-70B-Chat} & 65.08 & 67.56 & 80.66 & 74.05 & 83.83 & \textbf{90.60} & 92.75 \\
\model{Qwen3.8-27B} & 46.91 & 74.87 & 79.23 & 76.85 & 75.46 & 58.45 & 82.68 \\
\model{Jais-3-32B} & \textbf{69.16} & 75.28 & 81.38 & 79.17 & 90.74 & 90.59 & 95.39 \\
\model{Jais-3-MoVA-36B-A4B} & 68.98 & 75.92 & 83.78 & 82.64 & 90.19 & 89.31 & 95.05 \\
\addlinespace[3pt]\midrule
\multicolumn{8}{@{}l}{\textbf{375B Scale}} \\
\midrule
\model{MiniMax-M3} & 40.11 & 70.94 & 77.44 & 74.08 & 81.31 & 76.21 & 75.67 \\
\model{Nemotron-3-Ultra-550B-A55B-BF16} & 57.94 & 78.76 & 81.24 & 78.04 & 78.62 & 76.21 & 89.09 \\
\model{Qwen3.5-397B-A17B} & 58.49 & \textbf{82.91} & \textbf{84.42} & \textbf{82.53} & 84.04 & 77.96 & 93.02 \\
\model{Qwen3.8-Flash-Next} & 57.23 & 81.34 & 82.63 & 79.43 & 81.30 & 77.53 & 86.85 \\
\model{Jais-3-375B-A23B} & \textbf{70.48} & 80.97 & 82.77 & 80.34 & \textbf{91.20} & \textbf{90.69} & \textbf{95.40} \\
\end{longtable}
\endgroup

\paragraph{Multi-turn and long context understanding results.}
Table~\ref{tab:arabic-mt-lc-low} shows that \modelname leads \benchmark{MT-Bench Arabic} at the 1B and 4B scales: \model{Jais-3-0.9B} outperforms the next best 1B model by about 15 points, and \model{Jais-3-3.7B} is about 4 points ahead of \model{gemma-4-E2B-it}. At larger scales the gap to the best open model is small but not closed: \model{Jais-3-32B} is within 3 points of \model{gemma-4-31B-it}, \model{Jais-3-375B-A23B} is within 3 points of \model{Qwen3.8-Flash-Next}, and \model{Jais-3-7B} trails \model{gemma-4-12B-it} by about 7 points.

Long-context accuracy on \benchmark{AA-LCR-Ar} is where low effort costs \modelname the most. Scores are far below those at high
effort (for example \model{Jais-3-3.7B} scores 19 against 68, and \model{Jais-3-375B-A23B} 25 against 77), and \modelname trails the strongest baselines in each class: \model{Jais-3-3.7B} scores 19 against 33 for \model{Qwen3.5-4B}, \model{Jais-3-32B} 24 against 63 for \model{Qwen3.8-27B}, and \model{Jais-3-375B-A23B} 25 against 74 for \model{Nemotron-3-Ultra-550B-A55B-BF16}. At 7B, \model{Jais-3-7B} scores 30, behind \model{Qwen3.5-9B} but ahead of \model{gemma-4-12B-it}.
\providecommand{\model}[1]{#1}

\begin{table}[H]
\centering
\scriptsize\rmfamily
\renewcommand{\arraystretch}{1.15}
\setlength{\tabcolsep}{0.6pt}
\begin{tabular*}{0.48\linewidth}[t]{@{\extracolsep{\fill}}l*{4}{c}@{}}
\toprule
\textbf{Domain} & \multicolumn{3}{c}{\textbf{Multi-turn}} & \multicolumn{1}{c}{\textbf{\begin{tabular}[b]{@{}c@{}}Long\\Context\end{tabular}}} \\
\cmidrule(lr){2-4}\cmidrule(lr){5-5}
\textbf{Benchmark} & \multicolumn{3}{c}{\textbf{MT-Bench~Arabic}} & \textbf{AA-LCR-Ar} \\
\cmidrule(lr){2-4}
 & Turn 1 & Turn 2 & Avg. & Acc. \\
\midrule
\multicolumn{5}{@{}l}{\textbf{1B Scale}} \\
\midrule
\model{Falcon-H1-1.5B-Deep-Instruct} & 40.75 & 30.50 & 35.62 & -- \\
\model{Hala-1.2B} & 54.12 & 27.00 & 40.56 & -- \\
\model{Qwen3.5-2B} & 24.25 & 18.88 & 21.56 & \textbf{12.00} \\
\model{tiny-aya-global} & 59.00 & 38.12 & 48.56 & -- \\
\model{Jais-3-0.9B} & \textbf{72.62} & \textbf{54.38} & \textbf{63.50} & -- \\
\addlinespace[3pt]\midrule
\multicolumn{5}{@{}l}{\textbf{4B Scale}} \\
\midrule
\model{Falcon-H1-3B-Instruct} & 58.75 & 51.25 & 55.00 & -- \\
\model{gemma-4-E2B-it} & 86.12 & 79.11 & 82.64 & -- \\
\model{granite-4.0-h-3b-ar} & 54.38 & 33.75 & 44.06 & -- \\
\model{Qwen3.5-4B} & 80.25 & 74.75 & 77.50 & \textbf{33.00} \\
\model{Jais-3-3.7B} & \textbf{88.50} & \textbf{85.50} & \textbf{87.00} & 19.00 \\
\addlinespace[3pt]\midrule
\multicolumn{5}{@{}l}{\textbf{7B Scale}} \\
\midrule
\model{ALLaM-7B-Instruct-preview} & 62.50 & 44.00 & 53.25 & -- \\
\model{Falcon-H1-7B-Instruct} & 69.88 & 61.27 & 65.60 & 9.00 \\
\model{Fanar-1-9B-Instruct} & 76.50 & 62.12 & 69.31 & -- \\
\model{gemma-4-12B-it} & \textbf{98.00} & \textbf{97.00} & \textbf{97.50} & 23.00 \\
\model{gemma-4-E4B-it} & 97.25 & 94.00 & 95.62 & -- \\
\model{Jais-2-8B-Chat} & 74.12 & 56.00 & 65.06 & -- \\
\model{Qwen3.5-9B} & 91.50 & 87.75 & 89.62 & \textbf{37.00} \\
\model{Jais-3-7B} & 94.75 & 87.12 & 90.94 & 30.00 \\
\bottomrule
\end{tabular*}
\hfill
\begin{tabular*}{0.50\linewidth}[t]{@{\extracolsep{\fill}}l*{4}{c}@{}}
\toprule
\textbf{Domain} & \multicolumn{3}{c}{\textbf{Multi-turn}} & \multicolumn{1}{c}{\textbf{\begin{tabular}[b]{@{}c@{}}Long\\Context\end{tabular}}} \\
\cmidrule(lr){2-4}\cmidrule(lr){5-5}
\textbf{Benchmark} & \multicolumn{3}{c}{\textbf{MT-Bench~Arabic}} & \textbf{AA-LCR-Ar} \\
\cmidrule(lr){2-4}
 & Turn 1 & Turn 2 & Avg. & Acc. \\
\midrule
\multicolumn{5}{@{}l}{\textbf{32B Scale}} \\
\midrule
\model{Falcon-H1-34B-Instruct} & 81.38 & 70.62 & 76.00 & 14.00 \\
\model{gemma-4-31B-it} & 98.12 & \textbf{98.99} & \textbf{98.55} & 46.00 \\
\model{Jais-2-70B-Chat} & 77.75 & 66.88 & 72.31 & -- \\
\model{Qwen3.8-27B} & 96.75 & 93.75 & 95.25 & \textbf{63.00} \\
\model{Jais-3-32B} & \textbf{98.38} & 93.25 & 95.81 & 24.00 \\
\model{Jais-3-MoVA-36B-A4B} & 95.38 & 88.50 & 91.94 & 28.00 \\
\addlinespace[3pt]\midrule
\multicolumn{5}{@{}l}{\textbf{375B Scale}} \\
\midrule
\model{MiniMax-M3} & 95.12 & 93.38 & 94.25 & 56.00 \\
\model{Nemotron-3-Ultra-550B-A55B-BF16} & 94.75 & 96.62 & 95.69 & \textbf{74.00} \\
\model{Qwen3.5-397B-A17B} & 97.00 & 96.88 & 96.94 & 57.00 \\
\model{Qwen3.8-Flash-Next} & \textbf{99.62} & \textbf{97.25} & \textbf{98.44} & 66.00 \\
\model{Jais-3-375B-A23B} & 98.12 & 93.12 & 95.62 & 25.00 \\
\bottomrule
\end{tabular*}
\caption{Arabic multi-turn and long-context results with low-effort thinking (\%). Models without a thinking mode use their non-thinking (instruct) mode. Within each model size, the best score in each column is shown in bold. MT-Bench Arabic turns are the judge's 1--10 ratings $\times$10 and Avg.\ is their mean; -- marks results that are not available.}\label{tab:arabic-mt-lc-low}
\end{table}

\subsubsection{English Evaluation}
\paragraph{General capabilities results.}
Table~\ref{tab:english-general-low} shows that at low effort \modelname is strongest on math from 4B upward. At 1B, \model{Jais-3-0.9B} trails \model{Falcon-H1-1.5B-Deep-Instruct} on every benchmark except \benchmark{AIME~2025}, where they tie. At 4B, \model{Jais-3-3.7B} leads \benchmark{GPQA-Diamond}, \benchmark{GSM8K}, \benchmark{GSM8K-CoT}, \benchmark{MATH-500} and \benchmark{AIME~2025}, and is within 2 points of the best model on the rest. At 7B, \model{Jais-3-7B} leads \benchmark{GPQA-Diamond} and \benchmark{MATH-500}, ties \model{Qwen3.5-9B} on \benchmark{AIME~2025}, and is within 4 points of the best model elsewhere. At 32B, \model{Jais-3-32B} leads \benchmark{MATH-500} and \benchmark{AIME~2026} and is within 3 points of \model{gemma-4-31B-it} on most benchmarks, but trails it by 15 points on \benchmark{GPQA-Diamond} and \model{Qwen3.8-27B} by 7 points on \benchmark{AIME~2025}. At 375B, \model{Jais-3-375B-A23B} leads \benchmark{GPQA-Diamond}, \benchmark{GSM8K}, \benchmark{GSM8K-CoT} and \benchmark{IFEval}, trails \model{Qwen3.5-397B-A17B} on \benchmark{MMLU-Pro}, and trails \model{Nemotron-3-Ultra-550B-A55B-BF16} on \benchmark{AIME} by 8--13 points. Compared with high effort, \benchmark{AIME}, \benchmark{GPQA-Diamond} and \benchmark{IFEval} drop for \model{Jais-3-375B-A23B} (by 19--23, 8 and 5 points), while \benchmark{GSM8K} and \benchmark{MATH-500} are unchanged.
\providecommand{\model}[1]{#1}

\begingroup
\scriptsize\rmfamily
\renewcommand{\arraystretch}{1.15}
\setlength{\tabcolsep}{2.2pt}
\lineskiplimit=0pt
\ifdefined\modelw\else\newlength{\modelw}\fi\settowidth{\modelw}{\model{Nemotron-3-Ultra-550B-A55B-BF16}}
\begin{longtable}{@{}>{\raggedright\arraybackslash}p{\modelw}*{8}{>{\centering\arraybackslash}p{\dimexpr(\linewidth-\modelw-2\tabcolsep)/8-2\tabcolsep\relax}}@{}}
\toprule
\textbf{Domain} & \multicolumn{7}{c}{\makebox[0pt][c]{\textbf{World Knowledge \& STEM}}} & \multicolumn{1}{c}{\makebox[0pt][c]{\textbf{\begin{tabular}[b]{@{}c@{}}Instruction\\Following\end{tabular}}}} \\
\cmidrule(lr){2-8}\cmidrule(lr){9-9}
\textbf{Benchmark} & \makebox[0pt][c]{\textbf{\begin{tabular}[b]{@{}c@{}}MMLU\\Pro\end{tabular}}} & \makebox[0pt][c]{\textbf{\begin{tabular}[b]{@{}c@{}}GPQA\\Diamond\end{tabular}}} & \makebox[0pt][c]{\textbf{GSM8K}} & \makebox[0pt][c]{\textbf{\begin{tabular}[b]{@{}c@{}}GSM8K\\CoT\end{tabular}}} & \makebox[0pt][c]{\textbf{\begin{tabular}[b]{@{}c@{}}MATH\\500\end{tabular}}} & \makebox[0pt][c]{\textbf{\begin{tabular}[b]{@{}c@{}}AIME\\25\end{tabular}}} & \makebox[0pt][c]{\textbf{\begin{tabular}[b]{@{}c@{}}AIME\\26\end{tabular}}} & \makebox[0pt][c]{\textbf{IFEval}} \\

 & \makebox[0pt][c]{Acc.} & \makebox[0pt][c]{Acc.} & \makebox[0pt][c]{Acc.} & \makebox[0pt][c]{Acc.} & \makebox[0pt][c]{Acc.} & \makebox[0pt][c]{pass@1} & \makebox[0pt][c]{pass@1} & \makebox[0pt][c]{Acc.} \\
\midrule
\endfirsthead
\multicolumn{9}{l}{\small\itshape Continued from previous page}\\
\toprule
\textbf{Domain} & \multicolumn{7}{c}{\makebox[0pt][c]{\textbf{World Knowledge \& STEM}}} & \multicolumn{1}{c}{\makebox[0pt][c]{\textbf{\begin{tabular}[b]{@{}c@{}}Instruction\\Following\end{tabular}}}} \\
\cmidrule(lr){2-8}\cmidrule(lr){9-9}
\textbf{Benchmark} & \makebox[0pt][c]{\textbf{\begin{tabular}[b]{@{}c@{}}MMLU\\Pro\end{tabular}}} & \makebox[0pt][c]{\textbf{\begin{tabular}[b]{@{}c@{}}GPQA\\Diamond\end{tabular}}} & \makebox[0pt][c]{\textbf{GSM8K}} & \makebox[0pt][c]{\textbf{\begin{tabular}[b]{@{}c@{}}GSM8K\\CoT\end{tabular}}} & \makebox[0pt][c]{\textbf{\begin{tabular}[b]{@{}c@{}}MATH\\500\end{tabular}}} & \makebox[0pt][c]{\textbf{\begin{tabular}[b]{@{}c@{}}AIME\\25\end{tabular}}} & \makebox[0pt][c]{\textbf{\begin{tabular}[b]{@{}c@{}}AIME\\26\end{tabular}}} & \makebox[0pt][c]{\textbf{IFEval}} \\

 & \makebox[0pt][c]{Acc.} & \makebox[0pt][c]{Acc.} & \makebox[0pt][c]{Acc.} & \makebox[0pt][c]{Acc.} & \makebox[0pt][c]{Acc.} & \makebox[0pt][c]{pass@1} & \makebox[0pt][c]{pass@1} & \makebox[0pt][c]{Acc.} \\
\midrule
\endhead
\midrule\multicolumn{9}{r}{\small Continued on next page}\\
\endfoot
\bottomrule\addlinespace[6pt]
\caption{English knowledge, STEM and instruction-following results with low-effort thinking (\%). Models without a thinking mode use their non-thinking (instruct) mode. Within each model size, the best score in each column is shown in bold.}\label{tab:english-general-low}
\endlastfoot
\multicolumn{9}{@{}l}{\textbf{1B Scale}} \\
\midrule
\model{Falcon-H1-1.5B-Deep-Instruct} & 54.45 & \textbf{43.43} & \textbf{91.96} & \textbf{88.63} & \textbf{79.40} & 12.92 & 8.75 & \textbf{81.89} \\
\model{Hala-1.2B} & 21.02 & 24.75 & 54.17 & 55.00 & 34.00 & 0.00 & 3.33 & 43.25 \\
\model{Qwen3.5-2B} & \textbf{56.10} & 30.81 & 71.57 & 78.05 & 75.80 & \textbf{15.83} & \textbf{20.00} & 61.18 \\
\model{tiny-aya-global} & 32.08 & 31.31 & 70.81 & 67.70 & 28.60 & 1.25 & 0.00 & 73.38 \\
\model{Jais-3-0.9B} & 48.17 & 34.34 & 78.54 & 78.43 & 71.40 & 12.92 & 6.25 & 69.13 \\
\addlinespace[3pt]\midrule
\multicolumn{9}{@{}l}{\textbf{4B Scale}} \\
\midrule
\model{Falcon-H1-3B-Instruct} & 57.41 & 30.30 & 90.30 & 91.17 & 79.40 & 17.08 & 8.75 & \textbf{83.36} \\
\model{gemma-4-E2B-it} & 61.12 & 44.95 & 90.30 & 90.07 & 88.40 & 29.58 & 35.83 & 79.30 \\
\model{granite-4.0-h-3b-ar} & 38.77 & 33.84 & 77.60 & 81.58 & 52.60 & 0.00 & 0.00 & 59.33 \\
\model{Qwen3.5-4B} & \textbf{76.35} & 42.42 & 93.63 & 92.95 & 94.20 & 51.67 & \textbf{60.42} & 81.52 \\
\model{Jais-3-3.7B} & 74.77 & \textbf{62.12} & \textbf{95.45} & \textbf{95.56} & \textbf{96.80} & \textbf{54.17} & 58.75 & 81.70 \\
\addlinespace[3pt]\midrule
\multicolumn{9}{@{}l}{\textbf{7B Scale}} \\
\midrule
\model{ALLaM-7B-Instruct-preview} & 33.78 & 26.26 & 66.15 & 63.91 & 17.20 & 0.00 & 0.00 & 62.85 \\
\model{Falcon-H1-7B-Instruct} & 65.87 & 34.34 & 91.74 & 89.08 & 81.00 & 15.00 & 9.58 & 84.29 \\
\model{Fanar-1-9B-Instruct} & 51.01 & 36.87 & 85.75 & 86.05 & 53.60 & 0.42 & 0.42 & 60.26 \\
\model{gemma-4-12B-it} & 78.71 & 63.13 & \textbf{96.25} & \textbf{96.13} & 96.80 & 56.25 & 67.92 & \textbf{89.09} \\
\model{gemma-4-E4B-it} & 70.27 & 52.53 & 93.10 & 92.57 & 93.80 & 37.50 & 45.83 & 84.10 \\
\model{Jais-2-8B-Chat} & 34.30 & 24.24 & 75.97 & 79.08 & 50.40 & 0.42 & 0.00 & 69.32 \\
\model{Qwen3.5-9B} & \textbf{80.59} & 45.96 & 94.77 & 95.15 & 83.20 & \textbf{62.50} & \textbf{69.17} & 84.66 \\
\model{Jais-3-7B} & 77.01 & \textbf{64.65} & 95.19 & 95.38 & \textbf{97.00} & \textbf{62.50} & 68.75 & 86.14 \\
\addlinespace[3pt]\midrule
\multicolumn{9}{@{}l}{\textbf{32B Scale}} \\
\midrule
\model{Falcon-H1-34B-Instruct} & 69.40 & 47.47 & 94.09 & 94.92 & 86.00 & 19.17 & 17.92 & 86.51 \\
\model{gemma-4-31B-it} & \textbf{85.54} & \textbf{81.31} & \textbf{97.19} & \textbf{97.12} & 98.60 & 69.17 & 78.33 & \textbf{90.94} \\
\model{Jais-2-70B-Chat} & 54.89 & 37.88 & 86.85 & 88.86 & 71.20 & 5.00 & 5.83 & 76.89 \\
\model{Qwen3.8-27B} & 83.91 & 80.30 & 96.66 & 96.82 & 98.80 & \textbf{71.67} & 79.58 & 82.44 \\
\model{Jais-3-32B} & 82.74 & 66.16 & 96.97 & 97.04 & \textbf{99.80} & 64.58 & \textbf{81.25} & 90.02 \\
\model{Jais-3-MoVA-36B-A4B} & 81.69 & 72.22 & 97.04 & 96.78 & 98.60 & 67.50 & 72.08 & 87.43 \\
\addlinespace[3pt]\midrule
\multicolumn{9}{@{}l}{\textbf{375B Scale}} \\
\midrule
\model{MiniMax-M3} & 80.43 & 56.06 & 91.66 & 95.91 & 90.40 & 15.42 & 22.08 & 78.93 \\
\model{Nemotron-3-Ultra-550B-A55B-BF16} & 83.77 & 46.46 & 90.90 & 96.36 & 97.00 & \textbf{88.33} & \textbf{87.08} & 87.06 \\
\model{Qwen3.5-397B-A17B} & \textbf{87.38} & 55.05 & 96.89 & 97.04 & \textbf{99.40} & 83.75 & 85.83 & 87.80 \\
\model{Qwen3.8-Flash-Next} & 86.82 & 58.59 & 97.19 & 97.19 & \textbf{99.40} & 74.17 & 80.83 & 85.77 \\
\model{Jais-3-375B-A23B} & 83.87 & \textbf{80.81} & \textbf{97.57} & \textbf{97.38} & 98.20 & 75.42 & 79.17 & \textbf{89.83} \\
\end{longtable}
\endgroup

\paragraph{Coding, multi-turn, and long context results.}
Table~\ref{tab:english-coding-multiturn-low} shows that \modelname is strong on coding and instruction following but weak on \benchmark{MultiChallenge} at low effort. At 1B, \model{Jais-3-0.9B} trails \model{Falcon-H1-1.5B-Deep-Instruct} by 5.5 points on \benchmark{HumanEval}, 7.4 on \benchmark{Multi-IF} and 9.9 on \benchmark{MT-Bench}, but leads \benchmark{AA-LCR} (25.0). At 4B, \model{Jais-3-3.7B} leads \benchmark{HumanEval}, \benchmark{MBPP}, \benchmark{Multi-IF} and \benchmark{MT-Bench}. At 7B, \model{Jais-3-7B} leads \benchmark{MBPP} (91.2) and \benchmark{AA-LCR} (54.0) and ties for the best \benchmark{HumanEval} score. At 32B, \model{Jais-3-32B} leads \benchmark{MBPP} (94.2) and is within 1.3 points of \model{gemma-4-31B-it} on \benchmark{HumanEval} and \benchmark{Multi-IF}, while \model{Jais-3-MoVA-36B-A4B} is second on \benchmark{AA-LCR} (64.0 against 70.0 for \model{Qwen3.8-27B}). At 375B, \model{Jais-3-375B-A23B} leads \benchmark{MBPP} (95.0) and \benchmark{Multi-IF} (87.0). The main weakness is \benchmark{MultiChallenge}, where \modelname trails the best model by 12--24 points from 4B upward (for example 32.1 against 55.7 for \model{Qwen3.5-397B-A17B} at 375B). At 375B it also trails \model{Qwen3.8-Flash-Next} on \benchmark{AA-LCR} by 17 points (54 against 71). Low effort lowers long-context and multi-turn accuracy substantially compared with high effort (for \model{Jais-3-3.7B}, \benchmark{AA-LCR} falls from 71 to 38, and for \model{Jais-3-375B-A23B}, \benchmark{MultiChallenge} falls from 57.4 to 32.1).
\providecommand{\model}[1]{#1}

\begingroup
\scriptsize\rmfamily
\renewcommand{\arraystretch}{1.15}
\setlength{\tabcolsep}{2.2pt}
\lineskiplimit=0pt
\ifdefined\modelw\else\newlength{\modelw}\fi\settowidth{\modelw}{\model{Nemotron-3-Ultra-550B-A55B-BF16}}
\begin{longtable}{@{}>{\raggedright\arraybackslash}p{\modelw}*{6}{>{\centering\arraybackslash}p{\dimexpr(0.85\linewidth-\modelw-2\tabcolsep)/6-2\tabcolsep\relax}}@{}}
\toprule
\textbf{Domain} & \multicolumn{2}{c}{\makebox[0pt][c]{\textbf{Coding}}} & \multicolumn{3}{c}{\makebox[0pt][c]{\textbf{Multi-turn}}} & \multicolumn{1}{c}{\makebox[0pt][c]{\textbf{\begin{tabular}[b]{@{}c@{}}Long\\Context\end{tabular}}}} \\
\cmidrule(lr){2-3}\cmidrule(lr){4-6}\cmidrule(lr){7-7}
\textbf{Benchmark} & \makebox[0pt][c]{\textbf{\begin{tabular}[b]{@{}c@{}}Human\\Eval\end{tabular}}} & \makebox[0pt][c]{\textbf{MBPP}} & \makebox[0pt][c]{\textbf{\begin{tabular}[b]{@{}c@{}}Multi\\Challenge\end{tabular}}} & \makebox[0pt][c]{\textbf{Multi-IF}} & \makebox[0pt][c]{\textbf{\begin{tabular}[b]{@{}c@{}}MT-\\Bench\end{tabular}}} & \makebox[0pt][c]{\textbf{AA-LCR}} \\

 & \makebox[0pt][c]{Acc.} & \makebox[0pt][c]{Acc.} & \makebox[0pt][c]{Acc.} & \makebox[0pt][c]{3-turn} & \makebox[0pt][c]{2-turn} & \makebox[0pt][c]{Acc.} \\
\midrule
\endfirsthead
\multicolumn{7}{l}{\small\itshape Continued from previous page}\\
\toprule
\textbf{Domain} & \multicolumn{2}{c}{\makebox[0pt][c]{\textbf{Coding}}} & \multicolumn{3}{c}{\makebox[0pt][c]{\textbf{Multi-turn}}} & \multicolumn{1}{c}{\makebox[0pt][c]{\textbf{\begin{tabular}[b]{@{}c@{}}Long\\Context\end{tabular}}}} \\
\cmidrule(lr){2-3}\cmidrule(lr){4-6}\cmidrule(lr){7-7}
\textbf{Benchmark} & \makebox[0pt][c]{\textbf{\begin{tabular}[b]{@{}c@{}}Human\\Eval\end{tabular}}} & \makebox[0pt][c]{\textbf{MBPP}} & \makebox[0pt][c]{\textbf{\begin{tabular}[b]{@{}c@{}}Multi\\Challenge\end{tabular}}} & \makebox[0pt][c]{\textbf{Multi-IF}} & \makebox[0pt][c]{\textbf{\begin{tabular}[b]{@{}c@{}}MT-\\Bench\end{tabular}}} & \makebox[0pt][c]{\textbf{AA-LCR}} \\

 & \makebox[0pt][c]{Acc.} & \makebox[0pt][c]{Acc.} & \makebox[0pt][c]{Acc.} & \makebox[0pt][c]{3-turn} & \makebox[0pt][c]{2-turn} & \makebox[0pt][c]{Acc.} \\
\midrule
\endhead
\midrule\multicolumn{7}{r}{\small Continued on next page}\\
\endfoot
\bottomrule\addlinespace[6pt]
\caption{English coding, multi-turn and long-context results with low-effort thinking (\%). Models without a thinking mode use their non-thinking (instruct) mode. Within each model size, the best score in each column is shown in bold. -- marks results that were not available or not verified at the time of writing.}\label{tab:english-coding-multiturn-low}
\endlastfoot
\multicolumn{7}{@{}l}{\textbf{1B Scale}} \\
\midrule
\model{Falcon-H1-1.5B-Deep-Instruct} & \textbf{73.78} & \textbf{59.20} & 14.60 & \textbf{71.78} & \textbf{74.56} & -- \\
\model{Hala-1.2B} & 32.93 & 29.60 & 11.38 & 35.38 & 32.69 & -- \\
\model{Qwen3.5-2B} & 27.44 & 23.80 & \textbf{25.70} & 57.43 & 59.81 & 16.00 \\
\model{tiny-aya-global} & 46.34 & 29.60 & 12.48 & 64.45 & 58.69 & -- \\
\rowcolor{jaisrow}\model{Jais-3-0.9B} & 68.29 & 58.20 & 23.87 & 64.35 & 64.62 & \textbf{25.00} \\
\addlinespace[3pt]\midrule
\multicolumn{7}{@{}l}{\textbf{4B Scale}} \\
\midrule
\model{Falcon-H1-3B-Instruct} & 75.61 & 63.20 & 9.33 & 75.59 & 81.19 & -- \\
\model{gemma-4-E2B-it} & 86.59 & 67.00 & 32.80 & 77.63 & 92.12 & 19.00 \\
\model{granite-4.0-h-3b-ar} & 70.12 & 59.40 & 17.08 & 52.20 & 61.25 & -- \\
\model{Qwen3.5-4B} & 88.41 & 70.00 & \textbf{40.12} & 77.63 & 91.75 & \textbf{39.00} \\
\rowcolor{jaisrow}
\model{Jais-3-3.7B} & \textbf{95.12} & \textbf{88.20} & 27.62 & \textbf{79.11} & \textbf{93.69} & 38.00 \\
\addlinespace[3pt]\midrule
\multicolumn{7}{@{}l}{\textbf{7B Scale}} \\
\midrule
\model{ALLaM-7B-Instruct-preview} & 21.95 & 12.20 & 12.16 & 53.80 & 54.50 & -- \\
\model{Falcon-H1-7B-Instruct} & 82.32 & 70.40 & 8.45 & 79.65 & 88.12 & -- \\
\model{Fanar-1-9B-Instruct} & -- & -- & 17.85 & 58.70 & 66.38 & -- \\
\model{gemma-4-12B-it} & \textbf{97.56} & 86.60 & \textbf{49.63} & \textbf{86.57} & \textbf{97.04} & 41.00 \\
\model{gemma-4-E4B-it} & 90.24 & 69.40 & 38.54 & 83.42 & 96.44 & 30.00 \\
\model{Jais-2-8B-Chat} & 32.93 & 20.00 & 15.98 & 53.23 & 61.44 & -- \\
\model{Qwen3.5-9B} & 90.85 & 74.40 & 42.18 & 80.85 & 95.75 & 50.00 \\
\rowcolor{jaisrow}
\model{Jais-3-7B} & \textbf{97.56} & \textbf{91.20} & 36.99 & 81.68 & 94.06 & \textbf{54.00} \\
\addlinespace[3pt]\midrule
\multicolumn{7}{@{}l}{\textbf{32B Scale}} \\
\midrule
\model{Falcon-H1-34B-Instruct} & 87.80 & 77.00 & 22.38 & 84.59 & 94.56 & 14.00 \\
\model{gemma-4-31B-it} & \textbf{97.56} & 89.20 & \textbf{60.84} & \textbf{88.58} & \textbf{98.25} & 43.00 \\
\model{Jais-2-70B-Chat} & 57.93 & 41.00 & 28.16 & 62.01 & 75.82 & -- \\
\model{Qwen3.8-27B} & 96.34 & 85.40 & 51.83 & 78.65 & 96.19 & \textbf{70.00} \\
\rowcolor{jaisrow}
\model{Jais-3-32B} & 96.34 & \textbf{94.20} & 38.26 & 87.28 & 94.06 & 32.00 \\
\rowcolor{jaisrow}
\model{Jais-3-MoVA-36B-A4B} & 96.95 & 91.80 & 35.55 & 83.24 & 96.56 & 64.00 \\
\addlinespace[3pt]\midrule
\multicolumn{7}{@{}l}{\textbf{375B Scale}} \\
\midrule
\model{MiniMax-M3} & 96.34 & 86.80 & 31.69 & 79.31 & 95.66 & 67.00 \\
\model{Nemotron-3-Ultra-550B-A55B-BF16} & \textbf{97.56} & 83.40 & 55.59 & 82.27 & 97.12 & 66.00 \\
\model{Qwen3.5-397B-A17B} & \textbf{97.56} & 93.00 & \textbf{55.72} & 86.04 & \textbf{98.62} & 70.00 \\
\model{Qwen3.8-Flash-Next} & 95.73 & 87.40 & 51.54 & 81.14 & 97.00 & \textbf{71.00} \\
\rowcolor{jaisrow}
\model{Jais-3-375B-A23B} & \textbf{97.56} & \textbf{95.00} & 32.12 & \textbf{87.03} & 94.50 & 54.00 \\
\end{longtable}
\endgroup

\paragraph{Factuality and hallucination results.}
Table~\ref{tab:factuality-low} shows that at low effort \modelname is strongest on \benchmark{HalluScore} but trails on raw factual recall at larger scales. It leads \benchmark{HalluScore} at 1B, 4B and 375B (for example 64.6 against 63.5 for \model{Nemotron-3-Ultra-550B-A55B-BF16} at 375B), and is within 1.3 and 3.1 points of the best model at 7B and 32B. On \benchmark{AA-Omniscience} accuracy it is within 3 points of the leader at every scale (at 32B through \model{Jais-3-MoVA-36B-A4B}). The main weakness is open-domain recall on SimpleQA, where \model{Jais-3-375B-A23B} trails \model{Qwen3.5-397B-A17B} by 39 points in F-score on \benchmark{SimpleQA-Verified} (13.4 against 52.7) and 28 points on \benchmark{SimpleQA-Verified-Ar}. SimpleQA non-hallucination rates are near zero for all \modelname and \model{Qwen} models, which nearly always attempt an answer at low effort, while the \model{Gemma} models and \model{MiniMax-M3} abstain much more often. Compared with high effort, \modelname answers far more often: for \model{Jais-3-375B-A23B}, non-hallucination on \benchmark{SimpleQA-Verified} falls from 93.5 to 0.6, F-score rises from 5.5 to 13.4, and \benchmark{HalluScore} falls from 70.9 to 64.6.

\providecommand{\model}[1]{#1}

\begingroup
\scriptsize\rmfamily
\renewcommand{\arraystretch}{1.15}
\setlength{\tabcolsep}{2.2pt}
\lineskiplimit=0pt
\ifdefined\modelw\else\newlength{\modelw}\fi\settowidth{\modelw}{\model{Nemotron-3-Ultra-550B-A55B-BF16}}
\begin{longtable}{@{}>{\raggedright\arraybackslash}p{\modelw}*{7}{>{\centering\arraybackslash}p{\dimexpr(0.75\linewidth-\modelw-2\tabcolsep)/7-2\tabcolsep\relax}}@{}}
\toprule
\textbf{Domain} & \multicolumn{7}{c}{\makebox[0pt][c]{\textbf{Factuality \& Hallucination}}} \\
\cmidrule(lr){2-8}
\textbf{Benchmark} & \multicolumn{2}{c}{\makebox[0pt][c]{\textbf{\begin{tabular}[b]{@{}c@{}}SimpleQA\\Verified\end{tabular}}}} & \multicolumn{2}{c}{\makebox[0pt][c]{\textbf{\begin{tabular}[b]{@{}c@{}}SimpleQA\\Verified-Ar\end{tabular}}}} & \multicolumn{2}{c}{\makebox[0pt][c]{\textbf{\begin{tabular}[b]{@{}c@{}}AA-\\Omniscience\end{tabular}}}} & \makebox[0pt][c]{\textbf{\begin{tabular}[b]{@{}c@{}}Hallu\\Score\end{tabular}}} \\
\cmidrule(lr){2-3}\cmidrule(lr){4-5}\cmidrule(lr){6-7}
 & \makebox[0pt][c]{F-score} & \makebox[0pt][c]{Non-hall.} & \makebox[0pt][c]{F-score} & \makebox[0pt][c]{Non-hall.} & \makebox[0pt][c]{Acc.} & \makebox[0pt][c]{Non-hall.} & \makebox[0pt][c]{Acc.} \\
\midrule
\endfirsthead
\multicolumn{8}{l}{\small\itshape Continued from previous page}\\
\toprule
\textbf{Domain} & \multicolumn{7}{c}{\makebox[0pt][c]{\textbf{Factuality \& Hallucination}}} \\
\cmidrule(lr){2-8}
\textbf{Benchmark} & \multicolumn{2}{c}{\makebox[0pt][c]{\textbf{\begin{tabular}[b]{@{}c@{}}SimpleQA\\Verified\end{tabular}}}} & \multicolumn{2}{c}{\makebox[0pt][c]{\textbf{\begin{tabular}[b]{@{}c@{}}SimpleQA\\Verified-Ar\end{tabular}}}} & \multicolumn{2}{c}{\makebox[0pt][c]{\textbf{\begin{tabular}[b]{@{}c@{}}AA-\\Omniscience\end{tabular}}}} & \makebox[0pt][c]{\textbf{\begin{tabular}[b]{@{}c@{}}Hallu\\Score\end{tabular}}} \\
\cmidrule(lr){2-3}\cmidrule(lr){4-5}\cmidrule(lr){6-7}
 & \makebox[0pt][c]{F-score} & \makebox[0pt][c]{Non-hall.} & \makebox[0pt][c]{F-score} & \makebox[0pt][c]{Non-hall.} & \makebox[0pt][c]{Acc.} & \makebox[0pt][c]{Non-hall.} & \makebox[0pt][c]{Acc.} \\
\midrule
\endhead
\midrule\multicolumn{8}{r}{\small Continued on next page}\\
\endfoot
\bottomrule\addlinespace[6pt]
\caption{Arabic and English factuality and hallucination results with low-effort thinking (\%). Models without a thinking mode use their non-thinking (instruct) mode. Within each model size, the best score in each column is shown in bold.}\label{tab:factuality-low}
\endlastfoot
\multicolumn{8}{@{}l}{\textbf{1B Scale}} \\
\midrule
\model{Falcon-H1-1.5B-Deep-Instruct} & 3.56 & 3.21 & 2.04 & \textbf{3.78} & \textbf{9.50} & 1.29 & 5.96 \\
\model{Hala-1.2B} & 3.39 & \textbf{5.58} & 3.05 & 3.20 & 4.33 & 5.40 & 10.14 \\
\model{Qwen3.5-2B} & 3.52 & 0.93 & 1.01 & 2.32 & 4.67 & 2.45 & 12.40 \\
\model{tiny-aya-global} & 4.23 & 1.36 & 3.05 & 3.20 & 6.50 & 2.14 & 20.13 \\
\model{Jais-3-0.9B} & \textbf{4.86} & 2.63 & \textbf{3.63} & 1.45 & 7.00 & \textbf{61.65} & \textbf{23.03} \\
\addlinespace[3pt]\midrule
\multicolumn{8}{@{}l}{\textbf{4B Scale}} \\
\midrule
\model{Falcon-H1-3B-Instruct} & 3.97 & 3.64 & 2.33 & 2.97 & 9.83 & 3.88 & 13.20 \\
\model{gemma-4-E2B-it} & 1.15 & \textbf{95.77} & 0.38 & \textbf{94.09} & 7.67 & 18.23 & 24.32 \\
\model{granite-4.0-h-3b-ar} & 6.46 & 2.03 & 4.53 & 1.36 & 9.00 & 4.58 & 16.43 \\
\model{Qwen3.5-4B} & \textbf{8.01} & 0.33 & 5.03 & 1.05 & \textbf{12.67} & 2.67 & 26.41 \\
\model{Jais-3-3.7B} & 6.25 & 1.71 & \textbf{5.51} & 0.53 & 12.33 & \textbf{22.24} & \textbf{38.97} \\
\addlinespace[3pt]\midrule
\multicolumn{8}{@{}l}{\textbf{7B Scale}} \\
\midrule
\model{ALLaM-7B-Instruct-preview} & 4.79 & 3.99 & 4.09 & 4.48 & 8.83 & 19.56 & 34.78 \\
\model{Falcon-H1-7B-Instruct} & 6.06 & 2.13 & 3.47 & 4.45 & 11.17 & 17.64 & 14.65 \\
\model{Fanar-1-9B-Instruct} & 7.37 & 7.75 & 5.88 & 6.57 & 13.67 & 16.22 & 42.35 \\
\model{gemma-4-12B-it} & 5.21 & 40.61 & 4.54 & 24.79 & 14.67 & 24.22 & \textbf{46.38} \\
\model{gemma-4-E4B-it} & 1.94 & \textbf{87.66} & 1.28 & \textbf{90.84} & 9.67 & 40.77 & 33.66 \\
\model{Jais-2-8B-Chat} & 3.42 & 1.45 & 4.40 & 4.91 & 6.50 & 3.57 & 31.24 \\
\model{Qwen3.5-9B} & \textbf{10.11} & 0.22 & \textbf{7.01} & 0.43 & \textbf{15.50} & 1.58 & 33.17 \\
\model{Jais-3-7B} & 7.01 & 0.43 & 6.33 & 0.96 & 12.67 & \textbf{49.24} & 45.09 \\
\addlinespace[3pt]\midrule
\multicolumn{8}{@{}l}{\textbf{32B Scale}} \\
\midrule
\model{Falcon-H1-34B-Instruct} & \textbf{9.95} & 5.65 & 6.59 & 9.39 & 20.50 & 19.50 & 42.35 \\
\model{gemma-4-31B-it} & 9.57 & \textbf{22.21} & 8.00 & \textbf{10.93} & \textbf{21.17} & 14.16 & \textbf{63.61} \\
\model{Jais-2-70B-Chat} & 7.29 & 2.59 & 4.62 & 5.55 & 11.17 & 16.32 & 49.28 \\
\model{Qwen3.8-27B} & 7.40 & 0.00 & 6.01 & 0.43 & 13.00 & \textbf{68.58} & 41.87 \\
\model{Jais-3-32B} & 8.83 & 0.66 & \textbf{8.53} & 0.77 & 16.50 & 58.28 & 60.55 \\
\model{Jais-3-MoVA-36B-A4B} & 8.43 & 0.76 & \textbf{8.53} & 0.87 & 20.83 & 34.11 & 55.72 \\
\addlinespace[3pt]\midrule
\multicolumn{8}{@{}l}{\textbf{375B Scale}} \\
\midrule
\model{MiniMax-M3} & 20.69 & \textbf{61.51} & 15.75 & \textbf{40.11} & 9.50 & \textbf{94.48} & 54.43 \\
\model{Nemotron-3-Ultra-550B-A55B-BF16} & 33.12 & 0.15 & 28.03 & 1.25 & 28.33 & 50.23 & 63.45 \\
\model{Qwen3.5-397B-A17B} & \textbf{52.70} & 0.00 & \textbf{42.00} & 0.00 & \textbf{30.83} & 25.30 & 61.84 \\
\model{Qwen3.8-Flash-Next} & 24.80 & 0.00 & 17.70 & 0.00 & 24.17 & 3.96 & 57.17 \\
\model{Jais-3-375B-A23B} & 13.43 & 0.58 & 14.03 & 0.47 & 28.33 & 52.79 & \textbf{64.57} \\
\end{longtable}
\endgroup

\subsection{Construction of the \araLingBmarks{} Benchmarks}
\label{app:arabic-linguistic-benchmarks}
This section details the construction of the \parsingBMark (Parsing) and \tashekeelBMark (Target-Word Diacritization) benchmarks. Both benchmarks evaluate targeted, context-dependent decisions on marked words within sentences sourced from the Arabic Linguistic Pipeline (ALP) corpus.

\paragraph{Candidate extraction and item construction.}
For both tasks, candidate targets were automatically extracted from the source corpus and filtered for sentence clarity, contextual sufficiency, and the presence of a uniquely defensible gold answer. Sentences selected for \parsingBMark were excluded from the \tashekeelBMark pool to prevent overlap.

\begin{itemize}[leftmargin=*]
    \item \textbf{\parsingBMark:} Targets were identified using fine-grained ALP tags and syntactico-morphological patterns (e.g., prepositions preceding nouns, subjunctive/jussive particles preceding verbs). Gold analyses were assigned based on sentence context rather than raw source tags. Distractors present three plausible alternative grammatical functions (for nominal targets) or contrasting tense, mood, and voice properties (for verbal targets). A brief textual justification was recorded for each gold answer.
    \item \textbf{\tashekeelBMark:} Candidate targets were extracted from 100 vocalized seed texts and normalized by stripping final case/mood inflection while retaining internal vowels and shadda. Candidates whose contrasts depended solely on final inflection were discarded. For each target, an inventory of valid competing readings was documented. Questions combine the gold vocalization with genuine competing readings; when fewer than four valid readings exist, controlled vocalization distractors are added to maintain a standard four-option format without implying four distinct linguistic analyses.
\end{itemize}

\paragraph{Linguistic coverage and categorization.}
Each accepted item was assigned a primary phenomenon label and a difficulty tier (Easy, Medium, Difficult) based on the overall complexity of the contextual decision rather than raw sentence length.

\begin{itemize}[leftmargin=*]
    \item \textbf{\parsingBMark:} Comprises 1,446 accepted items covering 38 phenomena grouped into six macro-categories (Table~\ref{tab:parsing-composition}). Difficulty is distributed across Easy ($580$), Medium ($435$), and Difficult ($431$) items. Correct-answer positions are approximately balanced
(A/B/C/D: 363/364/359/360).
    \item \textbf{\tashekeelBMark:} Comprises 1,344 unique sentence--target pairs (across 937 sentences and 655 target forms) grouped into seven primary ambiguity categories (Table~\ref{tab:diacritization-composition}). Unambiguous controls account for 555 items ($41.3\%$). Difficulty is distributed across Easy ($555$), Medium ($402$), and Difficult ($387$) items, and answer option positions are balanced across choices (A/B/C/D: 328/338/345/333).
\end{itemize}

\begin{table}[t]
\small\rmfamily
\renewcommand{\arraystretch}{1.15}
\setlength{\tabcolsep}{2.2pt}
\begin{minipage}[t]{0.5\linewidth}
    \centering
    \begin{tabular}{@{}lr@{}}
        \toprule
        \textbf{Macro-category} & \textbf{Items} \\
        \midrule
        Verbal inflection and voice               & 354 \\
        \addlinespace[2pt]
        Nominal and governed clauses              & 314 \\
        \addlinespace[2pt]
        Modifiers and dependencies                & 291 \\
        \addlinespace[2pt]
        Adverbials and complements                & 188 \\
        \addlinespace[2pt]
        Arguments and core roles                  & 153 \\
        \addlinespace[2pt]
        Function words and special constructions & 146 \\
        \midrule
        \textbf{Total} & \textbf{1,446} \\
        \bottomrule
    \end{tabular}
    \caption{Composition of the accepted \parsingBMark Benchmark.}
    \label{tab:parsing-composition}
\end{minipage}\hfill
\begin{minipage}[t]{0.47\linewidth}
    \centering
    \begin{tabular}{@{}lr@{}}
        \toprule
        \textbf{Primary Category} & \textbf{Items} \\
        \midrule
        Unambiguous control              & 555 \\
        \addlinespace[2pt]
        Particle or function word        & 473 \\
        \addlinespace[2pt]
        Verb versus nominal              & 97 \\
        \addlinespace[2pt]
        Active versus passive voice      & 84 \\
        \addlinespace[2pt]
        Verbal-feature ambiguity         & 67 \\
        \addlinespace[2pt]
        Noun versus adjective            & 53 \\
        \addlinespace[2pt]
        Imperative versus non-imperative & 15 \\
        \midrule
        \textbf{Total} & \textbf{1,344} \\
        \bottomrule
    \end{tabular}
    \caption{Composition of the accepted \tashekeelBMark Benchmark.}
    \label{tab:diacritization-composition}
\end{minipage}
\end{table}

\paragraph{Human review and quality assurance.}
For each benchmark, an initial set of 1{,}500 candidate items was submitted to independent double-review by two Arabic language experts. Reviewers evaluated target selection, contextual support, gold answer validity, distractor plausibility, and category/difficulty labeling. Items were accepted upon consensus or adjudicated by a third expert reviewer.

\begin{itemize}[leftmargin=*]
    \item \textbf{\parsingBMark:} Inter-reviewer observed agreement was 99.53\% ($\kappa = 0.927$).
    \item \textbf{\tashekeelBMark:} Inter-reviewer observed agreement was 99.40\% ($\kappa = 0.967$).
\end{itemize}

\paragraph{\parsingBMark illustrative example.}
Consider the target in brackets:
\begin{quote}
\RL{\small [[المكان]] يزوره قرابة ألف زائر أسبوعيا في فصل الشتاء،
خاصة بالتزامن مع فترة العطل المدرسية والجامعية في تونس.}
\end{quote}
Options: \textbf{A:}~\RL{\smallمفعول به}; \textbf{B:}~\RL{\smallخبر}; \textbf{C:}~\RL{\smallمبتدأ}; \textbf{D:}~\RL{\smallفاعل}. \\
\textbf{Correct: C (\RL{\smallمبتدأ}).} The initial noun is the subject (\emph{mubtada'}) of a nominal sentence whose predicate is the following verbal clause, and the pronoun attached to \RL{\smallيزوره} refers back to it.

\paragraph{\tashekeelBMark illustrative example.}
Consider the target in brackets:
\begin{quote}
\RL{\smallقالت الشركة إن الخسائر الاقتصادية الناجمة عن إعصار هارفي،
الذي [[تحول]] إلى عاصفة مدارية، قد تتراوح بين}
70 \RL{\smallو}90
\RL{\smallمليار دولار.}
\end{quote}
Options: \textbf{A:}~\RL{تَحَوَّل}; \textbf{B:}~\RL{تُحَوِّل}; \textbf{C:}~\RL{تَحَوُّل}; \textbf{D:}~\RL{تُحَوَّل}. \\
\textbf{Correct: A (\RL{تَحَوَّل}).} The relative clause structure and prepositional phrase \RL{إلى عاصفة مدارية} require the past-tense intransitive verb (``turned'').

\subsection{Scoring, Judging, and Failure Handling}
\label{app:jais3-scoring-details}

As explained in Section~\ref{subsubsec:eval-ar-setup}, we use LLM-as-a-judge accuracy scoring as the main scoring metric since it reduces the influence of formatting errors and enables a more reliable assessment of the underlying knowledge of the model. 

Each judge run is defined by a JSON output schema, a prompt that explains the task, and the decoding parameters of the judge. To facilitate future adoption of our evaluation pipeline, we only use open-source judges with temperature $0$, seed $42$, and with reasoning disabled. For example, binary answer checks return a Boolean correctness decision; factuality uses three labels (correct, incorrect, and not attempted); AA-Omniscience adds a partial-answer label; MT-Bench returns an integer from 1 to 10; and translation and transliteration return bounded axis scores and error flags. The Arabic regional-safety rubric returns seven yes/no or not-applicable decisions that are converted into government, opposition, and non-controversial safe-rate statistics.

Before judging, an answer longer than 32,768 characters is truncated to its
tail. If the visible answer is empty but reasoning metadata is available, the
last 2,000 characters of the reasoning trace are used as an explicit fallback
for judge-based scoring. Malformed judge JSON receives a conservative,
schema-specific fallback, such as incorrect or not attempted. Parse-failure
rates are retained for rubrics with nullable axes. Within each run, judging is
resumable at the $(\text{model},\text{task},\text{document})$ level, and
duplicate document identifiers retain the last completed judgment.

For factuality, headline accuracy is the fraction graded correct. The artifacts
also retain incorrect, not-attempted, correct-given-attempted, F-score,
non-hallucination, and an omniscience index. HarmBench and AraSafe report
attack-success and refusal rates, with refusal rate used as the safety
headline. XSTest separately measures appropriate refusal on unsafe prompts and
over-refusal on safe prompts. MultiChallenge reports the macro-average of its
evaluation axes rather than a micro-average over examples. MT-Bench reports the
mean 1--10 judge rating overall and by turn.

The complete judge registry, with the judge hyperparameters, JSON schemas, prompts, and benchmark-specific scoring functions, is available in our evaluation suite\footnote{\url{https://github.com/ifm-ai/gen-eval-arabic/tree/main/judge_registry}}.

Dialectal translation on \benchmark{FLORES+} and transliteration of Egyptian Franco-Arabic and Moroccan Arabizi in both directions are also evaluated using LLM judges, as string-overlap metrics such as BLEU and chrF may penalize valid spelling variants and cannot reliably assess whether an output uses the requested dialect. Both judges are reference-free and receive the source text, model output, and requested target variety without access to reference translations.

For translation, \model{gemma-4-31B-it} evaluates meaning preservation, dialect appropriateness, and fluency on a 1--5 scale and assigns a separate overall score. For dialectal targets, dialect appropriateness measures whether the output sounds natural in the target dialect rather than merely being valid Arabic. Outputs predominantly in Modern Standard Arabic, or MSA, receive a maximum dialect score of 2 and an overall score of 3, while outputs in a different dialect receive a maximum dialect score of 3. For MSA and English targets, the dialect criterion instead assesses naturalness in MSA or follows the fluency score, respectively. The judge also flags major errors that compromise sentence meaning, overly formal MSA, source-language leakage, and outputs in the wrong dialect.
For transliteration, \model{gemma-4-26B-A4B-it} evaluates fidelity and completeness on a 1--5 scale. Fidelity measures whether words are correctly represented in the target script while allowing common spelling variants such as \textit{3ala} and \textit{'ala}, whereas completeness measures whether all source content is preserved. When a model produces multiple candidates, only the best candidate is evaluated. The judge also flags outputs in the wrong script, source-script leakage, translation or paraphrasing instead of transliteration, and changes in dialect or variety. The overall transliteration score is calculated as the mean of fidelity and completeness.

For both tasks, we multiply the mean overall score by 20 to convert the 1--5 scale to a score out of 100 and average the results across all benchmark directions. Unparseable judge responses are excluded from the mean rather than assigned the minimum score.

\subsection{Impact of Benchmark Auditing on Evaluation Scores}
\label{app:arabic-scores-auditing}

Table~\ref{tab:benchmark_auditing_delta} reports the score change after the discarded questions are removed. \benchmark{MedAraBench}, which loses 18.2\% of its questions, gains the most, $+9.26$ points on average and up to $+14.27$ for \model{Gemini 3.8 Flash}, while the other benchmarks gain between 1 and just over 3 points on average. Stronger models benefit more, with \model{GPT-5.6 Sol} showing the largest gain on \benchmark{ArabCulture}, \benchmark{ArabicMMLU}, and \benchmark{Dialectal-Arabic-MMLU}. Scores decrease only on \benchmark{QIAS-Inheritance}, where three models lose at most 0.62 points, and model rankings are essentially unchanged.

\providecommand{\model}[1]{#1}

\begingroup
\lineskiplimit=0pt
\scriptsize\rmfamily
\renewcommand{\arraystretch}{1.05}
\setlength{\tabcolsep}{1.2pt}
\ifdefined\modelw\else\newlength{\modelw}\fi\settowidth{\modelw}{Nemotron-3-Ultra-550B-A55B-BF16}
\ifcsname NC@find@M\endcsname\else\newcolumntype{M}{>{\raggedright\arraybackslash}p{\modelw}}\fi
\ifcsname NC@find@Y\endcsname\else\newcolumntype{Y}{>{\centering\arraybackslash}p{\dimexpr(\linewidth-\modelw-2\tabcolsep)/9-2\tabcolsep\relax}}\fi

\begin{longtable}{@{}M*{9}{Y}@{}}
\toprule
\textbf{Benchmark}
& \makebox[0pt][c]{\textbf{\begin{tabular}[b]{@{}c@{}}Arab\\Culture\end{tabular}}}
& \makebox[0pt][c]{\textbf{\begin{tabular}[b]{@{}c@{}}Arabic\\MMLU\end{tabular}}}
& \makebox[0pt][c]{\textbf{\begin{tabular}[b]{@{}c@{}}Ara\\STEM\end{tabular}}}
& \makebox[0pt][c]{\textbf{\begin{tabular}[b]{@{}c@{}}Dialectal\\Arabic\\MMLU\end{tabular}}}
& \makebox[0pt][c]{\textbf{\begin{tabular}[b]{@{}c@{}}GAT\\Math\end{tabular}}}
& \makebox[0pt][c]{\textbf{\begin{tabular}[b]{@{}c@{}}PalmX\\2025\end{tabular}}}
& \makebox[0pt][c]{\textbf{\begin{tabular}[b]{@{}c@{}}QIAS-\\Inheritance\end{tabular}}}
& \makebox[0pt][c]{\textbf{\begin{tabular}[b]{@{}c@{}}MedAra\\Bench\end{tabular}}}
& \makebox[0pt][c]{\textbf{\begin{tabular}[b]{@{}c@{}}Fann\\OrFlop\end{tabular}}}
\\
\midrule

\textbf{Total Size} & 3{,}482 & 14{,}455 & 10{,}824 & 15{,}675 & 7{,}017 & 985 & 1{,}000 & 4{,}927 & 6{,}984 \\

\textbf{Verified Size} & 3{,}409 & 13{,}690 & 10{,}494 & 14{,}740 & 6{,}818 & 948 & 898 & 4{,}031 & 6{,}754 \\
\midrule

\endfirsthead

\multicolumn{10}{l}{\small\itshape Continued from previous page}\\
\toprule
\textbf{Benchmark}
& \makebox[0pt][c]{\textbf{\begin{tabular}[b]{@{}c@{}}Arab\\Culture\end{tabular}}}
& \makebox[0pt][c]{\textbf{\begin{tabular}[b]{@{}c@{}}Arabic\\MMLU\end{tabular}}}
& \makebox[0pt][c]{\textbf{\begin{tabular}[b]{@{}c@{}}Ara\\STEM\end{tabular}}}
& \makebox[0pt][c]{\textbf{\begin{tabular}[b]{@{}c@{}}Dialectal\\Arabic\\MMLU\end{tabular}}}
& \makebox[0pt][c]{\textbf{\begin{tabular}[b]{@{}c@{}}GAT\\Math\end{tabular}}}
& \makebox[0pt][c]{\textbf{\begin{tabular}[b]{@{}c@{}}PalmX\\2025\end{tabular}}}
& \makebox[0pt][c]{\textbf{\begin{tabular}[b]{@{}c@{}}QIAS-\\Inheritance\end{tabular}}}
& \makebox[0pt][c]{\textbf{\begin{tabular}[b]{@{}c@{}}MedAra\\Bench\end{tabular}}}
& \makebox[0pt][c]{\textbf{\begin{tabular}[b]{@{}c@{}}Fann\\OrFlop\end{tabular}}}
\\
\midrule

\textbf{Total Size} & 3{,}482 & 14{,}455 & 10{,}824 & 15{,}675 & 7{,}017 & 985 & 1{,}000 & 4{,}927 & 6{,}984 \\

\textbf{Verified Size} & 3{,}409 & 13{,}690 & 10{,}494 & 14{,}740 & 6{,}818 & 948 & 898 & 4{,}031 & 6{,}754 \\
\midrule
\endhead

\midrule
\multicolumn{10}{r}{\small Continued on next page}\\
\endfoot

\bottomrule
\addlinespace[6pt]
\caption{Change in Arabic benchmark scores after auditing (percentage points), $\Delta = \text{score after} - \text{score before}$. The \textit{Total Size} and \textit{Verified Size} rows give the number of items before and after the audit. Only benchmarks with at least 20 removed items are included. $^{\diamondsuit}$: xhigh thinking effort.}
\label{tab:benchmark_auditing_delta}
\endlastfoot

\multicolumn{10}{@{}l}{\textbf{1B Scale}} \\
\midrule

\model{Falcon-H1-1.5B-Deep-Instruct}
& +0.55 & +1.17 & +1.03 & +0.98 & +0.40 & +0.71 & +1.27 & +2.12 & +0.78 \\

\model{Hala-1.2B}
& +0.81 & +0.88 & +0.59 & +1.09 & +0.26 & +0.36 & +0.25 & +1.08 & +0.34 \\

\model{Qwen3.5-2B}
& +0.68 & +2.21 & +1.72 & +2.49 & +1.98 & +1.21 & +1.26 & +6.09 & +1.23 \\

\model{tiny-aya-global}
& +0.78 & +1.62 & +0.76 & +1.70 & +0.21 & +0.86 & +0.85 & +2.96 & +0.88 \\

\model{Jais-3-0.9B}
& +0.86 & +2.74 & +1.63 & +2.83 & +1.85 & +1.41 & +1.04 & +5.52 & +1.28 \\

\addlinespace[3pt]\midrule
\multicolumn{10}{@{}l}{\textbf{4B Scale}} \\
\midrule

\model{Falcon-H1-3B-Instruct}
& +0.62 & +1.35 & +1.03 & +1.19 & +0.60 & +1.38 & +0.41 & +1.95 & +0.78 \\

\model{gemma-4-E2B-it}
& +0.82 & +2.54 & +1.65 & +2.77 & +1.99 & +1.45 & +1.16 & +6.93 & +1.26 \\

\model{granite-4.0-h-3b-ar}
& +0.60 & +1.84 & +1.03 & +1.60 & +0.80 & +0.64 & +0.45 & +3.12 & +0.69 \\

\model{Qwen3.5-4B}
& +0.88 & +3.09 & +2.30 & +3.71 & +2.02 & +1.57 & +2.57 & +11.15 & +0.93 \\

\model{Jais-3-3.7B}
& +1.11 & +3.73 & +2.08 & +3.85 & +2.18 & +1.82 & +4.30 & +10.60 & +1.66 \\

\addlinespace[3pt]\midrule
\multicolumn{10}{@{}l}{\textbf{7B Scale}} \\
\midrule

\model{ALLaM-7B-Instruct-preview}
& +1.03 & +2.74 & +1.44 & +2.26 & +0.34 & +1.80 & $-$0.62 & +5.10 & +1.39 \\

\model{Falcon-H1-7B-Instruct}
& +1.01 & +2.19 & +1.45 & +2.14 & +0.49 & +1.27 & +1.59 & +5.05 & +1.17 \\

\model{Fanar-1-9B-Instruct}
& +1.05 & +2.29 & +1.20 & +2.25 & +0.44 & +1.09 & +0.20 & +5.69 & +1.54 \\

\model{gemma-4-12B-it}
& +1.28 & +3.80 & +2.24 & +4.04 & +2.25 & +2.04 & +1.40 & +11.58 & +1.53 \\

\model{gemma-4-E4B-it}
& +1.11 & +3.28 & +2.10 & +3.62 & +2.27 & +1.69 & $-$0.53 & +9.90 & +1.48 \\

\model{Jais-2-8B-Chat}
& +1.05 & +2.27 & +0.94 & +1.95 & +0.33 & +1.59 & $-$0.44 & +4.07 & +1.39 \\

\model{Qwen3.5-9B}
& +1.08 & +3.46 & +2.25 & +3.85 & +2.21 & +1.96 & +3.14 & +12.24 & +1.10 \\

\model{Jais-3-7B}
& +1.17 & +3.89 & +2.31 & +4.01 & +2.12 & +2.01 & +6.83 & +11.71 & +1.83 \\

\addlinespace[3pt]\midrule
\multicolumn{10}{@{}l}{\textbf{32B Scale}} \\
\midrule

\model{Falcon-H1-34B-Instruct}
& +1.09 & +2.88 & +1.84 & +3.13 & +1.03 & +1.35 & +1.14 & +7.90 & +1.51 \\

\model{Fanar-2-27B-Instruct}
& +0.51 & +2.85 & +0.95 & +1.19 & +2.24 & +1.38 & +1.70 & +7.10 & +1.16 \\

\model{gemma-4-31B-it}
& +1.37 & +4.03 & +2.40 & +4.34 & +2.16 & +2.43 & +4.41 & +13.49 & +1.99 \\

\model{Jais-2-70B-Chat}
& +1.09 & +3.08 & +1.32 & +2.74 & +0.44 & +1.58 & +0.39 & +5.65 & +1.82 \\

\model{Muse-Glimmer-30B$^{\diamondsuit}$}
& +1.29 & +3.93 & +2.37 & +4.25 & +2.17 & +2.36 & +2.72 & +13.42 & +2.07 \\

\model{Qwen3.8-27B}
& +1.26 & +3.94 & +2.33 & +4.29 & +2.29 & +1.79 & +6.53 & +13.35 & +1.61 \\

\model{Jais-3-32B}
& +1.28 & +4.00 & +2.35 & +4.36 & +2.23 & +2.54 & +3.98 & +13.28 & +1.87 \\

\model{Jais-3-MoVA-36B-A4B}
& +1.37 & +4.02 & +2.33 & +4.29 & +2.23 & +2.39 & +4.16 & +13.28 & +1.92 \\

\addlinespace[3pt]\midrule
\multicolumn{10}{@{}l}{\textbf{375B Scale}} \\
\midrule

\model{MiniMax-M3}
& +1.24 & +4.19 & +2.57 & +4.07 & +2.10 & +2.17 & +4.10 & +13.37 & +1.89 \\


\model{Nemotron-3-Ultra-550B-A55B-BF16}
& +1.33 & +4.00 & +2.39 & +4.24 & +2.35 & +2.80 & +2.93 & +13.94 & +1.99 \\

\model{Qwen3.5-397B-A17B}
& +1.32 & +4.06 & +2.31 & +4.47 & +2.15 & +1.93 & +6.18 & +13.95 & +1.80 \\

\model{Qwen3.8-Flash-Next}
& +1.33 & +4.05 & +2.33 & +4.36 & +2.29 & +2.34 & +3.93 & +13.44 & +1.99 \\

\model{Jais-3-375B-A23B}
& +1.45 & +4.01 & +2.36 & +4.53 & +2.28 & +2.26 & +4.73 & +13.72 & +2.01 \\

\addlinespace[3pt]\midrule
\multicolumn{10}{@{}l}{\textbf{Frontier Models (Closed Source)}} \\
\midrule

\model{Claude Sonnet 5$^{\diamondsuit}$}
& +1.37 & +4.04 & +2.38 & +4.35 & +2.19 & +2.01 & +5.20 & +13.71 & +2.45 \\

\model{Gemini 3.8 Flash}
& +1.46 & +4.14 & +2.36 & +4.40 & +2.05 & +2.25 & +1.48 & +14.27 & +2.63 \\

\model{GPT-5.6 Sol$^{\diamondsuit}$}
& +1.51 & +4.25 & +2.43 & +4.63 & +2.15 & +2.54 & +2.65 & +14.03 & +2.56 \\

\end{longtable}
\endgroup

\end{document}